# Breaking the Curse of Dimensionality in Deep Neural Networks by Learning Invariant Representations

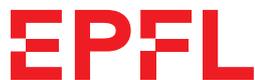

## Leonardo Petrini

acceptée sur proposition du jury :

Prof Hugo Dil, président du jury
Prof Matthieu Wyart, directeur de thèse
Prof Florent Krzakala, rapporteur
Prof Andrew Saxe, rapporteur
Prof Nathan Srebro, rapporteur

Tutto se pò fa'.
— Babbo

*Everything can be done.*
*(regarding practical problem-solving)*
*— Dad*

Ai miei genitori
*To my parents*

# Acknowledgements

I extend my deepest gratitude to Matthieu, my thesis advisor, for his invaluable guidance, mentorship, and thorough and honest feedback throughout the journey of this PhD.

I also thank the members of the jury—Prof. Florent Krzakala, Prof. Andrew Saxe, and Prof. Nathan Srebro—for their critical reading of my thesis and their insightful questions. Special acknowledgment goes to Prof. Hugo Dil, the president of the jury, for presiding over the evaluation process.

For my years at PCSL, I thank Mario for his teachings on problem-solving and coding and Francesco for being a mentor and collaborator on multiple projects. Alessandro, Antonio, and Umberto for the time spent together at Cubotron and outside. And Tom, I appreciate your understanding when we let our Italian conversations take over. A special mention goes to Corinne, our irreplaceable secretary, for her readiness to help and her untiring efforts. I thank all past group members of PCSL for the good times spent together.

After six fulfilling years in Lausanne, including four spent working towards this PhD, I owe a massive debt of gratitude to the friends I made along the way.[1] To Andre and Sté, we literally walked through countless adventures and thousands of kilometers side by side. To Giorgia, *l'amica geniale*, I owe you lots of laughs and advice. Even far apart, we always find ways to make wonderful memories together. To Diego, so deep in startup life he might as well be thru-hiking in New Zealand. You're missed, *caro*, but never forgotten. Ondine for the perpetual smile on your face; Ali, for having a WhatsApp group ready for every occasion; Niko, for all the gourmet meals you prepared us; and Eli, for sharing hikes, ski slopes, and office desks alike.

Being far from home never felt too distant, thanks to the friends I always find upon my return. Ale and Manu, despite our divergent paths in academics, our end-of-year reunions have been revitalizing exchanges. To my lifetime friends, Ale, Diego, Paco, Giulia, Lucrezia, Cecilia, Poch, Teo, Fil. Thank you for your enduring friendships and the precious moments we continue to share, whether near or far.

To my *nonni*, Angelo and Sara for constantly providing food and love. To my mum for teaching me how to organize work and get it done, to my dad for teaching me how to think out of the box and solve problems. Thank you for this, and for everything else. To Alle, for making life full of color.

*Lausanne, September 13, 2023*                                                    Leo

---

[1] Ordering in this paragraph is based on the number of shared WhatsApp groups (Andre and Sté 58, Giorgia 47, Diego 39, Ondine 33, Ali 23, Niko 17, Eli 12).

A Neural Network
*Drawing courtesy of A.P.*

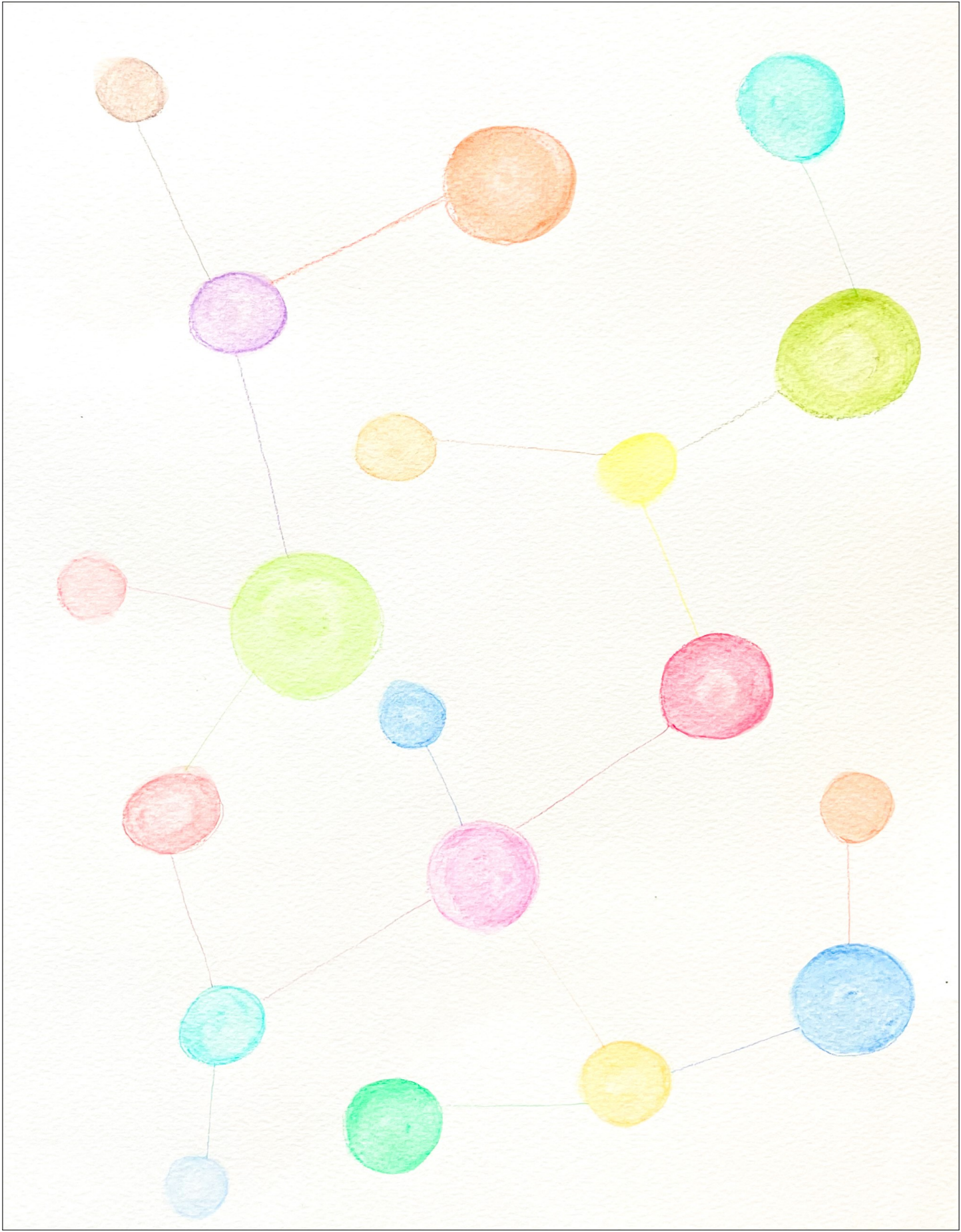

# Abstract

Artificial intelligence, particularly the subfield of machine learning, has seen a paradigm shift towards data-driven models that learn from and adapt to data. This has resulted in unprecedented advancements in various domains such as natural language processing and computer vision, largely attributed to *deep learning*, a special class of machine learning models. Deep learning arguably surpasses traditional approaches by learning the relevant features from raw data through a series of computational layers.

This thesis explores the theoretical foundations of deep learning by studying the relationship between the architecture of these models and the inherent structures found within the data they process. In particular, we ask: What drives the efficacy of deep learning algorithms and allows them to beat the so-called *curse of dimensionality*—i.e. the difficulty of generally learning functions in high dimensions due to the exponentially increasing need for data points with increased dimensionality? Is it their ability to learn relevant representations of the data by exploiting their structure? How do different architectures exploit different data structures? In order to address these questions, we push forward the idea that the structure of the data can be effectively characterized by its invariances——i.e. aspects that are irrelevant for the task at hand. Our methodology takes an empirical approach to deep learning, combining experimental studies with physics-inspired *toy models*. These simplified models allow us to investigate and interpret the complex behaviors we observe in deep learning systems, offering insights into their inner workings, with the far-reaching goal of bridging the gap between theory and practice.

Specifically, we compute tight generalization error rates of shallow fully connected networks demonstrating that they are capable of performing well by learning linear invariances, i.e. becoming insensitive to irrelevant linear directions in input space. However, we show that these network architectures can perform poorly in learning non-linear invariances such as rotation invariance or the invariance with respect to smooth deformations of the input. This result illustrates that, if a chosen architecture is not suitable for a task, it might overfit, making a kernel method, for which representations are not learned, potentially a better choice.

Modern architectures like convolutional neural networks, however, are particularly well-fitted to learn the non-linear invariances that are present in real data. In image classification, for example, the exact position of an object or feature might not be crucial for recognizing it. This property gives rise to an invariance with respect to small deformations. Our findings show that the neural networks that are more invariant to deformations tend to have higher performance, underlying the importance of exploiting such invariance.

Another key property that gives structure to real data is the fact that high-level features are

## Abstract

a hierarchical composition of lower-level features—a dog is made of a head and limbs, the head is made of eyes, nose, and mouth, which are then made of simple textures and edges. These features can be realized in multiple synonymous ways, giving rise to an invariance. To investigate the synonymic invariance that arises from the hierarchical structure of data, we introduce a toy data model that allows us to examine how features are extracted and combined to form increasingly higher-level representations. We show that deep neural networks, unlike their shallow counterparts, can learn this invariance layer by layer, and this allows beating the curse. Our analysis within this setting provides an estimate of the number of data samples needed for learning a task, given its hierarchical structure.

Overall, our research shows that deep learning is able to mitigate the curse of dimensionality by learning representations that are *invariant* to aspects of the data irrelevant for the task, effectively reducing the problem's dimensionality. Our quantitative characterizations of the generalization error as a function of the number of training points bring us closer to knowing a priori how many data points are needed to learn a task, given its structure.

# Résumé

L'intelligence artificielle, notamment le sous-domaine de l'apprentissage automatique, a connu un changement de paradigme vers des modèles basés sur les données qui apprennent et s'adaptent à ces données. Ceci a abouti à des avancées sans précédent dans divers domaines comme le traitement du langage naturel et la vision par ordinateur, principalement grâce à l'*apprentissage profond*, une classe spéciale de modèles d'apprentissage automatique. L'apprentissage profond surpasse apparemment les approches traditionnelles en apprenant les caractéristiques pertinentes directement à partir des données à travers une série de couches de calcul.

Cette thèse explore les fondements théoriques de l'apprentissage profond en étudiant la relation entre l'architecture de ces modèles et les structures inhérentes trouvées dans les données qu'ils traitent. En particulier, nous nous demandons : Quel est le facteur clé qui rend les algorithmes d'apprentissage profond efficaces et les aide à surmonter la difficulté d'apprendre des fonctions dans des espaces à haute dimension, connue sous le nom de *malédiction de la dimensionalité* ? Est-ce leur capacité à apprendre des représentations pertinentes des données en exploitant leur structure ? Comment différentes architectures exploitent-elles différentes structures de données ? Pour aborder ces questions, nous avançons l'idée que la structure des données peut être efficacement caractérisée par ses invariances—c'est-à-dire les aspects qui sont non pertinents pour la tâche en question.

Notre méthodologie adopte une approche empirique de l'apprentissage profond, combinant des études expérimentales avec des *modèles jouets* inspirés de la physique. Ces modèles simplifiés nous permettent d'investiguer et d'interpréter les comportements complexes que nous observons dans les systèmes d'apprentissage profond, offrant des perspectives sur leur fonctionnement interne, avec l'objectif à long terme de combler l'écart entre la théorie et la pratique.

Plus précisément, nous calculons les taux d'erreur de généralisation des réseaux a deux couches totalement connectés, démontrant qu'ils peuvent apprendre des invariances linéaires, c'est-à-dire devenir insensibles à des directions linéaires dans l'espace d'entrée qui sont non pertinents pour la tâche. Cependant, nous montrons que ces architectures peuvent mal fonctionner dans l'apprentissage d'invariances non linéaires comme l'invariance rotationnelle ou l'invariance par rapport aux petites déformations de l'entrée. Ce résultat illustre que si une architecture choisie n'est pas adaptée à une tâche donnée, elle pourrait être en overfitting, rendant une méthode de kernel, pour laquelle les représentations ne sont pas apprises, potentiellement un meilleur choix.

**Résumé**

Les architectures modernes telles que les réseaux de neurones convolutionnels, en revanche, sont particulièrement bien adaptées à l'apprentissage des invariances non linéaires présentes dans les données réelles. Dans la classification d'images, par exemple, la position exacte d'un objet ou d'une caractéristique peut ne pas être cruciale pour sa reconnaissance. Cette propriété donne lieu à une invariance par rapport aux petites déformations. Nos résultats montrent que les réseaux neuronaux les plus invariants aux déformations ont tendance à avoir des performances supérieures, soulignant l'importance d'exploiter une telle invariance. Une autre propriété clé qui donne de la structure aux données réelles est le fait que les caractéristiques de haut niveau sont une composition hiérarchique de caractéristiques de niveau inférieur—un chien est composé d'une tête et de membres, la tête est composée d'yeux, de nez et de bouche, qui sont ensuite composés de textures et de contours simples. Ces caractéristiques peuvent être réalisées de plusieurs manières synonymes, donnant lieu à une invariance. Pour étudier l'invariance synonymique qui découle de la structure hiérarchique des données, nous introduisons un modèle de données jouet qui nous permet d'examiner comment les caractéristiques sont extraites et combinées pour former des représentations de plus en plus élevées. Nous montrons que les réseaux de neurones profonds sont capable de apprendre avec succès ces structures hiérarchiques.

Finalement, nos travaux fournissent un aperçu des principaux mécanismes qui rendent les algorithmes d'apprentissage profond si efficaces. Par le biais d'études empiriques et de l'utilisation de modèles jouets, nous mettons en évidence le rôle de la structure des données dans les performances des algorithmes d'apprentissage profond, nous démontrons que des invariances spécifiques permettent d'obtenir des performances supérieures, et nous fournissons des orientations pour le choix de l'architecture de réseau la plus appropriée pour une tâche donnée.

**Mots-clés :** réseaux de neurones artificiels, apprentissage profond, malédiction de la dimensionalité, apprentissage de caractéristiques, apprentissage de représentations, structure de données.

# Sommario

L'intelligenza artificiale, e in particolare la branca dell'apprendimento automatico hanno subito una profonda trasformazione negli ultimi anni, convergendo verso modelli guidati dai dati, capaci di adattarsi e imparare da essi. Questa evoluzione ha generato progressi notevoli in diverse aree applicative, come l'elaborazione del linguaggio naturale e la visione artificiale. Tali avanzamenti sono in larga parte dovuti al deep learning (o *apprendimento profondo*), una categoria speciale di modelli di apprendimento.

Questa tesi esplora le basi teoriche del deep learning, con una particolare attenzione alla relazione tra l'architettura dei modelli e la struttura intrinseca dei dati su cui operano. In particolare, ci chiediamo: che ruolo svolge la struttura dei dati nel successo degli algoritmi di Deep Learning, specialmente nel superare la problematica nota come *maledizione della dimensionalità*? È nell'abilità di adattare le rappresentazioni interne ai dati il segreto del loro successo? Come diverse architetture sfruttano differenti tipologie di struttura nei dati? Per rispondere a queste domande, proponiamo di caratterizzare la struttura dei dati attraverso le loro invarianze, ovvero aspetti degli input che sono irrilevanti per il task in questione.

La nostra metodologia si basa su un approccio empirico al deep learning, integrando studi sperimentali con modelli teorici semplificati, ispirati alla fisica. Questi toy models (o *modelli giocattolo*) ci consentono di indagare ed interpretare i fenomeni complessi che si manifestano nei sistemi di deep learning. Con eventualmente l'obbiettivo di avvicinare teoria e pratica.

Nel dettaglio, quantifichiamo l'errore di generalizzazione in reti neurali non profonde, mostrando la loro capacità di imparare invarianze lineari. Al contrario, evidenziamo come queste architetture possano non essere ottimali nell'apprendere invarianze non lineari, come quelle per rotazioni o deformazioni dell'input. Ciò suggerisce che in certi scenari, metodi kernel, le cui rappresentazioni interne non si adattano alla struttura dei dati, potrebbero essere più efficaci.

Esaminiamo poi le architetture moderne, come le reti neurali convoluzionali, dimostrando la loro predisposizione a cogliere le invarianze non lineari presenti nei dati reali. Ad esempio, nella classificazione di immagini, una leggera deformazione dell'oggetto in questione non ne compromette il riconoscimento. Mostrando che reti più invarianti hanno prestazioni superiori, sottolineiamo l'importanza di questa proprietà.

Infine, discutiamo l'importanza della struttura gerarchica nei dati, dove le caratteristiche di alto livello emergono come combinazioni di quelle a livelli inferiori—un cane è composto da una testa e da arti, la testa è composta da occhi, naso e bocca, che sono poi composti da semplici texture e linee. Queste caratteristiche possono essere realizzate in più modi

## Sommario

equivalenti, o sinonimi, dando luogo a un'invarianza. Proponiamo un modello di dati che permette di studiare questa forma di invarianza e dimostriamo come solo le reti neurali profonde possano apprenderla, superando la maledizione della dimensionalità. In questo contesto, forniamo una stima quantitativa del numero di dati necessario per l'apprendimento di un dato compito, considerata la sua struttura gerarchica.

In conclusione, il nostro studio evidenzia come il deep learning sia capace di attenuare la maledizione della dimensionalità attraverso l'apprendimento di rappresentazioni invarianti. Le nostre analisi quantitative forniscono un'indicazione sul numero dati di addestramento necessari per il successo in un determinato task, in relazione alla sua struttura intrinseca.

**Parole Chiave:** reti neurali artificiali, apprendimento profondo, maledizione della dimensionalità, feature learning, apprendimento delle rappresentationi, struttura dei dati

# Contents

# List of Figures

# List of Tables

# Introduction

Machine learning, a cornerstone of artificial intelligence, is driven by algorithms that learn patterns in data to perform tasks without explicit instructions. This field has brought about significant advancements across various domains and given rise to specialized branches. Among these branches, deep learning is arguably the most successful. The goal of this thesis is to investigate the reasons behind this success.

To set the stage, we begin this introductory chapter with an overview of supervised learning fundamentals and the challenges inherent to it—particularly the bias-variance tradeoff and, more crucially for this thesis, the phenomenon known as the *curse of dimensionality*. It seems paradoxical that modern supervised learning algorithms perform well in high-dimensional tasks despite the curse of dimensionality typically impeding such learning. This suggests that high-dimensional data might be rich in structure in the form of invariances and symmetries. Hence, this thesis aims to unravel the nature of this structure and how deep learning algorithms, using suitable architectures, can exploit it.

To investigate this problem, we will review modern supervised learning algorithms, namely kernel methods and neural networks—readers familiar with these topics may wish to skip this discussion, as well as the preliminary one on supervised learning. One significant aspect we will emphasize is the ability of neural networks to adapt to the features of the data and possibly learn relevant representations, a capability absent in kernel methods. This leads us to the following key questions: *(i)* What features do neural networks learn, are they indeed relevant for the task at hand?, *(ii)* Does this feature learning contribute to some kind of dimensionality reduction, and therefore could it be a factor in overcoming the curse? and *(iii)* If so, can we quantify the impact of feature learning on performance in terms of how many data points are needed for learning a given task?

To address these questions, we first review empirical works, specifically *(i)* techniques that allow us to visualize the representations learned by deep networks and *(ii)* observables to characterize the learning of task relevant representations in deep network.

For answering the third question, we will go into the theoretical study of the infinite-width limit of neural networks, which allows establishing a strong link between neural networks and kernel methods. In fact, the same neural network architecture can be trained in two different regimes named *feature* and *lazy*, depending on the scale of parameters at initialization. The feature regime corresponds to standard neural network training where features can be learned, while the lazy regime to kernel methods, where features are dictated by the architecture and remain static during training. The joint study of these two training regimes allows us to dissect the impact of feature adaptation versus the influence of architecture choice, and thus to study whether feature adaptation truly underpins the success of deep learning.

Having formulated a framework for gauging the benefits of feature learning, the next step is to identify which features of the data are indeed learned by neural networks. This leads us to a more in-depth discussion on data structure, which we will examine in terms of data invariances. We will discuss three types of invariances. The first, *linear invariance*, is associated with the observation that some input coordinates may not be relevant for a given task. The second, *deformation invariance*, reflects the notion that the precise location of relevant features within the input might not affect the task. Finally, we will delve into *synonymic invariance*. This invariance builds on the notion that some tasks can be viewed as a hierarchical composition of features at varying levels, with the property that substituting these features with their synonyms does not alter the label. We conclude this introduction by providing an overview of the thesis structure and main results, pointing out chapters that delve into the topics introduced here. We finish with a note on our methodology.

## 1.1 Machine Learning

### 1.1.1 Supervised Learning

Supervised learning is a machine learning paradigm where a model learns to make predictions based on a set of labeled examples. More formally, given a set of $P$ pairs of data $(x_i, y_i)$, where $x_i \in X$ represents an input and $y_i \in Y$ represents the corresponding label or target, the objective of supervised learning is to approximate the true, yet unknown, function $f^* : X \rightarrow Y$ that maps the inputs to the outputs.

**Regression and Classification Tasks**  Supervised learning can be broadly divided into two categories: regression and classification. In regression, $Y = \mathbb{R}$, and the task is to predict a continuous target variable. In contrast, in classification, $Y = \{1, 2, ..., C\}$ represents a finite set of $C$ discrete classes, and the goal is to assign each input to one of these classes.

To give a concrete example of a regression setting, consider a scenario where we try to predict a person's height ($y$) based on their age ($x$). While there exists a general trend or a true function $f^*$ (children grow as they age), the exact height of a person at a certain age is often influenced by noise factors such as genetics, diet, and lifestyle. We could model this situation as follows:

$$y_i = f^*(x_i) + \epsilon_i, \tag{1.1}$$

where $f^*$ represents the true underlying function, and $\epsilon_i \sim \mathcal{N}(0, \sigma^2)$ is Gaussian noise with zero mean and $\sigma^2$ variance. This noise reflects the uncertainty or the individual differences that cannot be captured by age alone.

A classification task can be modeled similarly, except that the labels are categorical, e.g. $Y = \{0, 1\}$ and they are given by

$$y_i = \text{sign}(f^*(x_i) + \varepsilon_i) \tag{1.2}$$

where $\text{sign}(z)$ is a function that returns 1 if $z \geq 0$, and 0 otherwise.

**Building a Model, Training, and Testing**   In supervised learning, the objective is to find an accurate approximation to the true function $f^*$ typically through a parameterized model, denoted as $f_\theta$. The parameters $\theta$ could be coefficients in linear regression, weights in a neural network, etc., depending on the specific model used. These parameters are tuned using a set of *training* examples to produce an optimal function $f_\theta$ that can accurately predict the labels of unseen, *test* examples. This ability of a model to effectively predict unseen data is known as *generalization* and, for classification tasks, it is measured as the percentage of errors the model makes on the test data set, the *test error*.

**Test Error vs. Number of Training Points**   In this thesis, to evaluate a model's performance, we will typically examine its test error as a function of the number of training points, expressed as $\epsilon(P)$. The characterization of this function is practically valuable as it can provide insights into the necessary sample size to meet a specified accuracy goal for a given task. In many real-world settings Hestness et al. (2017); Spigler et al. (2020); Kaplan et al. (2020), this function exhibits a power-law behavior, that can be characterized by a scaling exponent $\beta$ with $\epsilon(P) \sim P^{-\beta}$, as illustrated in Figure 1.1(a). Alternatively, one can characterize the test error by a *sample complexity*, defined as the number of training points $P^*$ needed to achieve a given finite test error, e.g. $\epsilon^* = 1\%$. The use of sample complexity as a characterization is especially useful when the test error shows a transition in $P$ from random guessing to nearly zero error (Figure 1.1(b)), as it will be the case in the artificial task introduced in chapter 6, for example. In such cases, $P^*$ pinpoints the location of this transition as $P^*(\epsilon^*)$ is characterized by a mild dependence on $\epsilon^*$.

**Optimization**   The process of training involves adjusting the parameters $\theta$ in order to minimize the difference between the predictions of $f_\theta$ and the true output values $y$ on the training set. This is usually achieved through an optimization procedure that finds the $\theta$'s that minimize a training loss

$$\mathcal{L} = \frac{1}{P} \sum_i l(f_\theta(x_i), y_i),$$

where the loss function $l$ quantifies the dissimilarity between predictions $f_\theta(x_i)$ and true outputs $y_i$.

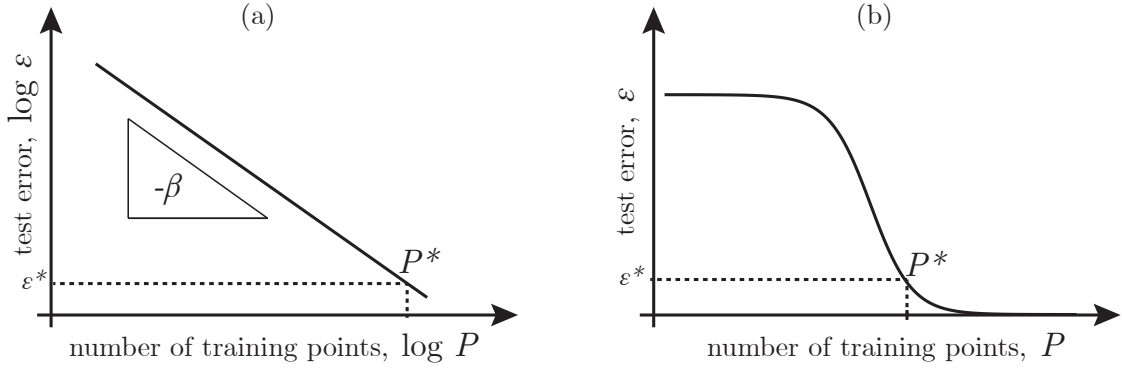

Figure 1.1: **Test error vs. number of training points: different scenarios and characterizations.** (a) When the curve follows a power law, $\epsilon(P) \sim P^{-\beta}$, it can be represented as a straight line in a log-log plot, where the exponent $-\beta$ defines the slope of the line. (b) Alternative scenario where $\epsilon(P)$ transitions from a large error, corresponding to random guessing, to nearly zero error. Both scenarios can be described using the sample complexity, $P^*$, which quantifies the number of training points required to achieve a specific finite error, $\epsilon^*$.

At the core of this training process is the definition of a suitable loss function. For instance, in regression tasks, the *mean square error (MSE)* is a popular choice. Given the predicted output $\hat{y}$ and the true output $y$, the MSE loss function is defined as:

$$l_{\text{MSE}} = (\hat{y} - y)^2.$$

For binary classification tasks, the *hinge loss* is frequently used

$$l_{\text{hinge}} = \max(0, 1 - y\hat{y}),$$

where $y = \pm 1$ and $\hat{y} \in \mathbb{R}$.

In the case of multi-class classification, a common choice of loss function is the cross-entropy loss. For a predicted vector of outputs $\hat{y}$ and a label $y$, the *cross-entropy loss* is defined as:

$$l_{\text{CE}} = -\sum_{c=1}^{C} y_c \log(\hat{y}_c),$$

where $C$ is the number of classes, $\boldsymbol{y}$ are the true labels encoded as a one-hot vector[2] and $\hat{y}$ represents the network output, normalized through a Softmax operation,

$$\sigma_{\text{softmax}}(f_\theta(x)) = \frac{e^{(f_\theta(x))_c}}{\sum_{c'=1}^{C} e^{(f_\theta(x))_{c'}}},$$

---

[2]One-hot encoding of a class refers to a representation method in which the class of interest is assigned a value of one, while all other class categories are assigned a value of zero, thus creating a binary vector that uniquely identifies the particular class.

with $(f_\theta(x))_c$ denoting the $c$-th component of the output vector.

The optimization process that reduces the loss, and thus improves the model's fit, commonly employs *gradient descent*, which iteratively adjusts the model parameters in the direction that minimizes the loss:

$$\theta_{t+1} = \theta_t - \eta \nabla \mathcal{L}(\theta_t),$$

where $\eta$ is the learning rate controlling the size of optimization steps, $t$ represents the step number, and $\nabla \mathcal{L}(\theta_t)$ is the gradient of the loss function with respect to the parameters, at the current parameter values.

In this manner, we adjust the function $f_\theta$ to align as closely as possible with the true function $f^*$ using the available training data. However, we stress again that the aim is not to perfectly fit the training data, but to generalize well to new, unseen data. This is where the challenges of supervised learning start to emerge.

**Bias-Variance Tradeoff** The bias-variance tradeoff Luxburg and Schölkopf (2011) is a key concept that helps us understand the generalization error of a predictive model. The generalization error can be decomposed into bias, variance, and an irreducible error term due to noise in the labels.

The *bias* of a model reflects the error introduced by approximating the true function (which may be highly complex) by a simpler model. High-bias models oversimplify the problem, leading to underfitting. The *variance*, on the other hand, quantifies the sensitivity of our model to fluctuations in the training data. Models with high variance are likely to be over-complex, and therefore susceptible to *overfitting*. As an example, let's consider the task of fitting a polynomial function

$$f_\theta(x) = \sum_{i=0}^{N} \theta_i x^i \tag{1.3}$$

to some data as in Figure 1.2. If we try to fit this data using a high-degree polynomial (a complex model with many parameters), we may obtain a model that passes exactly through every point of the training set but that fluctuates wildly in between (high variance, overfitting). However, if we fit a simple linear model (a low-degree polynomial) to the same data, we may find that our model does not capture the oscillating nature of the true underlying function at all (high bias, underfitting). In this scenario, a polynomial of intermediate degree may offer the best bias-variance tradeoff, capturing the broad trends in the data without being overly sensitive to noise. We see here that the number of parameters of the model affects its complexity.

**Regularization** Regularization techniques offer a pragmatic approach to managing this bias-variance tradeoff by controlling the complexity of the model. Adding a regularization term to the loss function, for example, places a constraint on the size or sparsity of the model

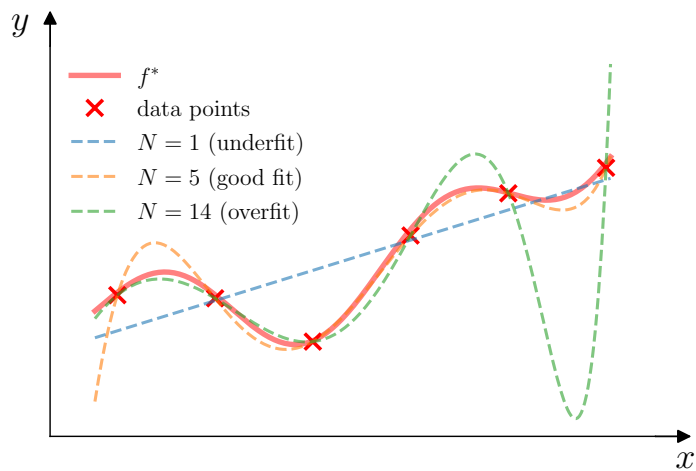

Figure 1.2: **Illustration of the bias-variance tradeoff in polynomial regression.** Red crosses represent data points $(x_i, y_i)$, with $y_i = f^*(x_i) + \varepsilon_i$. The red curve represents $f^*$, the true function. Three different polynomials of degrees $N = 1, 5, 14$ are considered. The linear polynomial (underfitting) exhibits high bias and low variance, failing to capture the complexity of the true function. The $N = 5$ polynomial achieves a good balance between bias and variance, fitting the true function. The $N = 14$ polynomial (overfitting) shows low bias but high variance, modeling the noise in the data and deviating significantly from the true function.

parameters. Regularization may allow us to find a balance between bias and variance.

Common regularization methods include L1 and L2 regularization, also known as Lasso Tibshirani (1996) and Ridge regularization (or weight decay), respectively. These techniques introduce a term to the loss function that penalizes the size of the model, with L1 regularization also favoring sparse solutions in which many of the parameters are zeroed. The regularization term is either the L1 or L2 norm of the parameters:

$$\lambda \sum_j |\theta_j|, \quad \text{or} \quad \lambda \sum_j \theta_j^2,$$

where $\lambda$ controls the regularization strength.

Depending on the model at hand, other regularization techniques may be used—e.g. dropout Srivastava et al. (2014), early stopping, etc.

Interestingly, the bias-variance tradeoff does not pose the same challenge for deep learning algorithms. As we will explore in greater detail in subsection 1.2.2, deep learning models are intrinsically biased towards finding simple, small-norm solutions, even when their number of parameters diverges.

### 1.1.2 The Curse of Dimensionality

While the bias-variance tradeoff manifests itself even in one-dimensional learning scenarios, a distinct and fundamental challenge arises when we consider high-dimensional data: the *curse of dimensionality* Bach (2017); Wainwright (2019). This term refers to the various difficulties and counterintuitive phenomena that arise when dealing with high-dimensional data.

For instance, consider a dataset containing measurements of $d$-dimensional vectors. Suppose we want to cover a fraction of the $d$-dimensional space with a grid of small hypercubes of side length $\epsilon$, one for each data point. The number of data points required to cover the space scales as $(1/\epsilon)^d$, which grows exponentially as the number of dimensions $d$ increases. This implies that even for a moderate value of $d$, we need a huge number of data points to populate the space.

The issue is that as the number of dimensions grows, the space itself expands so fast that our data points start to seem sparse, or spread out. This sparsity can be problematic for any method trying to estimate a function in this space. Any new data point we use to test our method will likely be quite far from any other sample. To properly cover the space as the dimensionality increases we would need an exponentially growing amount of data. More concretely, for regression tasks where the target is only assumed to be a Lipschitz continuous function, the test error decays only at a rate of $-1/d$ with respect to the number of training points Luxburg and Bousquet (2004), making learning impossible when $d \gg 1$.

**The Surprising Effectiveness of Deep Learning**   Strikingly, modern supervised learning algorithms—more specifically, deep learning—are able to beat the curse of dimensionality and learn tasks even in very high dimensions. Examples of deep learning success range from computer vision Voulodimos et al. (2018) and natural language processing OpenAI (2023) to computational biology Jumper et al. (2021) and game playing Silver et al. (2017). Despite these practical successes, a comprehensive theoretical understanding of how these algorithms overcome the curse of dimensionality remains elusive.

**The Intrinsic Dimensionality of Real Data**   One possible explanation for this puzzle could lie in the fact that although our machine learning tasks typically involve high-dimensional datasets, the actual data often exist in a lower-dimensional subspace.

The term *intrinsic dimensionality* (ID) is employed to describe the dimensionality of this subspace and can be thought of as the minimum number of variables needed to characterize it. For instance, consider a dataset of points lying on a line within a three-dimensional space. Although they exist in a 3D space, their intrinsic dimensionality is just one, as only one coordinate along the line accurately describes all points.

The intrinsic dimensionality of a dataset can be estimated by randomly drawing $P$ data points, and analyzing the typical distance $\delta$ between nearest neighbors, typically scaling as $\delta \sim P^{-1/d_{\text{ID}}}$.

**Introduction**

If these data points are embedded in a $d$-dimensional space but exist on a $d_{\text{ID}}$-dimensional manifold, the ratio $\log \delta / \log P$ provides an estimate of the intrinsic dimensionality. Advanced intrinsic dimension estimation methods further use information regarding the neighborhood of datapoints, other than the nearest neighbor distance alone Levina and Bickel (2004); Granata and Carnevale (2016); Facco et al. (2017). However, it's vital to approach these estimations with caution. Although they can offer insights into the intrinsic dimensionality of a dataset, they hinge on the assumption that real data reside on a smooth, continuous manifold of fixed intrinsic dimensionality. This assumption lacks empirical support, as the estimates regarding real-world datasets can only be based on discrete sets of data points. We will further discuss intrinsic dimension estimation in subsection 1.2.1, in the context of neural networks' internal representations.

With this understanding, we can discuss the intrinsic dimensionality of real datasets, and the implication on the curse of dimensionality. For a benchmark image classification dataset as ImageNet Deng et al. (2009), intrinsic dimensionality is estimated to be around fifty Pope et al. (2021), while it contains about $10^7$ data points. This number is orders of magnitude less than the expected $e^{50} \sim 10^{20}$ data points needed to adequately sample such a high-dimensional space and beat the curse. This fact suggests that lower intrinsic dimensionality alone isn't the key to deep learning's success, implying additional structural elements within the data.

**Invariances give Structure to Real Data**  A popular idea is that this additional structure is due to the presence of *invariances*, namely transformations of the input that leave the label unchanged Goodfellow et al. (2009); Bengio et al. (2013); Bruna and Mallat (2013); Mallat (2016). By developing representations that are *invariant* to these transformations the dimensionality of the problem could be effectively lowered and the curse beaten. This thesis centers around this concept of *learning invariances*—a phrase we use to describe the process of developing representations that are invariant to aspects of the data irrelevant to the task at hand. More specifically, our aim is to address the following set of questions:

- Which invariances are present in real data?

- Are deep learning algorithms able to learn these data invariances?

- If so, how many training examples do they need to achieve that?

- Is learning invariances indeed responsible for breaking the curse of dimensionality?

Before attacking these questions we need to introduce our main objects of study. We will start with kernel methods that we will use as a proxy for investigating the performance of algorithms in which features are fixed and hence learning invariances is not possible, and we will then introduce neural networks, in which feature adaptation allows the learning of data invariances.

### 1.1.3 Kernel methods

Kernel methods Scholkopf and Smola (2001) comprise a family of machine learning techniques that utilize a fixed feature representation, and determine the optimal weights for these features. The use of kernel functions facilitates the implicit mapping of data into a potentially higher-dimensional feature space, sometimes even infinite-dimensional, without having to compute the coordinates of the data in that space explicitly.

To illustrate this, let's consider the task of approximating a non-linear function with linear regression. A common approach to this problem involves creating a feature vector $\boldsymbol{\phi}(x)$ that projects each data point $x$ into a high-dimensional feature space, and fitting a linear model to these feature vectors:

$$f_{\boldsymbol{\theta}}(x) = \boldsymbol{\theta}^{\top}\boldsymbol{\phi}(x). \tag{1.4}$$

However, the direct computation of these feature vectors can be computationally challenging when the feature space is large or infinite-dimensional.

This is where the *kernel trick* comes in. This key technique in kernel methods bypasses the explicit computation of these high-dimensional feature vectors. It introduces a kernel function $k : X \times X \to \mathbb{R}$, which calculates the inner product between feature vectors in the high-dimensional feature space, $k(x, x') = \langle \phi(x), \phi(x') \rangle$. The *representer theorem* Schölkopf et al. (2001) tells us that, when solving an optimization problem with L2 regularization of the form

$$\min_{\boldsymbol{\theta}} \quad \frac{1}{P}\sum_i l(f_{\boldsymbol{\theta}}(x_i), y_i) + \lambda \|\boldsymbol{\theta}\|^2,$$

the predictor $f_{\boldsymbol{\theta}}(x)$ of Equation 1.4 can equivalently be written as:

$$f_{\boldsymbol{\alpha}}(x) = \sum_{i=1}^{P} \alpha_i k(x_i, x),$$

where $x_i$ are the training inputs, and $\alpha_i$ are the new (dual) parameters that can be learned from data. This formulation allows performing linear regression in infinite feature spaces and hence the application of linear techniques to non-linear problems, making kernel methods a powerful tool in machine learning.

**Smoothness and Curse of Dimensionality**    The understanding of the generalization capabilities of kernel methods has been a significant subject of investigation. This understanding involves examining the smoothness of the target function, that is related to its differentiability. The most favorable situation occurs when the target function resides in the *Reproducing Kernel Hilbert Space* (RKHS) of the chosen kernel Scholkopf and Smola (2001). In such a case, the error decays as $O(1/\sqrt{P})$ Smola et al. (1998); Rudi and Rosasco (2017). However, in the context of isotropic kernels of the form $k(x, x') = k(|x - x'|)$, that are commonly used in practice, for the target function to belong to the RKHS in dimension $d$, it must be $s$-times differentiable, with $s$ proportional to $d$. In high-dimensional settings, requiring the smoothness to grow with $d$ is unreasonable, and this can be seen as a manifestation of the curse of dimensionality. If the target falls outside of the RKHS, the error rate degrades to $O(P^{-s/d})$ Bach (2022), recovering the scaling of $O(P^{-1/d})$ when dealing with Lipschitz continuous functions ($s = 1$) Luxburg and Bousquet (2004).

**Exact Generalization Error Rates**   The exact asymptotics of the test error in between the RKHS and Lipschitz continuity extrema have been computed in Spigler et al. (2020); Bordelon et al. (2020); Canatar et al. (2021) in the case of isotropic kernel and noiseless target functions— i.e. function of the form Equation 1.1, without the noise term. More specifically, these works show the presence of a *spectral bias* in kernel regression: the projections of the target function on the eigenfunctions of the kernel with the largest eigenvalues are learned first. For isotropic kernels in dimension $d$, if the inputs $\boldsymbol{x}$ are distributed uniformly on the $d$-sphere, these eigenfunctions correspond to Fourier modes in $d = 1$, and to spherical harmonics in larger dimensions. In particular, $P$ training points allow for learning the first $P$ modes. Extensions of these results to the noisy case are discussed in Loureiro et al. (2022); Cui et al. (2022); Mei et al. (2022). Notice that these results rely on statistical assumptions on the predictor function and kernel features that are rarely satisfied in practice, and can cause the spectral bias prediction to fail in low-dimensional settings Tomasini et al. (2022b), while it appears to be accurate in high-dimensions.

In chapter 4, we will use the presence of a spectral bias to compute tight generalization error rates both for kernel methods and neural networks.

### 1.1.4   Neural Networks

In kernel methods, the selection of the kernel predetermines the feature vectors. We now transition to exploring neural networks, which offer the flexibility to learn features directly from the data.

Neural networks are at the heart of modern machine learning. They are built from interconnected nodes or *neurons* organized in layers and can learn to perform complex tasks. In this section, we introduce some fundamental types of neural networks.

**Perceptron**   The perceptron Rosenblatt (1958) is one of the simplest forms of a neural network and is the basic building block for more complex structures. A single perceptron takes a vector of inputs $\boldsymbol{x}$, applies a set of weights $\boldsymbol{w}$, adds a bias term $b$, and passes the result through a non-linear activation function $\sigma(\cdot)$ to produce an output. Mathematically, this can be written as

$$f(\boldsymbol{x}) = \sigma\left(\frac{\boldsymbol{w}^\top \boldsymbol{x}}{\sqrt{d}} + b\right),$$

where parameters are initialized with a standard Gaussian distribution $\mathcal{N}(0,1)$ and $d$ is the input dimension. The $\sqrt{d}$ factor serves as a normalization to keep the argument of the activation function $O(1)$ in the limit of large $d$. Classically, activation functions such as the step or sigmoid $\sigma(z) = (1 - e^{-z})^{-1}$ were employed as they reflect the active-inactive states of biological neurons. However, in contemporary practice, the Rectified Linear Unit (ReLU),

$$\sigma_{\text{ReLU}}(z) = \max(0, z),$$

is most commonly utilized.

**Two-layers Neural Networks**  A 2-layers (or one-hidden-layer) neural network extends the perceptron by introducing an additional layer of neurons. Each neuron in the hidden layer performs a similar operation to the perceptron, and their outputs are then linearly combined to produce the final output. Formally, the operation of a 2-layers neural network can be described by:

$$f(\boldsymbol{x}) = \frac{1}{\sqrt{h}} \boldsymbol{w}_2^\top \sigma\left(\frac{\boldsymbol{w}_1^\top \boldsymbol{x}}{\sqrt{d}} + \boldsymbol{b}_1\right), \tag{1.5}$$

where $\boldsymbol{w}_2$, $\boldsymbol{w}_1$, and $\boldsymbol{b}_1$, are the weights and biases of the network, $\sigma(\cdot)$ is the activation function and $h$ is the number of hidden neurons or network width. See also a graphical representation in Figure 1.3. This setup allows the model to learn more complex representations with respect to the perception. For example, a 2-layers can solve classification tasks that are not linearly separable. More generally, the *Universal Approximation Theorem* states that 2-layer neural networks can approximate any function to arbitrary precision, provided they have enough neurons Hornik et al. (1989); Barron (1993).

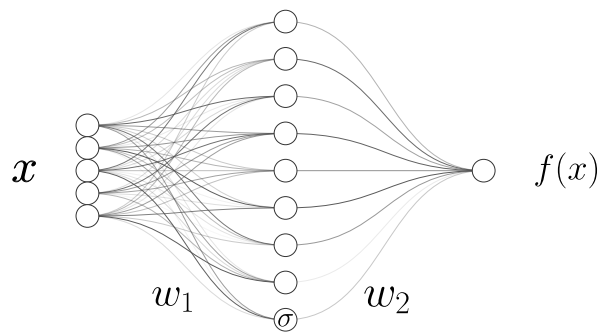

Figure 1.3: **Graphical representation of a 2-layer neural network** with an input dimension $d = 5$ and $h = 9$ hidden neurons. The network function, denoted by $f$, takes as input the vector $\boldsymbol{x}$. The activation function is denoted by $\sigma(\cdot)$ and applied to each hidden neuron independently. The weights for the two layers are denoted by $\boldsymbol{w}_2$, $\boldsymbol{w}_1$, respectively. The different strength of the lines depicts weights of different magnitude.

**Random Features**   Notice that usually all the network parameters of a 2-layers architecture are trained. A model in which only the second layer weights $\boldsymbol{w}_2$ are trained while the others are kept fixed is called *random features model* Rahimi and Recht (2007). This is a kernel method where the features are given by the first layer activations: $\phi(\boldsymbol{x}) = \sigma\left(\frac{\boldsymbol{w}_1^\top \boldsymbol{x}}{\sqrt{d}} + \boldsymbol{b}_1\right)$.

**Deep Fully Connected Networks (FCNs)**   Deep fully connected networks (also known as multilayer perceptrons) consist of multiple layers of neurons, with each layer fully connected to the next. A $L$-hidden-layers FCN can be described recursively as

$$\boldsymbol{a}_l = \sigma\left(\frac{\boldsymbol{w}_l^\top \boldsymbol{a}_{l-1}}{\sqrt{h_l}} + \boldsymbol{b}_l\right) \quad \text{for} \quad l = 1, \ldots, L, \tag{1.6}$$

where $l$ is the layer index, $\boldsymbol{w}_l$ and $\boldsymbol{b}_l$ are the weights and biases for the $l$-th layer, $\boldsymbol{a}_{l-1}$ is the output of the previous layer with $\boldsymbol{a}_0$ being the input $\boldsymbol{x}$, and $h_l$ the number of neurons at layer $l$, with $h_0$ being the input dimension $d$. The output of the network is then a linear combination of the last hidden layer activations,

$$f(\boldsymbol{x}) = \frac{1}{\sqrt{h_{L+1}}} \boldsymbol{w}_{L+1}^\top \boldsymbol{a}^L. \tag{1.7}$$

*Deep learning* generically refers to solving machine learning tasks with *deep* neural networks.

**Convolutional Neural Networks (CNNs)**   Convolutional Neural Networks are a type of neural network designed to process data in which locality is important in the sense that the relevance of each input coordinate or *pixel* is closely tied to its neighbors Lecun et al. (1998).

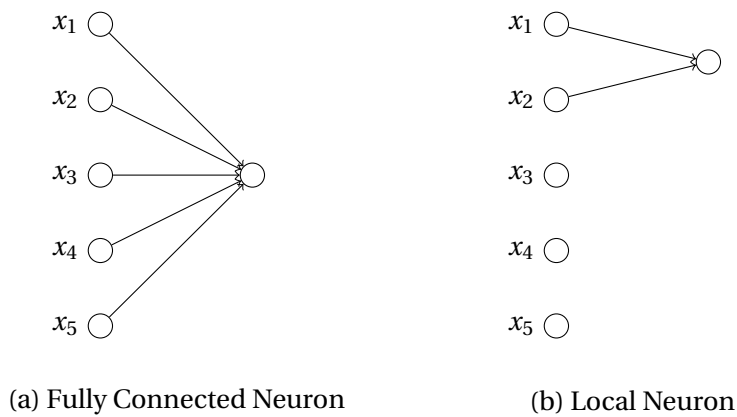

(a) Fully Connected Neuron

(b) Local Neuron

Figure 1.4: Comparison between (a) a Fully Connected Neuron that receives input from every element of the preceding layer, and (b) a Local Neuron, where each neuron processes data only from a specific part of the input (in this case, the patch composed by first two pixels $x_1$ and $x_2$).

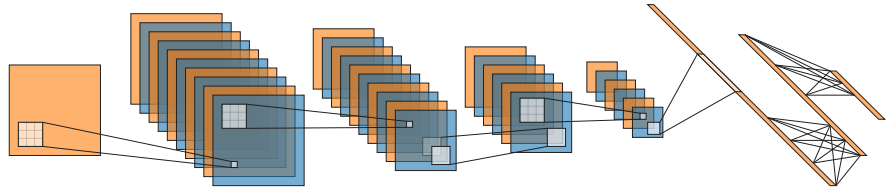

Figure 1.5: **Graphical representation of a convolutional neural network** (CNN) with 4 convolutional layers with many channels each and 2 fully-connected layers. The CNN acts on 2D images with filters of size 3 × 3 at each convolutional layer.

Unlike FCNs where neurons respond to all input locations, CNNs have local neurons that focus on specific patches of the input, see illustration in Figure 1.4. In addition, standard convolutional layers also use weight sharing, where the same weights are used by neurons examining different parts of the input. Weight sharing not only reduces the model's number of parameters but also introduces translation equivariance into the network architecture. Translation equivariance means that if the input is translated, the output will change in the same way. If $f$ is the activation of a particular layer of a CNN, and $T$ is a translation operator, then the layer is translation equivariant if:

$$f(T(\boldsymbol{x})) = T(f(\boldsymbol{x})) \tag{1.8}$$

where $\boldsymbol{x}$ is the input data. This property ensures that if a pattern is learned in one part of the image, it will be recognized in any other part. Translation invariance can be obtained by pooling together equivariant representations, i.e. by summing or averaging equivariant activations over all possible translations:

$$\sum_{\tau} f(T_{\tau}(x)). \tag{1.9}$$

Here, $\tau$ represents a translation, and $T_{\tau}$ is the corresponding translation operator. By summing over all translations, the response becomes independent of the specific location of the recognized feature in the input, leading to translation invariance.

Early implementations of CNNs often relied on pooling layers, such as max-pooling or average-pooling, to achieve *approximate* translation invariance by aggregating the responses of neurons within a *local* region. However, the use of pooling layers has diminished in modern practices and we will demonstrate that contemporary CNNs possess the capability to learn pooling directly from data, as also noted in Ruderman et al. (2018).

Convolutional layers have proven particularly effective in solving computer vision tasks, where locality is clearly important as neighboring pixels make sense together, and translation invariance as well, as the same object or feature can appear in different parts of the image frame. In these cases, 2D local filters are used, leading to CNNs as the one illustrated in Figure 1.5. In particular, with the publication of the *AlexNet* paper by Krizhevsky et al. (2012), CNNs started the modern deep learning revolution. AlexNet won the *ImageNet Large Scale*

*Visual Recognition Challenge* Deng et al. (2009), a prestigious image recognition competition, by a significant margin, outperforming the second-best entry by over 10%. This breakthrough demonstrated the potential of deep CNNs to handle high-dimensional data, and it set the stage for the widespread adoption of deep learning techniques in a variety of domains.

## 1.2   Feature Learning in Neural Networks

Our discussion so far has reviewed kernel methods and neural networks, touching upon the successes of the latter. While kernel methods rely on fixed feature representations, neural networks are characterized by their ability to adapt features to the data at hand. This adaptability is often seen as a fundamental component of their success.

In this section, we discuss the adaptability of neural networks more in depth. First, we focus on empirical studies to understand the nature of the features that these networks learn, and if these are relevant for the task at hand. Following that, we survey empirical tools in existing literature that elucidate how this feature learning process shapes representations in such a way of discarding irrelevant information. We then discuss how this representation learning can lead to dimensionality reduction, and to eventually beat the curse. To determine whether representation learning is indeed a key factor in overcoming the curse of dimensionality, and thereby contributing to the superior performance of neural networks, we require a framework that quantifies the impact of feature learning on performance.

As hinted at in the previous section, modern neural networks' successes also came with important architectural advancements. This raises the question: is it the adaptability that is primarily driving the success of deep learning, or is it more about the selection of the right architecture, or perhaps a combination of both? To answer these questions, we will explore the deeper ties that link neural networks and kernel methods beyond random feature models. We will focus on the two distinct training regimes that emerge when we consider the infinite-width limit of neural network architectures based on the scale of initialization: the Neural Tangent Kernel (NTK) or lazy regime, akin to kernel methods, and the mean-field or feature learning regime, mirroring the training of neural networks with feature adaptation. This discussion will provide a framework to disentangle the respective roles of architectural choice——which dictates the features for kernel methods——and adaptability in driving neural networks performance.

### 1.2.1   Empirical Studies of Feature Learning

**Visualizing the Learned Features**   In this paragraph, we discuss the questions: What kind of features of the data do neural networks learn thanks to their adaptability? Are they relevant for the task at hand? Empirical research offers some fascinating insights into this matter. In particular, it has been found that deep networks, especially CNNs, tend to learn relevant features in a hierarchical manner Zeiler and Fergus (2014); Yosinski et al. (2015); Olah et al.

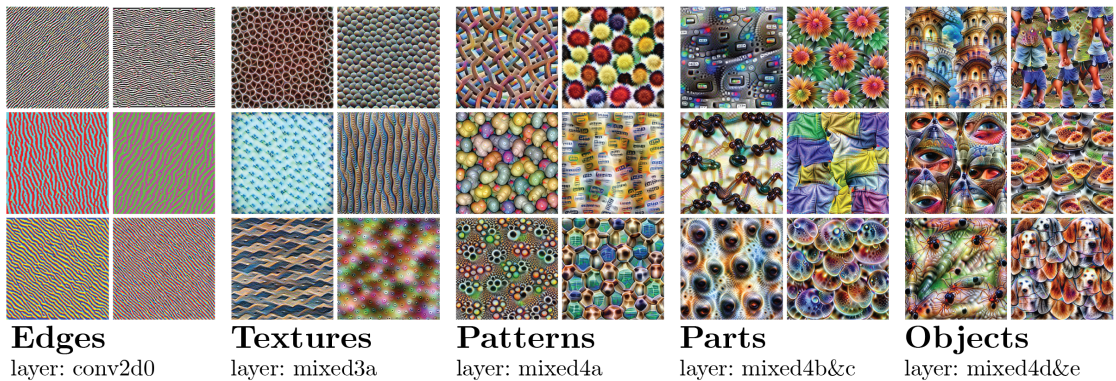

| **Edges** | **Textures** | **Patterns** | **Parts** | **Objects** |
| layer: conv2d0 | layer: mixed3a | layer: mixed4a | layer: mixed4b&c | layer: mixed4d&e |

Figure 1.6: **Feature visualization in deep networks.** This figure, adapted from Olah et al. (2017), uses feature visualization to showcase how GoogLeNet Szegedy et al. (2015), trained on the ImageNet dataset Deng et al. (2009), builds its understanding of images progressively with depth. These images are generated via gradient descent in input space such as to maximize the neurons' response at each layer. We can observe the images inducing maximal activation at increasingly deeper layers (left to right), growing in complexity and abstraction with depth.

(2017). Early layers tend to focus on simple, local features, such as edges in an image. As we move further into the depths of the network, these simple features are progressively combined to create more complex and abstract representations, including shapes or even entire objects, see illustration in Figure 1.6.

Several works tried to characterize this process of building more and more abstract representations empirically. In this context, we review here the information bottleneck framework of deep learning, and measurements of the intrinsic dimensionality of internal representations.

**The Information Bottleneck**  The information bottleneck provides a framework to analyze how to compress a random variable $X$ into an intermediate representation $T$, while keeping essential information about another related variable $Y$, given the joint probability distribution $P(X; Y)$ Tishby et al. (2000). This leads to the following optimization problem:

$$\min_{P(T|X)} I(T; X) - \beta I(T; Y) \tag{1.10}$$

where $I(T; X)$ is the mutual information[3] between the representation and the original data, $I(T; Y)$ is the mutual information between the representation and the target, and $\beta$ is a parameter that controls the trade-off between compression and preservation of relevant information.

In the context of deep learning, this framework has been used to explore the functioning of neural networks. Shwartz-Ziv and Tishby (2017) argue that the information bottleneck framework can be used to understand how deep networks progressively compress input data across successive layers, while maintaining task-relevant information, in order to build increasingly abstract internal representations. The authors suggest that deep neural networks internal representations (denoted by $T$ in this context) tend to solve a problem of the form in Equation 1.10 when trained with stochastic gradient descent. In particular, they argue that the training process consists in two distinct phases—a *fitting* phase where it captures the relevant information about the output and both $I(T; X)$ and $I(T; Y)$ increase, and a *compression* phase where it sheds irrelevant input details, characterized by a decrease of $I(T; X)$.

---

[3]Mutual Information is a measure that quantifies the amount of information gained about one random variable by observing another. It essentially describes how much knowing one variable reduces uncertainty about the other. If the variables are independent, the mutual information is zero. If they are identical, the mutual information is the same as the individual entropy of either variable.

However, Saxe et al. (2019) presents a critical examination of these claims. It posits that the conclusions drawn by Shwartz-Ziv and Tishby (2017) heavily depend on the activation function used in their experiments, making their findings not universally applicable. In particular, the compression phase could not be observed if instead of the sigmoidal activation function, the more common ReLU is deployed. Additionally, the critique highlights that the concept of mutual information, a cornerstone of the information bottleneck framework, isn't well-defined in deterministic settings. And the output or internal activations of a neural network is deterministically defined for a given input.

**The Intrinsic Dimension of Internal Representations**   Another way to characterize the process of learning features with depth is to idealize the data representation at each network layer as existing on a manifold in the activations space of that layer. By assessing the *intrinsic dimensionality* of these manifolds (cf. definition in subsection 1.1.2), we can gain insights into how the network reduces the dimensionality of the problem at each layer. Ansuini et al. (2019); Recanatesi et al. (2019) show that the intrinsic dimensionality of internal representations typically increases with training, in the first part of the network, while decreasing in the second part. Consequently, after training, the intrinsic dimensionality displays a concave or *hunchback* shape, as a function of depth. The reduction of dimensionality with training and depth in the second part of the networks fits with the idea that learning progressively filters out irrelevant details to construct increasingly abstract representations. The initial expansion, however, is harder to parse. The authors interpret it as the neural networks generating a broad set of features at early stages, and later keeping only those essential for the task.

Although this approach offers insightful observations, it comes with inherent difficulties. One key issue is that it assumes that real data exist on a smooth manifold, while in practice, the measurements are based on a discrete set of points. This leads to counter-intuitive results such as the increase in the intrinsic dimensionality with depth near the input, an effect that is impossible for continuous smooth manifolds. We resort to an example to illustrate how this increase with depth can result from spurious effects. Consider a manifold of a given intrinsic dimension that undergoes a transformation where one of the coordinates is multiplied by a large factor. This transformation would result in an elongated manifold that appears one-dimensional. The measured intrinsic dimensionality would consequently be one, despite the higher dimensionality of the manifold. In the context of neural networks, a network that operates on such an elongated manifold could effectively *reduce* this extra, spurious dimension. This operation could result in an increase in the observed intrinsic dimensionality as a function of network depth, even though the actual dimensionality of the manifold did not change. This phenomenon appears as a plausible explanation for what is observed in practice. Also, if this interpretation holds, the measures of intrinsic dimensionality of real data we discussed in subsection 1.1.2 might underestimate the true values.

While these tools provide valuable insights, they also underscore the need for new methodologies to probe the complex mechanisms responsible for building abstract and task-relevant representations in deep neural networks. In this thesis, we propose additional tools to this goal. These include the study of the neural tangent kernel after training (chapter 2), and, central to our work, the introduction of relative sensitivity measures for network activations in response to input transformations that leave the label unchanged (chapter 3, 4, 5 and 6). These sensitivity measures are centered on the concept of data invariances that we discuss more in depth in section 1.3.

Taking a broader view, the methods we discussed in this section seem to suggest that deep neural networks are able to learn increasingly abstract representations that are also lower and lower dimensional with depth, as they disregard irrelevant variability in the input. A key question arises from this observation:

- Does this dimensionality reduction via learning relevant features enable beating the curse of dimensionality?

To address this question we need to understand how feature adaptation affects performance—a matter that the empirical methods reviewed so far do not clarify. To investigate this point, we will introduce the two different training regimes of neural networks as they have been characterized in the literature. As already hinted at, this will allow us to isolate the contribution of feature learning to neural networks' performance, and to characterize it.

### 1.2.2   Training Regimes: *Feature* vs. *Lazy*

In this section, we will consider highly overparametrized neural networks, in particular, their infinite-width limit. The significance of this limit lies in the fact that it often corresponds in practice to the point of optimal performance. We discuss how to easily access this limit in practice via ensemble averaging, and its potential to elucidate various training regimes of neural networks, notably the *feature* and *lazy* regimes.

**Overparametrization and Double Descent**   Overparametrization in the context of neural networks refers to the scenario where the number of parameters in the model significantly exceeds the number of training data points. While we have seen that in traditional machine learning settings, such scenario can lead to overfitting due to the bias-variance tradeoff introduced in section 1.1 (see also illustration in Figure 1.2)—deep learning models have repeatedly shown that they can generalize well even when highly overparametrized Neyshabur et al. (2015); Zhang et al. (2017); Advani et al. (2020). Neyshabur et al. (2015) highlight that this may be the case thanks to an inductive bias or implicit regularization toward small norm solutions. In this context, a larger number of hidden neurons would allow for solutions of lower complexity to exist, and training would find them thanks to the inductive bias.

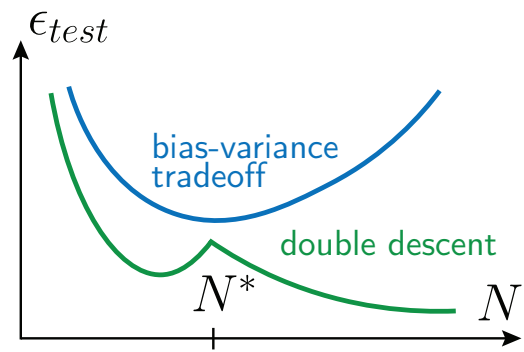

Figure 1.7: **Illustration of the bias-variance trade-off vs. double descent phenomenon.** Test error is reported on the *y*-axis, number of parameters of the model on the *x*-axis. The standard bias-variance trade-off that is present when e.g. fitting a function with a polynomial as in Figure 1.2, would predict the U-shaped behavior shown in blue (*N*, in this case, would correspond to the degree of the polynomial). Deep neural networks, instead, show a behavior like the one reported in green.

One of the most striking demonstrations of this seemingly counter-intuitive behavior is the phenomenon of *double descent*, as described by Spigler et al. (2019); Belkin et al. (2019); Advani et al. (2020); Nakkiran et al. (2021); Deng et al. (2020); Gerace et al. (2020); Hastie et al. (2020); Belkin et al. (2020); Mei and Montanari (2020); d'Ascoli et al. (2020a,b). The double descent curve refines our understanding of the bias-variance trade-off showing that the test error of a model does not always monotonically increase with overparametrization, as the U-shaped bias-variance curve would suggest. Instead, the curve exhibits a second descent, starting from the point where the model becomes overparametrized, see illustration in Figure 1.7. The double descent is a robust phenomenon that can be observed across a variety or architectures datasets and optimization procedures Nakkiran et al. (2021).

The second descent starts at the interpolation threshold, this is the minimal number of parameters for which the network can fit all training data, and is due to the noise coming from the random initialization of the network parameters Neal et al. (2019); Geiger et al. (2020a). Indeed, if one uses an ensemble of differently initialized networks as the predictor, the second peak disappears, and, when it can fit the training data, such a predictor often achieves the asymptotic test error a single network would display when the number of parameters tends to infinity Geiger et al. (2020a).

Having illustrated the significance of the infinite-width limit, and how to access it in practice, in the next paragraphs we will see how this limit serves as a foundation for the theoretical understanding of neural networks behavior, both at initialization and in the training process.

**Propagation at initialization** In the infinite-width limit, the signal propagation at initialization has been extensively explored Neal (1996); Williams (1997); Lee et al. (2018); de G. Matthews et al. (2018); Novak et al. (2019). Early foundational work by Neal (1996) and Williams (1997) demonstrated that a single-layer network with infinite hidden nodes behaves like a Gaussian process Görtler et al. (2019). This was later extended to deep networks Lee et al. (2018), showing that in the limit of infinite width, a deep neural network with random weights also behaves like a Gaussian process. These findings have paved the way for more recent studies investigating the infinite-width dynamics of neural networks.

**Neural Tangent Kernel and the Lazy Regime** Jacot et al. (2018) showed that in the limit of infinite width, the dynamics of deep neural networks during training can be exactly described by a kernel coined the Neural Tangent Kernel (NTK). They observed that the evolution under gradient descent of the network parameters in parameters space

$$\partial_t \theta = -\nabla_\theta \mathscr{L}, \tag{1.11}$$

$$= -\frac{1}{P}\sum_{i=1}^{P} \nabla_\theta f(\boldsymbol{x}_i)\nabla_f l(f, y_i) \tag{1.12}$$

could be rewritten in function space via the chain rule as

$$\partial_t f(\boldsymbol{x}) = \partial_\theta f(\boldsymbol{x})\partial_t \theta \tag{1.13}$$

$$= -\frac{1}{P}\sum_{i=1}^{P} \nabla_\theta f^\top(\boldsymbol{x})\nabla_\theta f(\boldsymbol{x}_i)\nabla_f l(f, y_i) \tag{1.14}$$

$$= -\frac{1}{P}\sum_{i=1}^{P} \Theta(\boldsymbol{x}, \boldsymbol{x}_i)\nabla_f l(f, y_i), \tag{1.15}$$

where we introduced the NTK:

$$\Theta(\boldsymbol{x}, \boldsymbol{x}') = \nabla_\theta f(\boldsymbol{x})^\top \nabla_\theta f(\boldsymbol{x}'). \tag{1.16}$$

Importantly, they discovered that, at large widths, the network's parameters move little with respect to initialization. For this reason, this regime has been also coined *lazy* Chizat and Bach (2018). As a consequence, the NTK remains constant during training in the infinite width limit, effectively evolving the network function in a linearized subspace around initialization Lee et al. (2019).

This work showed that, in this limit, neural networks effectively behave as kernel methods,

two machine learning approaches previously considered very different. Furthermore, this connection allowed to prove the convergence of gradient descent to zero loss solutions Li and Liang (2018); Du et al. (2019a,b); Allen-Zhu et al. (2019); Chizat and Bach (2018); Soltanolkotabi et al. (2019); Arora et al. (2019a); Zou et al. (2020). Arora et al. (2019b) extended the analysis to convolutional networks by computing the NTK of these architectures and the diagonalization Cagnetta et al. (2022) of such kernels allowed for understanding which functions they are able to efficiently learn.

**Feature or Active Regime**   While the NTK framework sheds light on the lazy training regime of deep learning, several works have highlighted a distinct, complementary aspect known as the *feature*, *active*, *rich* or *mean-field* regime Mei et al. (2018, 2019); Rotskoff and Vanden-Eijnden (2018); Chizat and Bach (2018); Sirignano and Spiliopoulos (2020); Nguyen (2019).

In this regime, we consider 2-layer neural networks whose width is taken to infinity but with a crucial modification to the output definition. Specifically, an additional factor $1/\sqrt{h}$ is introduced in the output of the model as defined in Equation 1.5, leading to a rescaling of the form:

$$f_\theta(x) \to \frac{1}{\sqrt{h}} f_\theta(x), \tag{1.17}$$

This rescaling makes the output $\ll 1$ at initialization, hence, to fit a target function of order one, the weights need to change significantly from their initialization. This limit gives rise to the so-called *feature learning regime* as neurons learn how to respond to different aspects of the input data in a substantial, rather than infinitesimal, manner.

In the feature learning regime and for $h \to \infty$, the neural network can be described entirely in terms of the density $\rho(\boldsymbol{w}_2, \boldsymbol{w}_1, \boldsymbol{b}_1)$ of parameters characterizing each neuron. An integral effectively replaces the sum in the network's output, which now takes the form:

$$f_\theta(\boldsymbol{x}) = \int d\boldsymbol{w}_2 \, d\boldsymbol{w}_1 \, d\boldsymbol{b}_1 \rho(\boldsymbol{w}_2, \boldsymbol{w}_1, \boldsymbol{b}_1) \boldsymbol{w}_2^\top \sigma\left(\frac{\boldsymbol{w}_1^\top \boldsymbol{x}}{\sqrt{d}} + \boldsymbol{b}_1\right). \tag{1.18}$$

Gradient descent dynamics under this limit leads to a dynamical equation on $\rho$ that is the typical one in physics for conserved quantities: i.e. the divergence of a flux

$$\partial_t \rho = -\nabla \cdot J, \tag{1.19}$$

where $J = \rho \Psi(\boldsymbol{w}_2, \boldsymbol{w}_1, \boldsymbol{b}_1; \rho_t)$ and $\Psi$ is a function that can be expressed in terms of the loss function Mei et al. (2018).

The term *hydrodynamic* comes into play due to the analogy between the evolution of $\rho$ and the hydrodynamics of interacting particles, each representing a neuron, in some external potential defined by the loss function. This formulation elegantly describes the collective behavior of the neurons as if they were fluid particles moving in a potential.

Despite their value, these studies have limitations. It's apparent from Figure 1.6 that depth plays a crucial role in feature learning. However, even though the mean-field limit has been investigated for deep networks Araújo et al. (2019); Sirignano and Spiliopoulos (2021); Nguyen and Pham (2023), incorporating depth into the mean-field framework fails to provide a compact description of the training dynamics, making the task of tracking it infeasible.

In sum, the feature learning regime provides a deeper understanding of why and when neural networks move beyond linear models, and it emphasizes the importance of understanding the interplay between these two regimes to fully grasp the behavior of deep learning. Specifically, examining these two regimes could help disentangle the influence of feature adaptation and architectural choice on the performance of neural networks. This perspective gives rise to a crucial question for this thesis: Is the future a return to kernel methods, as seen in the lazy regime, or does the magic of deep learning mainly stem from its ability to learn the relevant features in the data?

**Feature vs Lazy Regimes: Performance**   Many studies have focused on the transition between feature and lazy training regimes in deep learning Chizat and Bach (2018); Geiger et al. (2020b); Woodworth et al. (2020)—see also our review paper Geiger et al. (2021), omitted from this thesis. These works have shown that, even at finite widths, the scale of initialization controls the transition between the two regimes. This understanding has facilitated empirical investigations of performance in different settings as a function of the initialization scale. Chizat et al. (2019); Geiger et al. (2020b, 2021); Lee et al. (2020); Woodworth et al. (2020) has shown interesting differences in performance between the two regimes. When learning image tasks, fully-connected networks trained with gradient descent tend to do better in the lazy training regime. In contrast, for convolutional neural networks, feature learning generally performs better, see also Arora et al. (2019b); Shankar et al. (2020). This difference in performance leads to several important questions that we are going to address in the following chapters.

- What causes this performance gap between the two regimes?

- More specifically, can we design simple data models to make sense of the generalization error difference between feature learning and lazy training (chapter 2)?

- Why do FCNs struggle with image data in the feature learning regime when trained via gradient descent (chapter 4)?

- Why CNNs succeed in such setting (chapter 3 and 6)?

- Which mechanisms are responsible for successful feature learning in CNNs (chapter 5)?

To provide answers to these questions we need to discuss how to model and characterize the structure of the data that the feature learning regime can adapt to.

## 1.3 Data Structure

Data in the real world comes in various forms, each with its own unique properties and structures. The ability of a neural network to profitably learn features from data depends on its capacity to adapt to these particular structures. Therefore, understanding and characterizing the structure of data becomes a necessary step for assessing the benefits of feature learning quantitatively.

In particular, in this thesis we push forward the idea that this structure can be better characterized in terms of data invariances, that is aspects of the input data that leave the label unchanged. For instance, the stylized dog sketch in Figure 1.8 shows how a few lines can capture the essence of an object. This suggests that pixels at the corner of the frame, or the exact position of the relevant features, do not matter for recognizing the dog. Likewise, in hierarchical tasks like text, synonyms can be exchanged without altering the overall content of a sentence. Understanding which invariances rise from these properties of data, and their roles across different tasks, is a key step toward comprehending how deep learning models learn. In the sections that follow, we better define and investigate these different types of invariances, discussing how neural network architectures are able to exploit them. We will point to the chapters of this thesis that address these concepts in depth.

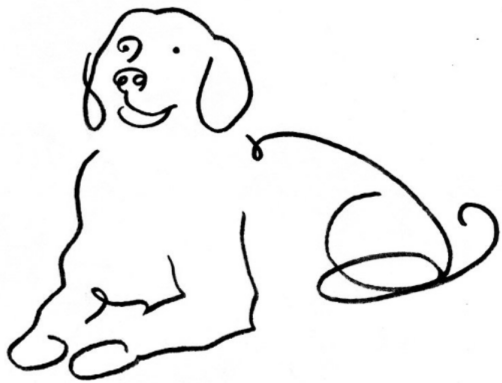

Figure 1.8: **Stylized representation of a dog.** This figure illustrates the idea of data invariances within image data, where recognizable features can be represented with a few lines, and small deformations often do not alter the overall semantic content.

### 1.3.1 Linear Invariance

Arguably the simplest invariance that gives structure to data is *linear invariance*. This manifests when the target function is highly anisotropic, in the sense that it depends only on a linear subspace of the input space—i.e. the target function is of the form

$$f^*(\boldsymbol{x}) = g(A\boldsymbol{x}) \quad \text{where} \quad A : \mathbb{R}^d \to \mathbb{R}^{d'} \quad \text{and} \quad d' \ll d. \tag{1.20}$$

This functional form for the target is often referred to as *single-index model* in the literature. An example of such linear invariance in real data could be pixels at the corner of an image, whose content may be irrelevant for the task.

**Shallow FCNs can learn Linear Invariance**    Several works have established that 2-layers fully connected networks can profitably exploit linear invariance in the feature learning regime Barron (1993); Bach (2017); Chizat and Bach (2020); Schmidt-Hieber (2020); Yehudai and Shamir (2019); Ghorbani et al. (2019, 2020); Wei et al. (2019). In particular, Barron (1993); Mei et al. (2016); Bach (2017) characterize the approximation properties of 2-layer networks, while Bach (2022); Schmidt-Hieber (2020); Chizat and Bach (2020); Ghorbani et al. (2020) discuss generalization, showing that these networks in the feature regime can beat the curse of dimensionality by adapting to the low-dimensional subspace, while this is not the case for kernel methods. Several results followed showing an advantage of the feature over the lazy regime, in various specific classification and regression settings with anisotropic target functions Ghorbani et al. (2019, 2020); Refinetti et al. (2021). Damian et al. (2022); Abbe et al. (2021, 2023); Bietti et al. (2022) have tackled the problem of determining the required number of samples to learn anisotropic tasks. They provide upper bounds for the sample complexity of the feature regime and highlight that it performs better than the lazy regime for similar problems Ghorbani et al. (2019). It remains unclear whether these bounds are tight in practice.

The work presented in **chapter 2** fits in this line of research. Specifically, we revisit the binary classification task introduced in Paccolat et al. (2021b) and show that, if the target function only depends on a linear subspace of input space, the weights associated with the relevant input subspace grow by $O(\sqrt{P})$ compared to the irrelevant weights, thereby building the right features for the task. This allows us to estimate the rates of generalization error with $P$ for both feature and lazy regimes. Notably, we demonstrate that these estimates are tight, in the sense that they accurately predict generalization error in practice. Importantly, our examination of a problem where test error versus $P$ follows a power law further enables us to draw qualitative comparisons with observations from real-world datasets, where power laws behaviors are ubiquitous Hestness et al. (2017); Spigler et al. (2020); Kaplan et al. (2020).

**Limitations of the study of Shallow Networks**    While this growing body of theoretical work allows for a better and better understanding of the advantages of learning features in 2-layers neural networks, we have seen that, for these architectures, learning features is sometimes in practice a drawback Geiger et al. (2020b); Lee et al. (2020). This implies that linear invariances, such as those concerning boundary pixels, may not be the most crucial in practical applications, and other forms of invariances might be at play for beating the curse of dimensionality.

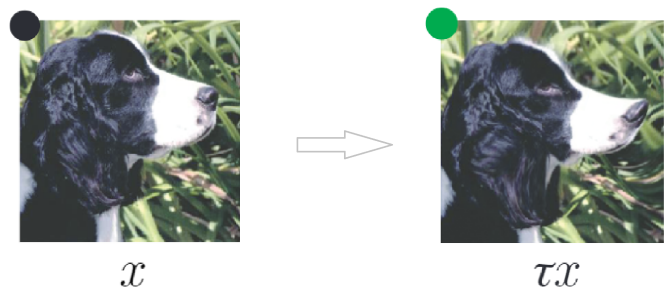

$x$ $\qquad$ $\tau x$

Figure 1.9: An image of a dog $x$ and its deformed version $\tau x$. The deformation $\tau$ is sampled from the distribution we introduce in chapter 3.

### 1.3.2 Deformation Invariance in Images

It is fascinating to note that we can often identify objects even from simple sketches containing only a few lines, as shown in Figure 1.8. This ability suggests that the important features in images are *sparse in space*, in the sense that they only occupy small portions of the image frame. As a consequence, the image frame can be slightly deformed, and hence the relevant features be moved, without altering the overall content of the image. Examples of such small deformations are shown in Figure 1.9. This hypothesis——that effective image classifiers should be stable to deformations——was advanced in Bruna and Mallat (2013); Mallat (2016). For a function $f$, being stable to deformations means that, given an image $\boldsymbol{x}$ and an operator $\tau$ that applies a smooth deformation, then $\|f(\boldsymbol{x}) - f(\tau \boldsymbol{x})\|$ is small, if the norm of $\tau$ is small. Bruna and Mallat (2013); Mallat (2016) further propose an architecture, the Scattering Transform, that is specifically designed to be stable to deformation by the use of localized filters at different scales and frequencies, providing insights into which kind of filters CNNs may learn to achieve such stability.

This picture naturally leads to an important question:

- Is this intuitive hypothesis verified in practice? In other words, do the high-performing modern neural networks actually succeed because they effectively leverage deformation invariance?

Empirical studies have shown that even slight shifts, rotations, or scale changes in images can significantly alter the network's output Azulay and Weiss (2018); Zhang (2019). This observation seems to contradict the hypothesis that CNNs are robust to minor deformations. However, the image transformations applied in these works often do not qualify as small deformations as they led to images with statistical properties dramatically different from the training set and involved procedures such as image cropping. A class of smooth deformations is introduced in Ruderman et al. (2018), suggesting that some level of deformation stability can be learn by deep neural networks. However, similarly to the previous studies, this work also did not investigate the impact of the learned stability on performance.

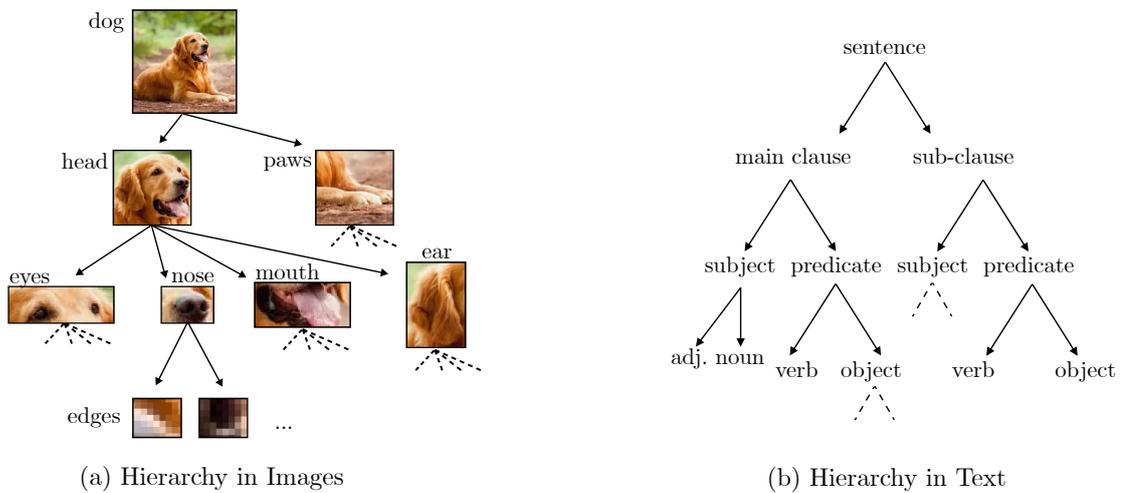

(a) Hierarchy in Images

(b) Hierarchy in Text

Figure 1.10: **Illustrating the hierarchical structure of real data.** (a) An example of the hierarchical structure of images: the class (dog) consists of high-level features (head, paws), that in turn can be represented as sets of lower-level features (eyes, nose, mouth, and ear for the head). (b) A similar hierarchical structure can be found in natural language: a sentence is made of clauses, each having different parts such as subject and predicate, which in turn may consist of several words.

Moving forward, we need a more comprehensive empirical examination of deformation stability in neural networks. This is the aim of chapter 3. In this chapter, we build an empirical framework to study deformation stability in neural networks. This framework includes both an ensemble of diffeomorphisms of controlled norm to be applied to images, and an observable to characterize deformation invariance. We will investigate whether deformation invariance is learned in neural network, or present from the start, and how different neural networks achieve different levels of invariance on benchmark image classification tasks. Finally, we establish a relationship between the learned deformation stability and performance of neural networks.

### 1.3.3 Hierarchical Tasks and Synonymic Invariance

The invariance discussed in the previous section is related to the fact that relevant features only occupy a small portion of the whole input. However, this does not address how these features need to be combined to accomplish a task, nor what is the role of depth in neural networks architectures. This leads us to another property that is likely relevant for the learnability of real data: hierarchical compositionality. This property prescribes how different low-level features are combined in order to produce higher-level features and eventually the label. This is a concept highlighted in multiple studies Lee et al. (2009); Bruna and Mallat (2013); Patel et al. (2015); Mossel (2018); Poggio et al. (2017); Mhaskar and Poggio (2016, 2019); Malach and Shalev-Shwartz (2018); Schmidt-Hieber (2020); Cagnetta et al. (2022), among others. To give a concrete example, consider an image of a dog—see illustration in Figure 1.10(a). At the highest level, we recognize the overall form of the dog. Breaking it down, we notice more specific features: the head, the body, the tail, and the legs. Each of these elements can be further divided into even lower-level features, such as the eyes, nose, and mouth on the head, or the fur patterns on the body. This hierarchical structure, from general to specific, mirrors how we recognize and interpret complex images. In fact, such a hierarchy can be found both in the visual cortex Gazzaniga et al. (2006) and in the layers of artificial networks trained on image classification (Figure 1.6), suggesting a fundamental relevance of these structures in the processing and understanding of natural visual data. Finally, this hierarchical structure is not exclusive to vision as it can be also found, for example, in natural language where a piece of text can be decomposed into paragraphs, sentences, sub-clauses, words, and syllables (Figure 1.10(b)).

- Can deep neural networks trained by gradient descent efficiently solve hierarchical tasks?

- How many data points do they need to achieve small generalization error?

- Furthermore, is there an associated invariance with the hierarchical structure, and can its understanding aid in answering the aforementioned questions?

**Previous Works**   Poggio et al. (2017); Mhaskar and Poggio (2016, 2019) study the approximation properties of shallow and deep networks of hierarchically local compositional functions as for example

$$f(\boldsymbol{x}) = g_4(g_1(x_1, x_2, x_3), g_2(x_1, x_4), g_3(x_5)). \tag{1.21}$$

Even if both shallow and deep networks can approximate any function with enough parameters, deep networks only need a number of parameters linear in $d$, while shallow networks need exponentially many. Moreover, deep networks allow for solutions whose generalization error is controlled by the maximum dimensionality of the constituent functions instead of the input dimension $d$ Schmidt-Hieber (2020). Gradient descent, though, can efficiently learn these kinds of tasks only if correlations between input features and the target label exist Shalev-Shwartz et al. (2017); Mossel (2018). To understand how the magnitude of correlations influence the sample complexity of gradient descent Malach and Shalev-Shwartz (2018, 2020) introduce a class tasks where inputs are a hierarchical composition of multiple levels of features, starting from a class down to low-level features. They further propose a sequential, layer-wise algorithm that alternates clustering with gradient descent steps, and are able to compute the sample complexity of such algorithm as a function of input-label correlations.

These seminal works start to address the questions we highlighted in this section. It still remains unclear though what is the sample complexity of the standard deep learning algorithms, i.e. vanilla or stochastic gradient descent on modern deep networks like convolutional neural networks, and if hierarchical compositionality gives rise to some kind of invariance. In **chapter 6** we answer these questions in the context of a generative model of hierarchical data that falls in the class of Malach and Shalev-Shwartz (2018), but for which we can explicitly compute input-output correlations in terms of the model parameters. We show what is the role of these correlations in determining the sample complexity of standard algorithms, we introduce the invariance with respect to the exchange of synonymous features as a way to characterize dimensionality reduction and show that deep networks are able to profitably learn it.

## 1.4   Overview of the Thesis

The core focus of this thesis is an exploration of deep neural networks' ability to adapt to data structures, specifically focusing on data invariances, and the consequential impacts of such adaptability on network performance. The various questions raised in this introduction serve as the central motivations guiding our research. In the first part of this section, we provide an overview of the content of the thesis, which is divided into four parts. We conclude this introductory chapter by describing the methodological framework behind our work (subsection 1.4.2).

### 1.4.1   Structure of the Thesis and Main Results

In **Part I** we start our investigation of the role of data invariances in the success of deep learning, focusing on linear invariance.

- In particular, in **chapter 2** we develop a quantitative framework to study how learning invariant representations can benefit performance, and then use this framework to gain qualitative understanding of real-world scenarios. We consider a simple model of linear invariance, where the data points lie in $d$ dimensions while the labels vary within a linear manifold of lower dimension $d' < d$. We provide evidence that in the feature learning regime, shallow neural networks can effectively adapt to the data structure, becoming invariant to the uninformative directions in input space by aligning weights towards the relevant directions. We quantify the magnitude of this alignment and show that it depends on the square root of the training set size. Contrarily, we show that this adaptation is absent in neural networks trained in the lazy regime, resulting in poorer performance. We quantify this gap with a prediction on the scaling exponent of generalization error vs. the number of training points, and we empirically verify that these exponents are tight in practice.

  To better understand the benefits of feature learning, we study the evolution of the Neural Tangent Kernel (NTK) over training time. We find that, as the NTK evolves, only a few of its eigenvalues become non-negligible, and the corresponding eigenvectors become more informative and align closely with the labels. This process leads to a kernel better suited for the task, with a performance matching the one of the feature regime. This finding applies to both shallow networks trained on the simple model of linear invariance and to deep CNNs trained on benchmark image datasets, highlighting the importance of invariants learning in both settings.

Motivated by the need to study more complex forms of invariances to understand the performance of deep neural networks on real data, **Part II** is devoted to the study of deformation invariance.

- In **chapter 3**, we aim to provide empirical evidence to the hypothesis that deep networks learn deformation invariance with training, hence reducing the dimensionality of the problem and leading to improved performance. To do so we introduced a maximum entropy distribution over diffeomorphisms. This allowed for generating typical diffeomorphisms with controlled norm.

  We then define the *relative stability to diffeomorphisms*, denoted as $R_f$, which is an empirical observable that characterizes the stability of the network function $f$ when transforming the input along a diffeomorphism, relative to the stability to additive noise of the same amplitude. The need for defining stability in relative terms arises from the observation that different network architectures exhibit varying levels of stability to additive noise, with better architectures typically being *less* stable.

  Before training, $R_f$ is found to be close to one for various datasets and architectures, suggesting that the initial network output is as sensitive to smooth deformations as it is to random perturbations of the image. However, after training, we observe a strong correlation between $R_f$ and the test error, with $R_f$ reducing significantly in modern architectures and benchmark image datasets with training. Contrarily, for FCNs $R_f$ even increases with training, highlighting their inability to learn deformation invariance. These results support the hypothesis that learning diffeomorphism invariance is key to achieving good generalization performance in image tasks.

  This naturally raises the question of why fully-connected neural networks perform so poorly, with an $R_f$ that even increases with training, and how this invariance is mechanistically achieved by deep neural networks that perform well. These two questions are addressed in chapter 4 and chapter 5, respectively.

- In **chapter 4**, we explore the limitations of feature learning in 2-layers neural networks for settings where non-linear invariances are present. In particular, we argue that if a given task presents a non-linear invariance, then it is better solved by a predictor that has little variations (i.e. is smooth) along directions of input space that correspond to the invariance. A fully-connected network would require a continuous distribution of neurons to represent such a task. However, in the feature regime and in the limit of small initialization or with regularization of the weights, neurons become sparse, i.e. orient themselves along a finite number of input directions Bach (2017); de Dios and Bruna (2020). In the case where linear invariances are not present, this sparsity can lead to a predictor that overfits some spurious directions that are not relevant for the task. We argued that this picture applies to: *(i)* regression of random functions of controlled smoothness on the sphere—where the target is stable to small rotations—and *(ii)* classification of benchmark image datasets—where the target is stable to small deformations.

In particular, we characterize the generalization error of neural networks trained in the feature and lazy regime in the first setting (*i*), and find that lazy training leads to smoother predictors than feature learning. As a result, lazy training outperforms feature learning when the target function is also sufficiently smooth. We derive generalization error rates through asymptotic arguments that we systematically back up with numerical studies.

In the second setting (*ii*), we build on our observation that good performance in image classification is associated with the stability or *smoothness* of the predictor function along diffeomorphisms. In fact, we show that lazy training, by maintaining a continuous distribution of neurons, results in predictors with smaller variations along diffeomorphisms compared to the feature regime, which leads to sparse solutions. This observation offers an explanation as to why lazy training outperforms feature learning in the context of image datasets.

This chapter focuses on the drawbacks of feature learning. Why this training regime is so effective when training modern architectures is still an open question.

- Building on the results of chapter 3, the goal of **chapter 5** is to elucidate some of the mechanisms by which CNNs learn the deformation invariance when classifying images, and how this surprisingly leads to instability to noise. We discuss two kinds of pooling mechanisms that can grant deformation invariance, and disentangle their role. The first, *spatial* pooling, integrates local patches within the image, thus losing the exact location of its features. The second, *channel* pooling, requires the interaction of different channels, which allows the network to become invariant to any local transformation by properly learning filters that are transformed versions of one another. Our experiments on benchmark image datasets reveal that both kinds of pooling are learned by deep networks and are responsible for improving deformation invariance. In particular, we show that spatial pooling is learned by making filters low-frequency.

We then focus on understanding how deep networks learn spatial pooling. To do so, we introduce idealized scale-detection tasks that can be solved by performing spatial pooling alone. Our findings suggest that the neural networks performing best on real data tend to excel in these tasks, highlighting their relevance in mirroring real-world scenarios. These tasks also open the way for theoretical analysis. In particular, they allow us to understand how deformation stability is achieved layerwise, by developing low-frequency filters, and how this naturally results in instability to additive noise.

After discussing how neural networks are able to handle the spatial sparsity of features, we go into the study of how these features are composed in order to solve hierarchically and compositionally local tasks (**Part III**).

- To this end, in **chapter 6** we introduce a model for hierarchical data that we name the *Random Hierarchy Model*, which is characterized by classes composed of combinations of high-level features, iteratively constructed from sub-features. Moreover, multiple combinations of sub-features can construct the same high-level feature. We refer to these different combinations as *synonyms*.

  One of our key findings is that the number of training data, referred to as sample complexity and denoted as $P^*$, required by deep convolutional neural networks trained by gradient descent to learn this task, contrary to what one might expect from dealing with high-dimensional data, grows only polynomially with the input dimensionality. This suggests that the curse of dimensionality can indeed be overcome for such hierarchical tasks. Our study further reveals that the sample complexity of deep CNNs $P^*$ is closely linked with the learning of invariant representations. Specifically, $P^*$ corresponds to the training set size at which the network's representations become invariant to exchanges of synonyms——in other words, it no longer matters for the internal representations which specific features are used to represent a class, as long as they are semantically equivalent. This is a critical aspect of learning hierarchically compositional tasks, as it shows that deep CNNs are not merely learning to recognize specific features, but are learning the underlying structure and relationships among the features. Notice that the picture is very different for shallow neural networks as they can only solve the task with a number of data points that is exponential in the dimension, hence incurring in the curse of dimensionality.

  Furthermore, $P^*$ coincides with the number of data at which the correlations between low-level features and classes become detectable. This finding suggests that deep CNNs leverage the correlations in the data in order to solve the task, enabling us to rationalize how the sample complexity depends on the hierarchical and compositional structure of the task.

This introductory chapter has attempted to highlight the primary outcomes of our research. The chapters that follow delve deeper into these findings, each representing an individual research paper that took shape during my doctoral journey. These chapters document my research progression and contributions to the field, with the aim of providing a basis to inspire and inform future investigations in this domain. The latter themes are further developed and expanded upon in the final concluding **chapter 7**.

### 1.4.2   Methodology: Empirical Science Approach to Deep Learning

In our research, we have adopted a methodology that combines the experimental approach used in the natural sciences with theoretical principles akin to those found in physics Zdeborová (2020). This method is well presented and advocated for in Nakkiran and Belkin (2022). We summarize their viewpoint in the following paragraph.

Machine learning research falls into two main areas. The first, *technological* research, works on improving learning systems, while the second, *scientific* research, aims to understand how learning works. Most current research is technological, focusing on improving practical performance. Scientific research, although essential, often receives less attention and is judged using similar standards as technological research, which may not be ideal. Moreover, while rigorous mathematical theories have gained significant attention in the community, they are just one part of the picture. The necessity for an empirical approach in machine learning stems from these shortcomings. Machine learning, especially in its application to deep learning, often deals with high levels of complexity that are beyond our current analytical capabilities. In these situations, it becomes essential to build systems and observe them, employing the principles of empirical inquiry commonly used in natural sciences.

In our research, we have aimed to embrace this empirical approach to machine learning, focusing on conducting experiments that reveal key insights into the behavior of deep learning algorithms, despite the scarcity of rigorous mathematical theorems regarding practical settings. Our goal has been to understand and interpret the patterns that emerge from these experiments, and subsequently develop theoretical frameworks sometimes in the form of simplified *toy models*, that elucidate these patterns. Toy models, borrowed from the practice of physics, are simplified theoretical constructs that, while not capturing all aspects of the complex systems they represent, isolate and illustrate critical features or behaviors. In the context of machine learning, such toy models can offer valuable qualitative insights into the workings of more complex deep learning algorithms.

This approach echoes the tradition of experimental physics, where empirical data and theoretical models, including toy models, engage in a continuous dialogue, each refining and informing the other. Such a feedback loop between empirical studies and theoretical modeling can greatly enhance our understanding of deep learning systems, providing a path to bridge the gap between theory and practice.

We conclude by discussing more practical considerations and challenges regarding our methodological approach to the study of deep learning systems.

**Hyperparameters Tuning**   Our methodology crucially involves the efficient handling of hyperparameters in deep learning models due to their significant impact on performance. We employ strategies to limit the number of hyperparameters needing tuning. One example of that is choosing the width of 2-layer networks. We set the width it sufficiently large to ensure results remain stable when further increased, the goal being to operate in the infinite-width limit. This choice is driven by observations that the infinite-width limit usually delivers superior performance in all training regimes—as we highlight in Geiger et al. (2021). Other examples involve the choice of loss function and training dynamics. For binary classification, we often use the hinge loss which provides a clear stopping criterion——zero training loss—that allows us to avoid tuning training time. Additionally, to avoid tuning the learning rate, we use a dynamics that mimics gradient flow, corresponding to the zero learning rate limit of gradient descent.

**Sensitivity to Hyperparameters**   While we have aimed to reduce the dependence of our results on hyperparameter tuning by adopting specific experimental settings, we appreciate the importance of understanding the sensitivity of our models to these choices and we aim to measure such sensitivity when computational resources allow for it. This is especially needed in cases where the choice of some of the parameters is not completely justified, or to check the generality of our findings. For instance, in chapter 3, to thoroughly test the relationship we find between stability to diffeomorphisms and test errors we examine it across multiple benchmark image datasets, across different values of the parameters that control the magnitude and spatial frequencies of the diffeomorphisms, and for varying number of training points.

**Statistical Significance**   Ensuring statistical significance is also an essential aspect of our research approach. We maintain full control over all sources of randomness, including the sampling of the training set, the initialization of network parameters, and the noise of SGD. To ensure that the variance of the results is under control, we perform multiple experiments in which we vary all sources of randomness simultaneously and average the results.

**Reproducibility**   Reproducibility is vital in scientific research, including empirical deep learning studies. Therefore, all the code used in our experiments is publicly available.[4] We also maintain consistency in our experiments by standardizing preprocessing steps of our datasets and computational constraints during model training. This uniformity ensures fair comparison across models and setups, minimizing biases from varied conditions. By doing so, we aim to enhance the transparency and reproducibility of our work, offering a stable framework within which our results can be replicated.

**Limitations of Empirical Research**   Although empirical research can yield relevant insights into deep learning systems, it's worth noting some inherent limitations. A comprehensive and rigorous validation of empirical claims, as seen in theoretical work, is not possible as no number of experiments can ever prove a scientific theory. Still, a reproducible experiment or observation can refute one, and this falsifiability is what gives validity to scientific theories Popper (1968). In this light, our aim is to support our hypotheses by exposing them to as diverse and challenging scenarios as possible in order to test their validity. However, this does not negate the existence of situations, contexts, or datasets where our findings may not apply, or the possibility of unaccounted confounding factors. As we aim to conduct responsible and robust research, we want to be transparent about these potential limitations. Moreover, we value the role of the peer review community in highlighting potential weaknesses and offering insightful critiques, which are essential to empirical research.

---

[4]The code to reproduce the experiments of this thesis is available online at github.com/leonardopetrini and github.com/pcsl-epfl.

# Linear Invariance Part I

# 2 Generalization Error Rates of Fully-Connected Networks when Learning Linear Invariances

The following paper is the preprint version of Paccolat et al. (2021a) published in *Journal of Statistical Mechanics: Theory and Experiment.*

**Candidate contributions**   The candidate contributed to discussions and was responsible for the second part of the paper (Sections 4 and 5).

# Geometric compression of invariant manifolds in neural nets

Jonas Paccolat[a], Leonardo Petrini[a], Mario Geiger[a], Kevin Tyloo[a], and Matthieu Wyart[a]

[a]Institute of Physics, École Polytechnique Fédérale de Lausanne, 1015 Lausanne, Switzerland

March 12, 2021

**Abstract**

We study how neural networks compress uninformative input space in models where data lie in $d$ dimensions, but whose label only vary within a linear manifold of dimension $d_\parallel < d$. We show that for a one-hidden layer network initialized with infinitesimal weights (i.e. in the *feature learning* regime) trained with gradient descent, the first layer of weights evolve to become nearly insensitive to the $d_\perp = d - d_\parallel$ uninformative directions. These are effectively compressed by a factor $\lambda \sim \sqrt{p}$, where $p$ is the size of the training set. We quantify the benefit of such a compression on the test error $\epsilon$. For large initialization of the weights (the *lazy training* regime), no compression occurs and for regular boundaries separating labels we find that $\epsilon \sim p^{-\beta}$, with $\beta_{\text{Lazy}} = d/(3d - 2)$. Compression improves the learning curves so that $\beta_{\text{Feature}} = (2d - 1)/(3d - 2)$ if $d_\parallel = 1$ and $\beta_{\text{Feature}} = (d + d_\perp/2)/(3d - 2)$ if $d_\parallel > 1$. We test these predictions for a stripe model where boundaries are parallel interfaces ($d_\parallel = 1$) as well as for a cylindrical boundary ($d_\parallel = 2$). Next we show that compression shapes the Neural Tangent Kernel (NTK) evolution in time, so that its top eigenvectors become more informative and display a larger projection on the labels. Consequently, kernel learning with the frozen NTK at the end of training outperforms the initial NTK. We confirm these predictions both for a one-hidden layer FC network trained on the stripe model and for a 16-layers CNN trained on MNIST, for which we also find $\beta_{\text{Feature}} > \beta_{\text{Lazy}}$. The great similarities found in these two cases support that compression is central to the training of MNIST, and puts forward kernel-PCA on the evolving NTK as a useful diagnostic of compression in deep nets.

## 1 Introduction and related works

Deep neural networks are successful at a variety of tasks, yet understanding why they work remains a challenge. Specifically, the data from which a rule or classes are learnt often lie in high dimension $d$ where the curse of dimensionality is expected. Quantitatively, this curse can be expressed on how the test error $\epsilon(p)$ depends on the training set size $p$. If mild assumptions are made on the task (for example regressing a Lipschitz continuous function), then $\epsilon(p)$ cannot be guaranteed to decay faster than $\epsilon \propto p^{-\beta}$ with an exponent $\beta = \mathcal{O}(1/d)$ [1]: learning is essentially impossible. In practice, $\beta$ is found to be much larger and to depend on the task, on the dataset and on the learning algorithm [2, 3], implying that learnable data are highly structured.

Accordingly, success of neural networks is often attributed to their ability to adapt to the structure of the data, which present many invariances [4]. For example in the context of classification, some pixels at the edge of the image may be unrelated to the class label. Likewise, smooth deformations of the image may leave the class unchanged. In that view, neural networks correspond to a succession of non-linear and linear operations where directions of neural representation for which the label does not vary are compressed. It is supported by the observations that kernels designed to perform such compression perform well [4]. Yet, there is no quantitative general framework to describe this compression and its effect on the exponent $\beta$. The information bottleneck framework for deep learning [5] proposes that information is compressed as it propagates deeper in the network. However, information in such a deterministic setting is ill-defined and conclusions can depend qualitatively on details of the architecture or on the estimation of information [6].

Still, more robust measures, such as the effective dimension of the neural representation of the data, support that compression occurs in deeper layers [7, 8].

Such a framework should include in which learning regime nets operate. Different regimes have recently been delineated by focusing on the infinite-width limits of neural networks, shown to converge to well-defined learning algorithms [9, 10, 11, 12]. These are practically useful limits to consider as performance generally improves with width [13, 14, 15, 16, 17], which simply comes from the fact that convergence to these asymptotic algorithms removes noise stemming from the random initialization of the weights [18, 19, 20]. Two limits are found, depending on how weights scale with width. In one limit [9], deep learning becomes equivalent to a kernel method coined Neural Tangent Kernel or NTK. Weights and neuron activities barely change and dimension reduction cannot occur. In the feature learning regime [10, 11], weights and neuron activities significantly change, the NTK evolves in time [10, 21] and compression can in principle occur. Yet understanding this dynamic and its effect on performance remains a challenge. For CNNs the feature learning regime tends to perform better [22, 19, 23] but it is not so for fully connected nets using vanilla gradient descent on various benchmarks of images [19]. This state of affairs calls for simple models of data in which the kernel evolution and its associated compression of invariants can be quantified, together with its effect on performance.

## 1.1 Our contribution

Here we consider binary classification and assume that the label does not vary along $d_\perp$ directions of input space. We will first focus on the *stripe model*, arguably the simplest model of invariant yet non linearly-separable data for which $d_\perp = d - 1$, and later show that our results holds for smaller $d_\perp$. Data consists of Gaussian random points $\underline{x}$ in $d$ dimensions, whose label is a function of a single coordinate $y(\underline{x}) = y(x_1)$, corresponding to parallel planes separating labels. In Section 3, we show for the stripe model that: (i) in the NTK limit, $\beta_{\text{Lazy}} = d/(3d-2)$ as we found earlier for isotropic kernels [24]. (ii) In the feature learning regime, if the weights are initialized infinitesimally a geometric compression along invariant directions of magnitude $\lambda \sim \sqrt{p}$ occurs at intermediate times. This weight compression is equivalent to a spatial compression of the data points as illustrated in Fig. 1. (iii) In the NTK limit if data are compressed by $\lambda$ before learning, performance closely matches that of the feature learning regime. This observation supports that the main gain of the latter regime is to perform this compression. Assuming that it is the case leads to the prediction $\beta_{\text{Feature}} = (2d-1)/(3d-2)$. In Section 4 we generalize this result to the case $d_\perp < d-1$, and argue that for sufficiently regular boundaries separating labels $\beta_{\text{Feature}} = (d + d_\perp/2)/(3d-2)$. We test this prediction when the boundaries separating labels is a cylinder with $d_\perp = 1$ and $d = 3$.

In Section 5, we argue that the evolution of the NTK is such that at the end of learning: (iv) The top eigenvectors of the associated Gram matrix become much more informative on the labels than at initialization. (v) The projection of the labels on these eigenvectors becomes large for the top eigenvectors and small otherwise, supporting that the performance of kernel methods using the NTK improves as it evolves during learning. We confirm these predictions empirically in the stripe model. Finally, we show that these points hold true in a multi-layer CNN applied to MNIST data, for which various observables are found to behave very similarly to the stripe model, including the fact that $\beta_{\text{Feature}} > \beta_{\text{Lazy}}$. These observations support that compression along invariant directions is indeed key to the success of this architecture, and underlines kernel PCA applied to the evolving NTK as a tool to characterize it.

The code used for this article is available online at `https://github.com/mariogeiger/feature_lazy/tree/compressing_invariant_manifolds`.

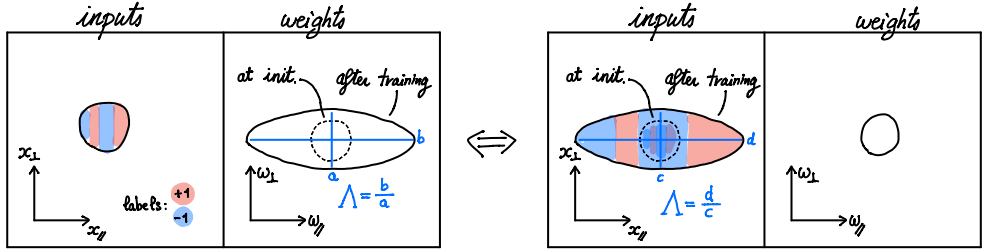

Figure 1: Illustration of data compression when the labels do not depend of $d_\perp$ directions in input space, as exemplified in the left panels. During training, the first layer weights inflate much more in the informative $d_\parallel$ directions. In relative terms, the network thus becomes much less sensitive to the $d_\perp$ uninformative directions. This effect is equivalent to a compression of uninformative directions in data space, as illustrated on the right panels.

### 1.2 Related works

In the physics literature, $\beta$ has been computed in regression or classification tasks for fixed kernels [25, 16, 26, 27, 28]. These results for classification generally consider linearly separable data and apply in the limit $d \to \infty$ and $p \to \infty$ with $\alpha = p/d$ fixed. In that limit for specific data it was shown for a regression task that feature learning can outperform the NTK regime [26]. Here we consider classification of non-linearly separable data, and take the limit of large training set size $p$ at fixed dimension $d$ which appears appropriate for common benchmarks [1].

Our work also connects to previous studies on how the anisotropy of the data distribution affects performance [3, 29, 24, 30]. For a large anisotropy, the effective dimension of the data is reduced, improving kernel methods [3]. The effect of a moderate anisotropy was investigated for kernel classification [24] and regression in neural nets [30]. Here we argue that in the presence of invariant, neural nets in the feature learning regime perform a compression equivalent to making the data anisotropic, and to our knowledge produce the first estimates of the training curves rate $\beta$ for both the lazy training and feature learning regime in the limit of large training set size $p$ at fixed dimension $d$.

Guarantees of performance for a one-hidden layer in the feature learning regime exist if some norm (characterizing the magnitude of the weight representing the function to be learnt) is finite, and if the dynamics penalizes this norm [10, 31]. In our model that norm is infinite (because there is no margin between labels of different classes). Instead we focus on vanilla gradient descent without special regularization (such regularizations are usually not used in practice). For gradient descent, with the logistic loss for a one-hidden layer can be shown to correspond to a max-margin classifier in a certain non-Hilbertian space of functions [32]. Dimension-independent guarantees on performance can be obtained if the data can be separated after projection in a low dimensional space, as occurs in our model. The analysis requires however to go to extremely long times. Here instead we focus on the hinge loss for which the dynamic stops after a reasonable time and we estimate the error and $\beta$ in specific cases instead of providing an upper bound to it.

On the empirical side, the alignment occurring during learning between the function being learnt and the top eigenvectors of the Gram matrix was noticed in [33] and observed more systematically in [34]. Our work offers an explanation for these findings in terms of the compression of invariant directions in data space.

## 2 General considerations on data and dynamics

### 2.1 Linear invariant data

We consider a binary classification task on data points lying in a $d$-dimensional space whose labels only depend on a linear subspace of dimension $d_\parallel < d$. Without loss of generality, we write the data points as $\underline{x} = (\underline{x}_\parallel, \underline{x}_\perp) \in \mathbb{R}^d$ with $\underline{x}_\parallel = (x_1, \ldots, x_{d_\parallel})$ and $\underline{x}_\perp = \underline{x} - \underline{x}_\parallel$, so that the label function only depends on the $d_\parallel$ first components: $y(\underline{x}) = y(\underline{x}_\parallel)$. In this work, we consider data points drawn from the standard normal distribution. In particular, we refer to the points of a training set of size $p$ as $\underline{x}^\mu \sim \mathcal{N}(0, I_d)$, for $\mu = 1, \ldots, p$.

---

[1]MNIST or CIFAR present an effective dimension $d_{\mathfrak{M}} \in [15, 35]$ [3] and $p \approx 6 \cdot 10^4$.

In Section 3, we shall focus on the simplest case where $d_\parallel = 1$, that we call the stripe model. In Section 4, we then generalize our findings to higher dimensional tasks and we confirm our results on a "cylindrical" model with $d_\parallel = 2$.

## 2.2 Learning algorithm

We consider the following fully-connected one-hidden layer neural network of ReLU activation,

$$f(\underline{x}) = \frac{1}{h} \sum_{n=1}^{h} \beta_n \, \sigma \left( \frac{\underline{\omega}_n \cdot \underline{x}}{\sqrt{d}} + b_n \right), \tag{1}$$

where $\sigma(x) = \sqrt{2} \max(0, x)$. In our simulations $h = 10000$. The trained parameters of the network are $\beta_n$, $\underline{\omega}_n$ and $b_n$. We use a vanilla gradient descent algorithm with the hinge loss on the predictor function $F(\underline{x}) = \alpha \left( f(\underline{x}) - f_0(\underline{x}) \right)$, where $f_0$ is the network function at initialisation and is not affected by gradient descent. With this trick, the amplitude of the network output is controlled by the scale $\alpha$. Varying it drives the network dynamics from the feature regime (small $\alpha$) to the lazy regime (large $\alpha$) [22]. The dynamical evolution of a generic weight $W \in \{\beta_n, b_n, \underline{\omega}_n\}_{n=1,\ldots,h}$ belonging to the network (1) thus follows the differential equation

$$\dot{W} = \frac{1}{p} \sum_{\mu=1}^{p} \partial_W f(\underline{x}^\mu) \, y(\underline{x}_\parallel^\mu) \, l' \left[ y(\underline{x}_\parallel^\mu) F(\underline{x}^\mu) \right], \tag{2}$$

where $l'(x) = \Theta(1-x)$ is the derivative of the hinge loss. All weights of the network are initialized according to the standard normal distribution. We show in Appendix A that the network output is statistically invariant under a rotation of the informative directions. Without loss of generality, we can thus choose the same basis for the data points as for the first layer weights. In particular, we introduce the following notation: $\underline{\omega}_n = (\underline{\omega}_{n,\parallel}, \underline{\omega}_{n,\perp})$.

## 2.3 Amplification factor

The effect of learning is quantified by the compression of the uninformative weights $\underline{\omega}_{n,\perp}$ with regard to the informative weights $\underline{\omega}_{n,\parallel}$. Mathematically, the neuron amplification factor $\lambda$ and the global amplification factor $\Lambda$ are defined as

$$\lambda_n = \frac{\left\| \underline{\omega}_{n,\parallel} \right\|_{d_\parallel}}{\left\| \underline{\omega}_{n,\perp} \right\|_{d_\perp}} \qquad \text{and} \qquad \Lambda = \sqrt{\frac{\sum_{n=1}^{h} \left\| \underline{\omega}_{n,\parallel} \right\|_{d_\parallel}^2}{\sum_{n=1}^{h} \left\| \underline{\omega}_{n,\perp} \right\|_{d_\perp}^2}}, \tag{3}$$

where $d_\perp = d - d_\parallel$ and the $d$-dimensional norm of a vector $\underline{v} = (v_1, \ldots, v_d)$ is defined as $\|\underline{v}\|_d^2 = \sum_{i=1}^{d} v_i^2 / d$, in order to remove the dimensional bias from the ratio.

## 2.4 Feature regime vs lazy regime

Throughout this work, it is assumed that the network width $h$ is sufficiently large for the algorithm to operate in the overparametrized regime [17, 18]. We define the Neural Tangent Kernel (NTK) $\Theta(\underline{x}_1, \underline{x}_2) = \partial_W f(\underline{x}_1) \cdot \partial_W f(\underline{x}_2)$, where the scalar product runs over all weights of the network. The gradient descent evolution (Eq. (2)) on the functional space then reads

$$\dot{f}(\underline{x}) = \frac{1}{p} \sum_{\mu=1}^{p} \Theta(\underline{x}, \underline{x}^\mu) \, y(\underline{x}_\parallel^\mu) \, l' \left[ y(\underline{x}_\parallel^\mu) F(\underline{x}^\mu) \right], \tag{4}$$

where the NTK can in principle evolve over time.

At initialization, the predictor function is zero. It then grows to fit the training set and doesn't stop until it is at least equal to one on all training points. The smaller the network scale $\alpha$ the more the weights need to evolve.

If $\alpha \gg 1$, the condition $F(\underline{x}) \sim 1$ can be fulfilled with infinitesimal weight increments $\delta W$, so that the predictor function is linear in $\delta W$. The dynamics thus reduces to a kernel method [9], meaning that the NTK is frozen to its initial state $\Theta_0$. For an isotropic distribution of the weights, the kernel $\Theta_0$ is isotropic and thus blind to the existence of many invariants in the data to be learned. This regime is coined the lazy regime for finite $h$ or the NTK regime if $h \to \infty$.

If $\alpha \ll 1$, the weights of the network need to evolve significantly in order to satisfy the condition $F(\underline{x}) \sim 1$ [19]. In that case, the NTK adapts to the data and we shall show that it becomes more and more sensitive to the informative directions. In particular, the first layer weights $\underline{\omega}_n$ aligns toward the informative linear subspace, as shown in Fig. 2 for the stripe model and in Fig. 7 for the cylinder model. This regime is coined the feature regime (or sometimes the rich regime) and we study it in the limit $h \to \infty$.

The transition between the two regimes is illustrated in Appendix B by learning the stripe model with different values of $\alpha$.

## 2.5 Learning timescales

We now give a general overview of the network evolution in time. We define the characteristic time $t^\star$ as the time when the predictor function first becomes of order one. Also, we introduce the neuron vector $\underline{z} = -\sqrt{d}b\underline{\omega}/\|\underline{\omega}\|^2$, which localizes the closest point of the ReLU hyperplane to the origin. We drop the neuron index for simplicity of notation. In the feature regime, we identify three temporal regimes:

○ Compressing regime: Before $t^\star$, all neuron vectors $\underline{z}$ converge toward a finite number of fixed points that we generically call $\underline{z}^\star$ [35]. We shall see that the individual weights all diverge exponentially with a time constant $\tau^\star \sim \tau = h\sqrt{d}/2$, which depends on the fixed point. As a consequence, at $t^\star$, the predictor function scales as $e^{t^\star/\tau}\alpha \sim 1$. In the mean field limit ($\alpha \to 0$), the characteristic time $t^\star \sim \tau \log(1/\alpha)$ thus diverges and all neurons effectively reach their fixed point. The logarithmic scaling of $t^\star$ is verified numerically in Appendix B.

  In the limit of infinite training set size ($p \to \infty$), all fixed points are located on the informative subspace, namely $\underline{z}^\star = (\underline{z}_\parallel^\star, \underline{0})$. We quantify this compression with the amplification factor $\lambda = \|\underline{z}_\parallel\|/\|\underline{z}_\perp\|$ which is divergent in this limit. For finite $p$, the compression is saturated by finite size effects: the data distribution is subject to fluctuations of the order of $1/\sqrt{p}$ compared to its population expectation. The fixed points are thus located at a distance of the order $\mathcal{O}(1/\sqrt{p})$ perpendicular to the informative subspace. In other words, as we show below the amplification factor saturates at $\lambda \sim \sqrt{p}$.

○ Fitting regime: After $t^\star$, a finite fraction of the training points satisfy the condition $y^\mu F(\underline{x}^\mu) > 1$. Because we consider the hinge loss these training points no longer contribute to the network evolution. In particular, they drop out of the sum in Eq. (2) [36]. The first points to be excluded are the furthest from the interfaces separating distinct labels. During this process, the fixed points move within the informative manifold such as to better fit the data. Relative fluctuations are still of order $\mathcal{O}(1/\sqrt{p})$, thus one expects the amplification factor to remain of the same order $\lambda \sim \sqrt{p}$, as we confirm empirically.

○ Over-fitting regime: When the number of points still remaining in the sum of Eq. (2) is of the order of one, the sum is dominated by fluctuations and the network overfits the remaining constraints. We check numerically that the previous predictions are not significantly altered during this final regime, which we don't study theoretically.

The neuron compression mechanism scales up to the whole network so that the global amplification factor also saturates with the fluctuations, namely $\Lambda \sim \sqrt{p}$. We expect this scaling to be a general property of linear invariant problems. In the next section, we describe this process in more details for the stripe model.

## 3 Stripe model

We consider the simplest model of linear invariant data, where the label function only depends on $d_\parallel = 1$ informative direction, namely $y(\underline{x}) = y(x_1)$. Layers of $y = +1$ and $y = -1$ regions alternate along the direction $\underline{e}_1$, separated by parallel planes. In particular, we define the single-stripe model, where the labels are negative if $x_{\min} < x_1 < x_{\max}$ and positive otherwise. In our numerical simulations, we use this model with the parameters $x_{\min} = -0.3$ and $x_{\max} = 1.18549$[2].

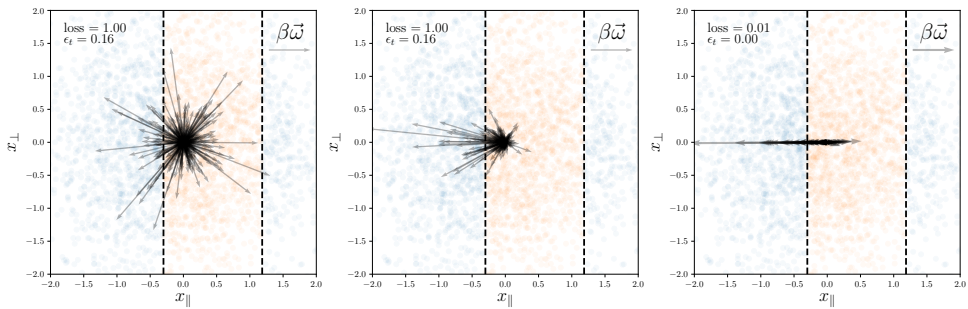

Figure 2: Representation of the weights alignment in the **single-stripe model**. An instance of the labelled training set is shown in the background. The arrows represent the quantity $\beta\underline{\omega}$ – properly rescaled to fit the figure – for a random subset of neurons. <u>Left</u>: At initialization, the weights are distributed isotropically. In the lazy regime, the same distribution persists during the training. <u>Center</u> / <u>Right</u>: During learning, in the feature regime, the weights tend to align with the direction $\underline{e}_1$. An animation of the current figure can be found at git.io/JJTS9.

### 3.1 Learning curves

We compare the lazy regime and the feature regime, by computing their respective learning curves, namely the test error vs the training set size $p$. Fig. 3 illustrates how the feature regime outperforms the lazy regime, when applied on the single-stripe model.

In the lazy regime, the algorithm reduces to a kernel method and one can rely on [24] to predict the learning curve exponent $\beta$. In that work, it is shown that for an isotropic kernel of bandwidth larger than the distance between nearest neighbours of the training set, the learning curve of the Support Vector Classifier (SVC) algorithm applied to the stripe model in dimension $d$ scales as $\epsilon \sim p^{-\beta}$, with $\beta = (d-1+\xi)/(3d-3+\xi)$, where $\xi \in (0, 2)$ is an exponent characterizing the kernel cusp at the origin. The NTK is isotropic on data lying on the sphere, has a bandwidth of order $\mathcal{O}(1)$ and its cusp is similar to the one of a Laplace kernel, namely $\xi = 1$. Hence, as the SVC algorithm minimizes the hinge loss, the learning curve of the lazy regime is expected to have an exponent $\beta_{\text{Lazy}} = d/(3d-2)$. This prediction is tested on Fig. 3.

In the same work, it is shown that if the uninformative directions of the data are compressed by a factor $\Lambda$, namely $\underline{x}_\perp \to \underline{x}_\perp/\Lambda$, the test error is improved by a factor $\Lambda^{-\frac{2(d-1)}{3d-2}}$ for $\xi = 1$. In the next section, we shall argue that, in the feature regime, the perpendicular weights $\underline{\omega}_{n,\perp}$ are suppressed by a factor $\sqrt{p}$ compared to the informative weights $\omega_{n,1}$ as their growth is governed by fluctuations of the data. Such a weight compression acts similarly as a data compression with $\Lambda \sim \sqrt{p}$ as depicted on Fig. 1. Assuming that the main effect of feature learning is this compression, we expect the learning curve exponent of the feature regime to be $\beta_{\text{Feature}} = (2d-1)/(3d-2)$. This scaling is again consistent with the numerical results of Fig. 3.

---

[2]The value $x_{\max} = \sqrt{2}\,\mathrm{erf}^{-1}(1 + \mathrm{erf}(x_{\min})) \approx 1.18549$ is chosen so that the two labels are equiprobable.

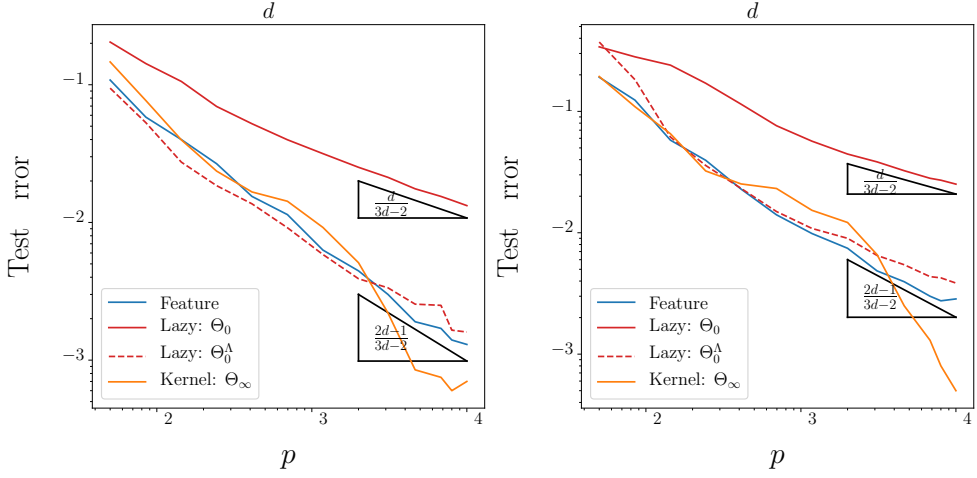

Figure 3: Test error vs the training set size $p$ for the **single-stripe model** in dimension $d = 5$ and $d = 10$. Two datasets are considered: points drawn from the standard normal distribution in dimension $d$ and its compression, where $\underline{x}_\perp \to \underline{x}_\perp / \Lambda^\star$, where $\Lambda^\star \sim \sqrt{p}$ is the global amplification factor at $t^\star$ (see Section 3.2 for the definitions). The labels are defined according to the single-stripe model with $x_{\min} = -0.3$ and $x_{\max} = 1.18549$. The task is learned following the dynamics of Eq. (2). In the feature regime (solid blue lines), the network scale is set to $\alpha = 10^{-6}$. In the lazy regime, learning is performed with a frozen Gram matrix ($\alpha \to \infty$), computed at initialization for both the original (solid red lines) and compressed (dashed red lines) datasets. The performance of the frozen Gram matrix at the end of feature training is also computed (solid orange lines). All results correspond to the median over 20 realizations of both the data distribution and the network initialization. The benchmark triangles represent the expected power laws.

## 3.2 Amplification effect

In this section we show that when learning the stripe model in the feature regime, the first layer weights align along the informative direction $\underline{e}_1$. In particular, we show that the ratio between the informative (or parallel) weights $\omega_{n,1}$ and the uninformative (or perpendicular) weights $\underline{\omega}_{n,\perp}$ scales as $\Lambda \sim \sqrt{p}$. This section being more technical can be skipped at first reading. For the interested reader, details are given in Appendix C.

### 3.2.1 Neuronal dynamics

We first consider the dynamics of a single generic neuron, whose dynamics is obtained from Eq. (2):

$$\dot{\underline{\omega}} = \frac{1}{hp} \sum_{\mu=1}^{p} \sigma' \left[ \frac{\underline{\omega} \cdot \underline{x}^\mu}{\sqrt{d}} + b \right] l' \left[ y^\mu F(\underline{x}^\mu) \right] \beta \frac{\underline{x}^\mu}{\sqrt{d}} y^\mu$$

$$\dot{b} = \frac{1}{hp} \sum_{\mu=1}^{p} \sigma' \left[ \frac{\underline{\omega} \cdot \underline{x}^\mu}{\sqrt{d}} + b \right] l' \left[ y^\mu F(\underline{x}^\mu) \right] \beta \, y^\mu ,$$

$$\dot{\beta} = \frac{1}{hp} \sum_{\mu=1}^{p} \sigma \left[ \frac{\underline{\omega} \cdot \underline{x}^\mu}{\sqrt{d}} + b \right] l' \left[ y^\mu F(\underline{x}^\mu) \right] y^\mu$$

(5)

where the neuron index $n$ is dropped. Because the ReLU activation is homogeneous, $\sigma(x) = x\sigma'(x)$, the equality $\beta\dot{\beta} - \underline{\omega} \cdot \dot{\underline{\omega}} - b\dot{b} = 0$ holds during the whole evolution. Following the discussion of Section 2.5, we now solve the above system in the limit $\alpha \to 0$, so that $t^\star \sim \tau \log(1/\alpha) \to \infty$. In the numerical experiments, we choose the network scale $\alpha = 10^{-6}$ and define $t^\star$ as the time when 10% of the training set satisfies the condition $y^\mu F(\underline{x}^\mu) > 1$.

**Compressing regime** As long as $t \ll t^\star$, the quantity $l'[y^\mu F(\underline{x}^\mu)] = 1$, $\forall \mu$, so that the system (5) only depends on the weights associated to the considered neuron. Each neuron thus evolves independently and only differs from the other neurons by its initial conditions.

We first consider the limit $p \to \infty$ and neglect the finite size effects. Applying the central-limit theorem, we carry out the integration over the perpendicular space in Appendix C.1. Defining the neuron amplification factor $\lambda = \omega_1/\omega_\perp$, where $\omega_\perp = \|\underline{\omega}_\perp\|$, the neuronal dynamics (5) becomes

$$
\begin{aligned}
\dot{\omega}_1 &= \frac{\beta}{\tau} \left\langle y(x_1)\, x_1\, g_\lambda(x_1 - \zeta_1) \right\rangle_{x_1} + \mathcal{O}\left(p^{-1/2}\right) \\
\dot{\underline{\omega}}_\perp &= \frac{\beta}{\tau} \frac{e^{-d b^2/2\omega^2}}{\sqrt{2\pi}} \frac{\omega_\perp}{\omega} \left\langle y\left(\frac{\omega_\perp}{\omega} x_1 + \frac{\omega_1^2}{\omega^2}\zeta_1\right)\right\rangle_{x_1} + \mathcal{O}\left(p^{-1/2}\right), \\
\dot{b} &= \frac{\sqrt{d}\beta}{\tau} \left\langle y(x_1)\, g_\lambda(x_1 - \zeta_1) \right\rangle_{x_1} + \mathcal{O}\left(p^{-1/2}\right)
\end{aligned}
\tag{6}
$$

where $\tau = h\sqrt{d/2}$, $\omega = \|\underline{\omega}\|$ and $\zeta_1 = -\sqrt{d}b/\omega_1$ is the intercept of the ReLU hyperplane with the $\underline{e}_1$ axis, while $g_\lambda(x) = \frac{1}{2}\left(1 + \mathrm{erf}(\lambda x/\sqrt{2})\right)$. The notation $\langle \cdot \rangle_{x_1}$ refers to the expectation over the Gaussian variable $x_1$.

We recall the definition of the neuron vector $\underline{z} = -\sqrt{d}b\underline{\omega}/\omega^2$. In [35], the authors show that the first layer weights of a one-hidden layer network of ReLU activation tend to align along a finite number of directions depending only on the dataset. Relying on the symmetries of the model, we seek solutions on the informative axis. We thus make the hypothesis that the fixed points are of the form $\underline{z}^\star = (z^\star, \underline{0})$, where $z^\star = z_1^\star = \zeta_1^\star$, which is equivalent to assuming that the amplification factor associated to such fixed points is diverging. In this limit, the system (6) simplifies: the expectation values only depend on the parameter $\zeta_1$ and the sign of $\lambda$. We respectively call them $C_1^\pm(\zeta_1)$, $C_\perp^\pm(\zeta_1)$ and $C_b^\pm(\zeta_1)$. As a consequence, the dynamics of $\zeta_1$,

$$
\dot{\zeta}_1 \xrightarrow{\lambda \to \pm\infty} -\frac{1}{\tau}\frac{\beta}{\omega_1}[d\, C_b^\pm(\zeta_1) + \zeta_1\, C_1^\pm(\zeta_1)],
\tag{7}
$$

yields the location of the fixed points as they lie where the above bracket vanishes. For the fixed points to be stable along the $\underline{e}_1$ axis, the second derivative of $\zeta_1$ needs to be negative. On a given fixed point $z^\star$ the expectation values $C_1^\pm(z^\star)$, $C_\perp^\pm(z^\star)$ and $C_b^\pm(z^\star)$ are constant and it is straight-forward to see that $\omega_1$, $b$ and $\beta$ all diverge exponentially with a time constant $\tau^\star \sim \tau$ given in Appendix C.2. Finally, we verify in Appendix C.2 that the perpendicular weights do not diverge as fast as $\omega_1$ as long as $\lambda C_1^\pm(z^\star)y(z^\star) < 0$ or $\sqrt{2\pi}|C_1^\pm(z^\star)| - e^{-z^{\star 2}/2} > 0$. Under these conditions, the amplification factor $\lambda$ thus diverges exponentially in time which justifies our initial hypothesis. We checked numerically that these conditions indeed hold for the considered models. The panel **b** of Fig. 4 illustrates $\dot{\zeta}$ for the single-stripe model in $d = 2$.

We now consider the finite $p$ corrections to a given fixed point $\underline{z}^\star$ and show that the amplification factor saturates at $\lambda^\star \sim \sqrt{p}$. The finite $p$ effects lead to an additional fluctuation term in each equation of the system (5). This correction is negligible for the dynamics of $\omega_1$, $b$ and $\beta$, however for the perpendicular weights it yields

$$
\dot{\underline{\omega}}_\perp = \frac{\beta}{\tau}\left[\frac{e^{-z^{\star 2}/2}}{\sqrt{2\pi}} \frac{\omega_\perp}{\omega} C_\perp^\pm(z^\star) + \frac{\underline{N}(z^\star)}{\sqrt{p}} D_\perp^\pm(z^\star)\right],
\tag{8}
$$

where $D_\perp^\pm(z^\star) = \langle \Theta(\pm(x_1 - z^\star)) \rangle_{x_1}$ and $\underline{N}(\underline{z}^\star)$ is a vector of random variables of variance one (see Appendix C.3). The first term in the above bracket is proportional to $1/\lambda$ and thus vanishes exponentially with time until it is of the order of the second term, namely $\mathcal{O}(1/\sqrt{p})$. We call $\tau_\perp$ the time when this crossover occurs. After $\tau_\perp$, $\dot{\underline{\omega}}_\perp$ is merely proportional to $\beta/\sqrt{p}$. Therefore, the perpendicular weights follow the same exponential growth as the other weights up to a $\mathcal{O}(1/\sqrt{p})$ prefactor and the amplification factor converges to a finite value $\lambda^\star$ that scales as

$$
\lambda^\star \sim \sqrt{p}.
\tag{9}
$$

We test numerically that all neurons converge to one of the above described fixed points by considering the single-stripe model. The panel **c** of Fig. 4 illustrates the trajectories of a random selection of neurons while training the network Eq. (1) until $t^\star$. Note that some neurons may not have yet reached a fixed point for two reasons. First, because $p$ is finite, a neuron initial position may lie too far from the training set domain. If no training point lies within the positive side of its associated ReLU hyperplane, it won't feel any gradient and will thus remain static. Second, the simulation is run with a finite network scale ($\alpha = 10^{-6}$), implying that the time $t^\star \sim \tau \log(1/\alpha)$ is also finite. Hence, some neurons may not have reached their asymptotic regime at $t^{\star 3}$.

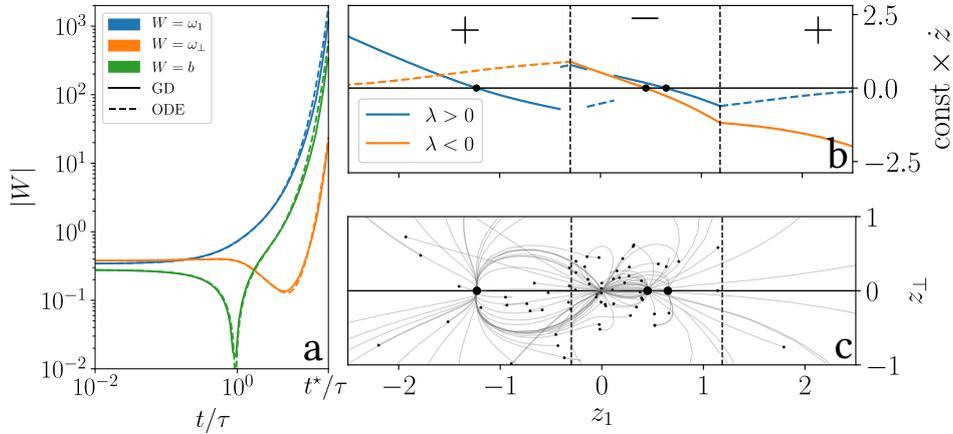

Figure 4: Numerical analysis of the **single-stripe model** with $x_{\min} = -0.3$ and $x_{\max} = 1.18549$ in dimension $d = 2$ with a training set of size $p = 10000$. The location of the two interfaces is illustrated by the vertical dashed lines. **(a)** Temporal evolution of the weights of a randomly chosen neuron. The solid lines illustrate the considered neuron dynamics in the neural network, while the dashed lines correspond to the numerical solutions of the ODE Eq. (6) and Eq. (8) obtained for the same initial conditions. The random variables $\underline{N}(\underline{z}^\star)$ is computed numerically. The curves are truncated at the time $t^\star$. **(b)** Function defining the location of the fixed points along $\zeta_1$ in the limit $\lambda \to \infty$. The two scenarios $\lambda > 0$ and $\lambda < 0$ are shown. The unstable regions, where the limit $\lambda \to \infty$ is inconsistent are represented with dashed lines. **(c)** Selection of neuronal trajectories in the $\underline{z}$-plane for $t < t^\star$. The small black dots mark the location of the initial conditions, while the large black dots lie on the predicted location of the three attractors of the compressing regime.

**Fitting regime** After $t^\star$, the loss derivative is zero on a finite fraction of the training set. As discussed in Section 2.5, these training points no longer contribute to the network dynamics. This long time evolution is beyond the scope of this work, but could be solved numerically in the limit $p \to \infty$ following the work of [36]. It requires to compute the network function at each step in order to decide which training points still contribute to the dynamics.

In this regime, the neurons are still sparsely distributed on the same number of fixed points [35] as in the previous regime. The location of the fixed points is however changing to fit the stripe. This process is shown on Fig. 5 for the stripe model in $d = 2$. Concerning the amplification factor, the $\sqrt{p}$ suppression of $\dot{\omega}_\perp$ compared to $\dot{\omega}_1$ remains true until the effective number of training points contributing to the dynamics becomes of order $\mathcal{O}(1)$, as shown on the top panel of Fig. 6.

---

[3]Because $\lambda$ initially grows exponentially, the definition of the perpendicular timescale yields $\tau_\perp \sim \tau^\star \log p$. For the amplification factor to reach its plateau $\lambda^\star \sim \sqrt{p}$ during the compressing regime, it is essential that $\tau_\perp < t^\star \sim \tau \log(1/\alpha)$. Hence the larger the training set size, the smaller $\alpha$ needs to be.

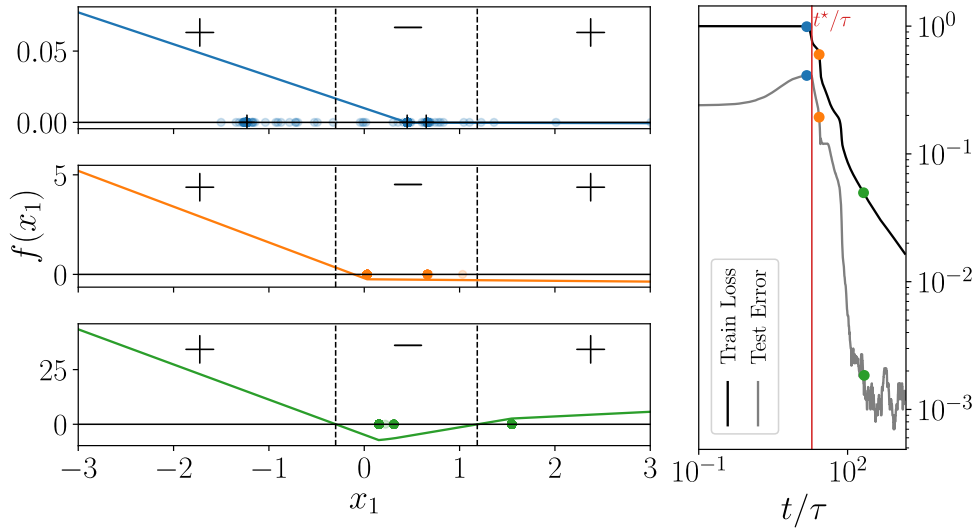

Figure 5: Evolution of the network while fitting the **single-stripe model** in $d = 2$ with a training set of size $p = 10000$. <u>Left</u>: Neural network decision function along the informative direction $\underline{e}_1$ at three different times. The $\zeta_1$ variable of each neuron is represented on the $x$-axis by colored dots. The darker the region, the larger the point density. On the top plot, the location of the predicted fixed points is marked by black crosses. The location of the two interfaces is illustrated by the vertical dashed lines. <u>Right</u>: Train loss and test error vs time. The three times considered on the left plot are indicated with the same color code. The characteristic time $t^\star$ is represented by the vertical red line.

### 3.2.2 Global amplification factor

In the previous discussion, we defined an amplification factor $\lambda = \omega_1/\omega_\perp$ for each neuron of the network. Following the definition Eq. (2.3) we now consider the global amplification factor $\Lambda$ averaged over all neurons, namely

$$\Lambda^2 = (d-1)\frac{\sum_{n=1}^{h} \omega_{n,1}^2}{\sum_{n=1}^{h} \omega_{n,\perp}^2}. \tag{10}$$

This definition compares the largest parallel weights to the largest perpendicular weights. The prefactor guarantees that $\Lambda(t = 0) = 1$. The top panel of Fig. 6 shows the exponential growth of $\Lambda$ toward the plateau at $\Lambda^\star = \Lambda(t^\star)$. The longer time evolution is subject to fluctuations but doesn't alter significantly the picture. On the bottom panel, we confirm the predicted scaling $\Lambda^\star \sim \sqrt{p}$. We also show that the same scaling applies to the maximum of the global amplification factor, $\Lambda_{\max} = \max_t \Lambda(t)$, which occurs during the fitting regime.

In this section we illustrated with a particular example how the neurons of the network converge to a finite set of fixed points. The associated amplification factors are shown to diverge with the dataset size: $\lambda \sim \sqrt{p}$. At the network scale this effect is equivalent to a data compression of the same amplitude. In the next section we extend this discussion to other linear invariant datasets.

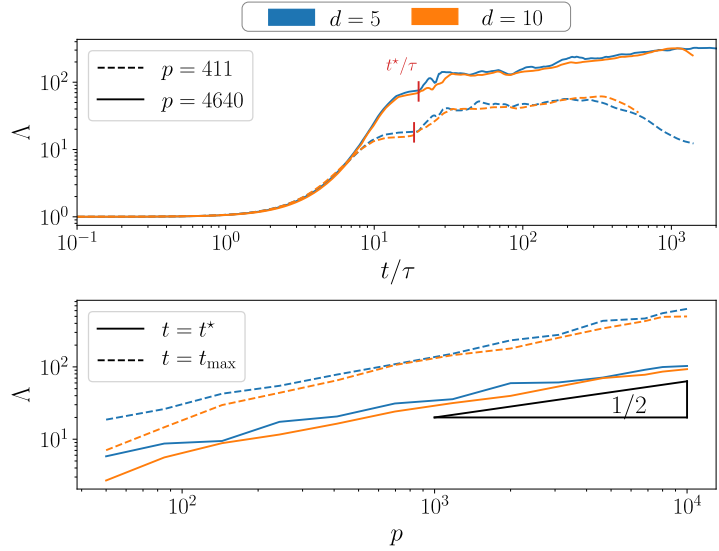

Figure 6: <u>Top</u>: Temporal evolution of the global amplification factor while learning the **single-stripe model**. Two dimensions $d = 5$ and $d = 10$ and two training set sizes $p = 411$ and $p = 4640$ are illustrated. The curves are averaged over 20 realizations of both the data distribution and the network initialization. The red vertical ticks mark the averaged critical time of the associated setup. <u>Bottom</u>: Global amplification factor vs the size of the training set $p$ for the **single-stripe model** in dimensions $d = 5$ and $d = 10$. Both the amplification factor $\Lambda^\star$ computed at $t^\star$ and the maximal amplification factor $\Lambda_{\max}$ are displayed. The curves correspond to the median over 20 realizations of both the data distribution and the network initialization. The benchmark triangle of slope $1/2$ confirms our scaling predictions for $\Lambda^\star$ and $\Lambda_{\max}$.

## 4 Generalization and cylinder model

**Compression mechanism** The compression mechanism illustrated in the stripe model is expected to occur generically in linear invariant models. If the label function were to depend on $d_\parallel$ directions, all neuron vectors $\underline{z}$ would converge toward fixed points located in the informative subspace of dimension $d_\parallel$. Similar finite $p$ effects as in the stripe model would saturate the resolution of the informative subspace, so that the informative weights $\underline{\omega}_\parallel$ would be larger than the perpendicular weights $\underline{\omega}_\perp$ by an amplification factor $\lambda \sim \sqrt{p}$.

**Advantage of feature regime** As the NTK is blind to the existence of invariants in the data, the performance of the lazy regime should not depend on $d_\parallel$. Indeed following the results of [24], the lazy regime learning curve follows an exponent $\beta_{\text{Lazy}} = d/(3d-2)$ for simple boundaries separating labels (such as plane, spheres or cylinders), a result conjectured to hold more generally for sufficiently smooth boundaries. The correspondence between the lazy training and the SVC considered in [24] is discussed in Section 3.1.

In [24], it is also shown that for linear invariant models with $d_\parallel > 1$, a compression of the perpendicular space by a factor $\Lambda$, $\underline{x}_\perp \to \underline{x}_\perp/\Lambda$, improves the performance of the SVC by a factor $\Lambda^{-d_\perp/(3d-2)}$, for a kernel of exponent $\xi = 1$. As discussed in Section 3.1, because in the feature regime such a compression occurs with $\Lambda \sim \sqrt{p}$, we expect the learning curve exponent of the feature regime to be $\beta_{\text{Feature}} = (d + d_\perp/2)/(3d-2)$.

**Cylinder model** We test our predictions by considering a cylinder model in $d = 3$. The data points are drawn from the standard normal distribution: $\underline{x} \sim \mathcal{N}(0, I_d)$, while the label function is a circle in the informative subspace of dimension $d_\parallel = 2$, namely $y(\underline{x}) = y(\|\underline{x}_\parallel\|) = +1$ if $\|\underline{x}_\parallel\| > R$ and negative otherwise. For the numerical simulations we use $R = 1.1774$[4]. We learn this model following the gradient descent algorithm described in Section 2.2.

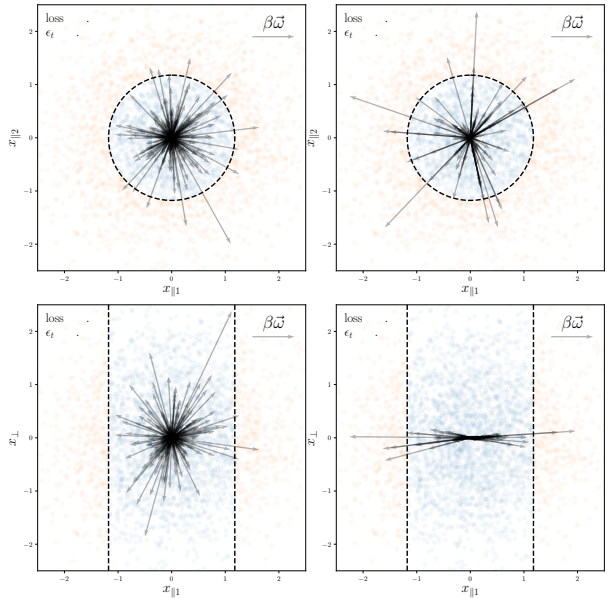

Figure 7: Representation of the amplification effect in the **cylinder model** with $d = 3$ and $d_\parallel = 2$. An instance of the labelled training set is shown in the background. The arrows represent the quantity $\beta \underline{\omega}$ – properly rescaled to fit the figure – for a random subset of neurons. We show the $x_\perp = 0$ section (first row) and the $x_{\parallel 2} = 0$ section (second row) of data-space. The first column reports the weights distribution at initialization, the second column at the end of training. An animated version of the current figure can be found at git.io/JJTS9.

The compression of the weight vectors $\underline{\omega}$ into the informative subspace displayed on Fig. 7 supports the previous general discussion. Also, we verify both the scaling of the amplification factor and the scaling of the learning curves on Fig. 8. As in the stripe model the time $t^\star$ is numerically defined as the time when the equality $y^\mu F(\underline{x}^\mu) > 1$ first holds for 10% of the training set. On the top panel, both the global amplification factor at $t^\star$ and the maximal global amplification factor are shown to scale as $\Lambda^\star \sim \Lambda_{\max} \sim \sqrt{p}$. The advantage of the feature regime over the lazy regime is displayed on the bottom panel. In particular, the predicted learning curve exponents $\beta_{\text{Lazy}} = d/(3d - 2) = 3/7$ and $\beta_{\text{Feature}} = (d + d_\perp/2)/(3d - 2) = 1/2$ are shown to be consistent with the numerical results.

---

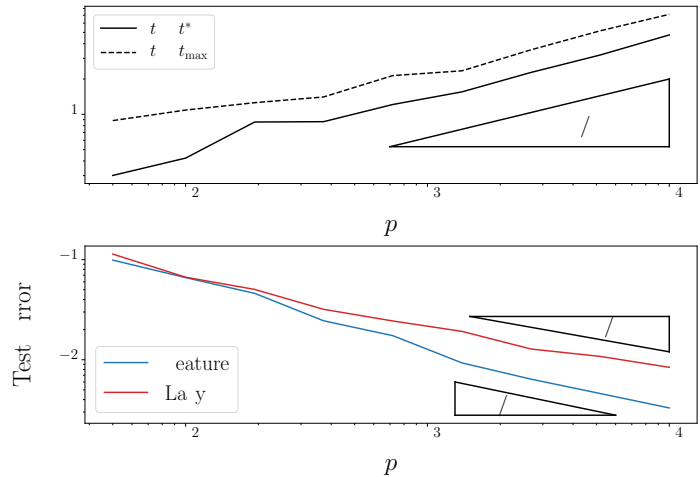

Figure 8:  <u>Top:</u> Global amplification factor vs the size of the training set $p$ for the **cylinder model** in dimension $d = 3$. Both the amplification factor $\Lambda^\star$ computed at $t^\star$ and the maximal amplification factor $\Lambda_{\max}$ are displayed. The curves correspond to the median over 49 realizations of both the data distribution and the network initialization. The benchmark triangle illustrates the expected power law. <u>Bottom:</u> Test error vs the training set size $p$ for the **cylinder model** in $d = 3$. In the feature regime (blue line), the network scale is set to $\alpha = 10^{-6}$. In the lazy regime (red line), learning is performed with the frozen Gram matrix computed at initialization ($\alpha \to \infty$). The curves correspond to the median over 9 realizations of both the data distribution and the network initialization. The benchmark triangles illustrate the power law predictions.

## 5  Signatures of compression in the temporal evolution of the NTK

Previous empirical studies of compression of uninformative directions in data space in neural nets have focused on the neural representations of the data layer by layer [5, 8]. Here instead we study how compression affects the evolution of the NTK as learning takes place, and show how this kernel becomes better suited for the considered task. We start from the stripe model and extend our analysis to a CNN trained on MNIST, and find striking similarities between the two cases.

### 5.1  Neural Tangent Kernel Principal Components

**General facts**  The neural tangent kernel reads $\Theta(\underline{x}, \underline{z}) = \psi(\underline{x}) \cdot \psi(\underline{z})$ where $\psi(\underline{x})$ is a vector of $N$ components $\psi_W(\underline{x}) := \partial_W f(\underline{x})$ and $W$ is one of the $N$ parameters of the model. The kernel can be expressed in terms of its eigenvalues and eigenfunctions (Mercer's Theorem) $\Theta(\underline{x}, \underline{z}) = \sum_\lambda \lambda \, \phi_\lambda(\underline{x}) \phi_\lambda(\underline{z})$. The functions $\phi_\lambda(\cdot)$ form an orthogonal basis on the space of functions, and satisfy the integral equation [37] $\int \Theta(\underline{x}, \underline{z}) \phi_\lambda(\underline{z}) \rho(\underline{z}) d\underline{z} = \lambda \phi_\lambda(\underline{x})$ where $\rho(\cdot)$ is the distribution of the data. In general, a kernel is expected to perform well if the RKHS norm $\|y\|_\theta$ of the function $y(\underline{x})$ being learnt is small [38]. It writes $\|y\|_\theta^2 = \sum_\lambda \omega_\lambda^2 / \lambda$ where $\omega_\lambda := \int y(\underline{x}) \phi_\lambda(\underline{x}) \rho(\underline{x}) d\underline{x}$. Thus, a kernel performs better if the large coefficients $\omega_\lambda$ in the eigenbasis of the kernel correspond to large $\lambda$. We will argue below that such a trend is enforced when the NTK evolves by compressing uninformative directions.

In practice, for a finite training set $\{\underline{x}^\mu\}_{\mu=1}^p$ of size $p$, the Gram matrix $K$ is accessible empirically. It is defined as the $p \times p$ matrix of scalar products $K_{\mu\nu} = \psi(\underline{x}^\mu) \cdot \psi(\underline{x}^\nu)$. Diagonalizing it corresponds to performing Kernel PCA [38], which identifies the principal components in the feature representation $\psi(\underline{x}^\mu)$ of the data: $K_{\mu\nu} = \sum_{\tilde{\lambda}} \tilde{\lambda} \, \tilde{\phi}_{\tilde{\lambda}}(\underline{x}^\mu) \tilde{\phi}_{\tilde{\lambda}}(\underline{x}^\nu)$. One has $\tilde{\lambda} \to \lambda$ and $\tilde{\phi}_{\tilde{\lambda}}(\underline{x}^\nu) \to \phi_\lambda(\underline{x}^\nu)$ as $p \to \infty$ for a fixed $\lambda$. Thus the coefficients $\omega_\lambda$ can be estimated as $\tilde{\omega}_\lambda := \frac{1}{p} \sum_{\mu=1...p} \tilde{\phi}_\lambda(\underline{x}^\mu) y(\underline{x}^\mu)$. In the following sections, we drop the tilde for ease of notation.

**Effect of compression on the evolution of the NTK**    At initialization, for fully connected nets the NTK is isotropic, and its eigenvectors are spherical harmonics [9]. For a fixed dimension $d_\parallel$ of the informative space, as the overall dimension $d$ grows, the value of a given spherical harmonics leads to vanishing information on the specific components $\underline{x}_\parallel$. As a consequence, we expect that even for large $\lambda$, $\phi_\lambda(\underline{x})$ contains little information on the label $y(\underline{x})$. It follows that the magnitude of the projected signal $\omega_\lambda$ is small in that limit.

By contrast, after learning in the limit $\Lambda \sim \sqrt{p} \to \infty$, the output function looses its dependence on the orthogonal space $\underline{x}_\perp$. The NTK can then generically be rewritten as:

$$\Theta(\underline{x}, \underline{z}) = \Theta_1(\underline{x}_\parallel, \underline{z}_\parallel) + \Theta_2(\underline{x}_\parallel, \underline{z}_\parallel)\underline{x}_\perp \cdot \underline{z}_\perp \tag{11}$$

where the second term comes from the derivative with respect to the first layer of weights (see Appendix D). For a Gaussian data density $\rho$ considered in this paper, eigenvectors with non-vanishing eigenvalues are then of two kind: $\phi_\lambda^1(\underline{x}_\parallel)$ – the eigenvectors of $\Theta_1$ – and $\phi_\lambda^2(\underline{x}_\parallel)\underline{u} \cdot \underline{x}_\perp$ where $\phi_\lambda^2(\underline{x}_\parallel)$ is an eigenvector of $\Theta_2$ and $\underline{u}$ any non-zero vectors. The null-space of the kernel then corresponds to all functions of the orthogonal space that are orthogonal to constant or linear functions. However for a finite $\Lambda$, we expect the associated eigenvalues to be small but different from zero.

Two qualitative predictions follow:

- The eigenvectors $\phi_\lambda^1$ only depend on $\underline{x}_\parallel$ and are thus generically more informative on the label $y(\underline{x}_\parallel)$ than spherical harmonics. It is also true, but to a lesser extent, for the eigenvectors $\phi_\lambda^2(\underline{x}_\parallel)\underline{u} \cdot \underline{x}_\perp$. Indeed for Gaussian data, they can be considered as a function of $\underline{x}_\parallel$ times a random Gaussian noise. Overall, we thus expect that for large eigenvalues the mutual information between $\phi_\lambda(\underline{x})$ and $y(\underline{x})$ to increase during learning.

- As a consequence, the magnitude of $\omega_\lambda$ associated to the top eigenvalues also tends to increase. We thus expect that the performance of kernel learning using the NTK at the end of training to be superior to that using the NTK at initialization.

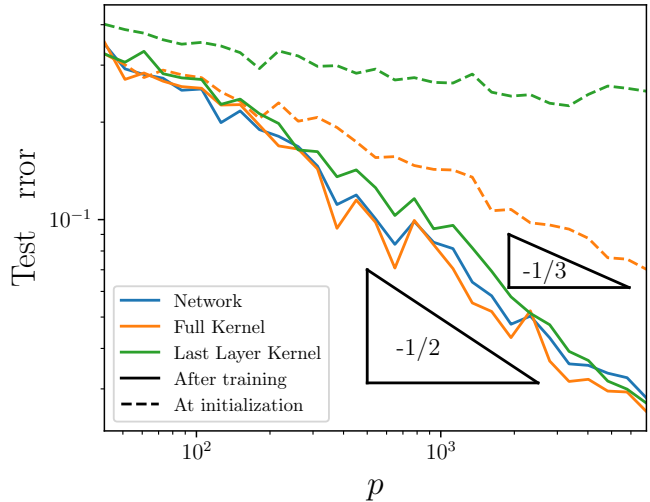

Figure 9: Performance of CNN trained to classify the parity of MNIST digits (binary class problem) as a function of the trainset size. The network is trained in the feature learning regime (blue line) using vanilla gradient descent with momentum, leading to $\beta_{\text{Feature}} \approx 0.5$. Before and after training the full kernel (i.e. with respect to all the parameters) and the kernel of the last layer are computed. These four frozen kernels are then used in a gradient descent algorithm using an independent trainset of the same size. All the measures are done 5 times with different initialization seeds and averaged. For the full kernel at initialization (dashed orange) we find $\beta_{\text{Lazy}} \approx 0.3$ and consequently $\beta_{\text{Lazy}} < \beta_{\text{Feature}}$.

### 5.2 Empirical tests

**Performance of kernel methods based on the NTK** In Fig.3 we test our prediction that kernel methods based on the NTK obtained at the end of training outperforms the NTK at initialization. We perform kernel learning using different data for the training set than those used to generate the NTK. We find that it is indeed the case: in fact, performance is found to be very similar to that of the neural net in the feature learning regime, except for the largest training set size where it even outperforms it. Note that this similarity is natural, since the features associated to the NTK contain the the last hidden layer of neurons, which can represent the network output with the last layer of weights.

We test the generality of this result in Fig.9 using a more modern CNN architecture on the MNIST data set. This architecture is inspired from MnasNet [39] with 16 convolutional layers. It distinguishes from MnasNet by the absence of batch-normalization. We again find that kernel methods based on the NTK at infinite time perform as well as the network in the feature learning regime, and even once again slightly better for the largest $p$.

Finally, it is interesting to compare this analysis with the kernel whose features correspond to the last layer of hidden neurons at the end of training. Training such a kernel simply corresponds to retraining the last layer of weights while fixing the activity of the last hidden neurons. Interestingly, this kernel performs well but generally less so than the network itself, as illustrated in Fig.9.

**Kernel PCA *v.s.* labels (Information and projection)** We now confirm that such improved performance of the NTK corresponds to the top kernel principal components becoming more informative on the task. As we argued in Section 5.1, we expected this to be the case, in the presence of compression. Specifically, we consider the $r$ largest eigenvalues $\lambda_{\max}, \ldots \lambda_r$ of the NTK Gram Matrix and their corresponding eigenvectors. We first compute the mutual information between a given eigenvector magnitude and the label $I(\phi_{\lambda_r}; y)$ – for details on the estimator see Appendix E. This mutual information is small and essentially independent of $r$ in the range studied for the NTK at initialization; both for the stripe model (Fig.10.a) and MNIST (Fig.10.c). However, at the end of learning, mutual information has greatly improved in both cases, a fact that holds true for the NTK and for the kernel obtained from the last layer of hidden neurons.

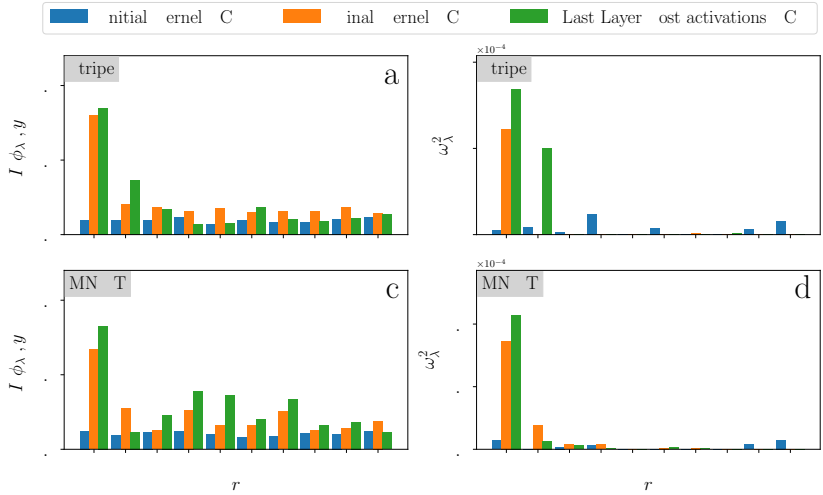

Figure 10: Mutual Information $I(\phi_{\lambda_r}; y)$ between each of the first ten NTK eigenvectors and the output label for the **stripe model** (a) and **MNIST** (c), respectively. The eigenvectors projection on the output labels $\omega_\lambda^2 = \frac{\langle \phi_{\lambda_r} | y \rangle^2}{p^2}$ is shown in panels (b) – **stripe model** – and (d) – **MNIST**. We show in blue the results for the NTK at initialization, in orange for the NTK after training in the feature regime and in green for the principal components of last layer post-activations.

As expected, the magnitude of the projection of each of the first $r$ eigenvectors onto the output labels $\omega_\lambda^2 = \langle \phi_{\lambda_r} | y \rangle^2 / p^2$ also greatly improves during learning. This effect is striking both for the stripe model (Fig.10.b) and for MNIST (Fig.10.d). At initialization, that projection does not show a significant trend with rank within the first 10 eigenvectors. Yet after learning, most of the projection occurs along the first mode of the NTK alone, with the second mode also showing a sizable projection for MNIST.

Overall, the similarities of these plots between MNIST and the stripe model support that compression is indeed a key effect characterizing learning for MNIST as well. To study further these similarities, we focus on the first two eigenvectors and plot data points (different labels appear as different colors) in the $(\phi_{\lambda_1}(\underline{x}), \phi_{\lambda_2}(\underline{x}))$ plane as shown in Fig. 11. As expected, these eigenvectors at initialization have essentially no information on the output label – the scatter plot looks like Gaussian noise both for the stripe model and MNIST (left column). By contrast, after learning data of different classes appear as well separated clouds of points in that plane (central column). Strikingly, performing the same analysis for the kernel obtained from the last layer of hidden neurons shows that data organize into a smaller manifold, which is approximately one-dimensional (right column). It is expected in the stripe model, since for $\Lambda \to \infty$ the hidden neurons activity can only depend on a single variable $x_1$. It is interesting that a similar dimension-reduction appears so clearly in MNIST as well, suggesting the importance of a nearly-one dimensional manifold in the representation of the last hidden layer. We have checked that such a one-dimensional structure is not apparent in the effective dimension of this representation [5].

---

[5] Computing the effective dimension (based on the scaling of the distance between points in terms of the number of points [40]) of that representation leads to $d_{\text{eff}} \approx 6$, possibly coming from the finite width of the nearly-one-dimensional manifold apparent in Fig. 11, bottom right.

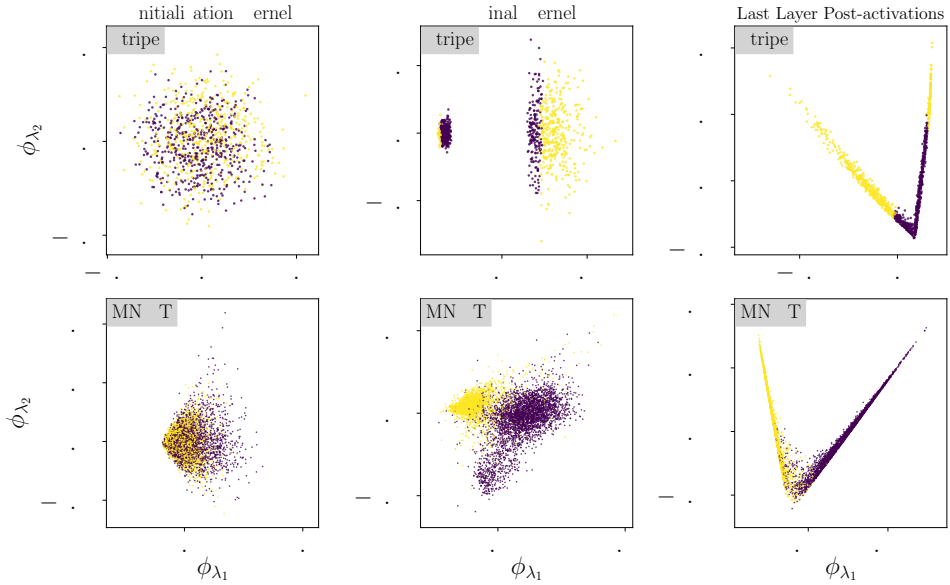

Figure 11: Scatter plot of the first two NTK eigenvectors – $\phi_{\lambda_1}$ and $\phi_{\lambda_2}$ – for the **stripe model** (first row) and **MNIST** (second row). Colors map class labels. Eigenvectors are computed for the NTK at initialization (first column) and after training (second column). The last column refers to the last layer post-activation principal components. These results are consistent with Fig. 10: (1) Before learning, the eigenvectors are not correlated to the labels, while they are after learning. (2) For the stripe model, only the first eigenvector of the final kernel contains information on the labels, as expected from panel b) of Fig. 10. (3) Two informative eigenvectors are necessary to linearly separate the stripe data as illustrated on the top-right panel as well as on panel b) of Fig. 10. The associated unidimensional representation is expected from the effective data compression for the stripe model where $d_\parallel = 1$. (4) For MNIST, the first two eigenvectors of the final kernel are not sufficient to classify completely the data as expected from panel d) of Fig. 10, but still suggest a compression along the uninformative directions. This last point is also motivated by the approximate unidimensional collapse observed in the bottom-right panel.

## 6 Conclusion

We have shown that in the presence of $d_\perp$ uninformative dimensions of the input, the weights of a one-hidden layer neural network become orthogonal to them. For a vanishingly small initialization of the weights and vanilla gradient descent, this effect is limited by the sample noise of the training set, and its magnitude is of order $\Lambda \sim \sqrt{p}$. For simple geometries of the boundaries separating labels, this effect increases the exponent $\beta$ characterizing learning curves with respect to the lazy training regime (in which the neuron orientation is nearly frozen). This increase depends on both $d_\perp$ and $d$. Both for the stripe and cylindrical model, the observed exponents $\beta$ are consistent with this prediction, supporting that for these models at least the main advantage of the feature learning regime is to compress invariant directions.

Next we have argued that such a compression shapes the evolution of the neural tangent kernel during learning, so that its principal components become more informative and display a larger projection on the label, effectively reducing the RKHS norm of the function being learnt. As a consequence, using gradient descent with the frozen NTK at the end of training leads to much better performance than at initialization, and we observe that it even outperforms the neural net in the feature learning regime. The analysis underlines that kernel PCA on the NTK is a valuable tool to characterize the compression of invariants. Overall we find striking similarities between a one-hidden layer FC network trained on the stripe model and a deep CNN trained on MNIST, supporting that compression is central to the performance of the latter as well.

One challenge for the future is to classify which conditions on the data can guarantee such an improvement of the NTK during learning – a question directly connected to the relative performance of lazy training *v.s.* feature learning, which appears to depend on the architecture for real data [19].

A second challenge is the development of quantitative models for the compression of other symmetries in the data, including the invariance of the label toward smooth deformations that characterize images. Is this compression ultimately responsible for the success of deep learning in beating the curse of dimensionality ? Answering this question presumably requires to focus on more modern architectures, in particular deep CNNs.

### Acknowledgments

We acknowledge G. Biroli, M. Gabrie, D. Kopitkov, S. Spigler, Y. Rouzaire and all members of the PCSL group for discussions. This work was partially supported by the grant from the Simons Foundation (#454953 Matthieu Wyart). M.W. thanks the Swiss National Science Foundation for support under Grant No. 200021-165509.

## A  Rotation invariance

In this appendix we prove that if we rotate the input of the network it doesn't affect its performance.

**Lemma**: For a group $G$ and a $G$-invariant function $f$, the gradient of $f$ is $G$-equivariant:

$$\nabla f(D(g)\underline{x}) = D(g)^{-T}\nabla f(\underline{x}) \quad \forall g \in G \text{ and } \forall \underline{x},$$

where $D$ is the representation of $G$ acting on the space of inputs $\underline{x}$ and $A^{-T}$ denotes the inverse transpose of the matrix $A$.

**Proof** The derivative of $f$ in the direction $u$ evaluated in $D(g)x$ is given by

$$\underline{u} \cdot \nabla f(D(g)\underline{x}) = \lim_{h \to 0} \frac{f(D(g)\underline{x} + h\underline{u}) - f(D(g)\underline{x})}{h} \tag{12}$$

$$= \lim_{h \to 0} \frac{f(\underline{x} + hD(g)^{-1}\underline{u}) - f(\underline{x})}{h} \tag{13}$$

$$= (D(g)^{-1}\underline{u}) \cdot \nabla f(\underline{x}) = u \cdot (D(g)^{-T}\nabla f(\underline{x})). \tag{14}$$

Since this formula holds for any direction $\underline{u}$, it proves the lemma.

In the context of a neural network, if the loss function of a neural network satisfies $\mathcal{L}(D_w(g)\underline{w}, D_x(g)\underline{x}) = \mathcal{L}(\underline{w}, \underline{x})$ with $D_w$ orthogonal, it is easy to see that the lemma applied to the loss reads $\nabla_w \mathcal{L}(D_w(g)\underline{w}, D_x(g)\underline{x}) = D_w(g)\nabla_w \mathcal{L}(\underline{w}, \underline{x})$. Here $\underline{w}$ refers to the weights of the network, whose dynamics is given by $\underline{\dot{w}}(t) = -\sum_\mu \nabla_w \mathcal{L}(\underline{w}, \underline{x}_\mu)$, where $\mu$ is the training set index. If we act with $G$ on $\underline{w}$ and on the training set, the derivative $\underline{\dot{w}}$ is transformed in the same way as $\underline{w}$. A network initialised to $D_w(g)\underline{w}_0$ instead of $\underline{w}_0$ and trained on $\{D_x(g)\underline{x}_\mu\}_\mu$ instead of $\{\underline{x}_\mu\}_\mu$ during a time $t$ will thus have its weights equal to $D_w(g)\underline{w}(t)$ instead of $\underline{w}(t)$.

In particular, this discussion holds for a network starting with a fully-connected layer: in this case $G$ is the orthogonal group, $D_x$ is the orthogonal matrix and $D_w$ is acting on the first weights with an orthogonal matrix and leaves the rest of the weights invariant.

In case of an initialisation distribution of the weights that satisfies $\rho(D_w(g)w) = \rho(w)$, the expected performance (averaged over the initialisations) will be independent of the global orientation of the inputs.

## B  $\alpha$ scan in the stripe model

We illustrate the transition from the feature regime to the lazy regime by considering the single-stripe model in dimension $d = 10$ with a training set of size $p = 1000$. We vary the network scale from $\alpha = 10^{-16}$ to $\alpha = 10^8$ (see Fig. 12). In the limit $\alpha \to \infty$, the test error converges to the one obtained by running the kernel dynamics with the NTK frozen at initialization, the characteristic time scales as $t^\star \sim 1/\alpha$ as expected from [19] and the global amplification factor equals one. In the opposite limit, $\alpha \to 0$, the test error converges to a plateau better than the lazy regime performance, the characteristic time grows logarithmically as discussed in Section 2.5 and the global amplification factor reaches a plateau.

## C  Stripe model dynamics

In this section, we give additional details to the computation carried in Section 3.2.1. We consider the large $p$ limit of the system (5), where it is well approximated by the central-limit theorem. For $t \ll t^\star$, the dynamics of each neuron is governed by the system

$$\dot{\omega}_1 = \frac{\beta}{\tau}\left(\mu_1 + \frac{\sigma_1}{\sqrt{p}}N_1\right)$$

$$\underline{\dot{\omega}}_\perp = \frac{\beta}{\tau}\left(\underline{\mu}_\perp + \frac{\sigma_\perp}{\sqrt{p}}\underline{N}_\perp\right)$$

$$\dot{b} = \frac{\beta}{\tau}\left(\mu_b + \frac{\sigma_b}{\sqrt{p}}N_b\right)$$

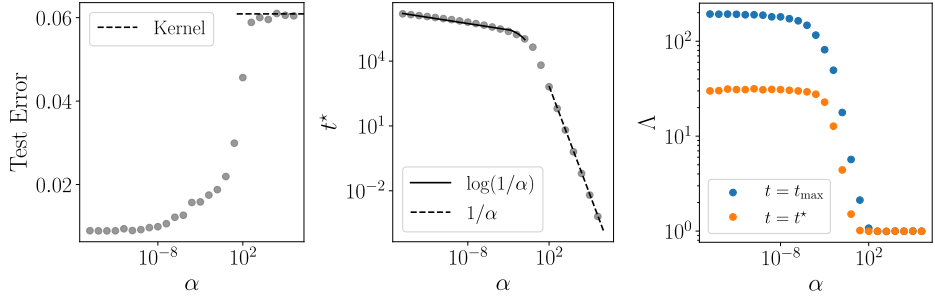

Figure 12: On each plot, the dots are obtained by averaging the gradient descent results over 5 different data realizations and network initialization. <u>Left</u>: Test error vs the network scale $\alpha$. The horizontal dashed line correspond to the test error of the frozen initial kernel dynamics, also averaged over 5 realizations. <u>Center</u>: Characteristic time $t^\star$ vs the network scale $\alpha$. <u>Right</u>: Global amplification factor vs the network scale $\alpha$. Both the amplification factor at $t^\star$ and the maximal amplification factor are represented.

up to $\mathcal{O}(p^{-1})$ corrections. The last layer weight is obtained from the constant of motion $\beta^2 - \|\underline{\omega}\|^2 - b^2 = \text{const}$. We compute the averages $\mu_1$, $\underline{\mu}_\perp$ and $\mu_b$ in Appendix C.1 and discuss the asymptotic solution in the limit $p \to \infty$ in Appendix C.2. The finite $p$ corrections and the associated standard deviations $\sigma_1$, $\sigma_\perp$ and $\sigma_b$ are considered in Appendix C.3.

## C.1  Computation of the averages

We compute the averages $\mu_1$, $\underline{\mu}_\perp$ and $\mu_b$ for data distributed according to the standard normal distribution: $\rho(\underline{x}) = \rho(\|\underline{x}\|) = (2\pi)^{-d/2} \exp\left(-\|\underline{x}\|^2/2\right)$. For the bias and the informative weights, we get

$$
\begin{aligned}
\mu_1 &= \int \mathrm{d}\underline{x}\rho(\underline{x})x_1 y(x_1)\Theta\left[\frac{\underline{\omega}\cdot\underline{x}}{\sqrt{d}}+b\right] \\
&= \int \mathrm{d}x_1\rho(x_1)x_1 y(x_1)\frac{1}{2}\left[1+\mathrm{erf}\left(\frac{b\sqrt{d}+\omega_1 x_1}{\sqrt{2}\omega_\perp}\right)\right] \\
\mu_b &= \sqrt{d}\int \mathrm{d}\underline{x}\rho(\underline{x})y(x_1)\Theta\left[\frac{\underline{\omega}\cdot\underline{x}}{\sqrt{d}}+b\right] \\
&= \sqrt{d}\int \mathrm{d}x_1\rho(x_1)y(x_1)\frac{1}{2}\left[1+\mathrm{erf}\left(\frac{b\sqrt{d}+\omega_1 x_1}{\sqrt{2}\omega_\perp}\right)\right].
\end{aligned}
$$

For the perpendicular weights, we treat each components independently, so that for $i > 1$:

$$
\begin{aligned}
\mu_i &= \int \mathrm{d}\underline{x}\rho(\underline{x})x_i y(x_1)\Theta\left[\frac{\underline{\omega}\cdot\underline{x}}{\sqrt{d}}+b\right] \\
&= \frac{\mathrm{sgn}(\omega_i)}{\sqrt{2\pi}}\int \mathrm{d}x_1\rho(x_1)y(x_1)\int \mathrm{d}x_\perp\rho(x_\perp)\mathrm{e}^{-\frac{(b\sqrt{d}+\omega_1 x_1+\tilde{\omega}_\perp x_\perp)^2}{2\omega_i^2}} \\
&= \frac{1}{\sqrt{2\pi}}\frac{\omega_i}{\omega}\mathrm{e}^{-\frac{db^2}{2\omega^2}}\int \mathrm{d}x_1\rho(x_1)y\left(\frac{\omega_\perp}{\omega}x_1 - \frac{\sqrt{d}b\omega_1}{\omega^2}\right)
\end{aligned}
$$

where we used the notation $\tilde{\omega}_\perp = \sqrt{\omega_\perp^2 - \omega_i^2}$. Using the definition of $\lambda$ and $\zeta_1$, one thus recovers the system (6).

### C.2 Infinite $p$

**Expectation values in the limit $|\lambda| \to \infty$**  In the limit $\lambda \to \pm\infty$, the function $g_\lambda$ becomes a Heaviside function whose direction depends on the sign of $\lambda$: $g_\lambda(x) \xrightarrow{\lambda \to \pm\infty} \Theta(\pm x)$. Consequently, the remaining integrals over the $x_1$ distribution in Appendix C.1 simplifies:

$$\langle y(x_1)\, x_1\, g_\lambda(x_1 - \zeta_1) \rangle_{x_1} \xrightarrow{\lambda \to \pm\infty} \langle y(x_1)\, x_1\, \Theta\left(\pm(x_1 - \zeta_1)\right) \rangle_{x_1} = C_1^\pm(\zeta_1)$$

$$\langle y(x_1)\, g_\lambda(x_1 - \zeta_1) \rangle_{x_1} \xrightarrow{\lambda \to \pm\infty} \langle y(x_1)\, \Theta\left(\pm(x_1 - \zeta_1)\right) \rangle_{x_1} = C_b^\pm(\zeta_1)$$

$$\left\langle y\left(\frac{\omega_\perp}{\omega}x_1 + \frac{\omega_1^2}{\omega^2}\zeta_1\right) \right\rangle_{x_1} \xrightarrow{\lambda \to \pm\infty} y(\zeta_1) = C_\perp(\zeta_1)$$

**Asymptotic solutions**  We assume that the neuron vector $\underline{z}$ is set constant and equal to $\underline{z}^\star = (z^\star, \underline{0})$. The dynamics of $\omega_1$, $\beta$ and $b$ thus no longer depend on the perpendicular weights. In the asymptotic regime, the sign of $\omega_1\beta$ is given by the sign of the constant $C_1^\pm(z^\star)$. In particular, using the constant of motion and the definition $b = -\omega_1 z^\star/\sqrt{d}$, we get $\beta = \text{sign}\left(C_1^\pm(z^\star)\right)\sqrt{1 + z^{\star 2}/d}\,\omega_1$, where we neglected the order one value of the constant of motion. Finally, one finds that the informative weights diverge as

$$\omega_1 \sim e^{t/\tau^\star}, \quad \text{with } \tau^\star = \frac{\tau}{|C_1^\pm(z^\star)|\sqrt{1 + z^{\star 2}/d}}. \tag{15}$$

Inserting the above relations into the perpendicular weights dynamics yields

$$\underline{\dot{\omega}}_\perp = \text{sign}\left[C_1^\pm(z^\star)\omega_1 y(z^\star)\right]\frac{\sqrt{1 + z^{\star 2}/d}}{2\pi\tau}e^{-\frac{z^{\star 2}}{2}}\,\underline{\omega}_\perp. \tag{16}$$

Hence, if $\text{sign}\left[C_1^\pm(z^\star)\omega_1 y(z^\star)\right] = -1$, the perpendicular weights all vanish exponentially. However, if $\text{sign}\left[C_1^\pm(z^\star)\omega_1 y(z^\star)\right] = +1$, they all diverge exponentially with a time constant

$$\tau_\perp^\star = \frac{2\pi\tau}{\sqrt{1 + z^{\star 2}/d}}e^{\frac{z^{\star 2}}{2}}$$

which still leads to a diverging amplification factor if $\tau_\perp^\star > \tau^\star$.

### C.3 Finite $p$

We assess the finite $p$ corrections of the asymptotic solutions given in Appendix C.2. Since the bias and the informative weights are divergent, they are not sensitive to finite $p$ corrections. However, for the perpendicular weights, it is essential to compute the standard deviations. Since the expectations have been computed previously, it is sufficient to look at the second non-central moments. For simplicity, we directly consider the limit $|\lambda| \to \infty$, so that for $i > 1$:

$$\mu_i^2 + \sigma_i^2 = \int d\underline{x}\rho(\underline{x})x_i^2\Theta\left[\frac{\underline{\omega}\cdot\underline{x}}{\sqrt{d}} + b\right] \xrightarrow{\lambda \to \pm\infty} \int dx_1\rho(x_1)\Theta\left[\pm(x_1 - z^\star)\right] = D_\perp^\pm(\zeta_1).$$

For each perpendicular direction, a random variable of variance one quantifies the discrepancy between the average $\mu_i$ and the exact sum over the dataset. Its value depends on the location of the ReLU hyperplane. In particular, once the considered neuron has reached its fixed point $\underline{z}^\star$, all random variables can be arranged into the constant perpendicular vector $\underline{N}_\perp(z^\star)$.

## D  NTK decomposition and eigenfunctions

In section 5.1 we argued that, for the setting considered in this paper, the NTK can be decomposed as

$$\Theta(\underline{x}, \underline{z}) = \Theta_1(\underline{x}_\|, \underline{z}_\|) + \Theta_2(\underline{x}_\|, \underline{z}_\|)\underline{x}_\perp \cdot \underline{z}_\perp. \tag{17}$$

In this appendix, we look at this decomposition more in details and derive the eigenfunctions functional form.

**NTK decomposition**  Recall the architecture considered in this paper,

$$f(\underline{x}) = \frac{1}{h} \sum_{n=1}^{h} \beta_n \, \sigma \left( \frac{\underline{\omega}_n \cdot \underline{x}}{\sqrt{d}} + b_n \right).$$

For this architecture, the NTK reads

$$\Theta(\underline{x}, \underline{z}) = \frac{1}{h^2} \sum_{n=1}^{h} \left[ \sigma \left( \frac{\underline{\omega}_n \cdot \underline{x}}{\sqrt{d}} + b_n \right) \sigma \left( \frac{\underline{\omega}_n \cdot \underline{z}}{\sqrt{d}} + b_n \right) + \beta_n^2 \sigma' \left( \frac{\underline{\omega}_n \cdot \underline{x}}{\sqrt{d}} + b_n \right) \sigma' \left( \frac{\underline{\omega}_n \cdot \underline{z}}{\sqrt{d}} + b_n \right) \left( 1 + \frac{\underline{x} \cdot \underline{z}}{d} \right) \right].$$

If the input space has only $d_\parallel$ informative directions, after feature learning ($\Lambda \to \infty$), the output function will only depend on $\underline{x}_\parallel$. This is because $\underline{\omega}_n \cdot \underline{x} \to \underline{\omega}_{n,\parallel} \cdot \underline{x}_\parallel$ and the NTK can be rewritten as

$$\Theta(\underline{x}, \underline{z}) = \frac{1}{h^2} \sum_{n=1}^{h} \left[ \sigma \left( \frac{\underline{\omega}_{n,\parallel} \cdot \underline{x}_\parallel}{\sqrt{d}} + b_n \right) \sigma \left( \frac{\underline{\omega}_{n,\parallel} \cdot \underline{z}_\parallel}{\sqrt{d}} + b_n \right) \right.$$
$$\left. + \beta_n^2 \sigma' \left( \frac{\underline{\omega}_{n,\parallel} \cdot \underline{x}_\parallel}{\sqrt{d}} + b_n \right) \sigma' \left( \frac{\underline{\omega}_{n,\parallel} \cdot \underline{z}_\parallel}{\sqrt{d}} + b_n \right) \left( 1 + \frac{\underline{x}_\parallel \cdot \underline{z}_\parallel}{d} + \frac{\underline{x}_\perp \cdot \underline{z}_\perp}{d} \right) \right],$$

where one can readily identify $\Theta_1(x_\parallel, \underline{z}_\parallel)$ and $\Theta_2(x_\parallel, \underline{z}_\parallel)$.

**NTK eigenfunctions**  Eigenfunctions satisfy the integral equation

$$\int \Theta(\underline{x}, \underline{z}) \phi_\lambda(\underline{z}) \rho(\underline{z}) d\underline{z} = \lambda \phi_\lambda(\underline{x}),$$

where $\rho(\cdot)$ is the distribution of the data. We assume here that $\rho(\underline{x}) = \rho_\parallel(\underline{x}_\parallel) \rho_\perp(\underline{x}_\perp) = \rho_\parallel(\underline{x}_\parallel) \Pi_i \rho_i(x_{\perp,i})$ with zero mean and the same variance in all directions. If we plug in the decomposition (17), we notice that eigenvectors are of two kinds, they are either eigenvectors of $\Theta_1(\underline{x}_\parallel, \underline{z}_\parallel)$ or of $\Theta_2(\underline{x}_\parallel, \underline{z}_\parallel) \underline{x}_\perp \cdot \underline{z}_\perp$ – i.e. they give zero when the other operator acts on them. The ones coming from $\Theta_1$ are solutions of

$$\int \Theta_1(\underline{x}_\parallel, \underline{z}_\parallel) \phi_\lambda^1(\underline{z}) \rho(\underline{z}) d\underline{z} = \lambda \phi_\lambda^1(\underline{x}).$$

Given that the l.h.s. only depends on $\underline{x}_\parallel$, we have $\phi_\lambda^1(\underline{x}) = \phi_\lambda^1(\underline{x}_\parallel)$. Integrating out $\underline{z}_\perp$ we get

$$\int \Theta_1(\underline{x}_\parallel, \underline{z}_\parallel) \phi_\lambda^1(\underline{z}_\parallel) \rho(\underline{z}_\parallel) d\underline{z}_\parallel = \lambda \phi_\lambda^1(\underline{x}_\parallel).$$

The second kind of eigenvectors satisfy

$$\int \Theta_2(\underline{x}_\parallel, \underline{z}_\parallel) \underline{x}_\perp \cdot \underline{z}_\perp \phi_\lambda^2(\underline{z}) \rho(\underline{z}) d\underline{z} = \lambda \phi_\lambda^2(\underline{x}).$$

Notice that $\underline{x}_\perp$ can be moved out of the integral. Consequently, eigenfunctions can only linearly depend on the perpendicular component – i.e. $\phi_\lambda^2(\underline{x}) = \phi_\lambda^2(\underline{x}_\parallel) \underline{u} \cdot \underline{x}_\perp$. The integral equation reads

$$\left( \int \Theta_2(\underline{x}_\parallel, \underline{z}_\parallel) \phi_\lambda^2(\underline{z}_\parallel) \rho(\underline{z}_\parallel) d\underline{z}_\parallel \right) (\underline{u} \cdot \underline{x}_\perp) \int z_\perp^2 \rho(z_\perp) dz_\perp = \lambda \phi_\lambda^2(\underline{x}_\parallel) \underline{u} \cdot \underline{x}_\perp,$$

where $\underline{u}$ can be any non-zero vector. To back what we stated previously – i.e. that eigenvectors are either of the two kinds – we show that no other eigenvector, different from linear combinations of $\phi_\lambda^1$ and $\phi_\lambda^2$, exists. Assume there exists $\phi_\lambda^\star(\underline{x}) \neq a \phi_{\lambda_1}^1(\underline{x}) + b \phi_{\lambda_2}^2(\underline{x})$, this would solve

$$\int \left[ \Theta_1(\underline{x}_\parallel, \underline{z}_\parallel) + \Theta_2(\underline{x}_\parallel, \underline{z}_\parallel) \underline{x}_\perp \cdot \underline{z}_\perp \right] \phi_\lambda^\star(\underline{z}) \rho(\underline{z}) d\underline{z} = \lambda \phi_\lambda^\star(\underline{x})$$

$$\int \Theta_1(\underline{x}_\parallel, \underline{z}_\parallel) \phi_\lambda^\star(\underline{z}) \rho(\underline{z}) d\underline{z} + \int \Theta_2(\underline{x}_\parallel, \underline{z}_\parallel) \underline{x}_\perp \cdot \underline{z}_\perp \phi_\lambda^\star(\underline{z}) \rho(\underline{z}) d\underline{z} = \lambda \phi_\lambda^\star(\underline{x})$$

$$\lambda_1 \phi_{\lambda_1}^1(\underline{x}) + \lambda_2 \phi_{\lambda_2}^2(\underline{x}) = \lambda \phi_\lambda^\star(\underline{x}),$$

resulting in a contradiction.

## E   Mutual Information Estimator

We propose a mutual information estimator $\widehat{I}(\underline{x}; y)$ that exploits the information we know about the binary labels distribution $P(y)$:

$$P(y = +) = P(y = -) = \frac{1}{2}.$$

The variable $\underline{x}$ is continuous and can live in high dimension. We define

$$q_+ = \frac{P(\underline{x}|y = +)}{2P(\underline{x})} = P(y = +|\underline{x}), \qquad q_- = \frac{P(\underline{x}|y = -)}{2P(\underline{x})} = P(y = -|\underline{x}).$$

We recall the definition of differential entropy for continuous variables,

$$H(\underline{x}) = -\int d^d\underline{x}\, P(\underline{x}) \log P(\underline{x})$$

Given that the mutual information can be expressed $I(\underline{x}; y) = H(\underline{x}) - H(\underline{x}|y)$, we compute the conditional entropy knowing $P(y)$ as[6]

$$
\begin{aligned}
H(\underline{x}|y) &= -\frac{1}{2}\int d^d\underline{x}\, P(\underline{x}|y = +) \log P(\underline{x}|y = +) - \frac{1}{2}\int d^d\underline{x}\, P(\underline{x}|y = -) \log P(\underline{x}|y = -)\\
&= -\int d^d\underline{x}\, P(\underline{x})q_+ \log(2P(\underline{x})q_+) - \int d^d\underline{x}\, P(\underline{x})q_- \log(2P(\underline{x})q_-)\\
&= -\int d^d\underline{x}\, P(\underline{x})(q_+ + q_-) \log(2P(\underline{x})) - \int d^d\underline{x}\, P(\underline{x})\big[q_+ \log(q_+) + q_- \log(q_-)\big]\\
&= H(\underline{x}) - 1 - \mathbb{E}_{\underline{x}}\big[q_+ \log(q_+) + q_- \log(q_-)\big].
\end{aligned}
$$

Finally, the mutual information is given by

$$
\begin{aligned}
I(\underline{x}; y) &= H(\underline{x}) - H(\underline{x}|y)\\
&= 1 + \mathbb{E}_{\underline{x}}\big[q_+ \log(q_+) + q_- \log(q_-)\big].
\end{aligned}
$$

We find the following estimator

$$
\begin{aligned}
\widehat{I}(\underline{x}; y) &= 1 + \frac{1}{p}\sum_{i=1}^{p} \widehat{q_+}(\underline{x}_i) \log \widehat{q_+}(\underline{x}_i) + \widehat{q_-}(\underline{x}_i) \log \widehat{q_-}(\underline{x}_i)\\
&= 1 - \frac{1}{p}\sum_{i=1}^{p} h_2(\widehat{q_+}(\underline{x}_i)),
\end{aligned}
$$

where $h_2(\cdot)$ is the binary entropy function[7].

We notice that we can rewrite

$$
\begin{aligned}
P(y|\underline{x}) &= \frac{P(\underline{x}|y)}{2\sum_y P(\underline{x}|y)P(y)}\\
&= \frac{P(\underline{x}|y)}{P(\underline{x}|y = +) + P(\underline{x}|y = -)},
\end{aligned}
$$

hence the MI estimation reduces to estimating $P(\underline{x}|y)$.

At this stage we propose the following approximation: suppose that $P(\underline{x}|y)$ is uniform in the ball containing the $k$ nearest neighbors of $\underline{x}$ which are labelled $y$, i.e.[8]

$$P(\underline{x}|y) \sim r_y^{-d}(\underline{x}_i).$$

---

[6]All the logarithms of this section are computed in base 2.
[7]$h_2(x) = -x\log x - (1 - x)\log(1 - x)$ .
[8]The estimation depends on the value of $k$ which is omitted to simplify the notation. For the estimations in this paper we use $k = 5$.

The estimation of $\widehat{q_+}$ finally reduces to

$$\widehat{q_+}(\underline{x}_i) = \frac{r_+^{-d}(\underline{x}_i)}{r_+^{-d}(\underline{x}_i) + r_-^{-d}(\underline{x}_i)} = \frac{1}{1 + \left(\frac{r_+(\underline{x}_i)}{r_-(\underline{x}_i)}\right)^d} \tag{18}$$

$$\widehat{q_-}(\underline{x}_i) = 1 - \widehat{q_+}(\underline{x}_i).$$

We tested the estimator on different datasets and identified two main flaws:

○ For large $d$, the estimator gets affected by the *curse of dimensionality*, distances between data-points become all similar to each other. As a result, the $\widehat{q_+}$ estimator gets biased towards ½.

○ If $\underline{x}$ lives on a manifold of dimension lower than the one of the embedding space, the use of $d$ in Eq. (18) – instead of the effective local dimension around $\underline{x}_i$ – biases the estimator towards its extrema.

Considering we employ the estimator only in $d = 1$, we skip the discussion on the possible ways to correct these flaws.

# Invariance toward Smooth Deformations

**Part II**

# **3** Deformation Invariance Strongly Correlates to Performance in Image Tasks

The following paper is the preprint version of Petrini et al. (2021) published in *Advances in Neural Information Processing Systems.*

**Candidate contributions**  The candidate contributed to all discussions. The candidate performed the experiments, in particular, discovered the correlation between test error and relative stability. The candidate drafted the first version of the paper.

# Relative stability toward diffeomorphisms indicates performance in deep nets

**Leonardo Petrini, Alessandro Favero, Mario Geiger, Matthieu Wyart**

Institute of Physics
École Polytechnique Fédérale de Lausanne
1015 Lausanne, Switzerland
{name.surname}@epfl.ch

## Abstract

Understanding why deep nets can classify data in large dimensions remains a challenge. It has been proposed that they do so by becoming stable to diffeomorphisms, yet existing empirical measurements support that it is often not the case. We revisit this question by defining a maximum-entropy distribution on diffeomorphisms, that allows to study typical diffeomorphisms of a given norm. We confirm that stability toward diffeomorphisms does not strongly correlate to performance on benchmark data sets of images. By contrast, we find that the *stability toward diffeomorphisms relative to that of generic transformations* $R_f$ correlates remarkably with the test error $\epsilon_t$. It is of order unity at initialization but decreases by several decades during training for state-of-the-art architectures. For CIFAR10 and 15 known architectures we find $\epsilon_t \approx 0.2\sqrt{R_f}$, suggesting that obtaining a small $R_f$ is important to achieve good performance. We study how $R_f$ depends on the size of the training set and compare it to a simple model of invariant learning.

## 1 Introduction

Deep learning algorithms LeCun et al. (2015) are now remarkably successful at a wide range of tasks Amodei et al. (2016); Huval et al. (2015); Mnih et al. (2013); Shi et al. (2016); Silver et al. (2017). Yet, understanding how they can classify data in large dimensions remains a challenge. In particular, the curse of dimensionality associated with the geometry of space in large dimension prohibits learning in a generic setting Luxburg and Bousquet (2004). If high-dimensional data can be learnt, then they must be highly structured.

A popular idea is that during training, hidden layers of neurons learn a representation Le (2013) that is insensitive to aspects of the data unrelated to the task, effectively reducing the input dimension and making the problem tractable Ansuini et al. (2019); Recanatesi et al. (2019); Shwartz-Ziv and Tishby (2017). Several quantities have been introduced to study this effect empirically. It includes (i) the mutual information between the hidden and visible layers of neurons Saxe et al. (2019); Shwartz-Ziv and Tishby (2017), (ii) the intrinsic dimension of the neural representation of the data Ansuini et al. (2019); Recanatesi et al. (2019) and (iii) the projection of the label of the data on the main features of the network Kopitkov and Indelman (2020); Oymak et al. (2019); Paccolat et al. (2021a), the latter being defined from the top eigenvectors of the Gram matrix of the neural tangent kernel (NTK) Jacot et al. (2018). All these measures support that the neuronal representation of the data indeed becomes well-suited to the task. Yet, they are agnostic to the nature of what varies in the data that need not being represented by hidden neurons, and thus do not specify what it is.

Recently, there has been a considerable effort to understand the benefits of learning features for one-hidden-layer fully connected nets. Learning features can occur and improve performance when the true function is highly anisotropic, in the sense that it depends only on a linear subspace of the input space Bach (2017); Chizat and Bach (2020); Ghorbani et al. (2019, 2020); Paccolat et al. (2021a); Refinetti et al. (2021); Yehudai and Shamir (2019). For image classification, such an anisotropy would occur for example if pixels on the edge of the image are unrelated to the task. Yet, fully-connected nets (unlike CNNs) acting on images tend to perform best in training regimes where features are not learnt Geiger et al. (2021, 2020); Lee et al. (2020), suggesting that such a linear invariance in the data is not central to the success of deep nets.

Instead, it has been proposed that images can be classified in high dimensions because classes are invariant to smooth deformations or diffeomorphisms of small magnitude Bruna and Mallat (2013); Mallat (2016). Specifically, Mallat and Bruna could handcraft convolution networks, the *scattering transforms*, that perform well and are stable to smooth transformations, in the sense that $\|f(x) - f(\tau x)\|$ is small if the norm of the diffeomorphism $\tau$ is small too. They hypothesized that during training deep nets learn to become stable and thus less sensitive to these deformations, thus improving performance. More recent works generalize this approach to more common CNNs and discuss stability at initialization Bietti and Mairal (2019a,b). Interestingly, enforcing such a stability can improve performance Kayhan and Gemert (2020).

Answering if deep nets become more stable to smooth deformations when trained and quantifying how it affects performance remains a challenge. Recent empirical results revealed that small shifts of images can change the output a lot Azulay and Weiss (2018); Dieleman et al. (2016); Zhang (2019), in apparent contradiction with that hypothesis. Yet in these works, image transformations (i) led to images whose statistics were very different from that of the training set or (ii) were cropping the image, thus are not diffeomorphisms. In Ruderman et al. (2018), a class of diffeomorphisms (low-pass filter in spatial frequencies) was introduced to show that stability toward them can improve during training, especially in architectures where pooling layers are absent. Yet, these studies do not address how stability affects performance, and how it depends on the size of the training set. To quantify these properties and to find robust empirical behaviors across architectures, we will argue that the evolution of stability toward smooth deformations needs to be compared relatively to that of any deformation, which turns out to vary significantly during training.

Note that in the context of adversarial robustness, attacks that are geometric transformations of small norm that change the label have been studied Alaifari et al. (2018); Alcorn et al. (2019); Athalye et al. (2018); Engstrom et al. (2019); Fawzi and Frossard (2015); Kanbak et al. (2018); Xiao et al. (2018). These works differ for the literature above and from out study below in the sense that they consider worst-case perturbations instead of typical ones.

### 1.1 Our Contributions

○ We introduce a *maximum entropy distribution* of diffeomorphisms, that allow us to generate typical diffeomorphisms of controlled norm. Their amplitude is governed by a "temperature" parameter $T$.

○ We define the *relative stability to diffeomorphisms index* $R_f$ that characterizes the square magnitude of the variation of the output function $f$ with respect to the input when it is transformed along a diffeomorphism, relatively to that of a random transformation of the same amplitude. It is averaged on the test set as well as on the ensemble of diffeomorphisms considered.

○ We find that at initialization, $R_f$ is close to unity for various data sets and architectures, indicating that initially the output is as sensitive to smooth deformations as it is to random perturbations of the image.

○ Our central result is that after training, $R_f$ correlates very strongly with the test error $\epsilon_t$: during training, $R_f$ is reduced by several decades in current State Of The Art (SOTA) architectures on four benchmark datasets including MNIST Lecun et al. (1998), FashionMNIST Xiao et al. (2017), CIFAR-10 Krizhevsky (2009) and ImageNet Deng et al. (2009). For more primitive architectures (whose test error is higher) such as fully connected nets or simple CNNs, $R_f$ remains of order unity. For CIFAR10 we study 15 known architectures and find empirically that $\epsilon_t \approx 0.2\sqrt{R_f}$.

○ $R_f$ decreases with the size of the training set $P$. We compare it to an inverse power $1/P$ expected in simple models of invariant learning Paccolat et al. (2021a).

The library implementing diffeomorphisms on images is available online at github.com/pcsl-epfl/diffeomorphism.

The code for training neural nets can be found at github.com/leonardopetrini/diffeo-sota and the corresponding pre-trained models at doi.org/10.5281/zenodo.5589870.

## 2 Maximum-entropy model of diffeomorphisms

### 2.1 Definition of maximum entropy model

We consider the case where the input vector $x$ is an image. It can be thought as a function $x(s)$ describing intensity in position $s = (u, v) \in [0, 1]^2$, where $u$ and $v$ are the horizontal and vertical coordinates. To simplify notations we consider a single channel, in which case $x(s)$ is a scalar (but our analysis holds for colored images as well). We denote by $\tau x$ the image deformed by $\tau$, i.e. $[\tau x](s) = x(s - \tau(s))$. $\tau(s)$ is a vector field of components $(\tau_u(s), \tau_v(s))$. The deformation amplitude is measured by the norm

$$\|\nabla \tau\|^2 = \int_{[0,1]^2} ((\nabla \tau_u)^2 + (\nabla \tau_v)^2) du dv. \tag{1}$$

To test the stability of deep nets toward diffeomorphisms, we seek to build *typical* diffeomorphisms of controlled norm $\|\nabla \tau\|$. We thus consider the distribution over diffeomorphisms that maximizes the entropy with a norm constraint. It can be solved by introducing a Lagrange multiplier $T$ and by decomposing these fields on their Fourier components, see e.g. Kardar (2007) or Appendix A. In this canonical ensemble, one finds that $\tau_u$ and $\tau_v$ are independent with identical statistics. For the picture frame not to be deformed, we impose fixed boundary conditions: $\tau = 0$ if $u = 0, 1$ or $v = 0, 1$. One then obtains:

$$\tau_u = \sum_{i,j \in \mathbb{N}^+} C_{ij} \sin(i \pi u) \sin(j \pi v) \tag{2}$$

where the $C_{ij}$ are Gaussian variables of zero mean and variance $\langle C_{ij}^2 \rangle = T/(i^2 + j^2)$. If the picture is made of $n \times n$ pixels, the result is identical except that the sum runs on $0 < i, j \leq n$. For large $n$, the norm then reads $\|\nabla \tau\|^2 = (\pi^2/2) n^2 T$, and is dominated by high spatial frequency modes. It is useful to add another parameter $c$ to cut-off the effect of high spatial frequencies, which can be simply done by constraining the sum in Eq.2 to $i^2 + j^2 \leq c^2$, one then has $\|\nabla \tau\|^2 = (\pi^3/8) c^2 T$.

Once $\tau$ is generated, pixels are displaced to random positions. A new pixelated image can then be obtained using standard interpolation methods. We use two interpolations, Gaussian and bi-linear[1], as described in Appendix C. As we shall see below, this choice does not affect our result as long as the diffeomorphism induced a displacement of order of the pixel size, or larger. Examples are shown in Fig.1 as a function of $T$ and $c$.

### 2.2 Phase diagram of acceptable diffeomorphisms

Diffeomorphisms are bijective, which is not the case for our transformations if $T$ is too large. When this condition breaks down, a single domain of the picture can break into several pieces, as apparent in Fig.1. It can be expressed as a condition on $\nabla \tau$ that must be satisfied in every point in space Lowe (2004), as recalled in Appendix B. This is satisfied locally with high probability if $\|\tau\|^2 \ll 1$, corresponding to $T \ll (8/\pi^3)/c^2$. In Appendix, we extract empirically a curve of similar form in the $(T, c)$ plane at which a diffeomorphism is obtained with probability at least $1/2$. For much smaller $T$, diffeomorphisms are obtained almost surely.

Finally, for diffeomorphisms to have noticeable consequences, their associated displacement must be of the order of magnitude of the pixel size. Defining $\delta^2$ as the average square norm of the pixel displacement at the center of the image in the unit of pixel size, it is straightforward to obtain from Eq.2 that asymptotically for large $c$ (cf. Appendix B for the derivation),

$$\delta^2 = \frac{\pi}{4} n^2 T \ln(c). \tag{3}$$

The line $\delta = 1/2$ is indicated in Fig.1, using empirical measurements that add pre-asymptotic terms to Eq.3. Overall, the green region corresponds to transformations that (i) are diffeomorphisms with high probability and (ii) produce significant displacements at least of the order of the pixel size.

---

[1] Throughout the paper, if not specified otherwise, bi-linear interpolation is employed.

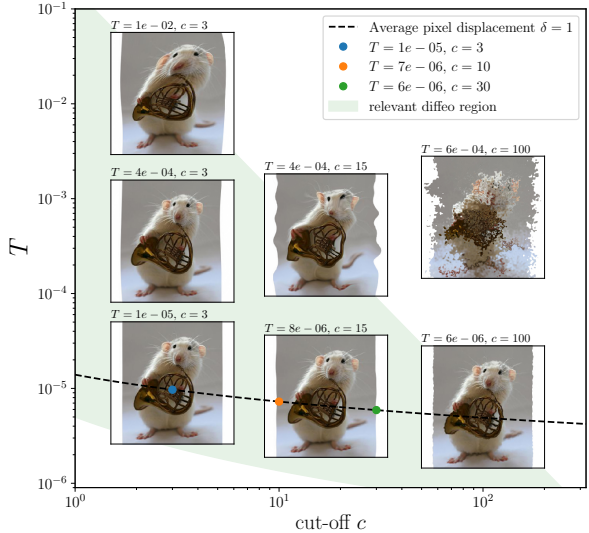

Figure 1: Samples of max-entropy diffeomorphisms for different temperatures $T$ and high-frequency cut-offs $c$ for an ImageNet datapoint of resolution $320 \times 320$. The green region corresponds to well behaving diffeomorphisms (see Section 2.2). The dashed line corresponds to $\delta = 1$. The colored points on the line are those we focus our study in Section 3.

## 3 Measuring the relative stability to diffeomorphisms

**Relative stability to diffeomorphisms** To quantify how a deep net $f$ learns to become less sensitive to diffeomorphisms than to generic data transformations, we define the relative stability to diffeomorphisms $R_f$ as:

$$R_f = \frac{\langle \|f(\tau x) - f(x)\|^2 \rangle_{x,\tau}}{\langle \|f(x + \eta) - f(x)\|^2 \rangle_{x,\eta}}. \tag{4}$$

where the notation $\langle \rangle_y$ can indicate alternatively the mean or the median with respect to the distribution of $y$. In the numerator, this operation is made over the test set and over the ensemble of diffeomorphisms of parameters $(T, c)$ (on which $R_f$ implicitly depends). In the denominator, the average is on the test set and on the vectors $\eta$ sampled uniformly on the sphere of radius $\|\eta\| = \langle \|\tau x - x\| \rangle_{x,\tau}$. An illustration of what $R_f$ captures is shown in Fig.2. In the main text, we consider median quantities, as they reflect better the typical values of distribution. In Appendix E.3 we show that our results for mean quantities, for which our conclusions also apply.

**Dependence of $R_f$ on the diffeomorphism magnitude** Ideally, $R_f$ could be defined for infinitesimal transformations, as it would then characterize the magnitude of the gradient of $f$ along smooth deformations of the images, normalized by the magnitude of the gradient in random directions. However, infinitesimal diffeomorphisms move the image much less than the pixel size, and their definition thus depends significantly on the interpolation method used. It is illustrated in the left panels of Fig.3, showing the dependence of $R_f$ in terms of the diffeomorphism magnitude (here characterised by the mean displacement magnitude at the center of the image $\delta$) for several interpolation methods. We do see that $R_f$ becomes independent of the interpolation when $\delta$ becomes of order unity. In what follows we thus focus on $R_f(\delta = 1)$, which we denote $R_f$.

**SOTA architectures become relatively stable to diffeomorphisms during training, but are not at initialization** The central panels of Fig.3 show $R_f$ at initialization (shaded), and after training (full) for two SOTA architectures on four benchmark data sets. The first key result is that, at initialization, these architectures are as sensitive to diffeomorphisms as they are to random transformations. Relative stability to diffeomorphisms at initialization (guaranteed theoretically in some cases Bietti and Mairal (2019a,b)) thus does not appear to be indicative of successful architectures.

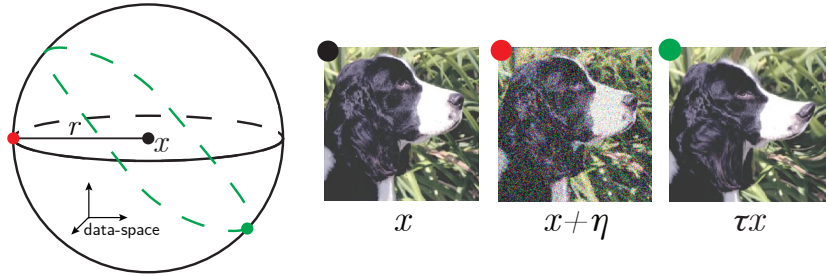

Figure 2: Illustrative drawing of the data-space $\mathbb{R}^{n \times n}$ around a data-point $x$ (black point). We focus here on perturbations of fixed magnitude – i.e. on the sphere of radius $r$ centered in $x$. The intersection between the images of $x$ transformed via typical diffeomorphisms and the sphere is represented in dashed green. By contrast, the red point is an example of random transformation. For large $n$, it is equivalent to adding an i.i.d. Gaussian noise to all the pixel values of $x$. Figures on the right illustrate these transformations, the color of the dot labelling them corresponds to that of the left illustration. The relative stability to diffeomorphisms $R_f$ characterizes how a net $f$ varies in the green directions, normalized by random ones.

By contrast, for these SOTA architectures, relative stability toward diffeomorphisms builds up during training on all the data sets probed. It is a significant effect, with values of $R_f$ after training generally found in the range $R_f \in [10^{-2}, 10^{-1}]$.

Standard data augmentation techniques (translations, crops, and horizontal flips) are employed for training. However, the results we find only mildly depend on using such techniques, see Fig.12 in Appendix.

**Learning relative stability to diffeos requires large training sets**    How many data are needed to learn relative stability toward diffeomorphisms? To answer this question, newly initialized networks are trained on different training-sets of size $P$. $R_f$ is then measured for CIFAR10, as indicated in the right panels of Fig.3. Neural nets need a certain number of training points ($P \sim 10^3$) in order to become relatively stable toward smooth deformations. Past that point, $R_f$ monotonically decreases with $P$. In a range of $P$, this decrease is approximately compatible with the an inverse behavior $R_f \sim 1/P$ found in the simple model of Section 6. Additional results for MNIST and FashionMNIST can be found in Fig.13, Appendix E.3.

**Simple architectures do not become relatively stable to diffeomorphisms**    To test the universality of these results, we focus on two simple architectures: (i) a 4-hidden-layer fully connected (FC) network (FullConn-L4) where each hidden layer has 64 neurons and (ii) LeNet LeCun et al. (1989) that consists of two convolutional layers followed by local max-pooling and three fully-connected layers.

Measurements of $R_f$ for these networks are shown in Fig.4. For the FC net, $R_f \approx 1$ at initialization (as observed for SOTA nets) but *grows* after training on the full data set, showing that FC nets do not learn to become relatively stable to smooth deformations. It is consistent with the modest evolution of $R_f(P)$ with $P$, suggesting that huge training sets would be required to obtain $R_f < 1$. The situation is similar for the primitive CNN LeNet, which only becomes slightly insensitive ($R_f \approx 0.6$) in a single data set (CIFAR10), and otherwise remains larger than unity.

**Layers' relative stability monotonically increases with depth**    Up to this point, we measured the relative stability of the output function for any given architecture. We now study how relative stability builds up as the input data propagate through the hidden layers. In Fig.14 of Appendix E.3, we report $R_f$ as a function of depth for both simple and deep nets. What we observe is $R_{f_0} \approx 1$ independently of depth at initialization, and monotonically decreases with depth after training. Overall, the gain in relative stability appears to be well-spread through the net, as is also found for stability alone Ruderman et al. (2018).

[2]With the only exception of the ImageNet results (central panel) in which only one trained network is considered.

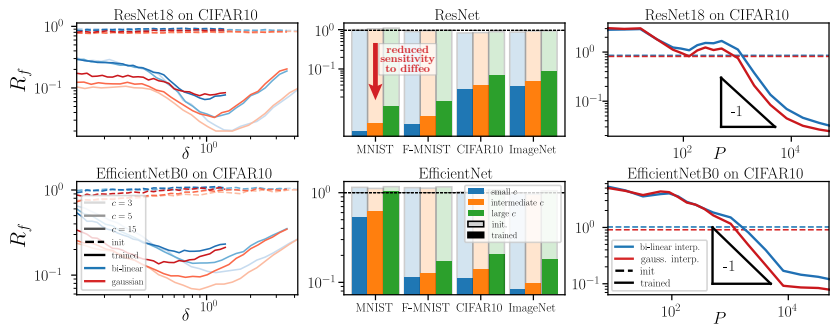

Figure 3: **Relative stability to diffeomorphisms $R_f$ for SOTA architectures.** Left panels: $R_f$ vs. diffeomorphism displacement magnitude $\delta$ at initialization (dashed lines) and after training (full lines) on the full data set of CIFAR10 ($P = 50k$) for several cut-off parameters $c$ and two interpolations methods, as indicated in legend. ResNet is shown on the top and EfficientNet on the bottom. Central panels: $R_f(\delta = 1)$ for four different data-sets ($x$−axis) and two different architectures at initialization (shaded histograms) and after training (full histograms). The values of $c$ (in different colors) are $(3, 5, 15)$ and $(3, 10, 30)$ for the first three data-sets and ImageNet, respectively. ResNet18 and EfficientNetB0 are employed for MNIST, F-MNIST and CIFAR10, ResNet101 and EfficientNetB2 for ImageNet. Right panels: $R_f(\delta = 1)$ vs. training set size $P$ at $c = 3$ for ResNet18 (top) and EfficientNetB0 (bottom) trained on CIFAR10. The value of $R_{f_0}$ at initialization is indicated with dashed lines. The triangles indicate the predicted slope $R_f \sim P^{-1}$ in a simple model of invariant learning, see Section 6. *Statistics:* Each point in the graphs[2] is obtained by training 16 differently initialized networks on 16 different subsets of the data-sets; each network is then probed with 500 test samples in order to measure stability to diffeomorphisms and Gaussian noise. The resulting $R_f$ is obtained by log-averaging the results from single realizations.

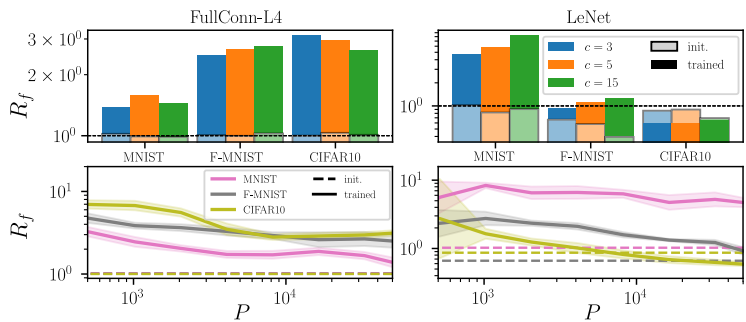

Figure 4: **Relative stability to diffeomorphisms $R_f$ in primitive architectures.** Top panels: $R_f$ at initialization (shaded) or for trained nets (full) for a fully connected net (left) or a primitive CNN (right) at $P = 50k$. Bottom panels: $R_f(P)$ for $c = 3$ and different data sets as indicated in legend. *Statistics:* see caption in the previous figure.

### 4 Relative stability to diffeomorphisms indicates performance

Thus, SOTA architectures appear to become relatively stable to diffeomorphisms after training, unlike primitive architectures. This observation suggests that high performance requires such a relative stability to build up. To test further this hypothesis, we select a set of architectures that have been relevant in the state of the art progress over the past decade; we systematically train them in order to compare $R_f$ to their test error $\epsilon_t$. Apart from fully connected nets, we consider the already cited LeNet (5 layers and $\approx 60k$ parameters); then AlexNet Krizhevsky et al. (2012) and VGG Simonyan and Zisserman (2015), deeper (8-19 layers) and highly over-parametrized (10-20M (million) params.) versions of the latter. We introduce *batch-normalization* in VGGs and *skip connections* with ResNets. Finally, we go to EfficientNets, that have all the advancements introduced in previous models and achieve SOTA performance with a relatively small number of parameters (<10M); this is accomplished by designing an efficient small network and properly scaling it up. Further details about these architectures can be found in Table 1, Appendix E.2.

The results are shown in Fig.5. The correlation between $R_f$ and $\epsilon_t$ is remarkably high (corr. coeff.[3] : 0.97), suggesting that generating low relative sensitivity to diffeomorphisms $R_f$ is important to obtain good performance. In Appendix E.3 we also report how changing the train set size $P$ affects the position of a network in the $(\epsilon_t, R_f)$ plane, for the four architectures considered in the previous section (Fig.18). We also show that our results are robust to changes of $\delta$, $c$ (Fig.21) and data sets (Fig.20).

What architectures enable a low $R_f$ value? The latter can be obtained with skip connections or not, and for quite different depths as indicated in Fig.5. Also, the same architecture (EfficientNetB0) trained by transfer learning from ImageNet – instead of directly on CIFAR10 – shows a large improvement both in performance and in diffeomorphisms invariance. Clearly, $R_f$ is much better predicted by $\epsilon_t$ than by the specific features of the architecture indicated in Fig.5.

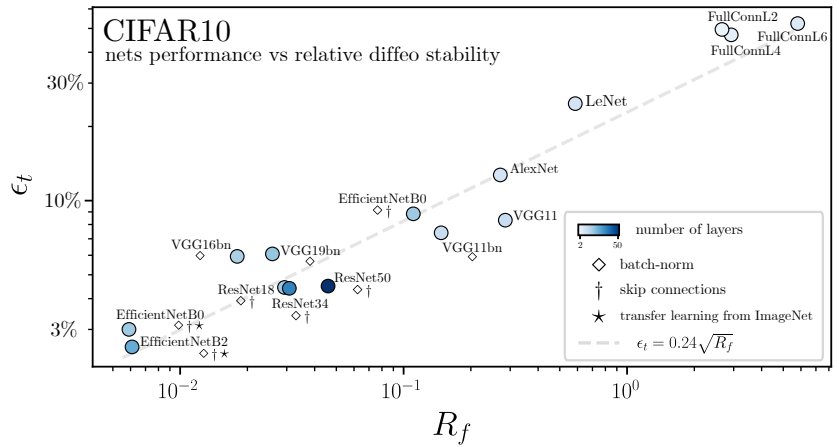

Figure 5: **Test error $\epsilon_t$ *vs.* relative stability to diffeomorphisms $R_f$** computed at $\delta = 1$ and $c = 3$ for common architectures when trained on the full 10-classes CIFAR10 dataset ($P = 50k$) with SGD and the cross-entropy loss; the EfficientNets achieving the best performance are trained by transfer learning from ImageNet ($\star$) – more details on the training procedures can be found in Appendix E.1. The color scale indicates depth, and the symbols the presence of batch-norm ($\diamond$) and skip connections ($\dagger$). Dashed grey line: power low fit $\epsilon_t \approx 0.2\sqrt{R_f}$. $R_f$ strongly correlates to $\epsilon_t$, much less so to depth or the presence of skip connections. *Statistics*: Each point is obtained by training 5 differently initialized networks; each network is then probed with 500 test samples in order to measure $R_f$. The results are obtained by log-averaging over single realizations. Error bars – omitted here – are shown in Fig.19, Appendix E.3.

## 5 Stability toward diffeomorphisms *vs.* noise

The relative stability to diffeomorphisms $R_f$ can be written as $R_f = D_f/G_f$ where $G_f$ characterizes the stability with respect to additive noise and $D_f$ the stability toward diffeomorphisms:

$$G_f = \frac{\langle \|f(x+\eta) - f(x)\|^2 \rangle_{x,\eta}}{\langle \|f(x) - f(z)\|^2 \rangle_{x,z}}, \qquad D_f = \frac{\langle \|f(\tau x) - f(x)\|^2 \rangle_{x,\tau}}{\langle \|f(x) - f(z)\|^2 \rangle_{x,z}}. \tag{5}$$

Here, we chose to normalize these stabilities with the variation of $f$ over the test set (to which both $x$ and $z$ belong), and $\eta$ is a random noise whose magnitude is prescribed as above. Stability toward additive noise has been studied previously in fully connected architectures Novak et al. (2018) and for CNNs as a function of spatial frequency in Tsuzuku and Sato (2019); Yin et al. (2019).

The decrease of $R_f$ with growing training set size $P$ could thus be due to an increase in the stability toward diffeomorphisms (i.e. $D_f$ decreasing with $P$) or a decrease of stability toward noise ($G_f$ increasing with $P$). To test these possibilities, we show in Fig.6 $G_f(P)$, $D_f(P)$ and $R_f(P)$ for MNIST, Fashion MNIST and CIFAR10 for two SOTA architectures. The central results are that (i) stability toward noise is always reduced for larger training sets. This observation is natural: when more data needs to be fitted, the function becomes rougher. (ii) Stability toward diffeomorphisms does not behave universally: it can increase with $P$ or decrease depending on the architecture and the training set. Additionally, $G_f$ and $D_f$ alone show a much smaller correlation with performance than $R_f$ – see Figs.15,16,17 in Appendix E.3.

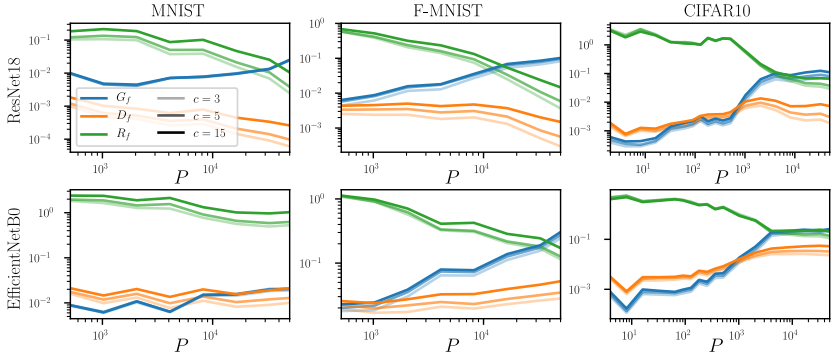

Figure 6: **Stability toward Gaussian noise ($G_f$) and diffeomorphisms ($D_f$) alone, and the relative stability** $R_f$. Columns correspond to different data-sets (MNIST, FashionMNIST and CIFAR10) and rows to architectures (ResNet18 and EfficientNetB0). Each panel reports $G_f$ (blue), $D_f$ (orange) and $R_f$ (green) as a function of $P$ and for different cut-off values $c$, as indicated in the legend. *Statistics:* cf. caption in Fig.3. Error bars – omitted here – are shown in Fig.22, Appendix E.3.

## 6 A minimal model for learning invariants

In this section, we discuss the simplest model of invariance in data where stability to transformation builds up, that can be compared with our observations of $R_f$ above. Specifically, we consider the "stripe" model Paccolat et al. (2021b), corresponding to a binary classification task for Gaussian-distributed data points $x = (x_\parallel, x_\perp)$ where the label function depends only on one direction in data space, namely $y(x) = y(x_\parallel)$. Layers of $y = +1$ and $y = -1$ regions alternate along the direction $x_\parallel$, separated by parallel planes. Hence, the data present $d-1$ invariant directions in input-space denoted by $x_\perp$ as illustrated in Fig.7-left.

When this model is learnt by a one-hidden-layer fully connected net, the first layer of weights can be shown to align with the informative direction Paccolat et al. (2021a). The projection of these weights on the orthogonal space vanishes with the training set size $P$ as $1/\sqrt{P}$, an effect induced by the sampling noise associated to finite training sets.

---

[3]Correlation coefficient: $\frac{\mathrm{Cov}(\log \epsilon_t, \log R_f)}{\sqrt{\mathrm{Var}(\log \epsilon_t)\mathrm{Var}(\log R_f)}}$.

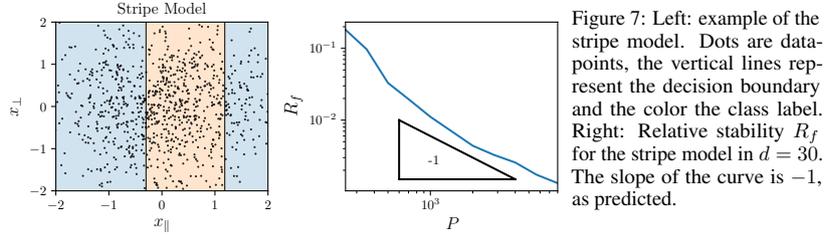

Figure 7: Left: example of the stripe model. Dots are datapoints, the vertical lines represent the decision boundary and the color the class label. Right: Relative stability $R_f$ for the stripe model in $d = 30$. The slope of the curve is $-1$, as predicted.

In this model, $R_f$ can be defined as:

$$R_f = \frac{\langle \|f(x_\parallel, x_\perp + \nu) - f(x_\parallel, x_\perp)\|^2 \rangle_{x,\nu}}{\langle \|f(x + \eta) - f(x)\|^2 \rangle_{x,\eta}}, \tag{6}$$

where we made explicit the dependence of $f$ on the two linear subspaces. Here, the isotropic noise $\nu$ is added only in the invariant directions. Again, we impose $\|\eta\| = \|\nu\|$. $R_f(P)$ is shown in Fig. 7-right. We observe that $R_f(P) \sim P^{-1}$, as expected from the weight alignment mentioned above.

Interestingly, Fig. 3 for CIFAR10 and SOTA architectures support that the $1/P$ behavior is compatible with the observations for some range of $P$. In Appendix E.3, Fig. 13, we show analogous results for MNIST and Fashion-MNIST. We observe the $1/P$ power-law scaling for ResNets. It suggests that for these architectures, learning to become invariant to diffeomorphisms may be limited by a naive measure of sampling noise as well. By contrast for EfficientNets, in which the decrease in $R_f$ is more limited, a $1/P$ behavior cannot be identified.

## 7 Discussion

A common belief is that stability to random noise (small $G_f$) and to diffeomorphisms (small $D_f$) are desirable properties of neural nets. Its underlying assumption is that the true data label mildly depends on such transformations when they are small. Our observations suggest an alternative view:

1. Figs. 6,16: better predictors are more sensitive to small perturbations in input space.

2. As a consequence, the notion that predictors are especially insensitive to diffeomorphisms is not captured by stability alone, but rather by the relative stability $R_f = D_f/G_f$.

3. We propose the following interpretation of Fig. 5: to perform well, the predictor must build large gradients in input space near the decision boundary – leading to a large $G_f$ overall. Networks that are relatively insensitive to diffeomorphisms (small $R_f$) can discover with less data that strong gradients must be there and generalize them to larger regions of input space, improving performance and increasing $G_f$.

This last point can be illustrated in the simple model of Section 6, see Fig. 7-left panel. Imagine two data points of different labels falling close to the – e.g. – left true decision boundary. These two points can be far from each other if their orthogonal coordinates differ. Yet, if $R_f = 0$ (now defined in Eq. 6), then the output does not depend on the orthogonal coordinates, and it will need to build a strong gradient – in input space – along the parallel coordinate to fit these two data. This strong gradient will exist throughout that entire decision boundary, improving performance but also increasing $G_f$. Instead, if $R_f = 1$, fitting these two data will not lead to a strong gradient, since they can be far from each other in input space. Beyond this intuition, in this model decreasing $R_f$ can quantitatively be shown to increase performance, see Paccolat et al. (2021b).

## 8    Conclusion

We have introduced a novel empirical framework to characterize how deep nets become invariant to diffeomorphisms. It is jointly based on a maximum-entropy distribution for diffeomorphisms, and on the realization that stability of these transformations relative to generic ones $R_f$ strongly correlates to performance, instead of just the diffeomorphisms stability considered in the past.

The ensemble of smooth deformations we introduced may have interesting applications. It could serve as a complement to traditional data-augmentation techniques (whose effect on relative stability is discussed in Fig.12 of the Appendix). A similar idea is present in Hauberg et al. (2016); Shen et al. (2020) but our deformations have the advantage of being easier to sample and data agnostic. Moreover, the ensemble could be used to build adversarial attacks along smooth transformations, in the spirit of Alaifari et al. (2018); Engstrom et al. (2019); Kanbak et al. (2018). It would be interesting to test if networks robust to such attacks are more stable in relative terms, and how such robustness affects their performance.

Finally, the tight correlation between relative stability $R_f$ and test error $\epsilon_t$ suggests that if a predictor displays a given $R_f$, its performance may be bounded from below. The relationships we observe $\epsilon_t(R_f)$ may then be indicative of this bound, which would be a fundamental property of a given data set. Can it be predicted in terms of simpler properties of the data? Introducing simplified models of data with controlled stability to diffeomorphisms beyond the toy model of Section 6 would be useful to investigate this key question.

## Acknowledgements

We thank Alberto Bietti, Joan Bruna, Francesco Cagnetta, Pascal Frossard, Jonas Paccolat, Antonio Sclocchi and Umberto M. Tomasini for helpful discussions. This work was supported by a grant from the Simons Foundation (#454953 Matthieu Wyart).

## A  Maximum entropy calculation

Under the constraint on the borders, $\tau_u$ and $\tau_v$ can be expressed in a real Fourier basis as in Eq.2. By injecting this form into $\|\nabla\tau\|^2$ we obtain:

$$\|\nabla\tau\|^2 = \frac{\pi^2}{4}\sum_{i,j\in\mathbb{N}^+}(C_{ij}^2 + D_{ij}^2)(i^2 + j^2) \qquad (7)$$

where $D_{ij}$ are the Fourier coefficients of $\tau_v$. We aim at computing the probability distributions that maximize their entropy while keeping the expectation value of $\|\nabla\tau\|^2$ fixed. Since we have a sum of quadratic random variables, the equipartition theorem Beale (1996) applies: the distributions are normal and every quadratic term contributes in average equally to $\|\nabla\tau\|^2$. Thus, the variance of the coefficients follows $\frac{T}{i^2+j^2}$ where the parameter $T$ determines the magnitude of the diffeomorphism.

## B  Boundaries of studied diffeomorphisms

**Average pixel displacement magnitude** $\delta$   We derive here the large-$c$ asymptotic behavior of $\delta$ (Eq.3). This is defined as the average square norm of the displacement field, in pixel units:

$$
\begin{aligned}
\delta^2 &= n^2 \int_{[0,1]^2} \|\tau(u,v)\|^2 du\,dv \\
&= 2Tn^2 \sum_{i^2+j^2\leq c^2} \frac{1}{i^2+j^2} \int_{[0,1]^2} \sin^2(i\pi u)\sin^2(j\pi v)du\,dv \\
&= \frac{Tn^2}{2} \sum_{i^2+j^2\leq c^2} \frac{1}{i^2+j^2} \\
&\approx \frac{Tn^2}{2} \int_{1\leq x^2+y^2\leq c^2} \frac{1}{x^2+y^2} dx\,dy \\
&= \frac{\pi Tn^2}{4} \int_1^c \frac{1}{r} dr \\
&= \frac{\pi}{4} n^2 T \log c,
\end{aligned}
$$

where we approximated the sum with an integral, in the third step. The asymptotic relations for $\|\nabla\tau\|$ that are reported in the main text are computed in a similar fashion. In Fig.8, we check the agreement between asymptotic prediction and empirical measurements. If $\delta \ll 1$, our results strongly depend on the choice of interpolation method. To avoid it, we only consider conditions for which $\delta \geq 1/2$, leading to

$$T > \frac{1}{\pi n^2 \log c}. \qquad (8)$$

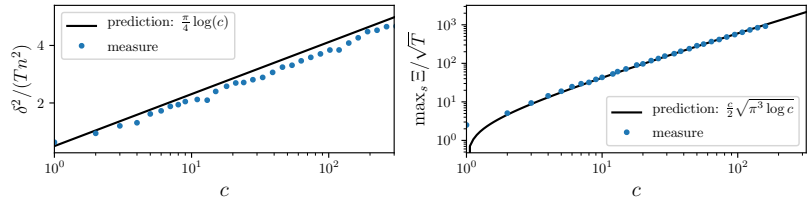

Figure 8: Left: The characteristic displacement $\delta(c,T)$ is observed to follow $\delta^2 \simeq \frac{\pi}{4}n^2T\log c$. Right: measurement of $\max_s \Xi$ supporting Eq.13.

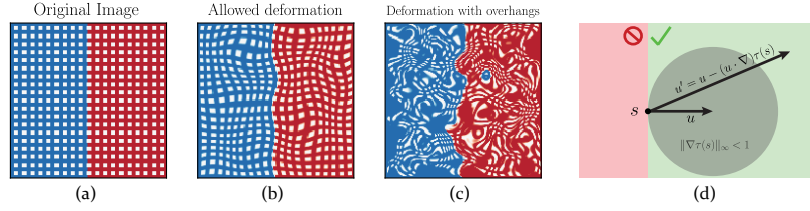

Figure 9: (a) Idealized image at $T = 0$. (b) Diffeomorphism of the image. (c) Deformation of the image at large $T$: colors get mixed-up together, shapes are not preserved anymore. (d) Allowed region for vector transformations under $\tau$. For any point in the image $s$ and any direction $u$, only displacement fields for which all the deformed direction $u'$ is non-zero generate diffeomorphisms. The bound in Eq.12 ($u' \cdot u > 0$) correspond to the green region. The gray disc corresponds to the bound $\|\nabla \tau\|_\infty < 1$.

**Condition for diffeomorphism in the $(T, c)$ plane** For a given value of $c$, there exists a temperature scale beyond which the transformation is not injective anymore, affecting the topology of the image and creating spurious boundaries, see Fig.9a-c for an illustration. Specifically, consider a curve passing by the point $s$ in the deformed image. Its tangent direction is $u$ at the point $s$. When going back to the original image ($s' = s - \tau(s)$) the curve gets deformed and its tangent becomes

$$u' = u - (u \cdot \nabla)\tau(s). \tag{9}$$

A smooth deformation is bijective iff all deformed curves remain curves which is equivalent to have non-zero tangents everywhere

$$\forall s, u \neq 0 \quad \|u'\| \neq 0. \tag{10}$$

Imposing $\|u'\| \neq 0$ does not give us any constraint on $\tau$. Therefore, we constraint $\tau$ a bit more and allow only displacement fields such that $u \cdot u' > 0$, which is a sufficient condition for Eq.10 to be satisfied – cf. Fig. 9d. By extremizing over $u$, this condition translates into

$$\frac{1}{2}\left( \sqrt{(\partial_x \tau_x - \partial_y \tau_y)^2 + (\partial_x \tau_y + \partial_y \tau_x)^2} - \partial_x \tau_x - \partial_y \tau_y \right) < 1 \tag{11}$$

or, equivalently,

$$\Xi = \frac{1}{2}\left( \sqrt{\|\nabla \tau\|^2 - 2\det(\nabla \tau)} - \text{Tr}(\nabla \tau) \right) < 1, \tag{12}$$

were we identified by $\Xi$ the l.h.s. of the inequality. We find that the median of the maximum of $\Xi$ over all the image ($\|\Xi(s)\|_\infty$) can be approximated by (see Fig.8b):

$$\max_s \Xi \simeq \frac{c}{2}\sqrt{\pi^3 T \log c}. \tag{13}$$

The resulting constraint on $T$ reads

$$T < \frac{4}{\pi^3 c^2 \log c}. \tag{14}$$

## C Interpolation methods

When a deformation is applied to an image $x$, each of its pixels gets mapped, from the original pixels grid, to new positions generally outside of the grid itself – cf. Fig. 9a-b. A procedure (interpolation method) needs to be defined to project the deformed image back into the original grid.

For simplicity of notation, we describe interpolation methods considering the square $[0, 1]^2$ as the region in between four pixels – see an illustration in Fig. 10a. We propose here two different ways to interpolate between pixels and then check that our measurements do not depend on the specific method considered.

**Bi-linear Interpolation** The bi-linear interpolation consists, as the name suggests, of two steps of linear interpolation, one on the horizontal, and one on the vertical direction – Fig. 10b. If we look at the square $[0, 1]^2$ and we apply a deformation $\tau$ such that $(0, 0) \mapsto (u, v)$, we have

$$x(u, v) = x(0, 0)(1 - u)(1 - v) + x(1, 0)u(1 - v) + x(0, 1)(1 - u)v + x(1, 1)uv. \tag{15}$$

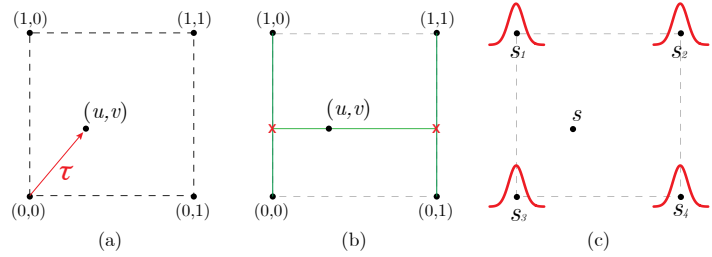

Figure 10: (a) We consider the region between four pixels as the square $[0, 1]^2$ where, after the application of a deformation $\tau$, the pixel $(0,0)$ is mapped into $(u, v)$. (b) **Bi-linear interpolation**: the value of $x$ in $(u, v)$ is computed by two steps of linear interpolation. First, we compute $x$ in the red crosses, by averaging values on the vertical axis. Then, a line interpolates horizontally the values in the red crosses to give the result. (c) **Gaussian interpolation**: we denote by $s_i$ the pixel positions in the original grid. The interpolated value of $s$ in any point of the image is given by a weighted sum of $n \times n$ Gaussian centered in each $s_i$ – in red.

**Gaussian Interpolation** In this case, a Gaussian function[4] is placed on top of each point in the grid – cf. Fig.10. The pixel intensity $x$ can be evaluated at any point outside the grid by computing

$$x(s) = \frac{\sum_i x(s_i) G(s - s_i)}{\sum_i G(s - s_i)}. \tag{16}$$

In order to fix the standard deviation $\sigma$ of $G$, we introduce the *participation ratio* $n$. Given $\Psi_i = G(s, s_i)|_{s=(0.5, 0.5)}$, we define

$$n = \frac{\left( \sum_i \Psi_i^2 \right)^2}{\sum_i \Psi_i^4}. \tag{17}$$

The participation ratio is a measure of how many pixels contribute to the value of a new pixel, which results from interpolation. We fix $\sigma$ in such a way that the participation ratio for the Gaussian interpolation matches the one for the bi-linear ($n = 4$), when the new pixel is equidistant from the four pixels around. This gives $\sigma = 0.4715$.

Notice that this interpolation method is such that it applies a Gaussian smoothing of the image even if $\tau$ is the identity. Consequently, when computing observables for $f$ with the Gaussian interpolation, we always compare $f(\tau x)$ to $f(\tilde{x})$, where $\tilde{x}$ is the smoothed version of $x$, in such a way that $f(\tau^{[T=0]} x) = f(\tilde{x})$.

**Empirical results dependence on interpolation** Finally, we checked to which extent our results are affected by the specific choice of interpolation method. In particular, blue and red colors in Figs.3, 13 correspond to bi-linear and Gaussian interpolation, respectively. The interpolation method only affects the results in the small displacement limit ($\delta \to 0$).

<u>Note</u>: throughout the paper, if not specified otherwise, bi-linear interpolation is employed.

---

[4] $G(s) = (2\pi\sigma^2)^{-1/2} e^{-s^2/2\sigma^2}$.

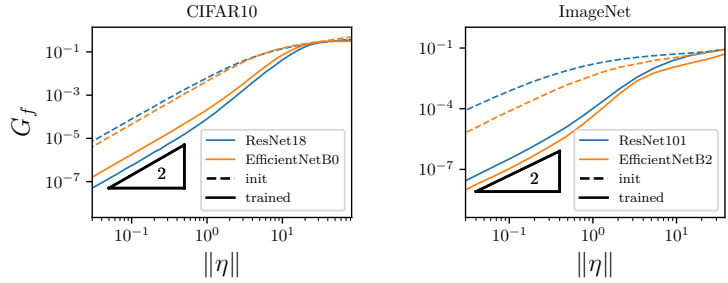

Figure 11: **Stability to isotropic noise** $G_f$ as a function of the noise magnitude $\|\eta\|$ for CIFAR10 (left) and ImageNet (right). The color corresponds to two different classes of SOTA architecture: ResNet and EfficientNet. The slope 2 at small $\|\eta\|$ identifies the linear regime. For larger noise magnitudes, non-linearities appear.

We introduced in Section 5 the stability toward additive noise:

$$G_f = \frac{\langle \|f(x+\eta) - f(x)\|^2 \rangle_{x,\eta}}{\langle \|f(x) - f(z)\|^2 \rangle_{x,z}}. \tag{18}$$

We study here the dependence of $G_f$ on the noise magnitude $\|\eta\|$. In the $\eta \to 0$ limit, we expect the network function to behave as its first-order Taylor expansion, leading to $G_f \propto \|\eta\|^2$. Hence, for small noise, $G_f$ gives an estimate of the average magnitude of the gradient of $f$ in a random direction $\eta$.

**Empirical results** Measurements of $G_f$ on SOTA nets trained on benchmark data-sets are shown in Figure 11. We observe that the effect of non-linearities start to be significant around $\|\eta\| = 1$. For large values of the noise – i.e. far away from data-points – the average gradient of $f$ does not change with training.

## E Numerical experiments

In this Appendix, we provide details on the training procedure, on the different architectures employed and some additional experimental results.

### E.1 Image classification training set-up:

- ○ Trainings are performed in `PyTorch`, the code can be found here github.com/leonardopetrini/diffeo-sota.
- ○ Loss function: cross-entropy.
- ○ Batch size: 128.
- ○ Dynamics:
  - – Fully connected nets: ADAM with `learning rate` $= 0.1$ and no scheduling.
  - – Transfer learning: SGD with `learning rate` $= 10^{-2}$ for the last layer and $10^{-3}$ for the rest of the network, `momentum` $= 0.9$ and `weight decay` $= 10^{-3}$. Both learning rates decay exponentially during training with a factor $\gamma = 0.975$.
  - – All the other networks are trained with SGD with `learning rate` $= 0.1$, `momentum` $= 0.9$ and `weight decay` $= 5 \times 10^{-4}$. The learning rate follows a cosine annealing scheduling Loshchilov and Hutter (2016).

- Early-stopping is performed – i.e. results shown are computed with the network obtaining the best validation accuracy out of 250 training epochs.

- For the experiments involving a training on a subset of the training date of size $P < P_{\max}$, the total number of epochs is accordingly re-scaled in order to keep constant the total number of optimizer steps.

- Standard data augmentation is employed: different random translations and horizontal flips of the input images are generated at each epoch. As a safety check, we verify that the invariance learnt by the nets is not purely due to such augmentation (Fig.12).

- Experiments are run on 16 GPUs NVIDIA V100. Individual trainings run in $\sim 1$ hour of wall time. We estimate a total of a few thousands hours of computing time for running the preliminary and actual experiments present in this work.

The stripe model is trained with an approximation of gradient flow introduced in Geiger et al. (2020), see Paccolat et al. (2021a) for details.

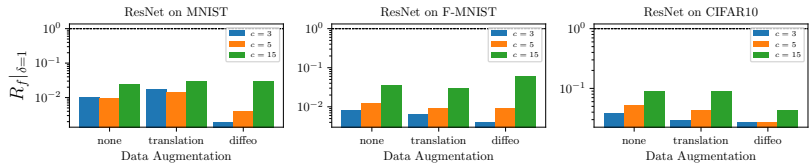

Figure 12: **Effect of data augmentation on $R_f$.** Relative stability to diffeomorphisms $R_f$ after training with different data augmentations: "none" (1st group of bars in each plot) for no data augmentation, "translation" (2nd bars) corresponds to training on randomly translated (by 4 pixels) and cropped inputs, and "diffeo" (3rd bars) to training on randomly deformed images with max-entropy diffeomorphisms ($T = 10^{-2}, c = 1$). Results are averaged over 5 trainings of ResNet18 on MNIST (left), FashionMNIST (center), CIFAR10 (right). Colors indicate different cut-off values when probing the trained networks. Different augmentations have a small quantitative, and no qualitative effect on the results. As expected, augmenting the input images with smooth deformations makes the net more invariant to such transformations.

**A note on computing stabilities at init. in presence of batch-norm**   We recall that batch-norm (BN) can work in either of two modes: *training* and *evaluation*. During training, BN computes the mean and variance on the current batch and uses them to normalize the output of a given layer. At the same time, it keeps memory of the running statistics on such batches, and this is used for the normalization steps at inference time (evaluation mode). When probing a network at initialization for computing stabilities, we put the network in evaluation mode, except for batch-norm (BN), which operates in train mode. This is because BN running mean and variance are initialized to 0 and 1, in such a way that its evaluation mode at initialization would correspond to not having BN at all, compromising the input signal propagation in deep architectures.

### E.2 Networks architectures

All networks implementations can be found at github.com/leonardopetrini/diffeo-sota/tree/main/models. In Table 1, we report salient features of the network architectures considered.

Table 1: **Network architectures, main characteristics.** We list here (columns) the classes of net architectures used throughout the paper specifying some salient features (depth, number of parameters, etc...) for each of them.

| features | FullConn | LeNet LeCun et al. (1989) | AlexNet Krizhevsky et al. (2012) |
|---|---|---|---|
| depth | 2, 4, 6 | 5 | 8 |
| num. parameters | 200k | 62k | 23 M |
| FC layers | 2, 4, 6 | 3 | 3 |
| activation | ReLU | ReLU | ReLU |
| pooling | / | max | max |
| dropout | / | / | yes |
| batch norm | / | / | / |
| skip connections | / | / | / |

| features | VGG Simonyan and Zisserman (2015) | ResNet He et al. (2016) | EfficientNetB0-2 Tan and Le (2019) |
|---|---|---|---|
| depth | 11, 16, 19 | 18, 34, 50 | 18, 25 |
| num. parameters | 9-20 M | 11-24 M | 5, 9 M |
| FC layers | 1 | 1 | 1 |
| activation | ReLU | ReLU | swish |
| pooling | max | avg. (last layer only) | avg. (last layer only) |
| dropout | / | / | yes + dropconnect |
| batch norm | if 'bn' in name | yes | yes |
| skip connections | / | yes | yes (inv. residuals) |

### E.3 Additional figures

We present here:

- Fig.13: $R_f$ as a function of $P$ for MNIST and FashionMNIST with the corresponding predicted slope, omitted in the main text.
- Fig.14: Relative diffeomorphisms stability $R_f$ as a function of depth for simple and deep nets.
- Figs15,16: diffeomorphisms and inverse of the Gaussian stability $D_f$ and $1/G_f$ *vs.* test error for CIFAR10 and the set of architectures considered in Section 4.
- Fig.17: $D_f$, $1/G_f$ and $R_f$ when using the mean in place of the median for computing averages $\langle \cdot \rangle$.
- Fig.18: curves in the $(\epsilon_t, R_f)$ plane when varying the training set size $P$ for FullyConnL4, LeNet, ResNet18 and EfficientNetB0.
- Figs19, 22: error estimates for the main quantities of interest – often omitted in the main text for the sake of figures' clarity.

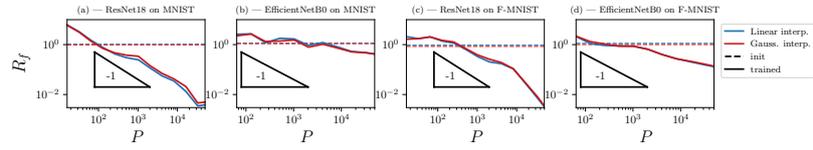

Figure 13: **Relative stability to diffeomorphisms** $R_f(P)$ **at** $\delta = 1$. Analogous to Figure 3-right but here we have MNIST (a-b) and FashionMNIST (c-d) in place of CIFAR10. Stability monotonically decreases with $P$. The triangles give a reference for the predicted slope in the stripe model – i.e. $R_f \sim P^{-1}$ – see Section 6. The slopes in case of ResNets are compatible with the prediction. For EfficientNets, the second panel of Fig.3 suggests that stability to diffeomorphisms is less important. Here, we also see that it builds up more slowly when increasing the training set size. Finally, blue and red colors indicate different interpolation methods used for generating image deformations, as discussed in Appendix C. Results are not affected by this choice.

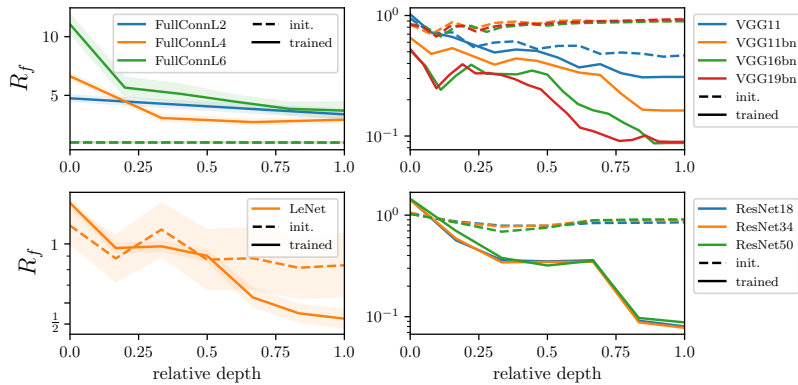

Figure 14: **Relative stability to diffeomorphisms as a function of depth.** $R_f$ as a function of the layers relative depth (i.e. $\frac{\text{current layer depth}}{\text{total depth}}$) where "0" identifies the output of the 1st layer and "1" the last. The relative stability is measured for the output of layers (or blocks of layers) inside the nets for simple architectures (1st column) and deep ones (2nd column) at initialization (dashed) and after training (full lines). All nets are trained on the full CIFAR10 dataset. $R_{f_0} \approx 1$ independently of depth at initialization while it decreases monotonically as a function of depth after training. *Statistics:* Each point is obtained by training 5 differently initialized networks; each network is then probed with 500 test samples in order to measure $R_f$. The results are obtained by log-averaging over single realizations.

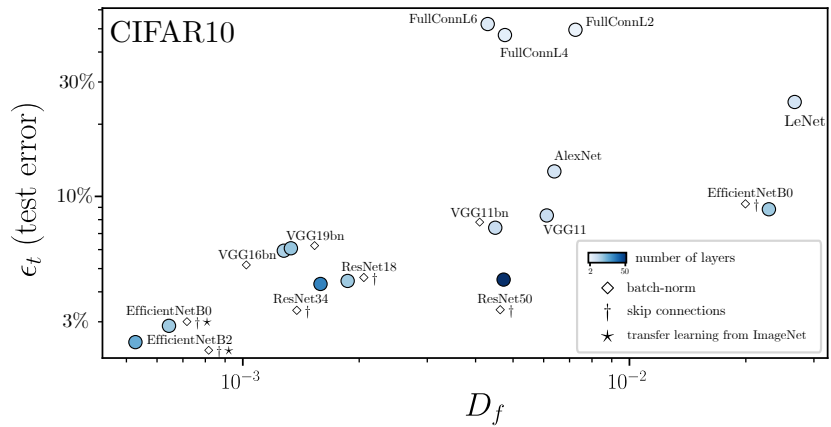

Figure 15: **Test error $\epsilon_t$ vs. stability to diffeomorphisms** $D_f$ for common architectures when trained on the full 10-classes CIFAR10 dataset ($P = 50k$) with SGD and the cross-entropy loss; the EfficientNets achieving the best performance are trained by transfer learning from ImageNet ($\star$) – more details on the training procedures can be found in Appendix E.1. The color scale indicates depth, and the symbols the presence of batch-norm ($\diamond$) and skip connections ($\dagger$). $D_f$ correlation with $\epsilon_t$ (corr. coeff.: 0.62) is much smaller than the one measured for $R_f$ – see Fig.3. *Statistics:* Each point is obtained by training 5 differently initialized networks; each network is then probed with 500 test samples in order to measure $D_f$. The results are obtained by log-averaging over single realizations.

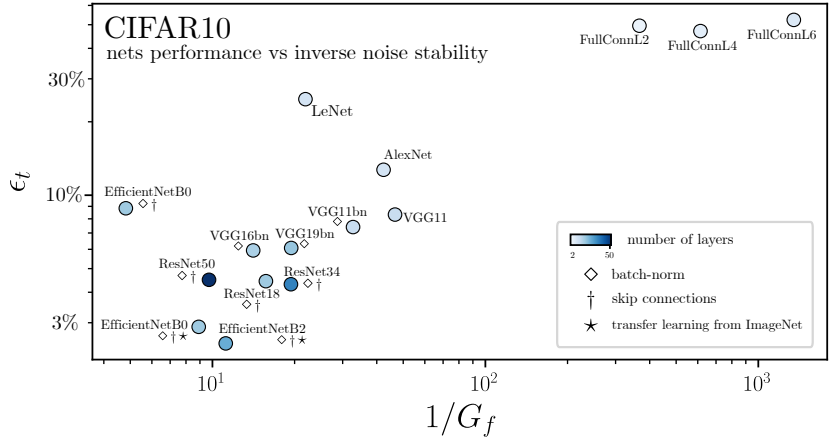

Figure 16: **Test error $\epsilon_t$ vs. inverse of stability to noise** $1/G_f$ for common architectures when trained on the full 10-classes CIFAR10 dataset ($P = 50k$) with SGD and the cross-entropy loss; the EfficientNets achieving the best performance are trained by transfer learning from ImageNet ($\star$) – more details on the training procedures can be found in Appendix E.1. The color scale indicates depth, and the symbols the presence of batch-norm ($\diamond$) and skip connections ($\dagger$). $G_f$ correlation with $\epsilon_t$ (corr. coeff.: 0.85) is less important than the one measured for $R_f$ – see Fig.3. *Statistics:* Each point is obtained by training 5 differently initialized networks; each network is then probed with 500 test samples in order to measure $G_f$. The results are obtained by log-averaging over single realizations.

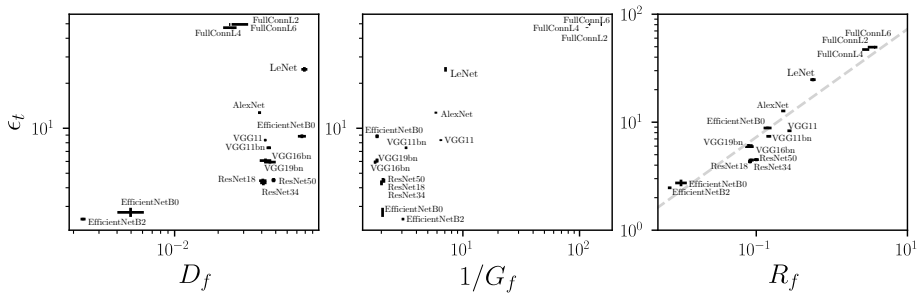

Figure 17: **Test error $\epsilon_t$ vs. $D_f$, $1/G_f$ and $R_f$ where $\langle \cdot \rangle$ is the mean.** Analogous to Figs 15-19, we use here the mean instead of the median to compute averages over samples and transformations.

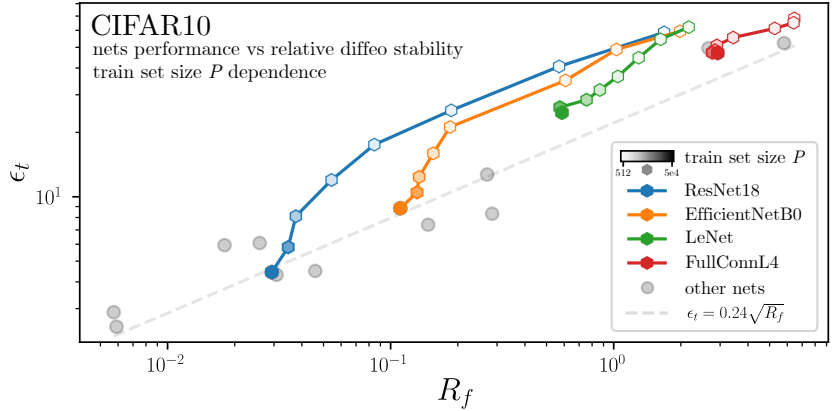

Figure 18: **Test error $\epsilon_t$ vs. relative stability to diffeomorphisms $R_f$ for different training set sizes $P$.** Same data as Fig.5, we report here curves corresponding to training on different set sizes for 4 architectures. The other architectures considered together with the power-law fit are left in background. For a small training set, CNNs behave similarly. *Statistics:* Each point is obtained by training 5 differently initialized networks; each network is then probed with 500 test samples in order to measure $R_f$. The results are obtained by log-averaging over single realizations.

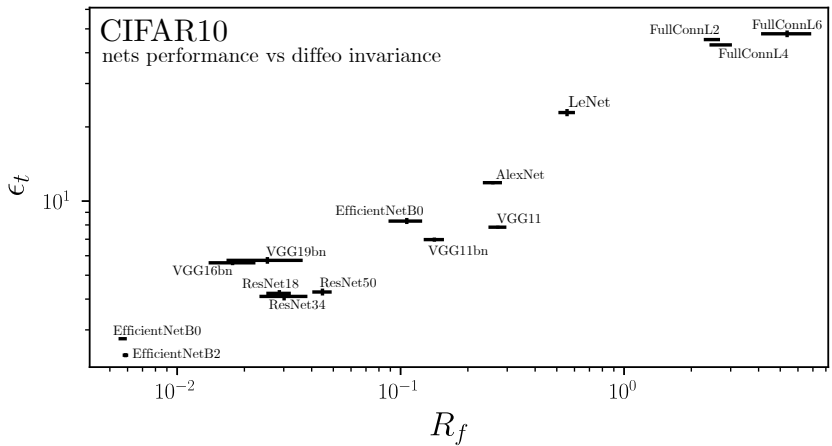

Figure 19: **Test error $\epsilon_t$ vs. relative stability to diffeomorphisms $R_f$ with error estimates.** Same data as Fig.5, we report error bars here. *Statistics:* Each point is obtained by training 5 differently initialized networks; each network is then probed with 500 test samples in order to measure $R_f$. The results are obtained by log-averaging over single realizations.

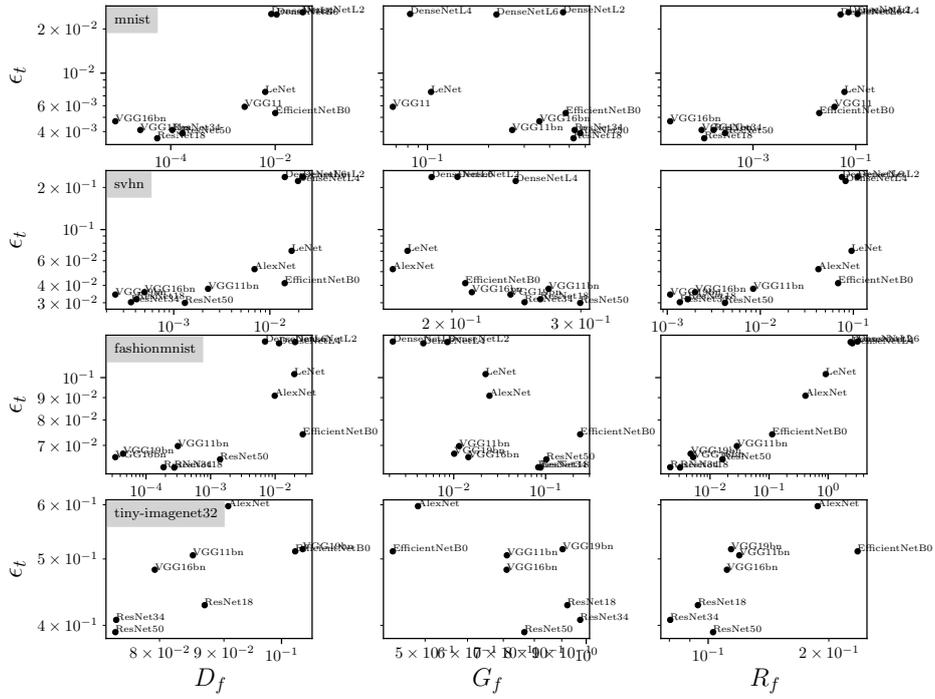

Figure 20: **Test error $\epsilon_t$ vs. $D_f$, $G_f$ and $R_f$ (on the columns) for different data sets (on the rows).** The corresponding correlation coefficients are shown in Table 2. Lines 1-2: MNIST and SVHN both contain images of digits and show a similar $\epsilon_t(R_f)$. Line 3: FashionMNIST results are comparable to the CIFAR10 ones shown in the main text. Line 4: Tiny ImageNet32 is a re-scaled (32x32 pixels) version of ImageNet with 200 classes and 100'000 training points. The task is harder than the other data sets and is such that we could not train simple networks (FC, LeNet) on it – i.e. the loss stays $\mathcal{O}(1)$ throughout training – so these are not reported here.

Table 2: **Test error vs. stability: correlation coefficients for different data sets.**

| data-set | $D_f$ | $G_f$ | $R_f$ |
|---|---|---|---|
| MNIST | 0.71 | -0.43 | 0.75 |
| SVHN | 0.87 | -0.28 | 0.81 |
| FashionMNIST | 0.72 | -0.68 | 0.94 |
| Tiny ImageNet | 0.69 | -0.66 | 0.74 |

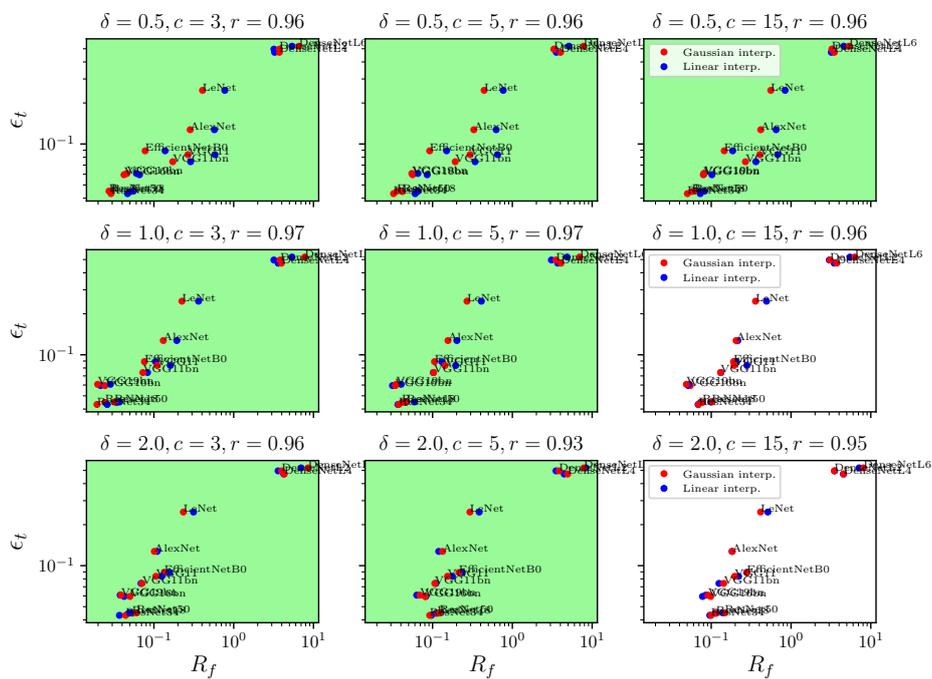

Figure 21: **Test error** $\epsilon_t$ **vs.** $D_f$, $G_f$ **and** $R_f$ **for CIFAR10 and varying** $\delta$ **and cut-off** $c$**.** Titles report the values of the varying parameters together with corr. coeffs. Parameters corresponding to allowed diffeo are indicated by the green background. Red and blue colors correspond to different interpolation methods. Overall, results are robust when varying these parameters.

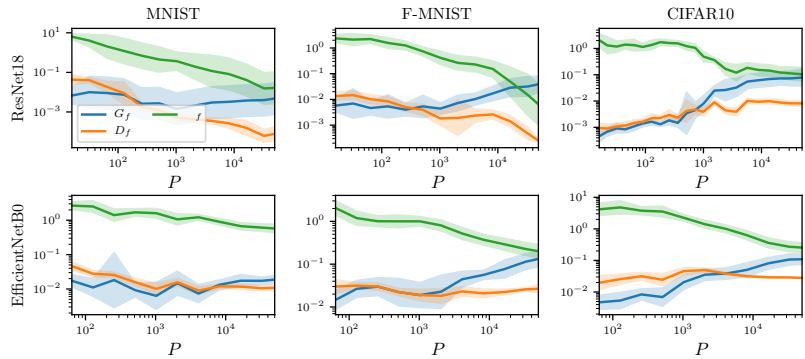

Figure 22: **Stability toward Gaussian noise ($G_f$) and diffeomorphisms ($D_f$) alone, and the relative stability $R_f$ with the relative errors.** Analogous to Fig.6 in which error estimates are omitted to favour clarity. Here we fix the cut-off to $c = 3$ and show error estimates instead. Columns correspond to different data-sets (MNIST, FashionMNIST and CIFAR10) and rows to architectures (ResNet18 and EfficientNetB0). Each panel reports $G_f$ (blue), $D_f$ (orange) and $R_f$ (green) as a function of $P$ and for different cut-off values $c$, as indicated in the legend. *Statistics*: Each point in the graphs is obtained by training 16 differently initialized networks on 16 different subsets of the data-sets; each network is then probed with 500 test samples in order to measure stability to diffeomorphisms and Gaussian noise. The resulting $R_f$ is obtained by log-averaging the results from single realizations. As we are plotting quantities in log scale, we report the relative error (shaded).

# 4 When Feature Learning Fails: Deformation Invariance deteriorates with training in Fully-Connected Networks

The following paper is the preprint version of Petrini et al. (2022) published in *Advances in Neural Information Processing Systems.*

**Candidate contributions**    The candidate contributed to all discussions and led the experimental part of the paper. F Cagnetta led the theoretical part.

# Learning sparse features can lead to overfitting in neural networks

**Leonardo Petrini** [*]
Institute of Physics
École Polytechnique Fédérale de Lausanne
`leonardo.petrini@epfl.ch`

**Francesco Cagnetta** [*]
Institute of Physics
École Polytechnique Fédérale de Lausanne
`francesco.cagnetta@epfl.ch`

**Eric Vanden-Eijnden**
Courant Institute of Mathematical Sciences
New York University
`eve2@cims.nyu.edu`

**Matthieu Wyart**
Institute of Physics
École Polytechnique Fédérale de Lausanne
`matthieu.wyart@epfl.ch`

## Abstract

It is widely believed that the success of deep networks lies in their ability to learn a meaningful representation of the features of the data. Yet, understanding when and how this feature learning improves performance remains a challenge: for example, it is beneficial for modern architectures trained to classify images, whereas it is detrimental for fully-connected networks trained on the same data. Here we propose an explanation for this puzzle, by showing that feature learning can perform worse than lazy training (via random feature kernel or the NTK) as the former can lead to a sparser neural representation. Although sparsity is known to be essential for learning anisotropic data, it is detrimental when the target function is constant or smooth along certain directions of input space. We illustrate this phenomenon in two settings: *(i)* regression of Gaussian random functions on the $d$-dimensional unit sphere and *(ii)* classification of benchmark datasets of images. For *(i)*, we compute the scaling of the generalization error with the number of training points and show that methods that do not learn features generalize better, even when the dimension of the input space is large. For *(ii)*, we show empirically that learning features can indeed lead to sparse and thereby less smooth representations of the image predictors. This fact is plausibly responsible for deteriorating the performance, which is known to be correlated with smoothness along diffeomorphisms.

## 1 Introduction

Neural networks are responsible for a technological revolution in a variety of machine learning tasks. Many such tasks require learning functions of high-dimensional inputs from a finite set of examples, thus should be generically hard due to the *curse of dimensionality* [1, 2]: the exponent that controls the scaling of the generalization error with the number of training examples is inversely proportional to the input dimension $d$. For instance, for standard image classification tasks with $d$ ranging in $10^3 \div 10^5$, such exponent should be practically vanishing, contrary to what is observed in practice [3]. In this respect, understanding the success of neural networks is still an open question. A popular explanation is that, during training, neurons adapt to features in the data that are relevant for the task [4], effectively reducing the input dimension and making the problem tractable [5, 6, 7]. However, understanding quantitatively if this intuition is true and how it depends on the structure of the task remains a challenge.

---
[*]Equal contribution (a coin was flipped).

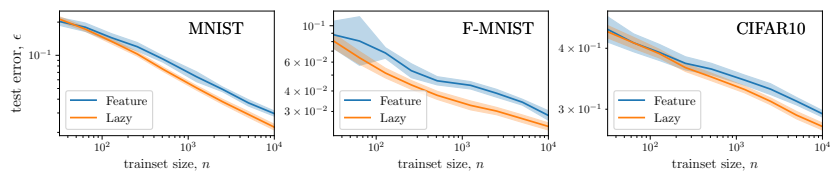

Figure 1: **Feature vs. Lazy in image classification.** Generalization error as a function of the training-set size $n$ for infinite-width fully-connected networks (FCNs) trained in the feature (blue) and lazy regime (orange). In the latter case the limit is taken exactly by training an SVC algorithm with the analytical NTK [23]. In the former case, the infinite-width limit can be accurately approximated for these datasets by considering very wide nets ($H = 10^3$), and performing ensemble averaging on different initial conditions of the parameters as shown in [24, 25]. Panels correspond to different benchmark image datasets [26, 27, 28]. Results are averaged over 10 different initializations of the networks and datasets.

Recently much progress was made in characterizing the conditions which lead to features learning, in the overparameterized setting where networks generally perform best. When the initialization scale of the network parameters is large [8] one encounters the *lazy training regime*, where neural networks behave as kernel methods [9, 10] (coined Neural Tangent Kernel or NTK) and features are not learned. By contrast, when the initialization scale is small, a *feature learning regime* is found [11, 12, 13] where the network parameters evolve significantly during training. This limit is much less understood apart from very simple architectures, where it can be shown to lead to sparse representations where a limited number of neurons are active after training [14]. Such sparse representations can also be obtained by regularizing the weights during training [2, 15].

In terms of performance, most theoretical works have focused on fully-connected networks. For these architectures, feature learning was shown to significantly outperform lazy training [16, 17, 18, 19, 11] for certain tasks, including approximating a function which depends only on a subset or a linear combination of the input variables. However, when such primitive networks are trained on image datasets, learning features is detrimental [20, 21], as illustrated in Fig. 1 (see [19, Fig. 3] for the analogous plot in the case of a target function depending on just one of the input variables, where learning features is beneficial). A similar result was observed in simple models of data [22]. These facts are unexplained, yet central to understanding the implicit bias of the feature learning regime.

## 1.1 Our contribution

Our main contribution is to provide an account of the drawbacks of learning sparse representations based on the following set of ideas. Consider, for concreteness, an image classification problem: *(i)* images class varies little along smooth deformations of the image; *(ii)* due to that, tasks like image classification require a continuous distribution of neurons to be represented; *(iii)* thus, requiring sparsity can be detrimental for performance. We build our argument as follows.

- In order to find a quantitative description of the phenomenon, we start from the problem of regression of a random target function of controlled smoothness on the $d$-dimensional unit sphere, and study the property of the minimizers of the empirical loss with $n$ observations, both in the lazy and the feature learning regimes. More specifically, we consider two extreme limits—the NTK limit and mean-field limit—as representatives of lazy and feature regimes, respectively (section 2). Both these limits admit a simple formulation that allows us to predict generalization performances. In particular, our results on feature learning rely on solutions having an atomic support. This property can be justified for one-hidden-layer neural networks with ReLU activations and weight decay. Yet, we also find such a sparsity empirically using gradient descent in the absence of regularization, if weights are initialized to be small enough.

- We find that lazy training leads to smoother predictors than feature learning. As a result, lazy training outperforms feature learning when the target function is also sufficiently smooth. Otherwise, the performances of the two methods are comparable, in the sense that they display the same asymptotic decay of generalization error with the number of training examples. Our predictions are obtained from asymptotic arguments that we systematically back up with numerical studies.

- For image datasets, it is believed that diffeomorphisms of images are key transformations along which the predictor function should only mildly vary to obtain good performance [29]. From the results above, a natural explanation as to why lazy beats feature for fully connected networks is that it leads to predictors with smaller variations along diffeomorphisms. We confirm that this is indeed the case empirically on benchmark datasets.

Numerical experiments are performed in PyTorch [30], and the code for reproducing experiments is available online at github.com/pcsl-epfl/regressionsphere.

### 1.2 Related Work

The property that training ReLU networks in the feature regime leads to a sparse representation was observed empirically [31]. This property can be justified for one-hidden-layer networks by casting training as a L1 minimization problem [32, 2], then using a representer theorem [33, 15, 34]. This is analogous to what is commonly done in predictive sparse coding [35, 36, 37, 38].

Many works have investigated the benefits of learning sparse representations in neural networks. [2, 16, 17, 18, 19, 39, 40] study cases in which the true function only depends on a linear subspace of input space, and show that feature learning profitably capture such property. Even for more general problems, sparse representations of the data might emerge naturally during deep network training—a phenomenon coined *neural collapse* [41]. Similar sparsification phenomena, for instance, have been found to allow for learning convolutional layers from scratch [42, 43]. Our work builds on this body of literature by pointing out that learning sparse features can be detrimental, if the task does not allow for it.

There is currently no general framework to predict rigorously the learning curve exponent $\beta$ defined as $\epsilon(n) = \mathcal{O}(n^{-\beta})$ for kernels. Some of our asymptotic arguments can be obtained by other approximations, such as assuming that data points lie on a lattice in $\mathbb{R}^d$ [44], or by using the non-rigorous replica method of statistical physics [45, 46, 47]. In the case $d = 2$, we provide a more explicit mathematical formulation of our results, which leads to analytical results for certain kernels. We systematically back up our predictions with numerical tests as $d$ varies.

Finally, in the context of image classification, the connection between performance and 'stability' or smoothness toward small diffeomorphisms of the inputs has been conjectured by [29, 48]. Empirically, a strong correlation between these two quantities was shown to hold across various architectures for real datasets [49]. In that reference, it was found that fully connected networks lose their stability over training: here we show that this effect is much less pronounced in the lazy regime.

### 2 Problem and notation

**Task** We consider a supervised learning scenario with $n$ training points $\{\boldsymbol{x}_i\}_{i=1}^n$ uniformly drawn on the $d$-dimensional unit sphere $\mathbb{S}^{d-1}$. We assume that the target function $f^*$ is an isotropic Gaussian random process on $\mathbb{S}^{d-1}$ and control its statistics via the spectrum: by introducing the decomposition of $f^*$ into spherical harmonics (see App. A for definitions),

$$f^*(\boldsymbol{x}) = \sum_{k \geq 0} \sum_{\ell=1}^{\mathcal{N}_{k,d}} f_{k,\ell}^* Y_{k,\ell}(\boldsymbol{x}) \quad \text{with} \quad \mathbb{E}\left[f_{k,\ell}^*\right] = 0, \quad \mathbb{E}\left[f_{k,\ell}^* f_{k',\ell'}^*\right] = c_k \delta_{k,k'} \delta_{\ell,\ell'}. \tag{2.1}$$

We assume that all the $c_k$ with $k$ odd apart from $c_1$ vanish: this is required to guarantee that $f^*$ can be approximated as well as desired with a one-hidden-layer ReLU network with no biases, as discussed in App. A. We also assume that the non-zero $c_k$ decay as a power of $k$ for $k \gg 1$, $c_k \sim k^{-2\nu_t - (d-1)}$. The exponent $\nu_t > 0$ controls the (weak) differentiability of $f^*$ on the sphere (see App. A) and also the statistics of $f^*$ in real space:

$$\mathbb{E}\left[|f^*(\boldsymbol{x}) - f^*(\boldsymbol{y})|^2\right] = O\left(|\boldsymbol{x} - \boldsymbol{y}|^{2\nu_t}\right) = O\left((1 - \boldsymbol{x} \cdot \boldsymbol{y})^{\nu_t}\right) \quad \text{as} \quad \boldsymbol{x} \to \boldsymbol{y}. \tag{2.2}$$

Examples of such a target function for $d = 3$ and different values of $\nu_t$ are reported in Fig. 2.

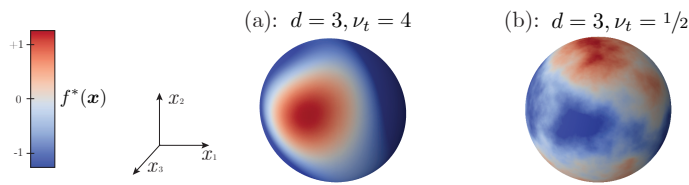

(a): $d = 3, \nu_t = 4$  (b): $d = 3, \nu_t = 1/2$

Figure 2: **Gaussian random process on the sphere.** We show here two samples of the task introduced in section 2 when the target function $f^*(\boldsymbol{x})$ is defined on the 3−dimensional unit sphere. (a) and (b) show samples of large and small smoothness coefficient $\nu_t$, respectively.

**Neural network representation in the feature regime**   In this regime we aim to approximate the target function $f^*(x)$ via a *one-hidden-layer neural network* of width $H$,

$$f_H(\boldsymbol{x}) = \frac{1}{H} \sum_{h=1}^{H} w_h \sigma(\boldsymbol{\theta}_h \cdot \boldsymbol{x}), \tag{2.3}$$

where $\{\boldsymbol{\theta}_h\}_{h=1}^{H}$ (the features) and $\{w_h\}_{h=1}^{H}$ (the weights) are the network parameters to be optimized, and $\sigma(x)$ denotes the ReLU function, $\sigma(x) = \max\{0, x\}$. If we assume that $\{\boldsymbol{\theta}_h, w_h\}_{h=1}^{H}$ are independently drawn from a probability measure $\mu$ on $\mathbb{S}^{d-1} \times \mathbb{R}$ such that the Radon measure $\gamma = \int_{\mathbb{R}} w \mu(\cdot, dw)$ exists, then as $H \to \infty$,

$$\lim_{H \to \infty} f_H(\boldsymbol{x}) = \int_{\mathbb{S}^{d-1}} \sigma(\boldsymbol{\theta} \cdot \boldsymbol{x}) d\gamma(\boldsymbol{\theta}) \qquad \text{a.e. on } \mathbb{S}^{d-1}. \tag{2.4}$$

This is the so-called mean-field limit [11, 12], and it is then natural to determine the optimal $\gamma$ via

$$\gamma^* = \arg\min_{\gamma} \int_{\mathbb{S}^{d-1}} |d\gamma(\boldsymbol{\theta})| \quad \text{subject to:} \quad \int_{\mathbb{S}^{d-1}} \sigma(\boldsymbol{\theta} \cdot \boldsymbol{x}_i) d\gamma(\boldsymbol{\theta}) = f^*(\boldsymbol{x}_i) \quad \forall i = 1, \dots, n. \tag{2.5}$$

In practice, we can approximate this minimization problem by using a network with large but finite width, constraining the feature to be on the sphere $|\boldsymbol{\theta}_h| = 1$, and minimizing the following empirical loss with L1 regularization on the weights,

$$\min_{\substack{\{w_h, \boldsymbol{\theta}_h\}_{h=1}^{H} \\ |\boldsymbol{\theta}_h| = 1}} \frac{1}{2n} \sum_{i=1}^{n} \left( f^*(\boldsymbol{x}_i) - \frac{1}{H} \sum_{h=1}^{H} w_h \sigma(\boldsymbol{\theta}_h \cdot \boldsymbol{x}_i) \right)^2 + \frac{\lambda}{H} \sum_{h=1}^{H} |w_h|. \tag{2.6}$$

This minimization problem leads to (2.5) when $H \to \infty$ and $\lambda \to 0$. Note that, by homogeneity of ReLU, (2.6) can be shown to be equivalent to imposing a regularization on the L2 norm of all parameters [32, Thm. 10], i.e. the usual weight decay.

To proceed we will make the following assumption about the minimizer $\gamma^*$:

**Assumption 1.** *The minimizer $\gamma^*$ of (2.5) is unique and atomic, with $n_A \leq n$ atoms, i.e. there exists $\{w_i^*, \boldsymbol{\theta}_i^*\}_{i=1}^{n_A}$ such that*

$$\gamma^* = \sum_{i=1}^{n_A} w_i^* \delta_{\boldsymbol{\theta}_i^*}. \tag{2.7}$$

The main component of the assumption is the uniqueness of $\gamma^*$; if it holds the sparsity of $\gamma^*$ follows from the representer theorem, see e.g. [33]. Both the uniqueness and sparsity of the minimizer can be justified as holding generically using asymptotic arguments involving recasting the L1 minimization problem 2.5 as a linear programming one: these arguments are standard (see e.g. [50]) and are presented in App. B for the reader convenience. In our arguments below to deduce the scaling of the generalization error we will mainly use that $n_A = O(n)$—we shall confirm this fact numerically even in the absence of regularization, if the weights are initialized to be small enough. Notice that from Assumption 1 it follows that the predictor in the feature regime corresponding to the minimizer $\gamma^*$ takes the following form

$$f^{\text{FEATURE}}(\boldsymbol{x}) = \sum_{i=1}^{n_A} w_i^* \sigma(\boldsymbol{\theta}_i^* \cdot \boldsymbol{x}). \tag{2.8}$$

**Neural network representation in the lazy regime.** In this regime we approximate the target function $f^*(x)$ via

$$f^{\text{NTK}}(\boldsymbol{x}) = \sum_{i=1}^{n} g_i K^{\text{NTK}}(\boldsymbol{x}_i \cdot \boldsymbol{x}), \tag{2.9}$$

where the weights $\{g_i\}_{i=1}^{n}$ solve

$$f^*(\boldsymbol{x}_j) = \sum_{i=1}^{n} g_i K^{\text{NTK}}(\boldsymbol{x}_i \cdot \boldsymbol{x}_j), \qquad j = 1, \ldots, n. \tag{2.10}$$

and $K^{\text{NTK}}(\boldsymbol{x} \cdot \boldsymbol{y})$ is the *Neural Tangent Kernel* (NTK) [9]

$$K^{\text{NTK}}(\boldsymbol{x} \cdot \boldsymbol{y}) = \int_{\mathbb{S}^{d-1} \times \mathbb{R}} \left( \sigma(\boldsymbol{\theta} \cdot \boldsymbol{x}) \sigma(\boldsymbol{\theta} \cdot \boldsymbol{y}) + w^2 \, \boldsymbol{x} \cdot \boldsymbol{y} \, \sigma'(\boldsymbol{\theta} \cdot \boldsymbol{x}) \sigma'(\boldsymbol{\theta} \cdot \boldsymbol{y}) \right) d\mu_0(\boldsymbol{\theta}, w). \tag{2.11}$$

Here $\mu_0$ is a fixed probability distribution which, in the NTK training regime [9], is the distribution of the features and weights at initialization. It is well-known [51] that the solution to kernel ridge regression problem can also be expressed via the kernel trick as

$$f^{\text{NTK}}(\boldsymbol{x}) = \int_{\mathbb{S}^{d-1} \times \mathbb{R}} (g_w(w, \boldsymbol{\theta}) \sigma(\boldsymbol{\theta} \cdot \boldsymbol{x}) + w \boldsymbol{x} \cdot \boldsymbol{g}_{\boldsymbol{\theta}}(\boldsymbol{\theta}, w) \sigma'(\boldsymbol{\theta} \cdot \boldsymbol{x})) \, d\mu_0(\boldsymbol{\theta}, w) \tag{2.12}$$

where $\boldsymbol{g}_{\boldsymbol{\theta}}$ and $g_w$ are the solutions of

$$\min_{g_w, \boldsymbol{g}_{\boldsymbol{\theta}}} \int_{\mathbb{S}^{d-1} \times \mathbb{R}} \left( g_w^2(w, \boldsymbol{\theta}) + |\boldsymbol{g}_{\boldsymbol{\theta}}(w, \boldsymbol{\theta})|^2 \right) d\mu_0(\boldsymbol{\theta}, w)$$

$$\text{subject to:} \int_{\mathbb{S}^{d-1} \times \mathbb{R}} (g_w(w, \boldsymbol{\theta}) \sigma(\boldsymbol{\theta} \cdot \boldsymbol{x}_i) + w \boldsymbol{x}_i \cdot \boldsymbol{g}_{\boldsymbol{\theta}}(w, \boldsymbol{\theta}) \sigma'(\boldsymbol{\theta} \cdot \boldsymbol{x}_i)) \, d\mu_0(\boldsymbol{\theta}, w) = f^*(\boldsymbol{x}_i)$$

$$\forall i = 1, \ldots, n. \tag{2.13}$$

Another lazy limit can be obtained equivalently by training only the weights while keeping the features to their initialization value. This is equivalent to forcing $\boldsymbol{g}_{\boldsymbol{\theta}}(\boldsymbol{\theta}, w)$ to vanish in Eq. 2.13, resulting again in a kernel method. The kernel, in this case, is called *Random Feature Kernel* ($K^{\text{RFK}}$), and can be obtained from Eq. 2.11 by setting $d\mu_0(\boldsymbol{\theta}, w) = \delta_{w=0} d\tilde{\mu}_0(\boldsymbol{\theta})$. The minimizer can then be written as in Eq. 2.9 with $K^{\text{NTK}}$ replaced by $K^{\text{RFK}}$.

## 3 Asymptotic analysis of generalization

In this section, we characterize the asymptotic decay of the generalization error $\bar{\epsilon}(n)$ averaged over several realizations of the target function $f^*$. Denoting with $d\tau^{d-1}(\boldsymbol{x})$ the uniform measure on $\mathbb{S}^{d-1}$,

$$\bar{\epsilon}(n) = \mathbb{E}_{f^*} \left[ \int d\tau^{d-1}(\boldsymbol{x}) \left( f^n(\boldsymbol{x}) - f^*(\boldsymbol{x}) \right)^2 \right] = \mathcal{A}_d n^{-\beta} + o(n^{-\beta}), \tag{3.1}$$

for some constant $\mathcal{A}_d$ which might depend on $d$ but not on $n$. Both for the lazy (see Eq. 2.9) and feature regimes (see Eq. 2.8) the predictor can be written as a sum of $\mathcal{O}(n)$ terms:

$$f^n(\boldsymbol{x}) = \sum_{j=1}^{\mathcal{O}(n)} g_j \varphi(\boldsymbol{x} \cdot \boldsymbol{y}_j) := \int_{\mathbb{S}^{d-1}} g^n(\boldsymbol{y}) \varphi(\boldsymbol{x} \cdot \boldsymbol{y}) d\tau(\boldsymbol{y}). \tag{3.2}$$

In the feature regime, the $g_j$'s ($\boldsymbol{y}_j$) coincide with the optimal weights $w_j^*$ (features $\boldsymbol{\theta}_j^*$), $\varphi$ with the activation function $\sigma$. In the lazy regime, the $\boldsymbol{y}_j$ are the training points $\boldsymbol{x}_j$, $\varphi$ is the neural tangent or random feature kernel the $g_j$'s are the weights solving Eq. 2. We have defined the density $g^n(\boldsymbol{x}) = \sum_j |\mathbb{S}^{d-1}| g_j \delta(\boldsymbol{x} - \boldsymbol{y}_j)$ so as to cast the predictor as a convolution on the sphere. Therefore, the projections of $f^n$ onto spherical harmonics $Y_{k,\ell}$ read $f_{k,\ell}^n = g_{k,\ell}^n \varphi_k$, where $g_{k,\ell}^n$ is the projection of $g^n(\boldsymbol{x})$ and $\varphi_k$ that of $\varphi(\boldsymbol{x} \cdot \boldsymbol{y})$. For ReLU neurons one has (as shown in App. A)

$$\varphi_k^{\text{LAZY}} \sim k^{-(d-1)-2\nu} \quad \text{with } \nu = 1/2 \text{ (NTK)}, 3/2 \text{ (RFK)}, \quad \varphi_k^{\text{FEATURE}} \sim k^{-\frac{d-1}{2}-3/2}. \tag{3.3}$$

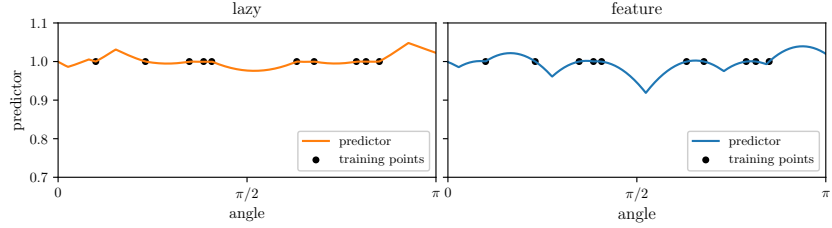

Figure 3: **Feature vs. Lazy Predictor.** Predictor of the lazy (left) and feature (right) regime when learning the constant function on the ring with 8 uniformly-sampled training points.

**Main Result**   Consider a target function $f^*$ with smoothness exponent $\nu_t$ as defined above, with data lying on $\mathbb{S}^{d-1}$. If $f^*$ is learnt with a one-hidden-layer network with ReLU neurons in the regimes specified above, then the generalization error follows $\bar{\epsilon}(n) \sim n^{-\beta}$ with:

$$\beta^{\text{LAZY}} = \frac{\min\{2(d-1)+4\nu, 2\nu_t\}}{d-1} \text{ with } \nu = \begin{cases} 1/2 \text{ for NTK,} \\ 3/2 \text{ for RFK,} \end{cases}, \qquad (3.4a)$$

$$\beta^{\text{FEATURE}} = \frac{\min\{(d-1)+3, 2\nu_t\}}{d-1}. \qquad (3.4b)$$

This is our central result. It implies that if the target function is a smooth isotropic Gaussian field (realized for large $\nu_t$), then lazy beats feature, in the sense that training the network in the lazy regime leads to a better scaling of the generalization performance with the number of training points.

**Strategy**   There is no general framework for a rigorous derivation of the generalization error in the ridgeless limit $\lambda \to 0$: predictions such as that of Eq. 3.4 can be obtained by either assuming that training points (for Eq. 3.4a) and neurons (for Eq. 3.4b) lie on a periodic lattice [44], or (for Eq. 3.4a) using the replica method from physics [45] as shown in App. F. Here we follow a different route, by first characterizing the form of the predictor for $d = 2$ (proof in App. C). This property alone allows us to determine the asymptotic scaling of the generalization error. We use it to analytically obtain the generalization error in the NTK case with a slightly simplified function $\varphi$ (details in App. D). This calculation motivates a simple ansatz for the form of $g^n(x)$ entering Eq. 3.2 and its projections onto spherical harmonics, which extends naturally to arbitrary dimension. We confirm the predictions resulting from this ansatz systematically in numerical experiments.

**Properties of the predictor in** $d = 2$   On the unit circle $\mathbb{S}^1$ all points are identified by a polar angle $x \in [0, 2\pi)$. Hence both target function and estimated predictor are functions of the angle, and all functions of a scalar product are in fact functions of the difference in angle. In particular, introducing $\tilde{\varphi}(x) = \varphi(\cos(x))$,

$$f^n(x) = \sum_j g_j \tilde{\varphi}(x - x_j) \equiv \int_0^{2\pi} \frac{dy}{2\pi} g^n(y) \tilde{\varphi}(x - y), \qquad (3.5)$$

where we defined

$$g^n(x) = \sum_{j=1}^n (2\pi g_j) \delta(y - x_j). \qquad (3.6)$$

Both for feature regime and NTK limit, the first derivative of $\tilde{\varphi}(x)$ is continuous except for two values of $x$ (0 and $\pi$ for lazy, $-\pi/2$ and $\pi/2$ for feature), so that $\tilde{\varphi}(x)''$ has a singular part consisting of two Dirac delta functions.

As a result, the second derivative of the predictor $(f^n)''$ has a singular part consisting of many Dirac deltas. If we denote with $(f^n)''_r$ the regular part, obtained by subtracting all the delta functions, we can show that (see App. C):

**Proposition 1.** *(informal) As $n \to \infty$, $(f^n)''_r$ converges to a function having finite second moment, i.e.*

$$\lim_{n \to \infty} \mathbb{E}_{f^*}[(f^n)''_r(x)]^2 = const. < \infty. \qquad (3.7)$$

Overfitting in Feature Learning

In the large $n$ limit, the predictor displays a singular second derivative at $O(n)$ points. Proposition 1 implies that outside of these singular points the second derivative is well defined. Thus, as $n$ gets large and the singular points approach each other, the predictor can be approximated by a chain of parabolas, as highlighted in Fig. 3 and noticed in [47] for a Laplace kernel. This property alone allows to determine the asymptotic scaling of the error in $d = 2$. In simple terms, Prop. 1 follows from the convergence of $g^n$ to the function satisfying $f^*(x) = \int \frac{dy}{2\pi} g(y) \tilde{\varphi}_r(x-y)$, which is guaranteed under our assumptions on the target function—a detailed proof is given in App. C.

**Decay of the error in $d = 2$ (sketch)**   The full calculation is in App. D. Consider a slightly simplified problem where $\tilde{\varphi}$ has a single discontinuity in its derivative, located at $x = 0$. In this case, $f^n(x)$ is singular if and only if $x$ is a data point. Consider then the interval $x \in [x_i, x_{i+1}]$ and set $\delta_i = x_{i+1} - x_i$, $x_{i+1/2} = (x_{i+1} + x_i)/2$. If the target function is smooth enough ($\nu_t > 2$), then a Taylor expansion implies $|f^*(x_{i+1/2}) - f^n(x_{i+1/2})| \sim \delta_i^2$. Since the distances $\delta_i$ between adjacent singular points are random variables with mean of order $1/n$ and finite moments, it is straightforward to obtain that $\bar{\epsilon}(n) \sim \sum_i (f^*(x_{i+1/2}) - f^n(x_{i+1/2}))^2 \sim \sum_i \delta_i^4 \sim n^{-4}$. By contrast if $f^*$ is not sufficiently smooth ($\nu_t \leq 2$), then $|f^*(x_{i+1/2}) - f^n(x_{i+1/2})| \sim \delta_i^{2\nu_t}$, leading to $\bar{\epsilon}(n) \sim n^{-2\nu_t}$. Note that for this asymptotic argument to apply to the feature learning regime, one must ensure that the distribution of the rescaled distance between adjacent singularities $n\delta_i$ has a finite fourth moment. This is obvious in the lazy regime, where the $\delta_i$'s are controlled by the position of the training points, but not in the feature regime, where the distribution of singular points is determined by that of the neuron's features. Nevertheless, we show that it must be the case in our setup in App. D.

**Interpretation in terms of spectral bias**   From the discussion above it is evident that there is a length scale $\delta$ of order $1/n$ such that $f^n(x)$ is a good approximation of $f^*(x)$ over scales larger than $\delta$. In terms of Fourier modes[2], one has: $i$) $\widehat{f^n}(k)$ matches $\widehat{f^*}(k)$ at long wavelengths, i.e. for $k \ll k_c \sim 1/n$. $ii$) In addition, since the phases $\exp(ikx_j)$ become effectively random phases for $k \gg k_c$, $\widehat{g^n}(k) = \sum_j g_j \exp(ikx_j)$ becomes a Gaussian random variable with zero mean and fixed variance and thus $iii$) $\widehat{f^n}(k) = \widehat{g^n}(k) \widehat{\tilde{\varphi}}(k)$ decorrelates from $f^*$ for $k \gg k_c$. Therefore

$$\bar{\epsilon}(n) \sim \sum_{|k|>k_c} \mathbb{E}_{f^*}\left[ \left( \widehat{g^n}(k)\widehat{\tilde{\varphi}}(k) - \widehat{f^n}(k) \right)^2 \right] \sim \sum_{|k|\geq k_c} \mathbb{E}_{f^*}\left[ (\widehat{g^n}(k))^2 \right] \widehat{\tilde{\varphi}}(k)^2 + \mathbb{E}_{f^*}\left[ \left( \widehat{f^n}(k) \right)^2 \right].$$

(3.8)

For $\nu_t > 2$, one has $\sum_j g_j^2 \sim n^{-1} \lim_{n \to \infty} \int g^n(x)^2 dx \sim n^{-1}$. It follows (see App. E for details) that the sum is dominated by the first term, hence entirely controlled by the Fourier coefficients of $\widehat{f^n}(k)$ at large $k$. A smoother predictor corresponds to a faster decay of $\widehat{f^n}(k)$ with $k$, thus a faster decay of the error with $n$. Plugging the relevant decays yields $\bar{\epsilon} \sim n^{-4}$ for feature regime and lazy regime with the NTK, and $n^{-6}$ for lazy regime with the RFK (which is smoother than the NTK). For $\nu_t \leq 2$, the two terms have comparable magnitude (see App. E), thus $\bar{\epsilon} \sim n^{-2\nu_t}$.

**Generalization to higher dimensions**   The argument above can be generalized for any $d$ by replacing Fourier modes with projections onto spherical harmonics. The characteristic distance between training points scales as $n^{-1/(d-1)}$, thus $k_c \sim n^{-1/(d-1)}$. Our ansatz is that, as in $d = 2$: $i$) for $k \ll k_c$, the predictor modes coincide with those of the target function, $f^n_{k,l} \approx f^*_{k,l}$ (this corresponds to the spectral bias result of kernel methods, stating that the predictor reproduces the first $O(n)$ projections of the target in the kernel eigenbasis [45]); $ii$) For $k \gg k_c$, $g^n_{k,l}$ is a sum of uncorrelated terms, thus a Gaussian variable with zero mean and fixed variance; $iii$) $f^n_{k,\ell} = g^n_{k,\ell} \tilde{\varphi}_k$ decorrelates from $f^*_{k,\ell}$ for $k \gg k_c$. $i$), $ii$) and $iii$) imply that:

$$\bar{\epsilon}(n) \sim \sum_{k \geq k_c} \sum_{l=1}^{\mathcal{N}_{k,d}} \mathbb{E}_{f^*}\left[ \left( f^n_{k,l} - f^*_{k,l} \right)^2 \right] \sim \sum_{k \geq k_c} \sum_{l=1}^{\mathcal{N}_{k,d}} \mathbb{E}_{f^*}\left[ (g^n_{k,l})^2 \right] \varphi_k^2 + k^{-2\nu_t - (d-1)}.$$

(3.9)

As shown in App. E, from this expression it is straightforward to obtain Eq. 3.4. Notice again that when the target is sufficiently smooth so that the predictor-dependent term dominates, the error is determined by the smoothness of the predictor. In particular, as $d > 2$, the predictor of feature learning is less smooth than both the NTK and RFK ones, due to the slower decay of the corresponding $\varphi_k$.

---

[2]The Fourier transform of a function $f(x)$ is indicated by the hat, $\hat{f}(k)$.

## 4 Numerical tests of the theory

We test successfully our predictions by computing the learning curves of both lazy and feature regimes when *(i)* the target function is constant on the sphere for varying $d$, see Fig. 4, and *(ii)* the target is a Gaussian random field with varying smoothness $\nu_t$, as shown in Fig. G.1 of App. G. For the lazy regime, we perform kernel regression using the analytical expression of the NTK [52] (see also Eq. A.19). For the feature regime, we find that our predictions hold when having a small regularization, although it takes unreachable times for gradient descent to exactly recover the minimal-norm solution—a more in-depth discussion can be found in App. G. An example of the atomic distribution of neurons found after training, which contrasts with the initial distribution, is displayed in Fig. 5a, left panel.

Another way to obtain sparse features is to initialize the network with very small weights [14], as proposed in [8]. As in the presence of an infinitesimal weights decay, this scheme also leads to sparse solutions with $n_A = \mathcal{O}(n)$ – an asymptotic dependence confirmed in Fig. G.3 of App. G. This observation implies that our predictions must apply in that case too, as we confirm in Fig. G.3.

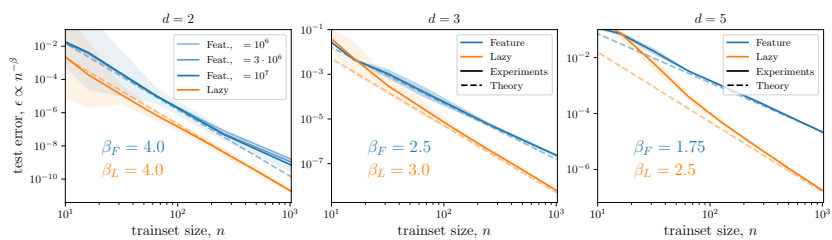

Figure 4: **Generalization error for a constant function** $f^*(\boldsymbol{x}) = 1$. Generalization error as a function of the training set size $n$ for a network trained in the feature regime with L1 regularization (blue) and kernel regression corresponding to the infinite-width lazy regime (orange). Numerical results (full lines) and the exponents predicted by the theory (dashed) are plotted. Panels correspond to different input-space dimensions ($d = 2, 3, 5$). Results are averaged over 10 different initializations of the networks and datasets. For $d = 2$ and large $n$, the gap between experiments and prediction for the feature regime is due to the finite training time $t$. Indeed our predictions become more accurate as $t$ increases, as illustrated in the left.

## 5 Evidence for overfitting along diffeomorphisms in image datasets

For fully-connected networks, the feature regime is well-adapted to learn anisotropic tasks [16]: if the target function does not depend on a certain linear subspace of input space, e.g. the pixels at the corner of an image, then neurons align perpendicularly to these directions [19]. By contrast, our results highlight a drawback of this regime when the target function is constant or smooth along directions in input space that require a continuous distribution of neurons to be represented. In such a case, the adaptation of the weights to the training points leads to a predictor with a sparse representation. Such a predictor would be less smooth than in the lazy regime and thus underperform.

Does this view hold for images, and explain why learning their features is detrimental for fully-connected networks? The first positive empirical evidence is that the neurons' distribution of networks trained on image data becomes indeed sparse in the feature regime, as illustrated in Fig. 5a, right, for CIFAR10 [28]. This observation raises the question of which are the directions in input space *i)* along which the target should vary smoothly, and *ii)* that are not easily represented by a discrete set of neurons. An example of such directions are global translations, which conserve the norm of the input and do not change the image class: the lazy regime predictor is indeed smoother than the feature one with respect to translations of the input (see App. H). Yet, these transformations live in a space of dimension 2, which is small in comparison with the full dimensionality $d$ of the data and thus may play a negligible role.

A much larger class of transformations believed to have little effect on the target are small diffeomorphisms [29]. A diffeomorphism $\tau$ acting on an image is illustrated in Fig. 5b, which highlights that

Overfitting in Feature Learning

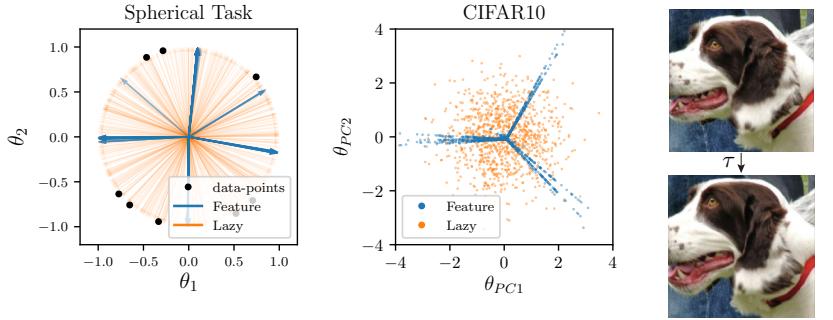

(a) **Features sparsification.** 1ˢᵗPanel: Distribution of neuron's feature for the task of learning a constant function on the sphere in 2D. Arrows represent a subset of the network features $\{\boldsymbol{\theta}_h\}_{h=1}^H$ after training in the lazy and feature regimes. Training is performed on $n = 8$ data-points (black dots). 2ⁿᵈPanel: FCN trained on CIFAR10. On the axes the first two principal components of the features $\{\boldsymbol{\theta}_h\}_{h=1}^H$ after training on $n = 32$ points in the feature (blue) and lazy (orange) regimes. Similarly to what is observed when learning a constant function, the $\boldsymbol{\theta}_h$ angular distribution becomes sparse with training in the feature regime.

(b) **Example of diffeomorphism.** Sample of a max-entropy deformation $\tau$ [49] when applied to a natural image, illustrating that it does not change the image class for the human brain.

Figure 5: **Features sparsification and example of a diffeomorphism.**

our brain still perceives the content of the transformed image as in the original one. Near-invariance of the task to these transformations is believed to play a key role in the success of deep learning, and in explaining how neural networks beat the curse of dimensionality [48]. Indeed, if modern architectures can become insensitive to these transformations, then the dimensionality of the problem is considerably reduced. In fact, it was found that the architectures displaying the best performance are precisely those which learn to vary smoothly along such transformations [49].

Small diffeomorphisms are likely the directions we are looking for. To test this hypothesis, following [49], we characterize the smoothness of a function along such diffeomorphisms, relative to that of random directions in input space. Specifically, we use the *relative sensitivity*:

$$R_f = \frac{\mathbb{E}_{x,\tau}\|f(\tau x) - f(x)\|^2}{\mathbb{E}_{x,\eta}\|f(x + \eta) - f(x)\|^2}. \tag{5.1}$$

In the numerator, the average is made over the test set and over an ensemble of diffeomorphisms, reviewed in App. I. The magnitude of the diffeomorphisms is chosen so that each pixel is shifted by one on average. In the denominator, the average runs over the test set and the vectors $\eta$ sampled uniformly on the sphere of radius $\|\eta\| = \mathbb{E}_{x,\tau}\|\tau x - x\|$, and this fixes the transformations magnitude.

We measure $R_f$ as a function of $n$ for three benchmark datasets of images, as shown in Fig. 6. We indeed find that $R_f$ is consistently smaller in the lazy training regime, where features are not learned. Overall, this observation supports the view that learning sparse features is detrimental when data present (near) invariance to transformations that cannot be represented sparsely by the architecture considered. Fig. 1 supports the idea that—for benchmark image datasets—this negative effect overcomes well-known positive effects of learning features, e.g. becoming insensitive to pixels on the edge of images (see App. H for evidence of this effect).

## 6 Conclusion

Our central result is that learning sparse features can be detrimental if the task presents invariance or smooth variations along transformations that are not adequately captured by the neural network architecture. For fully-connected networks, these transformations can be rotations of the input, but also continuous translations and diffeomorphisms.

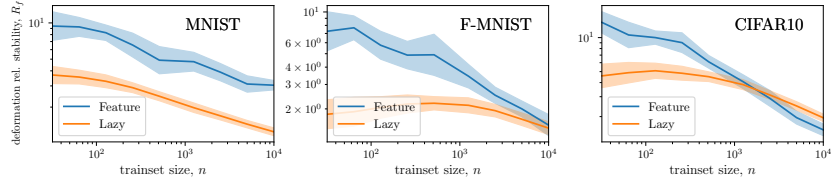

**Figure 6: Sensitivity to diffeomorphisms vs number of training points.** Relative sensitivity of the predictor to small diffeomorphisms of the input images, in the two regimes, for varying number of training points $n$ and different image datasets. Smaller values correspond to a smoother predictor, on average. Results are computed using the same predictors as in Fig. 1.

Our analysis relies on the sparsity of the features learned by a shallow fully-connected architecture: even in the infinite width limit, when trained in the feature learning regime such networks behave as $\mathcal{O}(n)$ neurons. The asymptotic analysis we perform for random Gaussian fields on the sphere leads to predictions for the learning curve exponent $\beta$ in different training regimes, which we verify. Such kind of results is scarce in the literature.

Note that our analysis focuses on ReLU neurons because *(i)* these are very often used in practice and *(ii)* in that case, $\beta$ will depend on the training regime, allowing for stringent numerical tests. If smooth activations (e.g. softplus) are considered, we expect that learning features will still be detrimental for generalization. Yet, the difference will not appear in the exponent $\beta$, but in other aspects of the learning curves (including numerical coefficients and pre-asymptotic effects) that are harder to predict.

Most fundamentally, our results underline that the success of feature learning for modern architectures still lacks a sufficient explanation. Indeed, most of the theoretical studies that previously emphasized the benefits of learning features have been considering fully-connected networks, for which learning features can be in practice a drawback. It is tempting to argue that in modern architectures, learning features is not at a disadvantage because smoothness along diffeomorphisms can be enforced from the start—due to the locally connected, convolutional, and pooling layers [53, 29]. Yet the best architectures often do not perform pooling and are not stable toward diffeomorphisms at initialization. *During training*, learning features leads to more stable and smoother solutions along diffeomorphisms [54, 49]. Understanding why building sparse features enhances stability in these architectures may ultimately explain the magical feat of deep CNNs: learning tasks in high dimensions.

## Acknowledgements

We thank Lénaïc Chizat, Antonio Sclocchi, and Umberto M. Tomasini for helpful discussions. The work of MW is supported by a grant from the Simons Foundation (#454953) and from the NSF under Grant No. 200021-165509. The work of EVE is supported by the National Science Foundation under awards DMR-1420073, DMS-2012510, and DMS-2134216, by the Simons Collaboration on Wave Turbulence, Grant No. 617006, and by a Vannevar Bush Faculty Fellowship.

# A  Quick recap of spherical harmonics

**Spherical harmonics**  This appendix collects some introductory background on spherical harmonics and dot-product kernels on the sphere [55]. See [56, 57] for an expanded treatment. Spherical harmonics are homogeneous polynomials on the sphere $\mathbb{S}^{d-1} = \{\boldsymbol{x} \in \mathbb{R}^d \mid \|\boldsymbol{x}\| = 1\}$, with $\|.\|$ denoting the L2 norm. Given the polynomial degree $k \in \mathbb{N}$, there are $\mathcal{N}_{k,s}$ linearly independent spherical harmonics of degree $k$ on $\mathbb{S}^{s-1}$, with

$$\mathcal{N}_{k,d} = \frac{2k+d-2}{k}\binom{d+k-3}{k-1}, \qquad \begin{cases} \mathcal{N}_{0,d} = 1 \quad \forall d, \\ \mathcal{N}_{k,d} \asymp A_d k^{d-2} \quad \text{for } k \gg 1, \end{cases} \tag{A.1}$$

where $\asymp$ means logarithmic equivalence for $k \to \infty$ and $A_d = \sqrt{2/\pi}(d-2)^{\frac{3}{2}-d}e^{d-2}$. Thus, we can introduce a set of $\mathcal{N}_{k,d}$ spherical harmonics $Y_{k,\ell}$ for each $k$, with $\ell$ ranging in $1, \ldots, \mathcal{N}_{k,d}$, which are orthonormal with respect to the uniform measure on the sphere $d\tau(\boldsymbol{x})$,

$$\{Y_{k,\ell}\}_{k \geq 0, \ell = 1, \ldots, \mathcal{N}_{k,d}}, \quad \langle Y_{k,\ell}, Y_{k,\ell'} \rangle_{\mathbb{S}^{d-1}} := \int_{\mathbb{S}^{d-1}} Y_{k,\ell}(\boldsymbol{x}) Y_{k,\ell'}(\boldsymbol{x}) \, d\tau(\boldsymbol{x}) = \delta_{\ell,\ell'}. \tag{A.2}$$

Because of the orthogonality of homogeneous polynomials with different degree, the set is a complete orthonormal basis for the space of square-integrable functions on $\mathbb{S}^{d-1}$. For any function $f : \mathbb{S}^{d-1} \to \mathbb{R}$, then

$$f(\boldsymbol{x}) = \sum_{k \geq 0} \sum_{\ell=1}^{\mathcal{N}_{k,d}} f_{k,\ell} Y_{k,\ell}(\boldsymbol{x}), \quad f_{k,\ell} = \int_{\mathbb{S}^{d-1}} f(\boldsymbol{x}) Y_{k,\ell}(\boldsymbol{x}) d\tau(\boldsymbol{x}). \tag{A.3}$$

Furthermore, spherical harmonics are eigenfunctions of the Laplace-Beltrami operator $\Delta$, which is nothing but the restriction of the standard Laplace operator to $\mathbb{S}^{d-1}$,

$$\Delta Y_{k,\ell} = -k(k+d-2)Y_{k,\ell}. \tag{A.4}$$

**Legendre polynomials**  By fixing a direction $\boldsymbol{y}$ in $\mathbb{S}^{d-1}$ one can select, for each $k$, the only spherical harmonic of degree $k$ which is invariant for rotations that leave $\boldsymbol{y}$ unchanged. This particular spherical harmonic is, in fact, a function of $\boldsymbol{x} \cdot \boldsymbol{y}$ and is called the Legendre polynomial of degree $k$, $P_{k,d}(\boldsymbol{x} \cdot \boldsymbol{y})$ (also referred to as Gegenbauer polynomial). Legendre polynomials can be written as a combination of the orthonormal spherical harmonics $Y_{k,\ell}$ via the addition theorem [56, Thm. 2.9],

$$P_{k,d}(\boldsymbol{x} \cdot \boldsymbol{y}) = \frac{1}{\mathcal{N}_{k,d}} \sum_{\ell=1}^{\mathcal{N}_{k,d}} Y_{k,\ell}(\boldsymbol{x}) Y_{k,\ell}(\boldsymbol{y}). \tag{A.5}$$

Alternatively, $P_{k,d}$ is given explicitly as a function of $t = \boldsymbol{x} \cdot \boldsymbol{y} \in [-1, 1]$ via the Rodrigues' formula [56, Thm. 2.23],

$$P_{k,d}(t) = \left(-\frac{1}{2}\right)^k \frac{\Gamma\left(\frac{d-1}{2}\right)}{\Gamma\left(k+\frac{d-1}{2}\right)} \left(1-t^2\right)^{\frac{3-d}{2}} \frac{d^k}{dt^k} \left(1-t^2\right)^{k+\frac{d-3}{2}}. \tag{A.6}$$

Here $\Gamma$ denotes the Gamma function, $\Gamma(z) = \int_0^\infty x^{z-1}e^{-x}\,dx$. Legendre polynomials are orthogonal on $[-1, 1]$ with respect to the measure with density $(1-t^2)^{(d-3)/2}$, which is the probability density function of the scalar product between points on $\mathbb{S}^{d-1}$.

$$\int_{-1}^{+1} P_{k,d}(t) P_{k',d}(t) \left(1-t^2\right)^{\frac{d-3}{2}} dt = \frac{|\mathbb{S}^{d-1}|}{|\mathbb{S}^{d-2}|} \frac{\delta_{k,k'}}{\mathcal{N}_{k,s}}. \tag{A.7}$$

Here $|\mathbb{S}^{d-1}| = 2\pi^{\frac{d}{2}}/\Gamma(\frac{d}{2})$ denotes the surface area of the $d$-dimensional unit sphere ($|\mathbb{S}^0| = 2$ by definition).

To sum up, given $\boldsymbol{x}, \boldsymbol{y} \in \mathbb{S}^{d-1}$, functions of $\boldsymbol{x}$ or $\boldsymbol{y}$ can be expressed as a sum of projections on the orthonormal spherical harmonics, whereas functions of $\boldsymbol{x} \cdot \boldsymbol{y}$ can be expressed as a sum of projections on the Legendre polynomials. The relationship between the two expansions is elucidated in the Funk-Hecke formula [56, Thm. 2.22],

$$\int_{\mathbb{S}^{d-1}} f(\boldsymbol{x} \cdot \boldsymbol{y}) Y_{k,\ell}(\boldsymbol{y}) \, d\tau(\boldsymbol{y}) = Y_{k,\ell}(\boldsymbol{x}) \frac{|\mathbb{S}^{d-2}|}{|\mathbb{S}^{d-1}|} \int_{-1}^{+1} f(t) P_{k,d}(t) \left(1-t^2\right)^{\frac{d-3}{2}} dt := f_k Y_{k,\ell}(\boldsymbol{x}). \tag{A.8}$$

Overfitting in Feature Learning

### A.1 Expansion of ReLU and combinations thereof

We can apply Eq. A.8 to have an expansion of neurons $\sigma(\boldsymbol{\theta} \cdot \boldsymbol{x})$ in terms of spherical harmonics [2, Appendix D]. After defining

$$\varphi_k := \frac{|\mathbb{S}^{d-2}|}{|\mathbb{S}^{d-1}|} \int_{-1}^{+1} \sigma(t) P_{k,d}(t) \left(1 - t^2\right)^{\frac{d-3}{2}} dt, \tag{A.9}$$

one has

$$\sigma(\boldsymbol{\theta} \cdot \boldsymbol{x}) = \sum_{k \geq 0} \mathcal{N}_{k,d} \varphi_k P_{k,d}(\boldsymbol{\theta} \cdot \boldsymbol{x}) = \sum_{k \geq 0} \varphi_k \sum_{\ell=1}^{\mathcal{N}_{k,d}} Y_{k,\ell}(\boldsymbol{\theta}) Y_{k,\ell}(\boldsymbol{x}). \tag{A.10}$$

For ReLU activations, in particular, $\sigma(t) = \max(0, t)$, thus

$$\varphi_k^{\mathrm{ReLU}} = \frac{|\mathbb{S}^{d-2}|}{|\mathbb{S}^{d-1}|} \int_0^{+1} t P_{k,d}(t) \left(1 - t^2\right)^{\frac{d-3}{2}} dt. \tag{A.11}$$

Notice that when $k$ is odd $P_{k,d}$ is an odd function of $t$, thus the integrand $t P_{k,d}(t)(1 - t^2)^{\frac{d-3}{2}}$ is an even function of $t$. As a result the integral on the right-hand side of Eq. A.11 coincides with half the integral over the full domain $[-1, 1]$,

$$\int_0^{+1} t P_{k,d}(t) \left(1 - t^2\right)^{\frac{d-3}{2}} dt = \frac{1}{2} \int_{-1}^{+1} t P_{k,d}(t) \left(1 - t^2\right)^{\frac{d-3}{2}} dt = 0 \text{ for } k > 1, \tag{A.12}$$

because, due to Eq. A.7, $P_{k,d}$ is orthogonal to all polynomials with degree strictly lower than $k$. For even $k$ we can use Eq. A.6 and get [2] (see Eq. 3.3, main text)

$$\int_0^{+1} t P_{k,d}(t) \left(1 - t^2\right)^{\frac{d-3}{2}} dt = \left(-\frac{1}{2}\right)^k \frac{\Gamma\left(\frac{d-1}{2}\right)}{\Gamma\left(k + \frac{d-1}{2}\right)} \int_0^1 t \frac{d^k}{dt^k} \left(1 - t^2\right)^{k + \frac{d-3}{2}} dt$$

$$= -\left(-\frac{1}{2}\right)^k \frac{\Gamma\left(\frac{d-1}{2}\right)}{\Gamma\left(k + \frac{d-1}{2}\right)} \frac{d^{k-2}}{dt^{k-2}} \left(1 - t^2\right)^{k + \frac{d-3}{2}} \Big|_{t=0}^{t=1} \tag{A.13}$$

$$\Rightarrow \varphi_k^{\mathrm{ReLU}} \sim k^{-\frac{d-1}{2} - \frac{3}{2}} \text{ for } k \gg 1 \text{ and even.}$$

Because all $\varphi_k^{\mathrm{ReLU}}$ with $k > 1$ and odd vanish, even summing an infinite amount of neurons $\sigma(\boldsymbol{\theta} \cdot \boldsymbol{x})$ with varying $\boldsymbol{\theta}$ does not allow to approximate any function on $\mathbb{S}^{d-1}$, but only those which have vanishing projections on all the spherical harmonics $Y_{k,\ell}$ with $k > 1$ and odd. This is why we set the odd coefficients of the target function spectrum to zero in Eq. 2.1.

### A.2 Dot-product kernels on the sphere

Also general dot-product kernels on the sphere admit an expansion such as Eq. A.10,

$$\mathcal{C}(\boldsymbol{x} \cdot \boldsymbol{y}) = \sum_{k \geq 0} \mathcal{N}_{k,d} c_k P_{k,d}(\boldsymbol{\theta} \cdot \boldsymbol{x}) = \sum_{k \geq 0} c_k \sum_{\ell=1}^{\mathcal{N}_{k,d}} Y_{k,\ell}(\boldsymbol{\theta}) Y_{k,\ell}(\boldsymbol{x}), \tag{A.14}$$

with

$$c_k = \frac{|\mathbb{S}^{d-2}|}{|\mathbb{S}^d|} \int_{-1}^{1} \mathcal{C}(t) P_{k,d}(t) \left(1 - t^2\right)^{\frac{d-3}{2}} dt. \tag{A.15}$$

The asymptotic decay of $c_k$ for large $k$ is controlled by the behaviour of $\mathcal{C}(t)$ near $t = \pm 1$, [58]. More precisely [58, Thm. 1], if $\mathcal{C}$ is infinitely differentiable in $(-1, 1)$ and has the following expansion around $\pm 1$,

$$\begin{cases} \mathcal{C}(t) = p_1(1 - t) + c_1(1 - t)^\nu + o\left((1 - t)^\nu\right) \text{ near } t = +1; \\ \mathcal{C}(t) = p_{-1}(-1 + t) + c_{-1}(-1 + t)^\nu + o\left((-1 + t)^\nu\right) \text{ near } t = -1, \end{cases} \tag{A.16}$$

where $p_{\pm 1}$ are polynomials and $\nu$ is not an integer, then

$$\begin{aligned} k \text{ even: } & c_k \sim (c_1 + c_{-1}) k^{-2\nu - (d-1)}; \\ k \text{ odd: } & c_k \sim (c_1 - c_{-1}) k^{-2\nu - (d-1)}, \end{aligned} \tag{A.17}$$

The result above implies that if $c_1 = c_{-1}$ ($c_1 = -c_{-1}$), then the eigenvalues with $k$ odd (even) decay faster than $k^{-2\nu - (d-2)}$. Moreover, if $\mathcal{C}$ is infinitely differentiable in $[-1, 1]$ then $c_k$ decays faster than any polynomial.

**NTK and RFK of one-hidden-layer ReLU networks**  Let $\mathbb{E}_{\boldsymbol{\theta}}$ denote expectation over a multivariate normal distribution with zero mean and unitary covariance matrix. For any $\boldsymbol{x}, \boldsymbol{y} \in \mathbb{S}^{d-1}$, the RFK of a one-hidden-layer ReLU network Eq. 2.3 with all parameters initialised as independent Gaussian random numbers with zero mean and unit variance reads

$$
\begin{aligned}
K^{\mathrm{RFK}}(\boldsymbol{x} \cdot \boldsymbol{y}) &= \mathbb{E}_{\boldsymbol{\theta}}\left[\sigma(\boldsymbol{\theta} \cdot \boldsymbol{x}) \sigma(\boldsymbol{\theta} \cdot \boldsymbol{y})\right] \\
&= \frac{(\pi - \arccos(t))t + \sqrt{1 - t^2}}{2\pi}, \text{ with } t = \boldsymbol{x} \cdot \boldsymbol{y}.
\end{aligned}
\tag{A.18}
$$

The NTK of the same network reads, with $\sigma'$ denoting the derivative of ReLU or Heaviside function,

$$
\begin{aligned}
K^{\mathrm{NTK}}(\boldsymbol{x} \cdot \boldsymbol{y}) &= \mathbb{E}_{\boldsymbol{\theta}}\left[\sigma(\boldsymbol{\theta} \cdot \boldsymbol{x}) \sigma(\boldsymbol{\theta} \cdot \boldsymbol{y})\right] + (\boldsymbol{x} \cdot \boldsymbol{y}) \mathbb{E}_{\boldsymbol{\theta}}\left[\sigma'(\boldsymbol{\theta} \cdot \boldsymbol{x}) \sigma'(\boldsymbol{\theta} \cdot \boldsymbol{y})\right] \\
&= \frac{2(\pi - \arccos(t))t + \sqrt{1 - t^2}}{2\pi}, \text{ with } t = \boldsymbol{x} \cdot \boldsymbol{y}.
\end{aligned}
\tag{A.19}
$$

As functions of a dot-product on the sphere, both NTK and RFK admit a decomposition in terms of spherical harmonics as Eq. A.15. For dot-product kernels, this expansion coincides with the Mercer's decomposition of the kernel [55], that is the coefficients of the expansion are the eigenvalues of the kernel. The asymptotic decay of the eigenvalues of such kernels $\varphi_k^{\mathrm{NTK}}$ and $\varphi_k^{\mathrm{RFK}}$ can be obtained by applying Eq. A.16 [58, Thm. 1]. Equivalently, one can notice that $K^{\mathrm{RFK}}$ is proportional to the convolution on the sphere of ReLU with itself, therefore $\varphi_k^{\mathrm{RFK}} = (\varphi_k^{\mathrm{ReLU}})^2$. Similarly, the asymptotic decay of $\varphi_k^{\mathrm{NTK}}$ can be related to that of the coefficients of $\sigma'$, derivative of ReLU: $\varphi_k(\sigma') \sim k\varphi(\sigma)$, thus $\varphi_k^{\mathrm{NTK}} \sim k^2(\varphi_k^{\mathrm{ReLU}})^2$. Both methods lead to Eq. 3.3 of the main text.

**Gaussian random fields and Eq. 2.2**  Consider a Gaussian random field $f^*$ on the sphere with covariance kernel $\mathcal{C}(\boldsymbol{x} \cdot \boldsymbol{y})$,

$$
\mathbb{E}[f^*(\boldsymbol{x})] = 0, \quad \mathbb{E}[f^*(\boldsymbol{x})f^*(\boldsymbol{y})] = \mathcal{C}(\boldsymbol{x} \cdot \boldsymbol{y}), \quad \forall \boldsymbol{x}, \boldsymbol{y} \in \mathbb{S}^{d-1}.
\tag{A.20}
$$

$f^*$ can be equivalently specified via the statistics of the coefficients $f_{k,\ell}^*$,

$$
\mathbb{E}[f_{k,\ell}^*] = 0, \quad \mathbb{E}[f_{k,\ell}^* f_{k',\ell'}^*] = c_k \delta_{k,k'} \delta_{\ell,\ell'},
\tag{A.21}
$$

with $c_k$ denoting the eigenvalues of $\mathcal{C}$ in Eq. A.15. Notice that the eigenvalues are degenerate with respect to $\ell$ because the covariance kernel is a function $\boldsymbol{x} \cdot \boldsymbol{y}$: as a result, the random function $f^*$ is isotropic in law.

If $c_k$ decays as a power of $k$, then such power controls the weak differentiability (in the mean-squared sense) of the random field $f^*$. In fact, from Eq. A.4,

$$
\left\|\Delta^{m/2} f^*\right\| = \sum_{k \geq 0} \sum_{\ell} \left(-k(k+d-2)\right)^m \left(f_{k,\ell}^*\right)^2.
\tag{A.22}
$$

Upon averaging over $f^*$ one gets

$$
\mathbb{E}\left[\left\|\Delta^{m/2} f^*\right\|\right] = \sum_{k \geq 0} \left(-k(k+d-2)\right)^m \sum_{\ell} \mathbb{E}\left[\left(f_{k,\ell}^*\right)^2\right] = \sum_{k \geq 0} \left(-k(k+d-2)\right)^m \mathcal{N}_{k,d} c_k.
\tag{A.23}
$$

From Eq. A.16 [58, Thm. 1], if $\mathcal{C}(t) \sim (1-t)^{\nu_t}$ for $t \to 1$ and/or $\mathcal{C}(t) \sim (-1+t)^{\nu_t}$ for $t \to -1$, then $c_k \sim k^{-2\nu_t - (d-1)}$ for $k \gg 1$. In addition, for finite but arbitrary $d$, $(-k(k+d-2))^m \sim k^{2m}$ and $\mathcal{N}_{k,s} \sim k^{d-2}$ (see Eq. A.1). Hence the summand in the right-hand side of Eq. A.23 is $\sim k^{2(m-\nu_t)-1}$, thus

$$
\mathbb{E}\left[\left\|\Delta^{m/2} f^*\right\|\right] < \infty \quad \forall m < \nu_t.
\tag{A.24}
$$

Alternatively, one can think of $\nu_t$ as controlling the scaling of the difference $\delta f^*$ over inputs separated by a distance $\delta$. From Eq. A.20,

$$
\begin{aligned}
\mathbb{E}\left[|f^*(\boldsymbol{x}) - f^*(\boldsymbol{y})|^2\right] &= 2\mathcal{C}(1) - 2\mathcal{C}(\boldsymbol{x} \cdot \boldsymbol{y}) = 2\mathcal{C}(1) + O((1 - \boldsymbol{x} \cdot \boldsymbol{y})^{\nu_t}) \\
&= 2\mathcal{C}(1) + O(|\boldsymbol{x} - \boldsymbol{y}|^{2\nu_t})
\end{aligned}
\tag{A.25}
$$

# B Uniqueness and Sparsity of the L1 minimizer

Recall that we want to find the $\gamma^*$ that solves

$$\gamma^* = \underset{\gamma}{\arg\min} \int_{\mathbb{S}^{d-1}} |d\gamma(\boldsymbol{\theta})| \quad \text{subject to} \quad \int_{\mathbb{S}^{d-1}} \sigma(\boldsymbol{\theta} \cdot \boldsymbol{x}_i) d\gamma(\boldsymbol{\theta}) = f^*(\boldsymbol{x}_i) \quad \forall i = 1, \dots, n. \quad \text{(B.1)}$$

In this appendix, we argue that the uniqueness of $\gamma^*$ which implies that it is atomic with at most $n$ atoms is a natural assumption. We start by discretizing the measure $\gamma$ into $H$ atoms, with $H$ arbitrarily large. Then the problem Eq. B.1 can be rewritten as

$$\boldsymbol{w}^* = \underset{\boldsymbol{w}}{\arg\min} \|\boldsymbol{w}\|_1, \quad \text{subject to} \quad \boldsymbol{\Phi}\boldsymbol{w} = \boldsymbol{y}, \quad \text{(B.2)}$$

with $\boldsymbol{\Phi} \in \mathbb{R}^{H \times n}$, $\Phi_{h,i} = \sigma(\boldsymbol{\theta}_h \cdot \boldsymbol{x}_i)$ and $y_i = f^*(\boldsymbol{x}_i)$.

Given $\boldsymbol{w} \in \mathbb{R}^H$, let $\boldsymbol{u} = \max(\boldsymbol{w}, 0) \geq \boldsymbol{0}$ and $\boldsymbol{v} = -\max(-\boldsymbol{w}, 0) \geq \boldsymbol{0}$ so that $\boldsymbol{w} = \boldsymbol{u} - \boldsymbol{v}$. It is well-known (see e.g. [50]) that the minimization problem in (B.2) can be recast in terms of $\boldsymbol{u}$ and $\boldsymbol{v}$ into a linear programming problem. That is, $\boldsymbol{w}^* = \boldsymbol{u}^* - \boldsymbol{v}^*$ with

$$(\boldsymbol{u}^*, \boldsymbol{v}^*) = \underset{\boldsymbol{u}, \boldsymbol{v}}{\arg\min} \, \boldsymbol{e}^T(\boldsymbol{u} + \boldsymbol{v}), \quad \text{subject to} \ \boldsymbol{\Phi}\boldsymbol{u} - \boldsymbol{\Phi}\boldsymbol{v} = \boldsymbol{y}, \quad \boldsymbol{u} \geq \boldsymbol{0}, \quad \boldsymbol{v} \geq \boldsymbol{0} \quad \text{(B.3)}$$

where $\boldsymbol{e} = [1, 1, \dots, 1]^T$. Assuming that this problem is feasible (i.e. there is at least one solution to $\boldsymbol{\Phi}\boldsymbol{u} - \boldsymbol{\Phi}\boldsymbol{v} = \boldsymbol{y}$ such that $\boldsymbol{u} \geq \boldsymbol{0}, \boldsymbol{v} \geq \boldsymbol{0}$), it is known that it admits extremal solution, i.e. solutions such that at most $n$ entries of $(\boldsymbol{u}^*, \boldsymbol{v}^*)$ (and hence of $\boldsymbol{w}^*$) are non-zero. The issue is whether such an extremal solution is unique. Assume that there are two, say $(\boldsymbol{u}_1^*, \boldsymbol{v}_1^*)$ and $(\boldsymbol{u}_2^*, \boldsymbol{v}_2^*)$. Then, by convexity,

$$(\boldsymbol{u}_t^*, \boldsymbol{v}_t^*) = (\boldsymbol{u}_1^*, \boldsymbol{v}_1^*)t + (\boldsymbol{u}_2^*, \boldsymbol{v}_2^*)(1 - t) \quad \text{(B.4)}$$

is also a minimizer of (B.3) for all $t \in [0, 1]$, with the same minimum value $\boldsymbol{u}_t^* + \boldsymbol{v}_t^* = \boldsymbol{u}_1^* + \boldsymbol{v}_1^* = \boldsymbol{u}_2^* + \boldsymbol{v}_2^*$. Generalizing this argument to the case of more than two extremal solutions, we conclude that all minimizers are global, with the same minimum value, and they live on the simplex where $\boldsymbol{e}^T(\boldsymbol{u} + \boldsymbol{v}) = \boldsymbol{e}^T(\boldsymbol{u}_1 + \boldsymbol{v}_1)$. Therefore, nonuniqueness requires that that this simplex has a nontrivial intersection with the feasible set where $\boldsymbol{\Phi}\boldsymbol{u} - \boldsymbol{\Phi}\boldsymbol{v} = \boldsymbol{y}$ with $\boldsymbol{u} \geq \boldsymbol{0}, \boldsymbol{v} \geq \boldsymbol{0}$. We argue that, generically, this will not be the case, i.e. the intersection will be trivial, and the extremal solution unique. In particular, since in our case we are in fact interested in the problem (B.1), we can always perturb slightly the discretization into $H$ atoms of $\gamma$ to guarantee that the extremal solution is unique. Since this is true no matter how large $H$ is, and any Radon measure can be approached to arbitrary precision using such discretization, we conclude that the minimizer of (B.1) should be unique as well, with at most $n$ atoms.

# C Proof of Proposition 1

In this section, we provide the formal statement and proof of Proposition 1. Let us recall the general form of the predictor for both lazy and feature regimes in $d = 2$. From Eq. 3.6,

$$f^n(x) = \sum_{j=1}^n g_j \tilde{\varphi}(x - x_j) = \int \frac{dy}{2\pi} g^n(y) \tilde{\varphi}(x - y). \quad \text{(C.1)}$$

where $n$ is the number of training points for the lazy regime and the number of atoms for the feature regime and, for $x \in (-\pi, \pi]$,

$$\tilde{\varphi}(x) = \begin{cases} \max\{0, \cos(x)\} & \text{(feature regime),} \\ \dfrac{2(\pi - |x|)\cos(x) + \sin(|x|)}{2\pi} & \text{(lazy regime, NTK),} \\ \dfrac{(\pi - |x|)\cos(x) + \sin(|x|)}{2\pi} & \text{(lazy regime, RFK).} \end{cases} \quad \text{(C.2)}$$

All these functions $\tilde{\varphi}$ have jump discontinuities on some derivative: the first for feature and NTK, the third for RFK. If the $l$-th derivative has jump discontinuities, the $l + 1$-th only exists in a distributional sense and it can be generically written as a sum of a regular function and a sequence of Dirac masses located at the discontinuities. With $m$ denoting the number of such discontinuities and $\{x_j\}_j$ their locations, $f^{(l)}$ denoting the $l$-th derivative of $f$, for some $c_j \in \mathbb{R}$,

$$f^{(l+1)}(x) = f_r^{(l+1)}(x) + \sum_{j=1}^{m} c_j \delta(x - x_j), \tag{C.3}$$

where $f_r$ denotes the *regular* part of $f$.

**Proposition 2.** *Consider a random target function $f^*$ satisfying Eq. 2.1 and the predictor $f^n$ obtained by training a one-hidden-layer ReLU network on $n$ samples $(x_i, f^*(x_i))$ in the feature or in the lazy regime (Eq. C.1). Then, with $\widehat{f}(k)$ denoting the Fourier transform of $f(x)$, one has*

$$\lim_{|k|\to\infty} \lim_{n\to\infty} \frac{\widehat{(f^n)''}(k)}{\widehat{f^*}(k)} = c, \tag{C.4}$$

*where $c$ is a constant (different for every regime). This result implies that as $n \to \infty$, $(f^n)''(x)$ converges to a function having finite second moment, i.e.*

$$\begin{aligned}
\lim_{n\to\infty} \mathbb{E}_{f^*} \left[(f^n)_r''(x)\right]^2 &= \lim_{n\to\infty} \mathbb{E}_{f^*} \left[\int dx \, ((f^n)_r'')^2(x)\right] \\
&= \lim_{n\to\infty} \mathbb{E}_{f^*} \left[\sum_k \widehat{(f^n)_r''}^2(k)\right] = const. < \infty,
\end{aligned} \tag{C.5}$$

*using the fact that $\mathbb{E}_{f^*} \left[(f^n)_r''(x)\right]^2$ does not depend on $x$ and $\mathbb{E}_{f^*}[\sum_k \widehat{f^*}^2(k)] = const.$*

*Proof:* Because our target functions are random fields that are in $L_2$ with probability one, and the RKHS of our kernels are dense in that space, we know that the test error vanishes as $n \to \infty$ [59]. As a result

$$f^*(x) = \lim_{n\to\infty} f^n(x) = \lim_{n\to\infty} \int \frac{dy}{2\pi} g^n(y) \tilde{\varphi}(x - y). \tag{C.6}$$

Consider first the feature regime and the NTK lazy regime. In both cases $\tilde{\varphi}$ has two jump discontinuities in the first derivative, located at $x = 0, \pi$ for the NTK and at $x = \pm \pi/2$, therefore we can write the second derivative as the sum of a regular function and two Dirac masses,

$$\begin{aligned}
(\tilde{\varphi}^{\text{FEATURE}})'' &= -\max\{0, \cos(x)\} + \delta(x - \pi/2) + \delta(x + \pi/2), \\
(\tilde{\varphi}^{\text{NTK}})'' &= \frac{-2(\pi - |x|)\cos(x) + 3\sin(|x|)}{2\pi} - \frac{1}{2\pi}\delta(x) + \frac{1}{2\pi}\delta(x - \pi).
\end{aligned} \tag{C.7}$$

As a result, the second derivative of the predictor can be written as the sum of a regular part $(f^n)_r''$ and a sequence of $2n$ Dirac masses. After subtracting the Dirac masses, both sides of Eq. C.1 can be differentiated twice and yield

$$(f^n)_r''(x) = \int \frac{dy}{2\pi} g^n(y) \tilde{\varphi}_r''(x - y). \tag{C.8}$$

Hence in the Fourier representation we have

$$\widehat{(f^n)_r''}(k) = \widehat{g^n}(k)(-k^2 \widehat{\tilde{\varphi}}_r(k)) \tag{C.9}$$

where we defined

$$\widehat{\tilde{\varphi}}(k) = \int_{-\pi}^{\pi} \frac{dx}{\sqrt{2\pi}} e^{ikx} \tilde{\varphi}(x), \qquad \widehat{\tilde{\varphi}_r}(k) = \int_{-\pi}^{\pi} \frac{dx}{\sqrt{2\pi}} e^{ikx} \tilde{\varphi}_r(x). \tag{C.10}$$

and used $\widehat{\tilde{\varphi}_r''}(k) = -k^2 \widehat{\tilde{\varphi}_r}(k)$. By universal approximation we have

$$\widehat{f^*}(k) = \int_{-\pi}^{\pi} \frac{dx}{\sqrt{2\pi}} e^{ikx} f^*(x) = \lim_{n\to\infty} \widehat{g^n}(k) \widehat{\tilde{\varphi}}(k) \qquad \Rightarrow \qquad \lim_{n\to\infty} \widehat{g^n}(k) = \frac{\widehat{f^*}(k)}{\widehat{\tilde{\varphi}}(k)}. \tag{C.11}$$

As a result by combining Eq. C.9 and Eq. C.11 we deduce

$$\lim_{n\to\infty} \widehat{(f^n)_r''}(k) = -\frac{k^2 \widehat{\tilde{\varphi}}_r(k)}{\widehat{\tilde{\varphi}}(k)} \widehat{f^*}(k). \tag{C.12}$$

Overfitting in Feature Learning

To complete the proof using this result it remains to estimate the scaling of $\widehat{\tilde{\varphi}}_r(k)$ and $\widehat{\tilde{\varphi}}(k)$ in the large $|k|$ limit.

For the feature regime, a direct calculation shows that $\tilde{\varphi}''_r = -\tilde{\varphi}$, implying that $\widehat{\tilde{\varphi}}_r(k) = -\widehat{\tilde{\varphi}}(k)$. This proves that Eq. C.4 is satisfied with $c = -1$.

For the NTK lazy regime $\tilde{\varphi}''_r$ and $-\tilde{\varphi}$ are different but they have similar singular expansions near $x = 0$ and $\pi$. Therefore their Fourier coefficients display the same asymptotic decay. More specifically, with $t = \cos(x)$ (or $x = \arccos(t)$), so that $\tilde{\varphi}(x) = \varphi(t)$, one has

$$\begin{cases} \varphi^{\text{NTK}}(t) = t - \dfrac{1}{\sqrt{2\pi}}(1-t)^{1/2} + O\left((1-t)^{3/2}\right) \text{ near } t = +1; \\[2mm] \varphi^{\text{NTK}}(t) = -\dfrac{1}{\sqrt{2\pi}}(-1+t)^{1/2} + O\left((-1+t)^{3/2}\right) \text{ near } t = -1, \end{cases} \tag{C.13}$$

and

$$\begin{cases} (\varphi^{\text{NTK}})''_r(t) = -t + \dfrac{5}{\sqrt{2\pi}}(1-t)^{1/2} + O\left((1-t)^{3/2}\right) \text{ near } t = +1; \\[2mm] (\varphi^{\text{NTK}})''_r(t) = +\dfrac{5}{\sqrt{2\pi}}(-1+t)^{1/2} + O\left((-1+t)^{3/2}\right) \text{ near } t = -1. \end{cases} \tag{C.14}$$

Therefore, due to Eq. A.17, Eq. C.4 is satisfied with $c = -5$. The same procedure can be applied to the RFK lazy regime, with the exception that it is the fourth derivative of $\tilde{\varphi}^{\text{RFK}}$ which can be written as a regular part plus Dirac masses, but one can still obtain the Fourier coefficients of the second derivative's regular part by dividing those of the fourth derivative's regular part by $k^2$.

## D  Asymptotics of generalization in $d = 2$

In this section we compute the decay of generalization error $\bar{\epsilon}$ with the number of samples $n$ in the following 2-dimensional setting:

$$f^n(x) = \sum_{j=1}^{n} g_j \tilde{\varphi}(x - x_j), \tag{D.1}$$

where the $x_j$'s are the training points (like in the NTK case) and $\varphi$ has a single discontinuity on the first derivative in 0.

Let us order the training points clockwise on the ring, such that $x_1 = 0$ and $x_{i+1} > x_i$ for all $i = 1, \ldots, n$, with $x_{n+1} := 2\pi$. On each of the $x_i$ the predictor coincides with the target,

$$f^n(x_i) = f^*(x_i) \quad \forall i = 1, \ldots, n. \tag{D.2}$$

For large enough $n$, the difference $x_{i+1} - x_i$ is small enough such that, within $(x_i, x_{i+1})$, $f^n(x)$ can be replaced with its Taylor series expansion up to the second order. In practice, the predictors appear like the cable of a suspension bridge with the pillars located on the training points. In particular, we can consider an expansion around $x_i^+ := x_i + \epsilon$ for any $\epsilon > 0$ and then let $\epsilon \to 0$ from above:

$$f^n(x) = f^n(x_i^+) + (x - x_i^+) f^{n\prime}(x_i^+) + \frac{(x - x_i^+)^2}{2}(f^n)''(x_i^+) + \mathcal{O}\left((x - x_i^+)^3\right). \tag{D.3}$$

By differentiability of $f^n$ in $(x_i, x_{i+1})$ the second derivative can be computed at any point inside $(x_i, x_{i+1})$ without changing the order of approximation in Eq. D.3, in particular we can replace $(f^n)''(x_i^+)$ with $c_i$, the mean curvature of $f^n$ in $(x_i, x_{i+1})$. Moreover, as $\epsilon \to 0$, $f^n(x_i^+) \to f^*(x_i)$ and $f^n(x_{i+1}^-) \to f^*(x_{i+1})$. By introducing the limiting slope $m_i^+ := \lim_{x \to 0^+} f^{n\prime}(x_i + x)$, we can write

$$f^n(x) = f^*(x_i) + (x - x_i) m_i^+ + \frac{(x - x_i)^2}{2} c_i + O\left((x - x_i^+)^3\right) \tag{D.4}$$

Computing Eq. D.4 at $x = x_{i+1}$ yields a closed form for the limiting slope $m_i^+$ as a function of the mean curvature $c_i$, the interval length $\delta_i := (x_{i+1} - x_i)$ and $\Delta f_i := f^*(x_{i+1}) - f^*(x_i)$. Specifically,

$$m_i^+ = \frac{\Delta f_i}{\delta_i} - \frac{\delta_i}{2} c_i. \tag{D.5}$$

The generalization error can then be split into contributions from all the intervals. If $\nu_t > 2$, A Taylor expansion leads to:

$$
\begin{aligned}
\epsilon(n) &= \int_0^{2\pi} \frac{dx}{2\pi} \left(f^n(x) - f^*(x)\right)^2 \\
&= \sum_{i=1}^n \int_{x_i}^{x_{i+1}} \frac{dx}{2\pi} \left[ (x - x_i)\left(m_i^+ - (f^*)'(x_i)\right) + \frac{(x - x_i)^2}{2}\left(c_i - (f^*)''(x_i)\right) + o\left((x - x_i^+)^2\right) \right]^2 \\
&= \sum_{i=1}^n \int_0^{\delta_i} \frac{d\delta}{2\pi} \left[ \delta\left(m_i^+ - (f^*)'(x_i)\right) + \frac{\delta^2}{2}\left(c_i - (f^*)''(x_i)\right) + o\left(\delta^2\right) \right]^2 \\
&= \sum_{i=1}^n \frac{1}{2\pi} \left[ \frac{\delta_i^3}{3}\left(m_i^+ - (f^*)'(x_i)\right)^2 + \frac{\delta_i^5}{20}\left(c_i - (f^*)''(x_i)\right)^2 \right. \\
&\qquad\qquad \left. + \frac{\delta_i^4}{4}\left(m_i^+ - (f^*)'(x_i)\right)\left(c_i - (f^*)''(x_i)\right) + o(\delta_i^5) \right].
\end{aligned}
\tag{D.6}
$$

In addition, as $\Delta f_i = (f^*)'(x_i)\delta_i + (f^*)''(x_i)\delta_i^2/2 + O(\delta_i^3)$,

$$
m_i^+ - (f^*)'(x_i) = \frac{\delta_i}{2}\left((f^*)''(x_i) - c_i\right) + o(\delta_i)^2,
\tag{D.7}
$$

thus

$$
\epsilon(n) = \frac{1}{2\pi} \sum_{i=1}^n \left[ \frac{\delta_i^5}{120}\left(c_i - (f^*)''(x_i)\right)^2 + o(\delta_i^5) \right].
\tag{D.8}
$$

implying:

$$
\bar{\epsilon}(n) = \frac{n^{-4}\left(n^{-1}\sum_{i=1}^n (n\delta_i)^5\right)}{240\pi} \lim_{n\to\infty} \int \mathbb{E}_{f^*}\left[ \left((f^n)''(x) - (f^*)''(x)\right)^2 \right] dx + o(n^{-4}) \sim \frac{1}{n^4}
\tag{D.9}
$$

where we used that *(i)* the integral converges to some finite value, due to proposition 2. From App. C, this integral can be estimated as $\sum_k \mathbb{E}_{f^*}\left[ \left(cf^*(k) - k^2 f^*(k)\right)^2 \right]$, that indeed converges for $\nu_t > 2$. *(ii)* $\left(n^{-1}\sum_{i=1}^n (n\delta_i)^5\right)$ has a deterministic limit for large $n$. It is clear for the lazy regime since the distance between adjacent singularities $\delta_i$ follows an exponential distribution of mean $\sim \frac{1}{n}$. We expect this to be also true for the feature regime in our set-up. Indeed, in the limit $n \to \infty$, the predictor approaches a parabola between singular points, which generically cannot fit more than three random points. There must thus be a singularity at least every two data-points with a probability approaching unity as $n \to \infty$, which implies that $\left(n^{-1}\sum_{i=1}^n (n\delta_i)^5\right)$ converges to a constant for large $n$.

Finally, for $\nu_t < 2$, the same decomposition in intervals applies, but a Taylor expansion to second order does not hold. The error is then dominated by the fluctuations of $f^*$ on the scale of the intervals, as indicated in the main text.

## E   Asymptotic of generalization via the spectral bias ansatz

According to the spectral bias ansatz, the first $n$ modes of the predictor $f_{k,\ell}^n$ coincide with the modes of the target function $f_{k,\ell}^*$. Therefore, the asymptotic scaling of the error with $n$ is entirely controlled by the remaining modes,

$$
\epsilon(n) \sim \sum_{k \geq k_c} \sum_{\ell=1}^{\mathcal{N}_{k,d}} \left(f_{k,\ell}^n - f_{k,\ell}^*\right)^2 \quad \text{with} \quad \sum_{k \leq k_c} \mathcal{N}_{k,d} \sim n.
\tag{E.1}
$$

Since $\mathcal{N}_{k,d} \sim k^{d-2}$ for $k \gg 1$, one has that, for large $n$, $k_c \sim n^{\frac{1}{d-1}}$. After averaging the error over target functions we get

$$
\bar{\epsilon}(n) \sim \sum_{k \geq k_c} \sum_{\ell=1}^{\mathcal{N}_{k,d}} \left\{ \mathbb{E}_{f^*}\left[ \left(f_{k,\ell}^n\right)^2 \right] + \mathbb{E}_{f^*}\left[ \left(f_{k,\ell}^*\right)^2 \right] - 2\mathbb{E}_{f^*}\left[ \left(f_{k,\ell}^n f_{k,\ell}^*\right) \right] \right\}.
\tag{E.2}
$$

Overfitting in Feature Learning

Let us recall that, with the predictor having the general form in [Eq. 3.2](#), then

$$f_{k,\ell}^n = g_{k,\ell}^n \varphi_k \quad \text{with} \quad g_{k,\ell}^n = \sum_{j=1}^n g_j Y_{k,\ell}(\boldsymbol{y}_j), \tag{E.3}$$

where the $\boldsymbol{y}_j$'s denote the training points for the lazy regime and the neuron features for the feature regime. For $k \ll k_c$, where $f_{k,\ell}^n = f_{k,\ell}^*$, $g_{k,\ell}^n = f_{k,\ell}^* / \varphi_k$. For $k \gg k_c$, due to the highly oscillating nature of $Y_{k,\ell}$, the factors $Y_{k,\ell}(\boldsymbol{y}_j)$ are essentially decorrelated random numbers with zero mean and finite variance, since the values of $(Y_{k,\ell}(\boldsymbol{y}_j))^2$ are limited by the addition theorem [Eq. A.5](#). Let us denote the variance with $\sigma_Y$. By the central limit theorem, $g_{k,\ell}^n$ converges to a Gaussian random variable with zero mean and finite variance $\sigma_Y^2 \sum_{j=1}^n g_j^2$. As a result,

$$\bar{\epsilon}(n) \sim \sum_{k \geq k_c} \sum_{\ell=1}^{\mathcal{N}_{k,d}} \left\{ \left( \sum_{j=1}^n g_j^2 \right) \varphi_k^2 + \mathbb{E}_{f^*} \left[ \left( f_{k,\ell}^* \right)^2 \right] \right\}$$

$$= \left( \sum_{j=1}^n g_j^2 \right) \sum_{k \geq k_c} \mathcal{N}_{k,d} \varphi_k^2 + \sum_{k \geq k_c} \mathcal{N}_{k,d} c_k, \tag{E.4}$$

where we have used the definition of $f^*$ ([Eq. 2.1](#)) to set the expectation of $(f_{k,\ell}^*)^2$ to $c_k$.

**Large $\nu_t$ case** When $f^*$ is smooth enough the error is controlled by the predictor term proportional to $\sum_{j=1}^n g_j^2$. More specifically, if

$$\sum_{k \geq 0} \sum_{\ell=1}^{\mathcal{N}_{k,d}} \frac{c_k}{\varphi_k^2} < +\infty, \tag{E.5}$$

then the function $g^n(\boldsymbol{x})$ converges to the square-summable function $g^*(\boldsymbol{x})$ such that $f^*(\boldsymbol{x}) = \int g^*(\boldsymbol{y}) \varphi(\boldsymbol{x} \cdot \boldsymbol{y}) \, d\tau(\boldsymbol{y})$. With $c_k \sim k^{-2\nu_t - (d-1)}$ and $\mathcal{N}_{k,d} \sim k^{d-2}$, in the lazy regime $\varphi_k \sim k^{-(d-1)-2\nu}$ [Eq. E.5](#) is satisfied when $2\nu_t > 2(d-1) + 4\nu$ ($\nu = 1/2$ for the NTK and $3/2$ for the RFK). In the feature regime $\varphi_k \sim k^{-(d-1)/2-3/2}$, [Eq. E.5](#) is satisfied when $2\nu_t > (d-1) + 3$. If $g^n(\boldsymbol{x})$ converges to a square-summable function, then

$$\sum_{j=1}^n g_j^2 = \frac{1}{n} \int g^n(\boldsymbol{x})^2 \, d\tau(\boldsymbol{x}) + o(n^{-1}) = \frac{1}{n} \sum_{k \geq 0} \mathcal{N}_{k,d} \frac{c_k}{\varphi_k^2} + o(n^{-1}), \tag{E.6}$$

which is proportional to $n^{-1}$. In addition, since $\mathcal{N}_{k,d} \sim k^{d-2}$ and $k_c \sim n^{\frac{1}{d-1}}$, one has

$$n^{-1} \sum_{k \geq k_c} \mathcal{N}_{k,d} \varphi_k \sim \begin{cases} n^{-1} k^{d-1} k^{-2(d-1)-4\nu} \Big|_{k=n^{\frac{1}{d-1}}} \sim n^{-2-\frac{4\nu}{d-1}} \text{ (Lazy)}, \\ n^{-1} k^{d-1} k^{-(d-1)-3} \Big|_{k=n^{\frac{1}{d-1}}} \sim n^{-1-\frac{3}{d-1}} \text{ (Feature)}, \end{cases} \tag{E.7}$$

and

$$\sum_{k \geq k_c} \mathcal{N}_{k,d} c_k \sim k^{d-1} k^{-2\nu_t - (d-1)} \Big|_{k=n^{\frac{1}{d-1}}} \sim n^{-\frac{2\nu_t}{d-1}}. \tag{E.8}$$

Hence, if $\nu_t$ is large enough so that [Eq. E.5](#) is satisfied, the asymptotic decay of the error is given by [Eq. E.7](#).

**Small $\nu_t$ case** If [Eq. E.7](#) does not hold then $g^n(\boldsymbol{x})$ is not square-summable in the limit $n \to \infty$. However, for large but finite $n$ only the modes up to the $k_c$-th are correctly reconstructed, therefore

$$\sum_{j=1}^n g_j^2 \sim \frac{1}{n} \sum_{k \leq k_c} \mathcal{N}_{k,d} \frac{c_k}{\varphi_k^2} \sim \begin{cases} n^{-1} k^{-2\nu_t} k^{2(d-1)+4\nu} \Big|_{k=n^{\frac{1}{d-1}}} \sim n^{-\frac{2\nu_t}{d-1}} n^{1+\frac{4\nu}{d-1}} \text{ (Lazy)}, \\ n^{-1} k^{-2\nu_t} k^{(d-1)+3} \Big|_{k=n^{\frac{1}{d-1}}} \sim n^{-\frac{2\nu_t}{d-1}} n^{\frac{3}{d-1}} \text{ (Feature)}, \end{cases} \tag{E.9}$$

Both for feature and lazy, multiplying the term above by $\sum_{k \geq k_c} \mathcal{N}_{k,d} \varphi_k$ from [Eq. E.7](#) yields $\sim n^{-2\nu_t/(d-1)}$. This is also the scaling of the target function term [Eq. E.8](#), implying that for small $\nu_t$ one has

$$\bar{\epsilon}(n) \sim n^{-\frac{2\nu_t}{d-1}} \tag{E.10}$$

both in the feature and in the lazy regimes.

# F  Spectral bias via the replica calculation

Due to the equivalence with kernel methods, the asymptotic decay of the test error in the lazy regime can be computed with the formalism of [45], which also provides a non-rigorous justification for the spectral bias ansatz. By ranking the eigenvalues from the biggest to the smallest, such that $\varphi_\rho$ denotes the $\rho$-th eigenvalue and denoting with $c_\rho$ the variance of the projections of the target onto the $\rho$-th eigenfunction, one has

$$\epsilon(n) = \sum_\rho \epsilon_\rho(n), \quad \epsilon_\rho(n) = \frac{\kappa(n)^2}{(\varphi_\rho + \kappa(n))^2} c_\rho, \quad \kappa(n) = \frac{1}{n} \sum_\rho \frac{\varphi_\rho \kappa(n)}{\varphi_\rho + \kappa(n)}. \quad (F.1)$$

It is convenient to introduce the eigenvalue density,

$$\mathcal{D}(\varphi) := \sum_{k \geq 0} \sum_{l=1}^{\mathcal{N}_{k,d}} \delta(\varphi - \varphi_k) = \sum_{k \geq 0} \mathcal{N}_{k,d} \delta(\varphi - \varphi_k) \sim \int_0^\infty k^{d-2} \delta(\varphi - k^{-(d-1)-2\nu}) \text{ for } k \gg 1. \quad (F.2)$$

After changing variables in the delta function, one finds

$$\mathcal{D}(\varphi) \sim \varphi^{-\frac{2(d-1)+2\nu}{(d-1)+2\nu}} \text{ for } \varphi \ll 1. \quad (F.3)$$

This can be used for inferring the asymptotics of $\kappa(n)$,

$$\begin{aligned}
\kappa(n) &= \frac{1}{n} \sum_\rho \frac{\varphi_\rho \kappa(n)}{\varphi_\rho + \kappa(n)} \sim \frac{1}{n} \int d\varphi \, \mathcal{D}(\varphi) \frac{\varphi \kappa(n)}{\varphi + \kappa(n)} \\
&\sim \frac{1}{n} \int_0^{\kappa(n)} d\varphi \, \mathcal{D}(\varphi) \varphi + \frac{\kappa(n)}{n} \int_{\kappa(n)}^{\varphi_0} d\varphi \, \mathcal{D}(\varphi) \\
&\sim \frac{1}{n} \kappa(n)^{1 - \frac{(d-1)}{(d-1)+2\nu}} \Rightarrow \kappa(n) \sim n^{-1 - \frac{2\nu}{d-1}}.
\end{aligned} \quad (F.4)$$

Once the scaling of $\kappa(n)$ has been determined, the modal contributions to the error can be split according to whether $\varphi_\rho \ll \kappa(n)$ or $\varphi_\rho \gg \kappa(n)$. The scaling of $\varphi_\rho$ with the rank $\rho$ is determined self-consistently,

$$\rho \sim \int_{\varphi_\rho}^{\varphi_1} d\varphi \, \mathcal{D}(\varphi) \sim \varphi_\rho^{-\frac{d-1}{(d-1)+2\nu}} \Rightarrow \varphi_\rho \sim \rho^{-1 - \frac{2\nu}{d-1}} \Rightarrow \varphi_\rho \gg (\ll) \kappa(n) \Leftrightarrow \rho \ll (\gg) n. \quad (F.5)$$

Therefore

$$\epsilon(n) \sim \kappa(n)^2 \sum_{\rho \ll n} \frac{c_\rho}{\varphi_\rho^2} + \sum_{\rho \gg n} c_\rho. \quad (F.6)$$

Notice that $\kappa(n)^2$ scales as $n^{-1} \sum_{k \geq k_c} \mathcal{N}_{k,s} \varphi_k$ in Eq. E.7, whereas $\sum_{\rho \ll n} c_\rho / \varphi_\rho^2$ corresponds to $n \sum_j g_j^2$ in Eq. E.9, so that the first term on the right-hand side of Eq. F.6 matches that of Eq. E.4. The same matching is found for the second term on the right-hand side of Eq. F.6, so that the replica calculation justifies the spectral bias ansatz.

# G  Training wide neural networks: does gradient descent (GD) find the minimal-norm solution?

In the main text we provided predictions for the asymptotics of the test error of the minimal norm solution that fits all the training data. Does the prediction hold when solution of Eq. 2.5 and Eq. 2.13 is approximately found by GD? More specifically, is the solution found by GD the minimal-norm one?

**Feature Learning**   We answer these questions by performing full-batch gradient descent in two settings (further details about the trainings are provided in the code repository, `experiments.md` file),

Overfitting in Feature Learning

1. **Min-L1.** Here we update weights and features of Eq. 2.3, with $\xi = 0$, by following the negative gradient of

$$\mathcal{L}_{\text{Min-L1}} = \frac{1}{2n} \sum_{i=1}^{n} \left( f^*(\boldsymbol{x}_i) - f(\boldsymbol{x}_i) \right)^2 + \frac{\lambda}{H} \sum_{h=1}^{H} |w_h|, \tag{G.1}$$

with $\lambda \to 0^+$. The weights $w_h$ are initialized to zero and the features are initialized uniformly and constrained to be on the unit sphere.

2. **$\alpha$-trick.** Following [8], here we minimize

$$\mathcal{L}_{\alpha\text{-trick}} = \frac{1}{2n\alpha} \sum_{i=1}^{n} \left( f^*(\boldsymbol{x}_i) - \alpha f(\boldsymbol{x}_i) \right)^2, \tag{G.2}$$

with $\alpha \to 0$. This trick allows to be far from the lazy regime by forcing the weights to evolve to $\mathcal{O}(1/\alpha)$, when fitting a target of order 1.

In both cases, the solution found by GD is sparse, in the sense that is supported on a finite number of neurons – in other words, the measure $\gamma(\boldsymbol{\theta})$ becomes atomic, satisfying Assumption 1. Furthermore, we find that

1. For **Min-L1**, the generalization error prediction holds (Fig. 4 and Fig. G.1) as the the minimal norm solution if effectively recovered, see Fig. G.2. Such clean results in terms of features position are difficult to achieve for large $n$ because the training dynamics becomes very slow and reaching convergence becomes computationally infeasible. Still, we observe the test error to plateau and reach its infinite-time limit much earlier than the parameters, which allows for the scaling predictions to hold.

2. $\alpha$-**trick**, however, does not recover the minimal-norm solution, Fig. G.2. Still, the solution found is of the type (2.7) as it is sparse and supported on a number of atoms that scales linearly with $n$, Fig. G.3, left. For this reason, we find that our predictions for the generalization error hold also in this case, see Fig. G.3, right.

**Lazy Learning**   In this case, the correspondence between the solution found by gradient descent and the minimal-norm one is well established [9]. Therefore, numerical experiments are performed here via kernel regression and the analytical NTK Eq. A.19: given a dataset $\{\boldsymbol{x}_i, y_i = f^*(\boldsymbol{x}_i)\}_{i=1}^{n}$, we define the gram matrix $\mathbf{K} \in \mathbb{R}^{n \times n}$ with elements $\mathbf{K}_{ij} = K(\boldsymbol{x}_i, \boldsymbol{x}_j)$ and the vector of target labels $\boldsymbol{y} = [y_1, y_2, \ldots, y_n]$. The $q_i$'s in Eq. 2.9 can be easily recovered by solving the linear system

$$\boldsymbol{y} = \frac{1}{n} \mathbf{K} \boldsymbol{q}. \tag{G.3}$$

**Experiments**   Numerical experiments are run with PyTorch on GPUs NVIDIA V100 (university internal cluster). Details for reproducing experiments are provided in the code repository, `experiments.md` file. Individual trainings are run in 1 minute to 1 hour of wall time. We estimate a total of a thousand hours of computing time for running the preliminary and actual experiments present in this work.

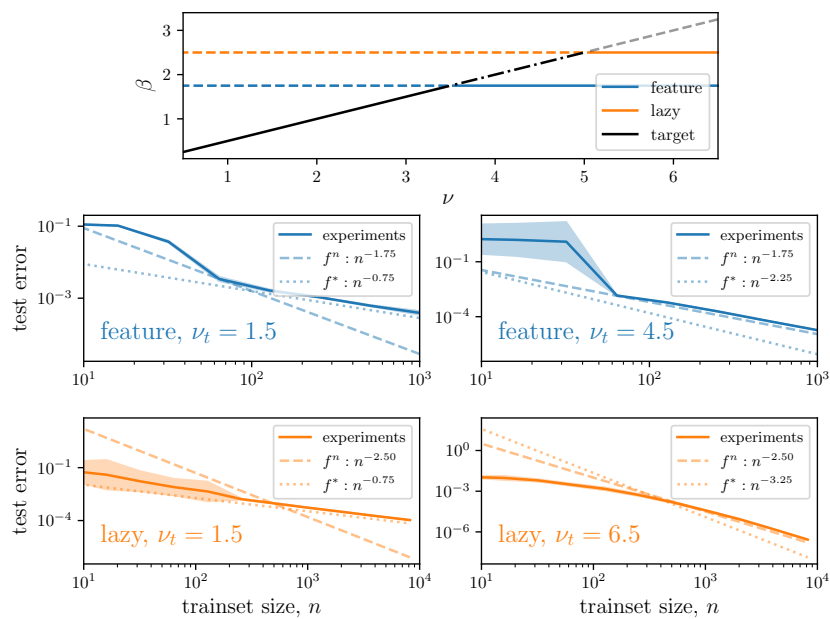

Figure G.1: **Gen. error decay vs. target smoothness and training regime.** Here, data-points are sampled uniformly from the spherical surface in $d = 5$ and the target function is an infinite-width FCN with activation function $\sigma(\cdot) = |\cdot|^{\nu_t - 1/2}$, corresponding to a Gaussian random process of smoothness $\nu_t$. 1st row: gen. error decay exponent as a function of the target smoothness $\nu_t$. The three curves correspond to the target contribution to the generalization error (black) and the predictor contribution in either feature (blue) or lazy (orange) regime. Full lines highlight the dominating contributions to the gen. error. 2nd row: agreement between predictions and experiments in the feature regime for a non-smooth (left) and smooth (right) target. In the first case, the error is dominated by the target $f^*$, in the second by the predictor $f^n$ – predicted exponents $\beta$ are indicated in the legends. 3rd row: analogous of the previous row for the lazy regime.

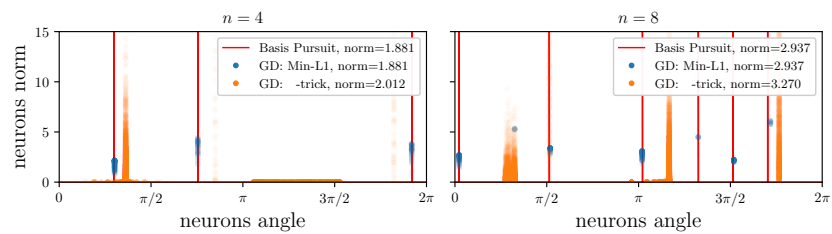

Figure G.2: **Comparing solutions.** Solutions to the spherically symmetric task in $d = 2$ for $n = 4$ (left) and $n = 8$ (right) training points. In red the minimal norm solution (Eq. 2.5) as found by Basis Pursuit [50]. Solutions found by GD in the Min-L1 and $\alpha$-trick setting are respectively shown in blue and orange. Dots correspond to single neurons in the network. The $x$-axis reports their angular position while the $y$-axis reports their norm: $|w_h| \|\boldsymbol{\theta}_h\|_2$. The total norm of the solutions, $\frac{\alpha}{H} \sum_{h=1}^{H} |w_h| \|\boldsymbol{\theta}_h\|_2$, is indicated in the legend.

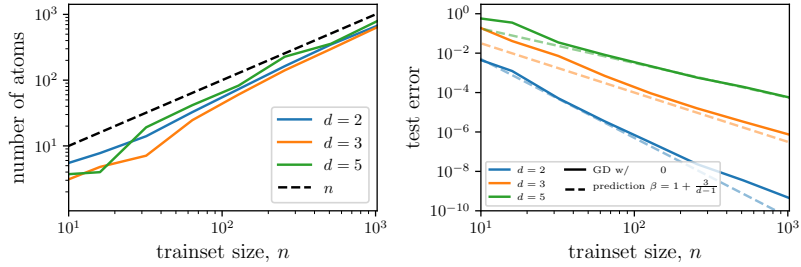

Figure G.3: **Solution found by the $\alpha$-trick.** We consider here the case of approximating the constant target function on $\mathbb{S}^{d-1}$ with an FCN. Training is performed starting from small initialization through the $\alpha$-trick. Left: Number of atoms $n_A$ as a function of the number of training points $n$. Neurons that are active on the same subset of the training set are grouped together and we consider each group a distinct atom for the counting. Right: Generalization error in the same setting (full), together with the theoretical predictions (dashed). Different colors correspond to different input dimensions. The case of $d = 2$ and large $n$ suffers from the same finite time effects discussed in Fig. 4. Results are averaged over 10 different initializations of the networks and datasets.

## H  Sensitivity of the predictor to transformations other than diffeomorphisms

This section reports experiments to integrate the discussion of section 5. In particular, we: *(i)* show that the lazy regime predictor is less sensitive to image translations than the feature regime one (as is the case for deformations, from Fig. 6); *(ii)* provide evidence of the positive effects of learning features in image classifications, namely becoming invariant to pixels at the border of images which are unrelated to the task.

To prove the above points we consider, as in Fig. 6, the relative sensitivity of the predictors of lazy and feature regime with respect to global translations for point *(i)* and corruption of the boundary pixels for point *(ii)*. The relative sensitivity to translations is obtained from Eq. 5.1 after replacing the transformation $\tau$ with a one-pixel translation of the image in a random direction. For the relative sensitivity to boundary corruption, the transformation consists in adding zero-mean and unit-variance Gaussian numbers to the boundary pixels. Both relative sensitivities are plotted in Fig. H.1, with translations on the left and boundary pixels corruption on the right.

In section 5 we then argue that differences in performance between the two training regimes can be explained by gaps in sensitivities with respect to input transformations that do not change the label. For *(i)*, the gap is similar to the one observed for diffeomorphisms (Fig. 6). Still, the space of translations has negligible size with respect to input space, hence we expect the diffeomorphisms to have a more prominent effect. In case *(ii)*, the feature regime is less sensitive with respect to irrelevant pixels corruption and this would give it an advantage over the lazy regime. The fact that the performance difference is in favor of the lazy regime instead, means that these transformations only play a minor role.

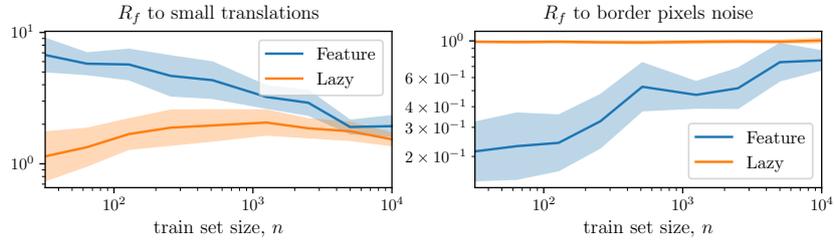

Figure H.1: **Sensitivity to input transformations vs number of training points.** Relative sensitivity of the predictor to (left) random 1-pixel translations and (right) white noise added at the boundary of the input images, in the two regimes, for varying number of training points $n$ and when training on FashionMNIST. Smaller values correspond to a smoother predictor, on average. Results are computed using the same predictors as in Fig. 1. Left: For small translations, the behavior is the same compared to applying diffeomorphisms. Right: The lazy regime does not distinguish between noise added at the boundary or on the whole image ($R_f = 1$), while the feature regime gets more insensitive to the former.

## I Maximum-entropy model of diffeomorphisms

We briefly review here the maximum-entropy model of diffeomorphisms as introduced in [49].

An image can be thought of as a function $x(s)$ describing intensity in position $s = (u, v) \in [0, 1]^2$, where $u$ and $v$ are the horizontal and vertical (pixel) coordinates. Denote $\tau x$ the image deformed by $\tau$, i.e. $[\tau x](s) = x(s - \tau(s))$. [49] propose an ensemble of diffeomorphisms $\tau(s) = (\tau_u, \tau_v)$ with i.i.d. $\tau_u$ and $\tau_v$ defined as

$$\tau_u = \sum_{i, j \in \mathbb{N}^+} C_{ij} \sin(i\pi u) \sin(j\pi v) \tag{I.1}$$

where the $C_{ij}$'s are Gaussian variables of zero mean and variance $T/(i^2 + j^2)$ and $T$ is a parameter controlling the deformation magnitude. Once $\tau$ is generated, pixels are displaced to random positions. See Fig. 5b for an example of such transformation.

# 5 When Feature Learning Succeeds: How Deformation Invariance is learned in Convolutional Neural Networks

The following paper is the preprint version of Tomasini et al. (2022a) appeared at the *ICLR 2023 Workshop on Physics for Machine Learning* and is currently under review.

**Candidate contributions**    The candidate contributed to all discussions and led research regarding the first part of the paper (Sections 2 and 3).

# HOW DEEP CONVOLUTIONAL NEURAL NETWORKS LOSE SPATIAL INFORMATION WITH TRAINING

**Umberto M. Tomasini** [*], **Leonardo Petrini** [*], **Francesco Cagnetta**, **Matthieu Wyart**
Institute of Physics
École Polytechnique Fédérale de Lausanne
`name.surname@epfl.ch`

### ABSTRACT

A central question of machine learning is how deep nets manage to learn tasks in high dimensions. An appealing hypothesis is that they achieve this feat by building a representation of the data where information irrelevant to the task is lost. For image datasets, this view is supported by the observation that after (and not before) training, the neural representation becomes less and less sensitive to diffeomorphisms acting on images as the signal propagates through the net. This loss of sensitivity correlates with performance, and surprisingly correlates with a *gain* of sensitivity to white noise acquired during training. These facts are unexplained, and as we demonstrate still hold when white noise is added to the images of the training set. Here, we *(i)* show empirically for various architectures that stability to image diffeomorphisms is achieved by both spatial and channel pooling, *(ii)* introduce a model scale-detection task which reproduces our empirical observations on spatial pooling and *(iii)* compute analitically how the sensitivity to diffeomorphisms and noise scales with depth due to spatial pooling. The scalings are found to depend on the presence of strides in the net architecture. We find that the increased sensitivity to noise is due to the perturbing noise piling up during pooling, after being rectified by ReLU units.

## 1 INTRODUCTION

Deep learning algorithms can be successfully trained to solve a large variety of tasks (Amodei et al., 2016; Huval et al., 2015; Mnih et al., 2013; Shi et al., 2016; Silver et al., 2017), often revolving around classifying data in high-dimensional spaces. If there was little structure in the data, the learning procedure would be cursed by the dimension of these spaces: achieving good performances would require an astronomical number of training data (Luxburg & Bousquet, 2004). Consequently, real datasets must have a specific internal structure that can be learned with fewer examples. It has been then hypothesized that the effectiveness of deep learning lies in its ability of building 'good' representations of this internal structure, which are insensitive to aspects of the data not related to the task (Ansuini et al., 2019; Shwartz-Ziv & Tishby, 2017; Recanatesi et al., 2019), thus effectively reducing the dimensionality of the problem.

In the context of image classification, Bruna & Mallat (2013); Mallat (2016) proposed that neural networks lose irrelevant information by learning representations that are insensitive to small deformations of the input, also called diffeomorphisms. This idea was tested in modern deep networks by Petrini et al. (2021), who introduced the following measures

$$D_f = \frac{\mathbb{E}_{x,\tau}\|f(\tau(x)) - f(x)\|^2}{\mathbb{E}_{x_1,x_2}\|f(x_1) - f(x_2)\|^2}, \qquad G_f = \frac{\mathbb{E}_{x,\eta}\|f(x+\eta) - f(x)\|^2}{\mathbb{E}_{x_1,x_2}\|f(x_1) - f(x_2)\|^2}, \qquad R_f = \frac{D_f}{G_f}, \quad (1)$$

to probe the sensitivity of a function $f$—either the output or an internal representation of a trained network—to random diffeomorphisms $\tau$ of $x$ (see example in Fig. 1, left), to large white noise perturbations $\eta$ of magnitude $\|\tau(x) - x\|$, and in relative terms, respectively. Here the input images $x$, $x_1$ and $x_2$ are sampled uniformly from the test set. In particular, the test error of trained networks is correlated with $D_f$ when $f$ is the network output. Less intuitively, the test error is anti-correlated with the sensitivity to white noise $G_f$. Overall, it is the relative sensitivity $R_f$ which correlates best with the error (Fig. 1, middle). This correlation is learned over training—as it is not seen at initialization—and built up layer by layer (Petrini et al., 2021). These phenomena are not simply due to benchmark data being noiseless, as they persist when input images are corrupted by some small noise (Fig. 1, right).

---

[*]Equal contribution.

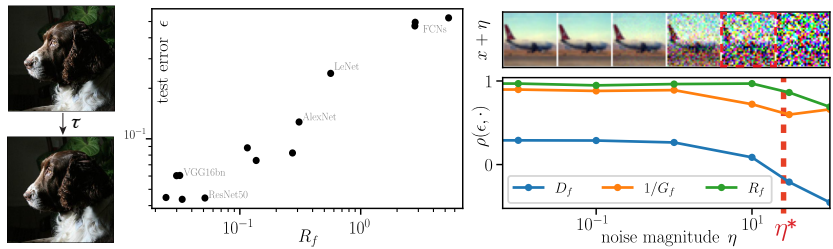

Figure 1: Left: example of a random diffeomorphism $\tau$ applied to an image. Center: test error vs relative sensitivity to diffeomorphisms of the predictor for a set of networks trained on CIFAR10, adapted from Petrini et al. (2021). Right: Correlation coefficient between test error $\epsilon$ and $D_f$, $G_f$ and $R_f$ when training different architectures on noisy CIFAR10, $\rho(\epsilon, X) = \mathrm{Cov}(\log \epsilon, \log X)/\sqrt{\mathrm{Var}(\log \epsilon)\mathrm{Var}(\log X)}$. Increasing noise magnitudes are shown on the $x$-axis and $\eta^* = \mathbb{E}_{\tau,x}\|\tau(x) - x\|^2$ is the one used for the computation of $G_f$. Samples of a noisy CIFAR10 datum are shown on top. Notice that $D_f$ and particularly $R_f$ are positively correlated with $\epsilon$, whilst $G_f$ is negatively correlated with $\epsilon$. The corresponding scatter plots are in Fig. 10 (appendix).

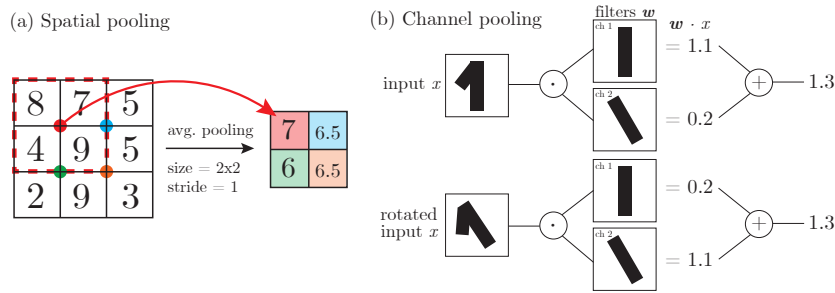

Figure 2: Spatial vs. channel pooling. (a) Spatial average pooling (size 2x2, stride 1) computed on a representation of size 3x3. One can notice that nearby pixel variations are smaller after pooling. (b) If the filters of different channels are identical up to e.g. a rotation of angle $\theta$, then, averaging the output of the application of such filters makes the result invariant to input rotations of $\theta$. This averaging is an example of channel pooling.

Operations that grant insensitivity to diffeomorphisms in a deep network have been identified previously (e.g. Goodfellow et al. (2016), section 9.3, sketched in Fig. 2). The first, *spatial* pooling, integrates local patches within the image, thus losing the exact location of its features. The second, *channel* pooling, requires the interaction of different channels, which allows the network to become invariant to any local transformation by properly learning filters that are transformed versions of one another. However, it is not clear whether these operations are actually learned by deep networks and how they conspire in building good representations. Here we tackle this question by unveiling empirically the emergence of spatial and channel pooling, and disentangling their role. Below is a detailed list of our contributions.

## Chapter 5. When Feature Learning Succeeds: How Deformation Invariance is learned in Convolutional Neural Networks

---

- We disentangle the role of spatial and channel pooling within deep networks trained on CIFAR10 (Section 2). More specifically, our experiments reveal the significant contribution of spatial pooling in decreasing the sensitivity to diffeomorphisms.

- In order to isolate the contribution of spatial pooling and quantify its relation with the sensitivities to diffeomorphism and noise, we introduce idealized scale-detection tasks (Section 3). In these tasks, data are made of two active pixels and classified according to their distance. We find the same correlations between test error and sensitivities of trained networks as found in Petrini et al. (2021). In addition, the neural networks which perform the best on real data tend to be the best on these tasks.

- We theoretically analyze how simple CNNs, made by stacking convolutional layers with filter size $F$ and stride $s$, learn these tasks (Section 4). We find that the trained networks perform spatial pooling for most of its layers. We show and verify empirically that the sensitivities $D_k$ and $G_k$ of the $k$-th hidden layer follow $G_k \sim A_k$ and $D_k \sim A_k^{-\alpha_s}$, where $A_k$ is the effective receptive field size and $\alpha_s = 2$ if there is no stride, $\alpha_s = 1$ otherwise.

The code and details for reproducing experiments are available online at github.com/leonardopetrini/relativestability/experiments_ICLR23.md.

### 1.2 RELATED WORK

In the neuroscience literature, the understanding of the relevance of pooling in building invariant representations dates back to the pioneering work of Hubel & Wiesel (1962). By studying the cat visual cortex, they identified two different kinds of neurons: simple cells responding to e.g. edges at specific angles and complex cells that *pool* the response of simple cells and detect edges regardless of their position or orientation in the receptive field. More recent accounts of the importance of learning invariant representations in the visual cortex can be found in Niyogi et al. (1998); Anselmi et al. (2016); Poggio & Anselmi (2016).

In the context of artificial neural networks, layers jointly performing spatial pooling and strides have been introduced with the early CNNs of Lecun et al. (1998), following the intuition that local averaging and subsampling would reduce the sensitivity to small input shifts. Ruderman et al. (2018) investigated the role of spatial pooling and showed empirically that networks with and without pooling layers converge to similar deformation stability, suggesting that spatial pooling can be learned in deep networks. In our work, we further expand in this direction by jointly studying diffeomorphisms and noise stability and proposing a theory of spatial pooling for a simple task.

The depth-wise loss of irrelevant information in deep networks has been investigated by means of the information bottleneck framework (Shwartz-Ziv & Tishby, 2017; Saxe et al., 2019) and the intrinsic dimension of the networks internal representations (Ansuini et al., 2019; Recanatesi et al., 2019). However, these works do not specify what is the irrelevant information to be disregarded, nor the mechanisms involved in such a process.

The stability of trained networks to noise is extensively studied in the context of adversarial robustness (Fawzi & Frossard, 2015; Kanbak et al., 2018; Alcorn et al., 2019; Alaifari et al., 2018; Athalye et al., 2018; Xiao et al., 2018a; Engstrom et al., 2019). Notice that our work differs from this literature by the fact that we consider typical perturbations instead of worst-case ones.

## 2 EMPIRICAL OBSERVATIONS ON REAL DATA

In this section we analyze the parameters of deep CNNs trained on CIFAR10 and ImageNet, so as to understand how they build representations insensitive to diffeomorphisms (details of the experiments in App. B). The analysis builds on two premises, the first being the assumption that insensitivity is built layer by layer in the network, as shown in Fig. 3. Hence, we focus on how each of the layers in a deep network contribute towards creating an insensitive representation. More specifically, let us denote with $f_k(x)$ the internal representation of an input $x$ at the $k$-th layer of the network. The entries of $f_k$ have three indices, one for the channel $c$ and two for the spatial location $(i, j)$. The relation between $f_k$ and $f_{k-1}$ is the following,

$$[f_k(x)]_{c;i,j} = \phi \left( b_c^k + \sum_{c'=1}^{H_{k-1}} \boldsymbol{w}_{c,c'}^k \cdot \boldsymbol{p}_{i,j}\left([f_{k-1}(x)]_{c'}\right) \right) \quad \forall\, c = 1, \dots, H_k, \qquad (2)$$

where: $H_k$ denotes the number of channels at the $k$-th layer; $b_c^k$ and $\boldsymbol{w}_{c,c'}^k$ the biases and *filters* of the $k$-th layer; each filter $\boldsymbol{w}_{c,c'}^k$ is a $F \times F$ matrix with $F$ the filter size; $\boldsymbol{p}_{i,j}\left([f_{k-1}(x)]_{c'}\right)$ denotes a $F \times F$-dimensional patch of $[f_{k-1}(x)]_{c'}$ centered at $(i,j)$; $\phi$ the activation function. The second premise is that a general diffeomorphism can be represented as a displacement field over the image, which indicates how each pixel moves in the transformation. Locally, this displacement field can be decomposed into a constant term and a linear part: the former corresponds to local translations, the latter to stretchings, rotations and shears.[1]

**Invariance to translations via spatial pooling.** Due to weight sharing, i.e. the fact that the same filter $\boldsymbol{w}_{c,c'}^k$ is applied to all the local patches $(i,j)$ of the representation, the output of a convolutional layer is *equivariant* to translations by construction: a shift of the input is equivalent to a shift of the output. To achieve an *invariant* representation it suffices to sum up the spatial entries of $f_k$—an operation called pooling in CNNs, we refer to it as *spatial* pooling to stress that the sum runs over the spatial indices of the representation. Even if there are no pooling layers at initialization, they can be realized by having homogeneous filters, i.e. all the $F \times F$ entries of $\boldsymbol{w}_{c,c'}^{k+1}$ are the same. Therefore, the closer the filters are to the homogeneous filter, the more they decrease the sensitivity of the representation to local translations.

**Invariance to other transformations via channel pooling.** The example of translations shows that building invariance can be performed by constructing an equivariant representation, and then pooling it. Invariance can also be built by pooling *across* channels. A two-channel example is shown Fig. 2, panel (b), where the filter of the second channel is built so as to produce the same output as the first channel when applied to a rotated input. The same idea can be applied more generally, e.g. to the other components of diffeomorphisms—such as local stretchings and shears. Below, we refer generically to any operation that build invariance to diffeomorphisms by assembling distinct channels as *channel pooling*.

**Disentangling spatial and channel pooling.** The relative sensitivity to diffeomorphisms $R_k$ of the $k$-th layer representation $f_k$ decreases after each layer, as shown in Fig. 3. This implies that spatial or channel pooling are carried out along the whole network. To disentangle their contribution we perform the following experiment: shuffle at random the connections between channels of successive convolutional layers, while keeping the weights unaltered. Channel shuffling amounts to randomly permuting the values of $c, c'$ in Eq. 2, therefore it breaks any channel pooling while not affecting single filters. The values of $R_k$ for deep networks after channel shuffling are reported in Fig. 3 as dashed lines and compared with the original values of $R_k$ in full lines. If only spatial pooling was present in the network, then the two curves would overlap. Conversely, if the decrease in $R_k$ was all due to the interactions between channels, then the shuffled curves should be constant. Given that neither of these scenarios arises, we conclude that both kinds of pooling are being performed.

**Emergence of spatial pooling after training.** To bolster the evidence for the presence of spatial pooling, we analyze the filters of trained networks. Since spatial pooling can be built by having homogeneous filters, we test for its presence by looking at the frequency content of learned filters $\boldsymbol{w}_{c,j}^k$. In particular, we consider the average squared projection of filters onto "Fourier modes" $\{\Psi_l\}_{l=1,\dots,F^2}$, taken as the eigenvectors of the discrete Laplace operator on the $F \times F$ filter grid. The square projections averaged over channels read

$$\gamma_{k,l} = \frac{1}{H_{k-1} H_k} \sum_{c=1}^{H_k} \sum_{c'=1}^{H_{k-1}} \left[ \Psi_l \cdot \boldsymbol{w}_{c,c'}^k \right]^2, \qquad (3)$$

---

[1] The displacement field around a pixel $(u_0, v_0)$ is approximated as $\tau(u,v) \simeq \tau(u_0, v_0) + J(u_0, v_0)[u - u_0, v - v_0]^T$, where $\tau(u_0, v_0)$ corresponds to translations and $J$ is the Jacobian matrix of $\tau$ whose trace, antisymmetric and symmetric traceless parts correspond to stretchings, rotations and shears, respectively.

and are shown in Fig. 4, 1st and 2nd row. When training a deep network such as VGG11 (with and without batch-norm) (Simonyan & Zisserman, 2015) on CIFAR10, filters of layers 2 to 6 become low-frequency with training, while layers 1, 7, 8 do not. Accordingly, larger gaps between dashed and full lines in Fig. 3 (right) open at layer 1, 7, 8: reduction in sensitivity is not due to spatial pooling in these layers. Moreover, the fact that the two dashed curves overlap is consistent with the frequency content of filters being the same for the two architectures after training. In the case of ImageNet, filters at all layers become low-frequency, except for $k = 1$.

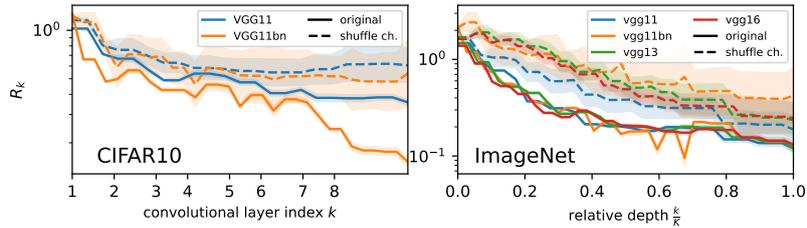

Figure 3: Relative sensitivity $R_k$ as a function of depth for VGG architectures trained on CIFAR10 (left) and ImageNet (right). Full lines refer to the original networks, dashed lines to the ones with shuffled channels. $K$ is the total depth of the networks. Experiments with different architectures are reported in App. C.

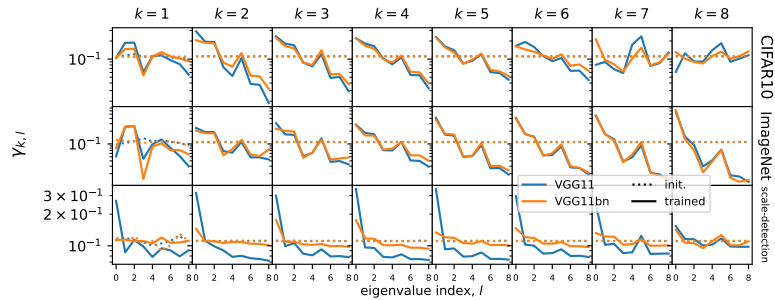

Figure 4: Projections of the network filters for VGG11 and VGG11bn onto the 9 eigenvectors of the $(3 \times 3)$-grid Laplacian when training on CIFAR10 (1st row), ImageNet, (2nd row) and the scale-detection task (3rd row): dotted and full lines correspond to initialization and trained networks, respectively. The $x$-axis reports low to high frequencies from left to right. Deeper layers are reported in rightmost panels. Low-frequency modes are the dominant components in layers 2-6 when training on CIFAR10, in layers 2-8 for ImageNet. The first (constant) mode has most of the power throughout the network for scale-detection task 1. An aggregate measure of the spatial frequency content of filters is reported in App. C, Fig. 12.

## 3  SIMPLE SCALE-DETECTION TASKS CAPTURE REAL-DATA OBSERVATIONS

To sum up, the empirical evidence presented in Section 2 indicates that *(i)* the generalization performance of deep CNNs correlates with their insensitivity to diffeomorphisms and sensitivity to Gaussian noise (Fig. 1); *(ii)* deep CNNs build their sensitivities layer by layer via spatial and channel pooling. We introduce now two idealized scale-detection tasks where the phenomena *(i)* and *(ii)* emerge again, and we can isolate the contribution of spatial pooling. Given the simpler structure of these tasks with respect to real data, we can understand quantitatively how spatial pooling builds up insensitivity to diffeomorphisms and sensitivity to Gaussian noise, as we show in Section 4.

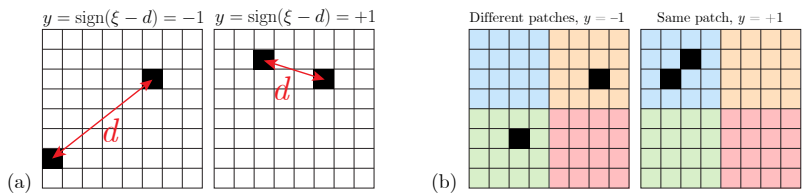

Figure 5: Example inputs for the scale-detection tasks. Task 1 (a): the label depends on whether the euclidean distance $d$ is larger (left) or smaller (right) than the characteristic scale $\xi$. Task 2 (b): the label depends on whether the active pixels belong to the same patch of size $\xi$ (right) or not (left)—patches are shown in different colors.

**Definition of scale-detection tasks.** Consider input images $x$ consisting of two active pixels on an empty background.

**Task 1:** Inputs are classified by comparing the euclidean distance $d$ between the two active pixels and some *characteristic scale* $\xi$, as in Fig. 5, left. Namely, the label is $y = \text{sign}(\xi - d)$.

Notice that a small diffeomorphism of such images corresponds to a small displacement of the active pixels. Specifically, each of the active pixels is moved to either of its neighboring pixels or left in its original position with equal probability.[2] By introducing a gap $g$ such that $d \in [\xi - \frac{g}{2}, \xi + \frac{g}{2}]$, task 1 becomes invariant to displacements of size smaller than $g$. Therefore, we expect that a neural network trained on task 1 will lose information on the exact location of the active pixels within the image, thus becoming insensitive to diffeomorphisms. Intuitively, spatial pooling up to the scale $\xi$ is the most direct mean to achieve such insensitivity. The result of the integration depends on whether none, one or both the active pixels lie within the pooling window, thus it is still informative of the task. We will show empirically that this is indeed the solution reached by trained CNNs.

**Task 2:** Inputs are partitioned into nonoverlapping patches of size $\xi$, as in Fig. 5, right. The label $y$ is $+1$ if the active pixels fall within the same patch, $-1$ otherwise.

In task 2, the irrelevant information is the location of the pixels within each of the non-overlapping patches. The simplest means to lose such information requires to couple spatial pooling with a stride of the size of the pooling window itself.

**Same phenomenology as in real image datasets.** Although these scale-detection tasks are much simpler than standard benchmark datasets, deep networks trained on task 1 display the same phenomenology highlighted in Section 2 for networks trained on CIFAR10 and ImageNet. First, the test error is positively correlated with the sensitivity to diffeomorphisms of the network predictor (Fig. 8, left panel, in App. C) and negatively correlated with its sensitivity to Gaussian noise (middle panel) for a whole range of architectures. As a result, the error correlates well with the relative sensitivity $R_f$ (right panel). Secondly, the internal representations of trained networks $f_k$ become progressively insensitive to diffeomorphisms and sensitive to Gaussian noise through the layers, as shown in Fig. 9 of App. C. Importantly, the curves relating sensitivities to the relative depth remain essentially unaltered if the channels of the networks are shuffled (shown as dashed lines in Fig. 9). We conclude that, on the one hand channel pooling is negligible, and, on the other hand, all channels are approximately equal to the mean channel. Finally, direct inspection of the filters (Fig. 4, bottom row) shows that the 0-frequency component grows much larger than the others over training for layers 1-7, which are the layers where $R_k$ decreases the most in Fig. 9. Filters are thus becoming nearly homogeneous, which means that the convolutional layers become effectively pooling layers.

---

[2] We fix the length of these displacements to 1 pixel because *(i)* is the smallest value that prevents the use of pixel interpolation, which would make one active pixel an extended object *(ii)* allows for the analysis of Section 4.

# Chapter 5.   When Feature Learning Succeeds: How Deformation Invariance is learned in Convolutional Neural Networks

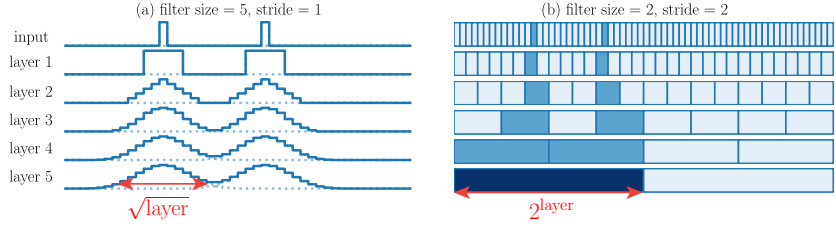

Figure 6: Hidden layers representations of simple CNNs for a scale-detection input for stride $s = 1$ and filter size $F = 5$ (left) and $s = F = 2$ (right) when having homogeneous filters at every layer. The effective receptive field size of the last layer in the two different cases is shown in red. (Left) every active pixel in the input becomes a Gaussian profile whose width increases throughout the network. (Right) every neuron in layer $k$ has activity equal to the number of active pixels which are present in its receptive field of width $2^k$. The dark blue in the last layer indicates that there are 2 active pixels in its receptive field, while the lighter blue of the precedent layers indicates that there is just 1.

## 4   THEORETICAL ANALYSIS OF SENSITIVITIES IN SCALE-DETECTION TASKS

We now provide a scaling analysis of the sensitivities to diffeomorphisms and noise in the internal representations of simple CNNs trained on the scale-detection tasks of Section 3. It allows to quantitatively understand how spatial pooling makes the internal representations of the network progressively more insensitive to diffeomorphisms and sensitive to Gaussian noise.

**Setup.**   We consider simple CNNs made by stacking $\tilde{K}$ identical convolutional layers with generic filter size $F$, stride $s = 1$ or $F$ and ReLU activation function $\phi(x) = \max(0, x)$. In particular, we train CNNs with stride 1 on task 1 and CNNs with stride $F$ on task 2. For the sake of simplicity, we consider the one-dimensional version of the scale-detection tasks, but our analysis carries unaltered to the two-dimensional case. Thus, input images are sequences $x = (x_i)_{i=1,\dots,L}$ of $L$ pixels, where $x_i = 0$ for all pixels except two. For the active pixels $x_i = \sqrt{L/2}$, so that all input images have $\|x\|^2 = L$. We will also consider single-pixel data $\delta_j = (\delta_{j,i})_{i=1,\dots,L}$. If the active pixels in $x$ are the $i$-th and the $j$-th, then $x = \sqrt{L/2}\,(\delta_i + \delta_j)$. For each layer $k$, the internal representation $f_k(x)$ of the trained network is defined as in Eq. 2. The *receptive field* of the $k$-th layer is the number of input pixels contributing to each component of $f_k(x)$. We define the *effective* receptive field $A_k$ as the typical size of the representation of a single-pixel input, $f_k(\delta_i)$, as illustrated in red in Fig. 6. We denote the sensitivities of the $k$-th layer representation with a subscript $k$ ($D_k$ for diffeomorphisms, $G_k$ for noise, $R_k$ for relative).

**Assumptions.**   All our results are based on the assumption that the first few layers of the trained network behave effectively as a single channel with a homogeneous positive filter and no bias. The equivalence of all the channels with their mean is supported by Fig. 9, which shows how shuffling channels does not affect the internal representations of VGGs. In addition, Fig. 4 (bottom row) shows that the mean filters of the first few layers are nearly homogeneous. We set the homogeneous value of each filter so as to keep the norm of representations constant over layers. Moreover, we implement a deformation of the input $x$ of our scale-detection tasks as a random displacement of each active pixel at either left or wight with probability 1/2.

### 4.1   TASK 1, STRIDE 1

For a CNN with stride 1, under the homogeneous filter assumption, the size of the effective receptive field $A_k$ grows as $\sqrt{k}$. A detailed proof is presented in App. A and Fig. 6, left panel, shows an illustration of the process. Intuitively, applying a homogeneous filter to a representation is equivalent to making each pixel diffuse, i.e. distributing its intensity uniformly over a neighborhood of size $F$. With a single-pixel input $\delta_i$, the effective receptive field of the $k$-th layer $f_k(\delta_i)$ is equivalent to a $k$-step diffusion of the pixel, thus it approaches a Gaussian distribution of standard deviation $\sqrt{k}$ centered at $i$. The size $A_k$ is the standard deviation, thus $A_k \sim \sqrt{k}$. The proof we present in [App. A](#) requires large depth $\tilde{K} \gg 1$ and large image width $L \gg F\tilde{K}^{1/2}$ and the empirical studies of [Section 3](#) satisfy these contraints ($F \sim 3$, $L \sim 32$ and $\tilde{K} \sim 10$).

We remark that at initialization, $f_k(x)$ behave, in the limit of large number of channels and width (and small bias), as Gaussian random fields with correlation matrix $\mathbb{E}\left[f_k(x)f_k(y)\right] \approx \delta(x - y)$, with $\delta$ the Dirac delta ([Schoenholz et al., 2017](#); [Xiao et al., 2018b](#)). This spiky correlation matrix implies that for any perturbation $y = x + \varepsilon$, the representation $f_k(y)$ changes with respect to $f_k(x)$ independently on $\varepsilon$. This behavior is remarkably different to the smooth case achieved by the diffusion, since after training. Consequently, both $D_k$ and $G_k$ are constant with respect to $k$ at initialization . This is consistent with the observations reported in [Fig. 7](#).

**Sensitivity to diffeomorphisms.** Let $i$ and $j$ denote the active pixels locations, so that $x \propto \delta_i + \delta_j$. Since both the elements of the inputs and those of the filters are non-negative, the presence of ReLU nonlinearities is irrelevant and the first few hidden layers are effectively linear layers. Hence the representations are linear in the input, so that $f_k(x) = f_k(\delta_i + \delta_j) = f_k(\delta_i) + f_k(\delta_j)$. In addition, since the effect of a diffeomorphism is just a 1-pixel translation of the representation irrespective of the original positions of the pixels, the normalized sensitivity $D_k$ can be approximated as follows

$$D_k \sim \frac{\|f_k(\delta_{i+1}) - f_k(\delta_i)\|_2^2}{\|f_k(\delta_i)\|_2^2}. \tag{4}$$

The denominator in [Eq. 4](#) is the squared norm of a Gaussian distribution of width $\sqrt{k}$, $\|f_k(v_i)\|_2^2 \sim k^{-1/2}$. The numerator compares $f_k$ with a small translation of itself, thus it can be approximated by the squared norm of the derivative of the Gaussian distribution, $\|f_k(\delta_{i+1}) - f_k(\delta_i)\|_2^2 \sim k^{-3/2}$. Consequently, we have

$$D_k \sim k^{-1} \sim A_k^{-2}. \tag{5}$$

**Sensitivity to Gaussian noise.** To analyze $G_k$ one must take into account the rectifying action of ReLU, which sets all the negative elements of its input to zero. The first ReLU is applied after the first homogeneous filters, thus the zero-mean noise is superimposed on a patch of $F$ active pixels. Outside such a patch, only positive noise terms survive. Within the patch, being summed to a positive background, also negative terms can survive the rectification of ReLU. Nevertheless, if the size of the image is much larger than the filter size, the contribution from active pixels to $G_k$ is negligible and we can approximate the difference between noisy and original representations $f_1(x + \eta) - f_1(x)$ with the rectified noise $\phi(\eta)$. After the first layer, the representations consist of non-negative numbers, thus we can forget again the ReLU and write

$$G_k \sim \frac{\mathbb{E}_\eta \|f_k(\phi(\eta))\|_2^2}{\|f_k(\delta_i)\|_2^2}. \tag{6}$$

Repeated applications of homogeneous filters to the rectified noise $\phi(\eta)$ result again in a diffusion of the signal. Since $\phi(\eta)$ has different independent and identically distributed non-zero entries for different realizations of $\eta$, averaging over $\eta$ is equivalent to considering a homogeneous profile for $f_k(\phi(\eta))$. As a result, the numerator in [Eq. 6](#) is a constant independent of $k$. The denominator is the same as in [Eq. 4](#), $\|f_k(\delta_i)\|_2^2 \sim k^{-1/2}$, hence

$$G_k \sim k^{1/2} \sim A_k, \tag{7}$$

i.e. the sensitivity to Gaussian noise grows as the size of the effective receptive fields. From the ratio of [Eq. 5](#) and [Eq. 7](#), we get $R_k \sim A_k^{-3}$.

### 4.2  TASK 2, STRIDE EQUAL FILTER SIZE

When the stride $s$ equals to the filter size $F$ the number of pixels of the internal representations is reduced by a factor $F$ at each layer, thus $f_k$ consists of $L/F^k$ pixels. Meanwhile, the effective size of the receptive fields grows exponentially at the same rate: $A_k = F^k$ (see [Fig. 6](#), left for an illustration).

**Sensitivity to diffeomorphisms.** For a given layer $k$, consider a partition of the input image into $L/F^k$ patches. Each pixel of $f_k$ only looks at one such patch and its intensity coincides with the number of active pixels within the patch. As a result, the only diffeomorphisms that change $f_k$ are those which move one of the active pixels from one patch to another. Since active pixels move by 1, this can only occur if one of the active pixels was originally located at the border of a patch, which in turn occurs with probability $\sim 1/F^k$. In addition, the norm $\|f_k(\delta_i)\|_2$ at the denominator does not scale with $k$, so that

$$D_k \sim F^{-k} \sim A_k^{-1}. \qquad (8)$$

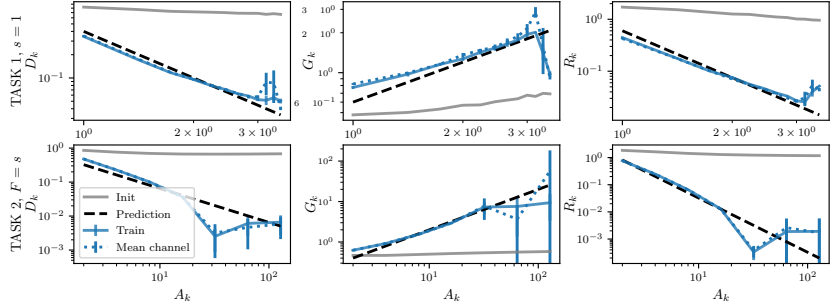

Figure 7: Sensitivities of internal representations $f_k$ of simple CNNs against the $k$-th layer receptive field size $A_k$ for trained networks (solid blue) and at initialization (solid gray). The top row refers to task 1 with $s = 1$ and $F = 3$; the bottom row to task 2 with $F = s = 2$. For a first large part of the network, the sensitivities obtained by replacing each layer with the mean channel (blue dotted) overlap with the original sensitivities. Predictions Eq. 5, Eq. 7 for task 1 and Eq. 8, Eq. 9 for task 2 are shown as black dashed lines.

**Sensitivity to Gaussian noise.** Each pixel of $f_k$ looks at a patch of the input of size $F^k$, thus $f_k$ is affected by the sum of all the noises acting on such patch. Since these noises have been rectified by ReLU, by the Central Limit Theorem the sum scales as the number of summands $F_k$. Thus, the contribution of each pixel of $f_k$ to the numerator of $G_k$ scales as $(F^k)^2$. As there are $L/F^k$ pixels in $f_k$, one has

$$G_k \sim (F^k)^2 \left( L/F^k \right) \sim F^k \sim A_k. \qquad (9)$$

Without rectification, the sum of $F^k$ independent noises would scale as the square root of the number of summands $F^k$, yielding a constant $G_k$. We conclude that the rectifying action of ReLU is crucial in building up sensitivity to noise. $R_k \sim A_k^{-2}$ follows from the ratio of Eq. 8 and Eq. 9.

### 4.3 COMPARING PREDICTIONS WITH EXPERIMENTS

We test our scaling predictions (Eq. 5 to Eq. 9) in Fig. 7, for stride 1 CNNs trained on task 1 and stride $F$ CNNs trained on task 2 in the top and bottom panels, respectively. Notice that if all the filters at a given layer are replaced with their average, the behavior of the sensitivities as a function of depth does not change (compare solid and dotted blue curves in the figure). This confirms our assumption that all channels behave like the mean channel. In addition, Tables 1 and 2 show that the mean filters are approximately homogeneous. Further details on the experiments are provided in App. B.

## 5 CONCLUSION

The meaning of an image often depends on sparse regions of the data, as evidenced by the fact that artists only need a small number of strokes to represent a visual scene. The exact locations of the features determining the image class are flexible, and indeed diffeomorphisms of limited magnitude leave the class unchanged. Here, we have shown that such an invariance is learned in deep networks by performing spatial pooling and channel pooling. Modern architectures learn these pooling operations—as they are not imposed by the architecture—suggesting that it is best to let the pooling adapt to the specific task considered. Interestingly, spatial pooling comes together with an increased sensitivity to random noise in the image, as captured in simple artificial models of data.

It is commonly believed that the best architectures are those that extract the features of the data most relevant for the task. The pooling operations studied here, which allow the network to forget the exact locations of these features, are probably more effective when features are better extracted. This point may be responsible for the observed strong correlations between the network performance and its stability to diffeomorphisms. Designing synthetic models of data whose features are combinatorial and stable to smooth transformations is very much needed to clarify this relationship, and ultimately understand how deep networks learn high-dimensional tasks with limited data.

APPENDIX

## A  TASK 1, STRIDE 1: PROOFS

In Section 4.1 we consider a simple CNN with stride $s = 1$ and filter size $F$ trained on scale-detection task 1. We fix the total depth of these networks to be $\tilde{K}$. We postulated in Sec. 4 that this network displays a one-channel solution with homogeneous filter $[1/F, ..., 1/F]$ and no bias. We can understand the representation $f_k(x)$ at layer $k$ of an input datum $x$ by using single-pixel inputs $\delta_i$. Let us recall that these inputs have all components to 0 except the $i$-th, set to 1. Then, we have that a general datum $x$ is given by $x \propto (\delta_i + \delta_j)$, where $i$ and $j$ are the locations of the active pixel in $x$. We have argued in the main text that the representation $f_k(\delta_i)$ is a Gaussian distribution with width $\sqrt{k}$. In this Appendix we prove this statement.

First, we observe that in this solution, since both the elements of the filters and those of the inputs are non-negative, the networks behaves effectively as a linear operator. In particular, each layer corresponds to the application of a $L \times L$ circulant matrix $M$, which is obtained by stacking all the $L$ shifts of the following row vector,

$$[\underbrace{1, 1, ..., 1}_{F} \underbrace{0, 0, 0, ..., 0}_{L-F}]. \tag{10}$$

with periodic boundary conditions. The first row of such a matrix is fixed as follows. If $F$ is odd the patch of size $F$ is centered on the first entry of the first row, while if $F$ is even we choose to have $(F/2)$ ones at left of the first entry and $(F/2) - 1$ at its right. The output $f_k$ of the layer $k$ is then the following: $f_k(\delta_i) = M^k \delta_i$.

**Proposition A.1** *Let's consider the $L \times L$ matrix $M$ and a given $L$ vector $\delta_i$, as defined above. For odd $F \geq 3$, in the limit of large depth $\tilde{K} \gg 1$ and large width $\tilde{L} \gg F\sqrt{K}$, we have that*

$$(M^k)_{ab}\delta_i = \frac{1}{2\sqrt{\pi}\sqrt{D^{(1)}}\sqrt{k}}e^{-\frac{(a-i)^2}{4D^{(1)}k}}, \qquad D^{(1)} = \frac{1}{12F}(F-1)^3, \tag{11}$$

*while for even $F$:*

$$(M^k)_{ab}\delta_i = \frac{1}{2\sqrt{\pi}\sqrt{D^{(2)}}\sqrt{k}}e^{-\frac{(v^{(2)}k + a - i)^2}{4D^{(2)}k}}, \qquad D^{(2)} = \frac{1}{12F}\left(F^3 - 3F^2 + 6F - 4\right), \tag{12}$$

*with $v^{(2)} = (1 - F)/(2F)$.*

**Proof:**  The matrix $M$ can be seen as the stochastic matrix of a Markov process, where at each step the random walker has uniform probability $1/F$ to move in a patch of width $F$ around itself. We write the following recursion relation for odd $F$,

$$p_{a,i}^{(k+1)} = \frac{1}{F}\left(p_{a-(F-1)/2,i}^{(k)} + ... + p_{a,i}^{(k)} + ... + p_{a+(F-1)/2,i}^{(k)}\right), \tag{13}$$

and even $F$,

$$p_{a,i}^{(k+1)} = \frac{1}{F}\left(p_{a-F/2,i}^{(k)} + ... + p_{a,i}^{(k)} + ... + p_{a+(F/2-1),i}^{(k)}\right). \tag{14}$$

In any of these two cases, this is the so-called master equation of the random walk (Risken, 1996). In the limit of large image width $L$ and large depth $\tilde{K}$, we can write the related equation for the continuous process $p_i(a, k)$, which is called Fokker-Planck equation in physics and chemistry (Risken, 1996) or forward Kolmogorov equation in mathematics (Saloff-Coste & Bremaud, 2000),

$$\partial_k p_{a,i}^{(k)} = v\partial_a p_{a,i}^{(k)} + D\partial_a^2 p_{a,i}^{(k)}. \tag{15}$$

where the drift coefficient $v$ and the diffusion coefficient $D$ are defined in terms of the probability distribution $W_i(x)$ of having a jump $x$ starting from the location $i$

$$v = \int dx W_i(x)x, \qquad D = \int dx W_i(x)x^2. \tag{16}$$

In our case we have $W_i(x) = 1/F$ for $x \in [i - (F - 1)/2, i + (F - 1)/2]$ for odd $F$ and $x \in [i - F/2, i + F/2 - 1]$ for even $F$, yielding the solutions for the Fokker-Planck equations for even and odd $F$ reported in Eq. 11 and Eq. 12.

We can better characterize the limits of large image width $L$ and large network depth $\tilde{K}$ as follows. The proof relies on the fact that a random walk, after a large number of steps, converges to a diffusion process. Here the number of steps is given by the depth $\tilde{K}$ of the network. Consequently, we need $\tilde{K} \gg 1$. Moreover, we want that the diffusion process is not influenced by the boundaries of the image, of width $L$. The average path walked by the random walker after $\tilde{K}$ steps is given by $F\sqrt{\tilde{K}}$. Then, we require $F\sqrt{\tilde{K}} \ll L$.

$\square$

## B   Experimental setup

All experiments are performed in PyTorch. The code with the instructions on how to reproduce experiments are found here: github.com/leonardopetrini/relativestability/experiments_ICLR23.md.

### B.1   Deep networks training

In this section, we describe the experimental setup for the training of the deep networks deployed in Sections 1, 2 and 3.

For CIFAR10, fully connected networks are trained with the ADAM optimizer and learning rate $= 0.1$ while for CNNs SGD, learning rate $= 0.1$ and momentum $= 0.9$. In the latter case, the learning rate follows a cosine annealing scheduling. In all cases, the networks are trained on the cross-entropy loss, with a batch size of 128 and for 250 epochs. Early stopping at the best validation error is performed for selecting the networks to study. During training, we employ standard data augmentation consisting of random translations and horizontal flips of the input images. On the scale-detection task, we perform SGD on the hinge loss and halve the learning rate to $0.05$. All results are averaged when training on 5 or more different networks initializations.

For ImageNet, we used pretrained models from Pytorch, `torchvision.models`.

### B.2   Simple CNNs training

In this section we present the experimental setup for the training of simple CNNs introduced in Section 4, whose sensitivities to diffeomorphisms and Gaussian noise are shown in Fig. 7.

To learn task 1 we use CNNs with stride $s = 1$ and filter size $F = 3$. The width of the CNN is fixed to 1000 channels, while the depth to 12 layers. We use the Scale-Detection task in the version of Fig. 5 (b), with $\xi = 11$ and gap $g = 4$ and image size $L = 32$. For the training, we use $P = 48$ training points and Stochastic Gradient Descent (SGD) with learning rate $0.01$ and batch size 8. We use weight decay for the $L_2$ norm of the filters weights with ridge $0.01$. We stop the training after 500 times the interpolation time, which is the time required by the network to reach zero interpolation error of the training set. The goal of this procedure is to reach the solution with minimal norm. The generalization error of the trained CNNs is exactly zero: they learn spatial pooling perfectly. We show the sensitivities of the trained CNNs, averaged over 4 seeds, in the top panels of Fig. 7, where we also successfully test the predictions (Eq. 5, Eq. 7). We remark that to compute $G_k$ we inserted Gaussian noise with already the ReLU applied on, since we observe that without it we would see a pre-asymptotic behaviour for $G_k$ with respect to $A_k$.

Task 2 is learned using CNNs with stride equal to filter size $s = F = 2$. For the dataset, we use the block-wise version of the Scale-Detection task shown in Fig. 5 (c), fixing $\xi = 2^5$ and $L = 2^7$. We use 7 layers and 1000 channels for the CNNs. The training is performed using SGD and weight decay with the same parameters as in task 1, with $P = 2^{10}$ training points. In the bottom panels of Fig. 7 we show that the predictions (Eq. 8, Eq. 9) capture the experimental results, averaged over 10 seeds.

To support the assumption done in Section 4 that the trained CNNs are effectively behaving as one channel with homogeneous positive filters, we report the numerical values of the average filter over channels per layer in Table 1 for Task 1 and Table 2 for Task 2. They are positive in the first 9 hidden layers, where channel pooling is most pronounced.

|        | Init. | After training |
|--------|-------|----------------|
| $k=1$  | $[0.0132, 0.0023, -0.0068]$ | $[0.2928, 0.2605, 0.2928]$ |
| $k=2$  | $[0.0014, -0.0007, -0.0009]$ | $[0.0039, 0.0035, 0.0039]$ |
| $k=3$  | $[-0.0006, -0.0001, 0.0010]$ | $[0.0043, 0.0038, 0.0043]$ |
| $k=4$  | $[3.4610e-05, 6.5687e-04, -9.1634e-04]$ | $[0.0039, 0.0033, 0.0038]$ |
| $k=5$  | $[-0.0006, 0.0002, -0.0009]$ | $[0.0038, 0.0032, 0.0038]$ |
| $k=6$  | $[0.0012, -0.0011, -0.0003]$ | $[0.0038, 0.0031, 0.0038]$ |
| $k=7$  | $[-0.0006, 0.0004, 0.0003]$ | $[0.0041, 0.0032, 0.0040]$ |
| $k=8$  | $[0.0005, -0.0012, 0.0010]$ | $[0.0036, 0.0024, 0.0035]$ |
| $k=9$  | $[0.0005, -0.0012, 0.0010]$ | $[0.0021, 0.0016, 0.0017]$ |
| $k=10$ | $[-0.0025, 0.0015, -0.0006]$ | $[-0.0013, -0.0008, -0.0010]$ |
| $k=11$ | $[-0.0006, 0.0005, 0.0009]$ | $0.0002, 0.0002, 0.0002$ |
| $k=12$ | $[3.3418e-04, 3.3521e-05, 1.3936e-03]$ | $[0.0009, 0.0008, 0.0009]$ |

Table 1: Average over channels of filters in layer $k$, before and after training, for simple CNNs with $s = 1$ and $F = 3$ trained on task 1. The network learns filters which are much more homogeneous than initialization.

|        | Init. | After training |
|--------|-------|----------------|
| $k=1$  | $[-0.0559, -0.0291]$ | $[0.3828, 0.3737]$ |
| $k=2$  | $[-0.0022, 0.0010]$ | $[0.0060, 0.0059]$ |
| $k=3$  | $[0.0006, -0.0010]$ | $[0.0064, 0.0065]$ |
| $k=4$  | $[-0.0020, 0.0009]$ | $[0.0059, 0.0060]$ |
| $k=5$  | $[0.0002, 0.0008]$ | $[9.9935e-05, 2.1380e-04]$ |
| $k=6$  | $[-0.0003, -0.0010]$ | $[-0.0028, -0.0029]$ |
| $k=7$  | $[-7.4610e-04, 8.4595e-05]$ | $[-0.0009, -0.0009]$ |

Table 2: Average over channels of filters in layer $k$, before and after training, for simple CNNs with $s = F = 2$ trained on task 2. The network learns filters which are much more homogeneous than initialization.

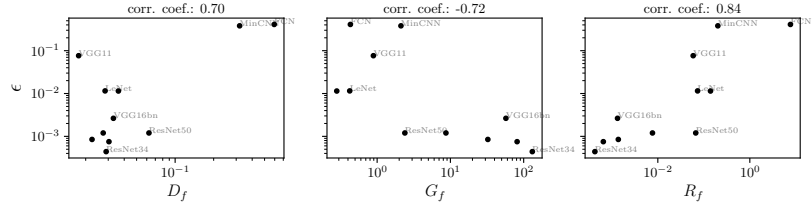

Figure 8: Generalization error $\epsilon$ versus sensitivity to diffeomorphisms $D_f$ (left), noise $G_f$ (center) and relative sensitivity $R_f$ (right) for a wide range of architectures trained on scale-detection task 1 (train set size: 1024, image size: 32, $\xi = 14$, $g = 2$). As in real data, $\epsilon$ is positively correlated with $D_f$ and negatively correlated with $G_f$. The correlation is the strongest for the relative measure $R_f$.

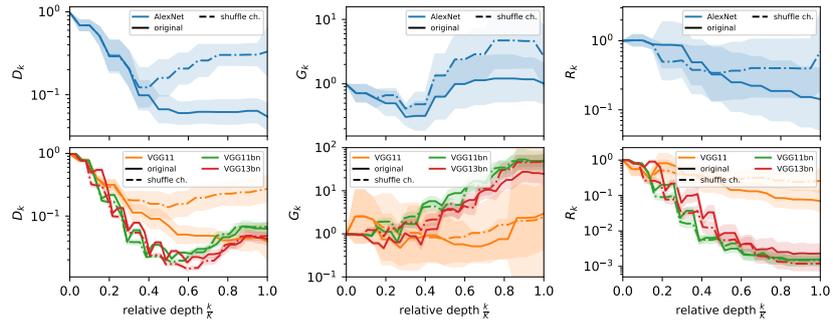

Figure 9: Sensitivities ($D_k$ left, $G_k$ middle and $R_k$ right) of the internal representations vs relative depth for AlexNet (1st row) and VGG networks (2nd row) trained on scale-detection task 1. Dot-dashed lines show the sensitivities of networks with shuffled channels.

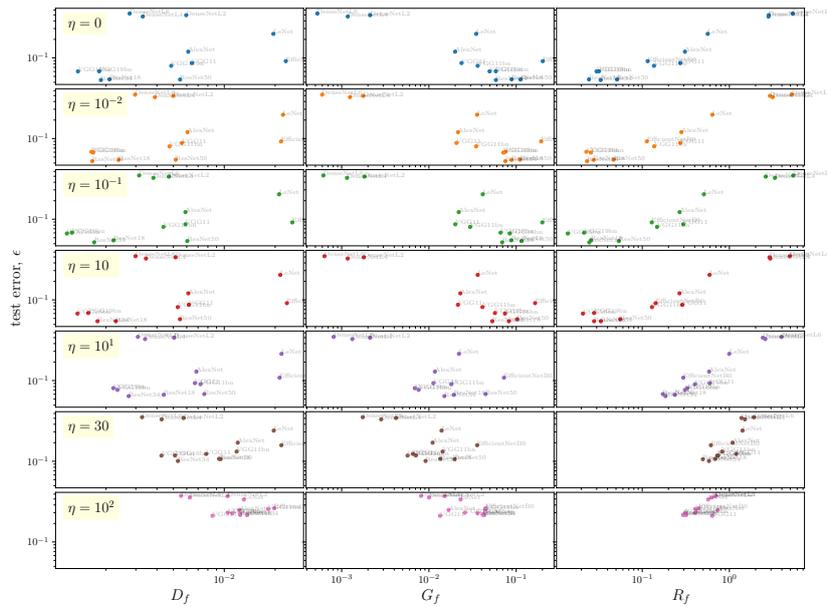

Figure 10: Test error vs. sensitivities (columns) when training on noisy CIFAR10. The different rows correspond to increasing noise magnitude $\eta$. Different points correspond to networks architectures, see gray labels. The content of this figure is also represented in compact form in Fig. 1, right.

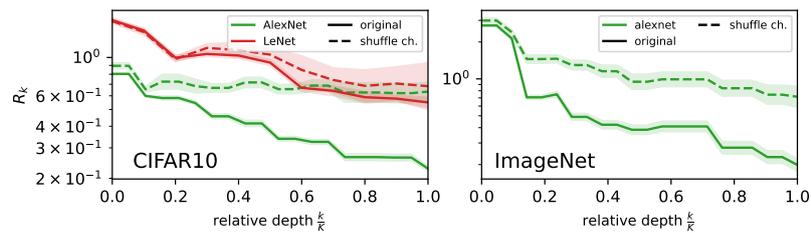

Figure 11: Analogous of Fig. 3 for different network architectures: relative sensitivity $R_k$ as a function of depth for LeNet and AlexNet architectures trained on CIFAR10 (left) and ImageNet (right). Full lines indicate experiments done on the original networks, dashed lines the ones after shuffling channels. $K$ indicates the networks total depth.

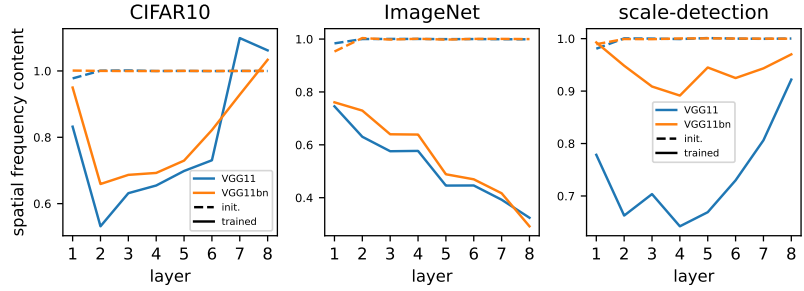

Figure 12: Spatial frequency content of filters for CIFAR10 (left), ImageNet (center) and the scale-detection task (right). The $y$-axis reports an aggregate measure among spatial frequencies: $N(\sum_{i=1}^{N}\lambda_l)^{-1}\langle\|\boldsymbol{w}_c^k\|^2\rangle_c^{-1}\sum_{l=1}^{F^2}\lambda_l\langle(\boldsymbol{\Psi}_l\cdot\boldsymbol{w}_c^k)^2\rangle_c$, where $\boldsymbol{\Psi}_l$ are the $3\times3$ Laplacian eigenvectors and $\lambda_l$ the corresponding eigenvalues, $\boldsymbol{w}_c^k$ the $c$-th filter of layer $k$ and $\langle\cdot\rangle_c$ denotes the average over $c$. This is an aggregate measure over frequencies, the frequencies distribution is reported in the main text, Fig. 4.

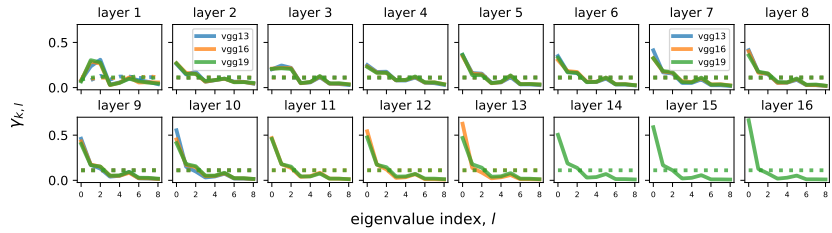

Figure 13: Analogous of Fig. 4 for deep VGGs trained on ImageNet. Dotted and full lines respectively correspond to initialization and trained networks. The $x$-axis reports low to high frequencies from left to right. Deeper layers are reported in rightmost panels.

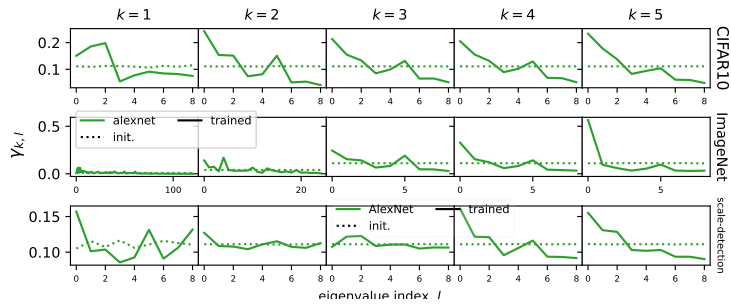

Figure 14: Analogous of Fig. 4 for AlexNet trained on CIFAR10 (1st row), ImageNet (2nd row) and the scale detection task (3rd row). Dotted and full lines respectively correspond to initialization and trained networks. The $x$-axis reports low to high frequencies from left to right. Deeper layers are reported in rightmost panels.

# Synonymic Invariance Part III

# 6 A Toy Model for the Hierarchical Compositionality of Real Data

The following paper is the preprint version of Petrini et al. (2023) that is currently under review.

**Candidate contributions**     The candidate contributed to all discussions, implemented the Random Hierarchy Model, designed and performed the experiments.

# How Deep Neural Networks Learn Compositional Data: The Random Hierarchy Model

Leonardo Petrini[a*], Francesco Cagnetta[a†*], Umberto M. Tomasini[a], Alessandro Favero[a,b], and Matthieu Wyart[a]

[a]Institute of Physics, EPFL, Lausanne, Switzerland
[b]Institute of Electrical Engineering, EPFL, Lausanne, Switzerland

June 28, 2023

## Abstract

Learning generic high-dimensional tasks is notably hard, as it requires a number of training data exponential in the dimension. Yet, deep convolutional neural networks (CNNs) have shown remarkable success in overcoming this challenge. A popular hypothesis is that learnable tasks are highly structured and that CNNs leverage this structure to build a low-dimensional representation of the data. However, little is known about how much training data they require, and how this number depends on the data structure. This paper answers this question for a simple classification task that seeks to capture relevant aspects of real data: the Random Hierarchy Model. In this model, each of the $n_c$ classes corresponds to $m$ synonymic compositions of high-level features, which are in turn composed of sub-features through an iterative process repeated $L$ times. We find that the number of training data $P^*$ required by deep CNNs to learn this task *(i)* grows asymptotically as $n_c m^L$, which is only polynomial in the input dimensionality; *(ii)* coincides with the training set size such that the representation of a trained network becomes invariant to exchanges of synonyms; *(iii)* corresponds to the number of data at which the correlations between low-level features and classes become detectable. Overall, our results indicate how deep CNNs can overcome the curse of dimensionality by building invariant representations, and provide an estimate of the number of data required to learn a task based on its hierarchically compositional structure.

The achievements of deep learning algorithms [1] are outstanding. These methods exhibit superhuman performances in areas ranging from image recognition [2] to Go-playing [3], and large language models such as GPT4 [4] can generate unexpectedly sophisticated levels of reasoning. However, despite these accomplishments, we still lack a fundamental understanding of the underlying factors. Indeed, Go configurations, images, and patches of text lie in high-dimensional spaces, which are hard to sample due to the *curse of dimensionality* [5]: the distance $\delta$ between neighboring data points decreases very slowly with their number $P$, as $\delta = \mathcal{O}(P^{-1/d})$ where $d$ is the space dimension. A generic task such as regression of a continuous function [6] requires a small $\delta$ for high performance, implying that $P$ must be *exponential* in the dimension $d$. Such a number of data is unrealistically large: for example, the benchmark dataset ImageNet [7], whose effective dimension is estimated to be $\approx 50$ [8], consists of only $\approx 10^7$ data, significantly smaller than $e^{50} \approx 10^{20}$. This immense difference implies that learnable tasks are not generic, but highly structured. What is then the nature of this structure, and why are deep learning methods able to exploit it? Without a quantitative answer, it is impossible to predict even the order of magnitude *of the order of magnitude* of the number of data necessary to learn a specific task.

A popular idea attributes the efficacy of deep learning methods to their ability to build a useful representation of the data, which becomes increasingly complex across the layers. In simple terms, neurons closer to the input learn to detect simple features like edges in a picture, whereas those deeper in the network learn to recognize more abstract features, such as faces [9, 10]. Intuitively, if these representations are also invariant to aspects of the data unrelated to the task, such as the exact position of an object in a frame for image classification [11], they may effectively reduce the dimensionality of the problem and make it tractable. This view is supported by several empirical studies of the hidden representations of trained networks. In particular, measures such as *(i)* the mutual information between such representations and the input [12, 13], *(ii)* their intrinsic dimensionality [14, 15], and *(iii)* their sensitivity toward transformations that do not affect the task (e.g. smooth deformations for image classification [16,17]), all eventually decay with the layer depth. In some cases, the magnitude of this decay correlates with performance [16]. However, these studies do not indicate how much data is required to learn such representations, and thus the task.

Here we study this question for tasks which are hierarchically compositional—arguably a key property for the learnability of real data [18–25]. To provide a concrete example, consider the picture of a dog (see Fig. 1). The image consists of several high-level features like head, body, and limbs, each composed of sub-features like ears, mouth, eyes, and nose for the head. These sub-features can be further thought of as combinations of low-level features such as edges. Recent studies have revealed that: *(i)* deep networks represent hierarchically compositional tasks more efficiently than shallow networks [21]; *(ii)* the minimal number of data that contains enough information to reconstruct such tasks is polynomial in the input dimension [24], although extracting this information remains impractical with standard optimization algorithms; *(iii)* correlations between the input data and the task are critical for learning [19,26] and can be exploited by algorithms based on the iteration of clustering methods [22,27]. While these seminal works offer important insights, they do not directly address practical settings, specifically deep convolutional neural networks (CNNs) trained using gradient descent. Consequently, we currently don't know how the hierarchically compositional structure of the task influences the *sample complexity*, i.e., the number of data necessary to learn the task.

In this work, we adopt the physicist's approach [28–31] of introducing a simplified model of data, which we then investigate quantitatively via a combination of theoretical arguments and numerical experiments. The task we consider, introduced in Section 1, is a multinomial classification where the class label is determined by the hierarchical composition of input features into progressively higher-level features (see Fig. 1). This model belongs to the class of generative models introduced in [22,27], corresponding to the specific choice of random composition rules. More specifically, we consider a classification problem with $n_c$ classes, where the class label is expressed as a hierarchy of $L$ *randomly-chosen* composition rules. In each rule, $m$ distinct tuples of $s$ adjacent low-level features are grouped together and assigned the same high-level feature taken from a finite vocabulary of size $v$ (see Fig. 1). Then, in Section 3, we show empirically that the sample complexity $P^*$ of deep CNNs trained with gradient descent scales as $n_c m^L$. Furthermore, we find that $P^*$ coincides with both *a)* the number of data that allows for learning a representation that is invariant to exchanging the $m$ semantically equivalent low-level features (subsection 3.1) and *b)* the size of the training set for which the correlations between low-level features and class label become detectable (Section 4). Via *b)*, $P^*$ can be derived under our assumption on the randomness of the composition rules.

# 1  The Random Hierarchy Model

In this section, we introduce our model task, which is a multinomial classification problem with $n_c$ classes, where the input-output relation is *compositional*, *hierarchical*, and *local*. To build the dataset, we let each class label $\alpha = 1, \ldots, n_c$ generate the set of input data with label $\alpha$ as follows.

i) Each label generates $m$ distinct representations consisting of $s$-tuples of *high-level features* (see Fig. 2 for an example with $s = 2$ and $m = n_c = 3$). Each of these features belongs to a finite vocabulary of size $v$ ($v = 3$ in the figure), so that there are $v^s$ possible representations and $n_c m \leq v^s$. We call the assignment of $m$ distinct $s$-tuples to each label a *composition rule*;[1]

ii) Each of the $v$ high-level feature (level-$L$) generates $m$ distinct representations of $s$ sub-features (level-$(L-1)$), out of the $v^s$ possible ones. Thus, $m \leq v^{s-1}$. After two generations, labels are represented as $s^2$-tuples and there are $m \times m^s$ data per class;

iii) The input data are obtained after $L$ generations (level-1 representation) so that each datum $\boldsymbol{x}$ consists of $d = s^L$ input features $x_j$. We apply one-hot encoding to the input features: each of the $x_j$'s is a $v$-dimensional sequence with one element set to 1 and the others to 0, the index of the non-zero component representing the encoded feature. The number of data per class is

$$m \times m^s \times \cdots \times m^{s^{L-1}} = m^{\sum_{i=0}^{L-1} s^i} = m^{\frac{s^L-1}{s-1}}, \quad (1)$$

hence the total number of data $P_{\max}$ reads

$$P_{\max} \equiv n_c m^{\frac{s^L-1}{s-1}} = n_c m^{\frac{d-1}{s-1}}. \quad (2)$$

A generic classification task is thus specified by $L$ composition rules and can be represented as a $s$-ary tree—an example with $s = 2$ and $L = 3$ is shown in Fig. 1(c) as a binary tree. The tree representation highlights that the class label $\alpha(\boldsymbol{x})$ of a datum $\boldsymbol{x}$ can be written as a hierarchical composition of $L$ local functions of $s$ variables [20,21]. For instance, with $s = L = 2$ ($\boldsymbol{x} = (x_1, x_2, x_3, x_4)$),

$$\alpha(x_1, \ldots, x_4) = g_2\left(g_1(x_1, x_2), g_1(x_3, x_4)\right), \quad (3)$$

where $g_1$ and $g_2$ represent the 2 composition rules.

---

[1]Composition rules are called *production rules* in formal language theory [32].

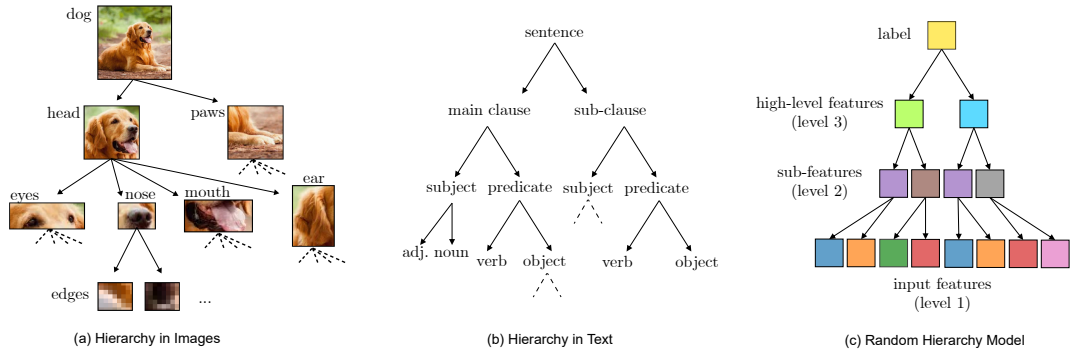

Figure 1: Illustrating the hierarchical structure of real-world and artificial data. (a) An example of the hierarchical structure of images: the class (dog) consists of high-level features (head, paws), that in turn can be represented as sets of lower-level features (eyes, nose, mouth, and ear for the head). Notice that, at each level, there can be multiple combinations of low-level features giving rise to the same high-level feature. (b) A similar hierarchical structure can be found in natural language: a sentence is made of clauses, each having different parts such as subject and predicate, which in turn may consist of several words. (c) An illustration of the artificial data structure we propose. The samples reported here were drawn from an instance of the Random Hierarchy Model for depth $L = 3$ and tuple length $s = 2$. Different features are shown in different colors.

In the *Random Hierarchy Model* (RHM) the $L$ composition rules are chosen uniformly at random over all the possible assignments of $m$ low-level representations to each high-level feature. As sketched in Fig. 2, the random choice induces correlations between low- and high-level features. In simple terms, each of the high-level features—1, 2 or 3 in the figure—is more likely to be represented with a certain low-level feature in a given position—blue on the left for 1, yellow for 2 and green for 3. These correlations are crucial for our predictions and are analyzed in detail in Appendix B.

Let us remark that the $L$ composition rules can be chosen such that the low-level features are homogeneously distributed across high-level features for all positions, as sketched in Fig. 3. We refer to this choice as the *Homogeneous Features Model*. In this model, none of the low-level features is predictive of the high-level feature. With $s = 2$ and Boolean features $v = m = 2$, the Homogeneous Features Model reduces to the problem of learning a parity function [33].

Finally, note that we only consider the case where the parameters $s$, $m$ and $v$ are constant through the hierarchy levels for clarity of exposition. It is straightforward to extend the model, together with the ensuing conclusions, to the case where all the levels of the hierarchy have different parameters.

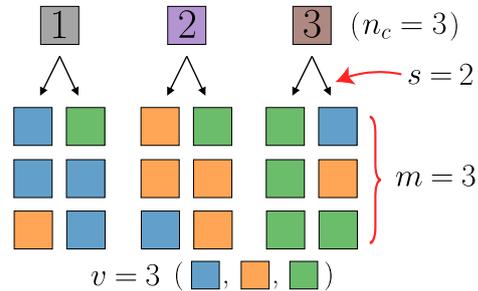

Figure 2: Label to lower-level features mapping in the Random Hierarchy Model (RHM). Each of the $n_c = 3$ classes (numbered boxes at the top) corresponds to $m = 3$ distinct couples (unnumbered boxes at the bottom) of features. These features belong to a finite vocabulary (blue, orange and green, with size $v = 3$). Iterating this mapping $L$ times with the lower-level features as high-level features of the next step yields the full dataset. Notice that some features appear more often in the representation of a certain class than in those of the others, e.g. blue on the left appears twice in class 1, once in class 2 and never in class 3. As a result, low-level features are generally correlated with the label.

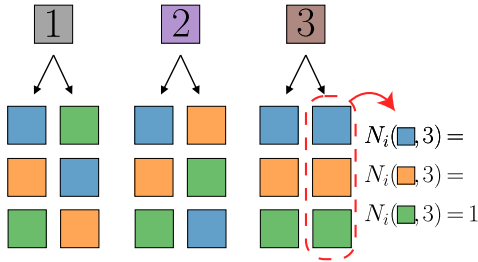

Figure 3: Label to lower-level features mapping in the Homogeneous Feature Model with $v = m = n_c = 3$ and $s = 2$. In contrast with the case illustrated in Fig. 2, this mapping is such that each of the 3 possible low-level features appears exactly once in each of the 2 elements of the representation of each class. In maths, denoting with $N_i(\mu; \alpha)$ the number of times that the low-level feature $\mu$ appears in the $i$-th position of the representation of class $\alpha$, one has that $N_i(\mu; \alpha) = 1$ for all $i = 1, 2$, for all $\mu =$ green, blue, orange and for all $\alpha = 1, 2, 3$.

## 2 Characteristic Sample Sizes

The main focus of our work is the answer to the following question:

**Q:** *How much data is required to learn a typical instance of the Random Hierarchy Model?*

In this section, we first discuss two characteristic scales of the number of training data for an RHM with $n_c$ classes, vocabulary size $v$, multiplicity $m$, depth $L$, and tuple size $s$. The first, related to the curse of dimensionality, represents the sample complexity of methods that are not able to learn the hierarchical structure of the data. The second, which comes from information-theoretic considerations, represents the minimal number of data necessary to reconstruct an instance of the RHM. These two sample sizes can be thought of as an upper and lower bound to the sample complexity of deep CNNs, which indeed lies between the two bounds (cf. Section 3).

### 2.1 Curse of Dimensionality ($P_{\max}$)

Let us recall that the curse of dimensionality predicts an exponential growth of the sample complexity with the input dimension $d = s^L$. Fig. 4 shows the test error of a one-hidden-layer fully-connected network trained on instances of the RHM while varying the number of training data $P$ (see methods for details of the training procedure) in the maximal $m$ case, $m = v^{s-1}$. The bottom panel demonstrates that the sample complexity is proportional to the total dataset size $P_{\max}$. Since, from Eq. 2, $P_{\max}$ grows exponentially with $d$, we conclude that shallow fully-connected networks suffer from the curse of dimensionality. By contrast, we will see that using CNNs results in a much gentler growth (i.e. polynomial) of the sample complexity with $d$.

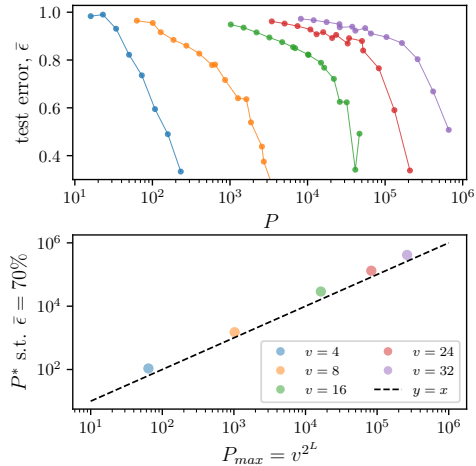

Figure 4: **Sample complexity for one-hidden-layer fully-connected networks,** $v = n_c = m$ **and** $s = 2$. Top: Test error vs. the number of training data. Different colors correspond to different vocabulary sizes $v$. Bottom: number of data corresponding to test error $\epsilon = 0.7$ as a function of $P_{\max}$. The black dashed line indicates a linear relationship: one-hidden-layer fully-connected networks achieve a small test error only when trained on a finite fraction of the whole dataset, thus their sample complexity grows exponentially with the input dimension.

### 2.2 Information-Theoretic Limit ($P_{\min}$)

An algorithm with full prior knowledge of the generative model can reconstruct an instance of the RHM with a number of points $P_{\min} \ll P_{\max}$. For instance, we can consider an extensive search within the hypothesis class of all possible hierarchical models with fixed $n_c$, $m$, $v$, and $L$. Then, if $n_c = v$ and $m = v^{s-1}$, so that the model generates all possible input data, we can use a classical result of the PAC (Probably Approximately Correct) framework of statistical learning theory [34] to relate $P_{\min}$ with the logarithm of the cardinality of the hypothesis class, that is the number of possible instances of the hierarchical model. The number of possible composition rules equals the number of ways of allocating $v^{s-1}$ of $v^s$ possible tuples to each of the $v$ classes/features, i.e., a multinomial coefficient,

$$\#\,\{\text{rules}\} = \frac{(v^s)!}{((v^{s-1})!)^v} \qquad (4)$$

Since an instance consists of $L$ independently chosen composition rules, we have

$$\#\,\{\text{instances}\} = (\#\,\{\text{rules}\})^L \left(\frac{1}{v!}\right)^{L-1} \qquad (5)$$

where the additional multiplicative factor $(v!)^{1-L}$ takes into account that the input-label mapping is invariant for relabeling of the features of the $L-1$ internal representations. Upon taking the logarithm and approximating the factorials for large $v$ via Stirling's formula,

$$P_{\min} = \log\left(\#\,\{\text{instances}\}\right) \xrightarrow{v \gg 1} Lv^s, \qquad (6)$$

Intuitively, the problem boils down to understanding the $L$ composition rules, each needing $m \times v$ examples ($v^s$ for $m = v^{s-1}$). $P_{\min}$ grows only linearly with the depth $L$—hence logarithmically in $d$—whereas $P_{\max}$ is exponential in $d$. Having used full knowledge of the generative model, $P_{\min}$ can be thought of as a lower bound for the sample complexity of a generic supervised learning algorithm which is agnostic of the data structure.

## 3 Sample Complexity of Deep CNNs

In this section, we focus on deep learning methods. In particular, we ask

**Q:** *How much data is required to learn a typical instance of the Random Hierarchy Model with a deep CNN?*

Thus, after generating an instance of the RHM with fixed parameters $n_c$, $s$, $m$, $v$, and $L$, we train a deep CNN with $L$ hidden layers, filter size and stride equal to $s$ (see Fig. 5 for an illustration) with stochastic gradient descent (SGD) on $P$ training points selected at random among the RHM data. Further details of the training are in *Materials and Methods*.

By looking at the test error of trained networks as a function of the training set size (top panels of Fig. 6 and Fig. 7, see also Fig. 15 in Appendix G for a study with varying $n_c$), we notice the existence of a characteristic value of $P$ where the error decreases dramatically, thus the task is learned. In order to study the behavior of this threshold with the parameters of the RHM, we define the sample complexity as the smallest $P$ such that the test error $\epsilon(P)$ is smaller than $\epsilon_{\text{rand}}/10$, with $\epsilon_{\text{rand}} = 1 - n_c^{-1}$ denoting the average error when choosing the label uniformly at random. The bottom panels of Fig. 6 (for the case $n_c = m = v$) and Fig. 7

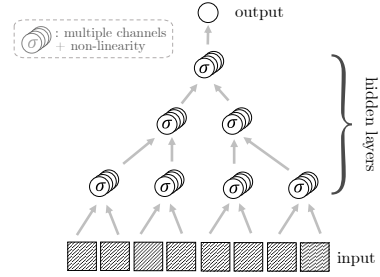

Figure 5: Neural network architecture that matches the RHM hierarchy. This is a deep CNN with $L$ hidden layers, and stride and filter size equal to the tuple length $s$. Filters that act on different input patches are the same (weight sharing). The number of input channels equals $v$ and the output is a vector of size $n_c$.

(with $m < v$, see Appendix G for varying $n_c$) show that the sample complexity scales as

$$P^* = n_c m^L \Leftrightarrow \frac{P^*}{n_c} = d^{\ln(m)/\ln(s)}, \qquad (7)$$

independently of the vocabulary size $v$. Eq. 7 shows that deep CNNs only require a number of samples that scales as a power of the input dimension $d = s^L$ to learn the RHM: the curse of dimensionality is beaten. This evidences the ability of CNNs to harness the hierarchical compositionality inherent to the task. The question then becomes: what mechanisms do these networks employ to achieve this feat?

### 3.1 Emergence of Synonymic Invariance in Deep CNNs

A natural approach to learning the RHM would be to identifying the sets of $s$-tuples of input features that correspond to the same higher-level feature. Examples include the pairs of low-level features in Fig. 2 and Fig. 3 which belong to the same column. In general, we refer to $s$-tuples that share the same higher-level representation as *synonyms*. Identifying synonyms at the first level would allow us to replace each $s$-dimensional patch of the input with a single symbol, reducing the dimensionality of the problem from $s^L$ to $s^{L-1}$. Repeating this procedure $L$ times would lead to the class labels and, consequently, to the solution of the task.

In order to test if deep CNNs trained on the RHM resort to a similar solution, we introduce the *synonymic sensitivity*, which is a measure of the invariance of any given function of the input with respect to the exchange of synonymic $s$-tuples. We define $S_{k,l}$ as the sensitivity of the $k$-th layer representation of a trained network with respect to exchanges of synonymous tuples of level-$l$ features.

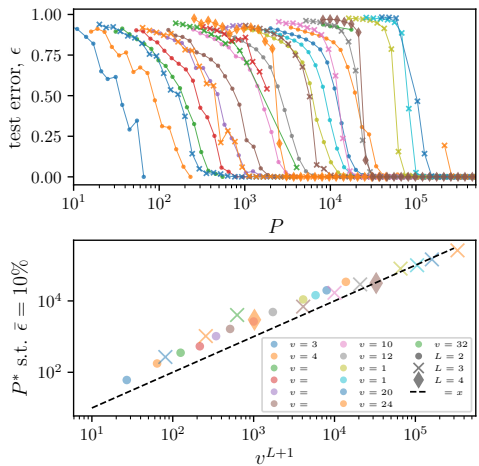

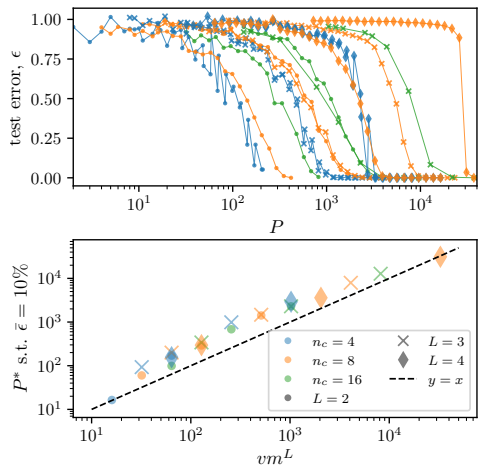

Figure 6: **Sample complexity for deep CNNs, $m = n_c = v$ and $s = 2$.** Top: Test error vs number of training points. Different colors correspond to different vocabulary sizes $v$. Markers to hierarchy depths $L$. Deep CNNs are able to achieve zero generalization error when enough training points are provided. Bottom: sample complexity $P^*$ corresponding to a test error $\epsilon^* = 0.1$. Remarkably, the neural networks can generalize with a number of samples $P^* = v^{L+1} \ll P_{\max}$.

Figure 7: **Sample complexity for deep CNNs, $m < v$, $n_c = v$ and $s = 2$.** Top: Test error vs number of training points. Different colors correspond to different vocabulary sizes $v$. Markers to hierarchy depths $L$. Bottom: sample complexity $P^*$ corresponding to a test error $\epsilon^* = 0.1$. Similarly to the previous plot, this confirms that the sample complexity of deep CNNs scales as $P^* = n_c m^L$.

Namely,

$$S_{k,l} = \frac{\langle \|f_k(x) - f_k(P_l x)\|^2 \rangle_{x, P_l}}{\langle \|f_k(x) - f_k(z)\|^2 \rangle_{x,z}}, \quad (8)$$

where: $f_k$ is the vector of the activations of the $k$-th layer in the network; $P_l$ is an operator that replaces all the level-$l$ tuples with synonyms selected uniformly at random; $\langle \cdot \rangle$ with subscripts $x, z$ denote an average over all the inputs in an instance of the RHM; the subscript $P_l$ denotes average over all the exchanges of synonyms.

In particular, $S_{k,1}$ quantifies the invariance of the hidden representations learned by the network at layer $k$ with respect to exchanges of synonymic tuples of input features. Fig. 8 reports $S_{2,1}$ as a function of the training set size $P$ for different combinations of the model parameters. We focused on $S_{2,1}$—the sensitivity of the second layer of the deep CNN to permutations at the first level of the hierarchy—since synonymic invariance can generally be achieved at all layers $k$ starting from $k = l + 1$, and not before [2] Notice that all curves display a sigmoidal shape, signaling the existence of a characteristic sample size which marks the emergence of synonymic sensitivity in the learned representations. Remarkably, by rescaling the $x$-axis by the sample complexity of Eq. 7 (bottom panel), curves corresponding to different parameters collapse. We conclude that the generalization ability of a network relies on the synonymic invariance of its hidden representations.

Measures of the synonymic sensitivity $S_{k,1}$ for different layers $k$ are reported in Fig. 9 (blue lines), showing indeed that small values of $S_{k,1}$ are achieved for $k \geq 2$. Fig. 9 also shows the sensitivities to exchanges of synonyms in the higher-level representations of the RHM: all levels are learned together as $P$ increases, and invariance to level-$l$ exchanges is achieved at layer $k = l + 1$, as expected. The figure displays the test error too (gray dashed), to further emphasize its correlation with synonymic invariance.

---

[2]To illustrate this, consider a hierarchy of depth $L = 2$, $s = 2$, and a two-hidden-layers CNN. In the general case, synonymic invariance to permutations at level one, cannot be achieved at the first layer of the network. This is because, say a level-1 feature can be represented at the input as $(\alpha, \beta)$, $(\alpha, \alpha)$ and $(\beta, \alpha)$, but not as $(\beta, \beta)$. Then, it is impossible to build a neuron that would have the same response to the first three pairs but not the fourth. Instead, a simple solution exists for layer 2 to become invariant to exchanges at level 1. This consists in building $v^2$ neurons at the first layer $k = 1$, each responding to one input pair. Clearly, the representation at $k = 1$ is not invariant to the substitution of synonyms. The second layer, though, can assign identical weights to all the $v$ neurons that encode for the same feature, hence becoming invariant to permutations at $l = 1$.

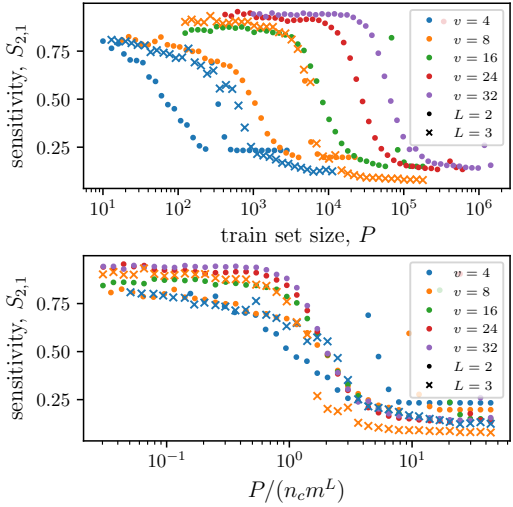

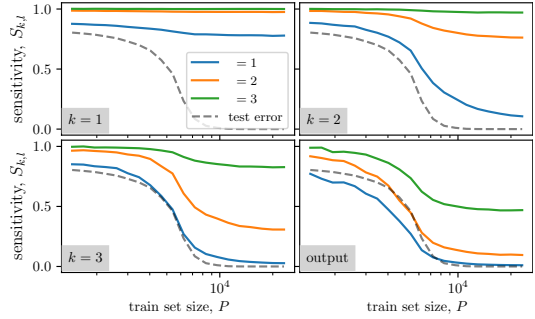

Figure 8: Sensitivity $S_{2,1}$ of the second layer of a deep CNN to permutations in the first level of the RHM with $L = 2, 3$, $s = 2$, $n_c = m = v$, as a function of the training set size (top) and after rescaling by $P^* = n_c m^L$ (bottom). Sensitivity decreases from 1 to approximately zero, i.e. deep CNNs are able to learn synonymic invariance with enough training points. The collapse after rescaling highlights that this can be done with $P^*$ training points.

## 4 Correlations Govern Synonymic Invariance

We now provide a theoretical argument for understanding the scaling of $P^*$ of Eq. 7 with the parameters of the RHM. First, we compute a third characteristic sample size $P_c$, defined as the size of the training set for which the *local* correlation between any of the input patches and the label becomes detectable. Remarkably, $P_c$ coincides with $P^*$ of Eq. 7. Secondly, we demonstrate how a one-hidden-layer neural network acting on a single patch can use such correlations to build a synonymic invariant representation in a single step of gradient descent, so that $P_c$ and $P^*$ also correspond to the emergence of an invariant representation.

### 4.1 Identify Synonyms by Counting

The invariance of the RHM labels with respect to exchanges of synonymous input patches can be inferred by counting the occurrences of such patches in all the data belonging to a given class $\alpha$. Intuitively, tuples of features that appear with identical frequencies are likely synonyms.

Figure 9: Permutation sensitivity $S_{k,l}$ of the layers of a deep CNN trained on the RHM with $L = 3$, $s = 2$, $n_c = m = v = 8$, as a function of the training set size $P$. The permutation of synonyms is performed at different levels, as indicated in colors. The different panels correspond to the sensitivity of different layers' activations, indicated by the gray box. Synonymic invariance is learned at the same time for all layers, and most of the invariance to level $l$ is obtained at layer $k = l + 1$.

More specifically, let us denote an $s$-dimensional input patch with $\boldsymbol{x}_j$ for $j$ in $1, \ldots, s^{L-1}$, a $s$-tuple of input features with $\boldsymbol{\mu} = (\mu_1, \ldots, \mu_s)$, and the number of data in class $\alpha$ which display $\boldsymbol{\mu}$ in the $j$-th patch with $N_j(\boldsymbol{\mu}; \alpha)$. Normalizing this number by $N_j(\boldsymbol{\mu}) = \sum_\alpha N_j(\boldsymbol{\mu}; \alpha)$ yields the conditional probability $f_j(\alpha|\boldsymbol{\mu})$ for a datum to belong to class $\alpha$ conditioned on displaying the $s$-tuple $\boldsymbol{\mu}$ in the $j$-th input patch,

$$f_j(\alpha|\boldsymbol{\mu}) := \Pr\{\boldsymbol{x} \in \alpha|\boldsymbol{x}_j = \boldsymbol{\mu}\} = \frac{N_j(\boldsymbol{\mu}; \alpha)}{N_j(\boldsymbol{\mu})}. \text{[3]} \quad (9)$$

If the low-level features are homogeneously spread across classes, as in the Homogeneous Features Model of Fig. 3, then $f = n_c^{-1}$, independently of and $\alpha$, $\boldsymbol{\mu}$ and $j$. In contrast, due to the aforementioned correlations, the probabilities of the RHM are all different from $n_c^{-1}$ (see Fig. 2). We refer to this difference as *signal*.[4] Distinct level-1 tuples $\boldsymbol{\mu}$ and $\boldsymbol{\nu}$ yield a different $f$ (and thus a different signal) with high probability unless they share the same level-2 representation. Therefore, this signal can be used to identify synonymous level-1 tuples.

### 4.2 Signal vs. Sampling Noise

When measuring the conditional class probabilities with only $P$ training data, the occurrences in the right-hand side of Eq. 24 are replaced with empirical occurrences, which induce a sampling *noise* on the $f$'s. For the identification of synonyms to be possible, this noise must be smaller in magnitude than the aforementioned signal—a visual representation of the comparison between signal and noise is depicted in Fig. 10.

---

[3]The notation $\boldsymbol{x}_j = \boldsymbol{\mu}$ means that the elements of the patch $\boldsymbol{x}_j$ encode the tuple of features $\boldsymbol{\mu}$

[4]Cases in which all features are homogeneously spread across classes can also appear in the RHM, but with vanishing probability in the limit of large $n_c$ and $m$, see Appendix E.

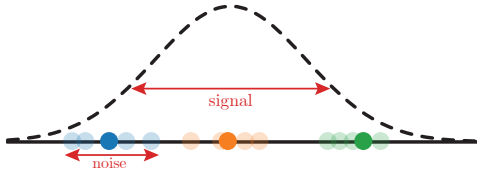

Figure 10: **Signal vs. noise illustration.** The dashed function represents the distribution of $f(\alpha|\boldsymbol{\mu})$ resulting from the random sampling of the RHM rules. The solid dots illustrate the *true* frequencies $f(\alpha|\boldsymbol{\mu})$ sampled from this distribution, with different colors corresponding to different groups of synonyms. The typical spacing between the solid dots, given by the width of the distribution, represents the *signal*. Transparent dots represent the empirical frequencies $\hat{f}_j(\alpha|\boldsymbol{\mu})$, with dots of the same color corresponding to synonymous features. The spread of transparent dots of the same color, which is due to the finiteness of the training set, represents the *noise*.

The magnitude of the signal can be computed as the ratio between the standard deviation and mean of $f_j(\alpha|\boldsymbol{\mu})$ over realizations of the RHM. The full calculation is presented in Appendix B: here we present a simplified argument based on an additional independence assumption. Given a class $\alpha$, the tuple $\boldsymbol{\mu}$ appearing in the $j$-th input patch is determined by a sequence of $L$ choices—one choice per level of the hierarchy—of one among $m$ possible lower-level representations. These $m^L$ possibilities lead to all the $mv$ distinct $s$-tuples. $N_j(\boldsymbol{\mu};\alpha)$ is proportional to how often the tuple $\boldsymbol{\mu}$ is chosen—$m^L/(mv)$ times on average. Under the assumption of independence of the $m^L$ choices, the fluctuations of $N_j(\boldsymbol{\mu};\alpha)$ relative to its mean are given by the central limit theorem and read $(m^L/(mv))^{-1/2}$ in the limit of large $m$. If $n_c$ is sufficiently large, the fluctuations of $N_j(\boldsymbol{\mu})$ are negligible in comparison. Therefore, the relative fluctuations of $f_j$ are the same as those of $N_j(\boldsymbol{\mu};\alpha)$: the size of the signal is $(m^L/(mv))^{-1/2}$.

The magnitude of the noise is given by the ratio between the standard deviation and mean, over independent samplings of a training set of fixed size $P$, of the empirical conditional probabilities $\hat{f}_j(\alpha|\boldsymbol{\mu})$. Only $P/(n_c mv)$ of the training points will, on average, belong to class $\alpha$ while displaying feature $\mu$ in the $j$-th patch. Therefore, by the convergence of the empirical measure to the true probability, the sampling fluctuations of $\hat{f}$ relative to the mean are of order $[P/(n_c mv)]^{-1/2}$—see Appendix B for details.

Balancing signal and noise yields the characteristic $P_c$ for the emergence of correlations. For large $m$, $n_c$ and $P$,

$$P_c = n_c m^L, \tag{10}$$

which coincides with the empirical sample complexity of deep CNNs discussed in Section 3.

### 4.3 Learning Synonymic Invariance by the Gradients

To complete the argument, we consider a simplified one-step gradient descent setting [35, 36], where $P_c$ marks the number of training examples required to learn a synonymic invariant representation. In this setting (details presented in Appendix C), we train a one-hidden layer fully-connected network on the first $s$-dimensional patches of the data. This network cannot fit data which have the same features on the first patch while belonging to different classes. Nevertheless, the hidden representation of the network can become invariant to exchanges of synonymous patches.

More specifically, as we show in Appendix C, with identical initialization of the hidden weights and orthogonalized inputs, the update of the hidden representation $f_h(\boldsymbol{\mu})$ of the $s$-tuple of low-level features $\boldsymbol{\mu}$ after one step of gradient descent follows

$$\Delta f_h(\boldsymbol{\mu}) = \frac{1}{P} \sum_{\alpha=1}^{n_c} a_{h,\alpha} \left( \hat{N}_1(\boldsymbol{\mu};\alpha) - \frac{1}{n_c} \sum_{\beta=1}^{n_c} \hat{N}_1(\boldsymbol{\mu};\beta) \right), \tag{11}$$

where $f_h(\boldsymbol{\mu})$ coincides the pre-activation of the $h$-th neuron and $\boldsymbol{a}_h = (a_{h,1}, \ldots, a_{h,n_c})$ denotes the associated $n_c$ dimensional readout weight. $\hat{N}_1$ is used to denote the empirical estimate of the occurrences in the first input patch. Hence, by the result of the previous section, the hidden representation becomes insensitive to the exchange of synonymic features for $P \gg P_c$.

This prediction is confirmed empirically in Fig. 11, which shows the sensitivity $S_{1,1}$ of the hidden representation of shallow fully-connected networks trained in the setting of this section, as a function of the number $P$ of training data for different combinations of the model parameters. The bottom panel, in particular, highlights that the sensitivity is close to 1 for $P \ll P_c$ and close to 0 for $P \gg P_c$. In addition, notice that the collapse of the pre-activations of synonymic tuples onto the same, synonymic invariant value, implies that the rank of the hidden weights matrix tends to $v$—the vocabulary size of higher-level features. This low-rank structure is typical in the weights of deep networks trained on image classification [37–40].

**Using all patches via weight sharing.** Notice that using a one-hidden-layer CNN which looks at all patches via weight sharing and global average pooling would yield the same result since the average over patches reduces both the signal and the noise by the same factor–see subsection C.1 for details.

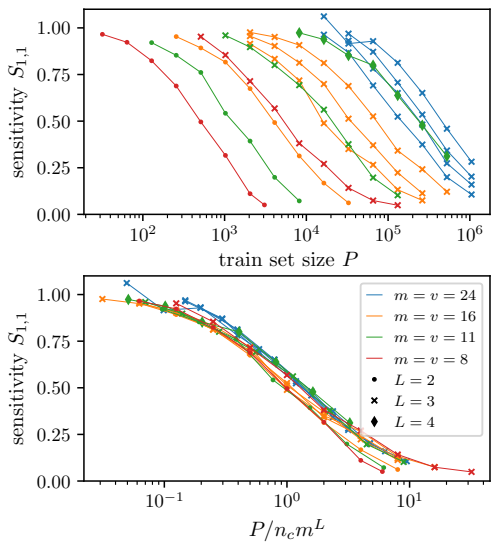

Figure 11: Synonymic sensitivity of the hidden representation vs $P$ for a one-hidden-layer fully-connected network trained on the first patch of the inputs of an RHM with $s = 2$ and $m = v$, for several values of $L$, $v$, and $n_c \leq v$. The top panel shows the bare curves whereas, in the bottom panel, the x-axis is rescaled by $P_c = n_c m^L$. The collapse of the rescaled curves highlights that $P_c$ coincides with the threshold number of training data for building a synonymic invariant representation.

**Improved Performance via Clustering.** Note that our signal-vs-noise argument is based on a single class $\alpha$, as it considers the scalar quantity $\hat{f}(\alpha | \boldsymbol{\mu})$. However, an observer seeking to identify synonyms could in principle use the information from all classes, represented by the $n_c$-dimensional vector of empirical frequencies $(\hat{f}(\alpha | \boldsymbol{\mu}))_{\alpha=1,...,n_c}$. Following this idea, one can devise a layer-wise algorithm where the representations of each layer are first updated with a single step of gradient descent (as in Eq. 82), then clustered into synonymic groups [22, 27]. Such an algorithm can solve the RHM with less than $n_c m^L$ training points—$\sqrt{n_c} m^L$ in the maximal dataset case $n_c = v$ and $m = v^{s-1}$, as we show empirically and justify theoretically in Appendix D. Notably, the dependence on the dimensionality $m^L$ is unaffected by the change of algorithm, although the prefactor reveals the advantage of the dedicated clustering algorithm over standard CNNs.

## 5 Conclusions

We have introduced a hierarchical model of classification task, where each class is identified by a number of equivalent high-level features (synonyms), themselves consisting of a number of equivalent sub-features according to a hierarchy of random composition rules. First, we established via a combination of extensive experiments and theoretical arguments that the sample complexity of deep CNNs is a simple function of the number of classes $n_c$, the number of synonymic features $m$ and the depth of the hierarchy $L$. This result provides a rule of thumb for estimating the order of magnitude of the sample complexity of real datasets. In the case of CIFAR10 [41], for instance, having 10 classes, taking reasonable values for the RHM parameters such as $m \in [5, 15]$ and $L = 3$, yields $P^* \in [10^5, 3 \times 10^4]$,comparable with the sample complexity of modern architectures (see Fig. 16 in Appendix G).

Secondly, our results indicate a separation between shallow networks, which are cursed by the input dimensionality, and sufficiently deep CNNs, which are not. We thus complement previous analyses based on expressivity [21] or information-theoretical considerations [24] with a generalization result.

Last but not least, we proposed to characterize the quality of internal representations with their sensitivity toward transformations of the data which leave the task invariant. This analysis bypasses the issues of previous characterizations. For example, approaches based on mutual information [12] that is ill-defined when the network representations are deterministic functions of the input [13]. Approaches based on intrinsic dimension [14,15] can display counter-intuitive results, refer to Appendix F for a more in-depth discussion on the intrinsic dimension, and on how this quantity behaves in our setup. Interestingly, our approach supports that performance should strongly correlate with the invariance toward synonyms of the internal representation. This prediction could in principle be tested in natural language processing models, but also for image data sets by performing discrete changes to images that leave the class unchanged.

Looking forward, the Random Hierarchy Model is a rich but minimal model where open questions in the theory of deep learning could be clarified. For instance, a formidable challenge such as the description of the gradient descent dynamics of deep networks, becomes significantly simpler for the RHM, owing to the simple structure of the target representations. Other important questions, including the ability of fully-connected networks to learn local connections [30, 42, 43], the benefits of residual connections [44] or the advantages of deep learning over kernel methods [25, 45–47] can be studied quantitatively within this model, as functions of the multiple parameters that define the hierarchical structure of the task.

## Materials and Methods

### Experimental Setup

The experiments are performed using the PyTorch deep learning framework [48]. The code used for the experiments is available online at https://github.com/pcsl-epfl/hierarchy-learning.

### RHM implementation

The code implementing the RHM is available online at https://github.com/pcsl-epfl/hierarchy-learning/blob/master/datasets/hierarchical.py. The inputs sampled from the RHM are represented as a one-hot encoding of low-level features. This makes each input of size $s^L \times v$. The inputs are whitened so that the average pixel value over channels is equal to zero.

### Model Architecture

One-hidden-layer fully-connected networks have input dimension equal to $s^L \times v$, $H = 10^4$ hidden neurons, and $n_c$ outputs. The deep convolutional neural networks (CNNs) have weight sharing, stride equal to filter size equal to $s$ and $L$ hidden layers. In this case, we set the width $H$ to be larger than the number of possible $s$-tuples that can exist at a given layer, $H \gg v^s$.

### Training Procedure

Neural networks are trained using stochastic gradient descent (SGD) on the cross-entropy loss, with a batch size of 128 and a learning rate equal to 0.3. Training is stopped when the training loss decreases below a certain threshold fixed to $10^{-3}$.

### Measurements

The performance of the models is measured as the percentage error on a test set. The test set size is chosen to be $\min(P_{\max} - P, 20'000)$. Synonymic sensitivity, as defined in Eq. 8, is measured on a test set of size $\min(P_{\max} - P, 1'000)$. Reported results for a given value of RHM parameters are averaged over 10 jointly different instances of the RHM and network initializations.

# Appendix

## A  Statistics of The Composition Rules

In this section, we consider a single composition rule, that is the assignment of $m$ $s$-tuples of low-level features to each of the $v$ high-level features. In the RHM these rules are chosen uniformly at random over all the possible rules, thus their statistics are crucial in determining the correlations between the input features and the class label.

### A.1  Statistics of a single rule

For each rule, we call $N_i(\mu_1; \mu_2)$ the number of occurrences of the low-level feature $\mu_1$ in position $i$ of the $s$-tuples generated by the higher-level feature $\mu_2$. The probability of $N_i(\mu_1; \mu_2)$ is that of the number of successes when drawing $m$ (number of $s$-tuples associated with the high-level feature $\mu_2$) times without replacement from a pool of $v^s$ (total number of $s$-tuples with vocabulary size $v$) objects where only $v^{s-1}$ satisfy a certain condition (number of $s$-tuples displaying feature $\mu_1$ in position $i$):

$$\Pr\{N_i(\mu_0; \mu_1) = k\} = \binom{v^{s-1}}{k} \binom{v^s - v^{s-1}}{m-k} \Big/ \binom{v^s}{m},\tag{12}$$

which is a hypergeometric distribution $\mathrm{Hg}_{v^s, v^{s-1}, m}$, with mean

$$\langle N \rangle = m \frac{v^{s-1}}{v^s} = \frac{m}{v},\tag{13}$$

and variance

$$\sigma_N^2 := \left\langle (N - \langle N \rangle)^2 \right\rangle = m \frac{v^{s-1}}{v^s} \frac{v^s - v^{s-1}}{v^s} \frac{v^s - m}{v^s - 1} = \frac{m}{v} \frac{v-1}{v} \frac{v^s - m}{v^s - 1} \xrightarrow{m \gg 1} \frac{m}{v},\tag{14}$$

independently of the position $i$ and the specific low- and high-level features. Notice that, since $m \le v^{s-1}$ with $s$ fixed, large $m$ implies also large $v$.

### A.2  Joint statistics of a single rule

**Shared high-level feature.**  For a fixed high-level feature $\mu_2$, the joint probability of the occurrences of two different low-level features $\mu_1$ and $\nu_1$ is a multivariate Hypergeometric distribution,

$$\Pr\{N_i(\mu_1; \mu_2) = k; N_i(\nu_1; \mu_2) = l\} = \binom{v^{s-1}}{k} \binom{v^{s-1}}{l} \binom{v^s - 2v^{s-1}}{m-k-l} \Big/ \binom{v^s}{m},\tag{15}$$

giving the following covariance,

$$c_N := \langle (N_i(\mu_1; \mu_2) - \langle N \rangle)(N_i(\nu_1; \mu_2) - \langle N \rangle) \rangle = -\frac{m}{v^2} \frac{v^s - m}{v^s - 1} \xrightarrow{m \gg 1} -\left(\frac{m}{v}\right)^2 \frac{1}{m}.\tag{16}$$

The covariance can also be obtained via the constraint $\sum_{\mu_1} N_i(\mu_1; \mu_2) = m$. For any finite sequence of identically distributed random variables $X_\mu$ with a constraint on the sum $\sum_\mu X_\mu = m$,

$$\sum_{\mu=1}^{v} X_\mu = m \Rightarrow \sum_{\mu=1}^{v} (X_\mu - \langle X_\mu \rangle) = 0 \Rightarrow$$

$$(X_\mu - \langle X_\nu \rangle) \sum_{\mu=1}^{v} (X_\mu - \langle X_\mu \rangle) = 0 \Rightarrow \sum_{\mu=1}^{v} \langle (X_\nu - \langle X_\nu \rangle)(X_\mu - \langle X_\mu \rangle) \rangle = 0 \Rightarrow$$

$$\mathrm{Var}\,[X_\mu] + (v-1)\mathrm{Cov}\,[X_\mu, X_\nu] = 0.\tag{17}$$

In the last line, we used the identically distributed variables hypothesis to replace the sum over $\mu \ne \nu$ with the factor $(v-1)$. Therefore,

$$c_N = \mathrm{Cov}\,[N_i(\mu_1; \mu_2), N_i(\nu_1; \mu_2)] = -\frac{\mathrm{Var}\,[N_i(\mu_1; \mu_2)]}{v-1} = -\frac{\sigma_N^2}{v-1}.\tag{18}$$

**Shared low-level feature.** The joint probability of the occurrences of the same low-level feature $\mu_1$ starting from different high-level features $\mu_2 \neq \nu_2$ can be written as follows,

$$\Pr\{N(\mu_1; \mu_2) = k; N(\mu_1; \nu_2) = l\} = \Pr\{N(\mu_1; \mu_2) = k | N(\mu_1; \nu_2) = l\} \times \Pr\{N(\mu_1; \nu_2) = l\} \tag{19}$$

$$= \mathrm{Hg}_{v^s - m, v^{s-1} - l, m}(k) \times \mathrm{Hg}_{v^s, v^{s-1}, m}(l), \tag{20}$$

resulting in the following 'inter-feature' covariance,

$$c_{if} := \mathrm{Cov}\left[N_i(\mu_1; \mu_2), N_i(\mu_1; \nu_2)\right] = -\left(\frac{m}{v}\right)^2 \frac{v-1}{v^s - 1}. \tag{21}$$

**No shared features.** Finally, by multiplying both sides of $\sum_{\mu_1} N(\mu_1; \mu_2) = m$ with $N(\nu_1; \nu_2)$ and averaging, we get

$$c_g := \mathrm{Cov}\left[N_i(\mu_1; \mu_2), N_i(\nu_1; \nu_2)\right] = -\frac{\mathrm{Cov}\left[N_i(\mu_1; \mu_2), N_i(\nu_1; \nu_2)\right]}{v - 1} = \left(\frac{m}{v}\right)^2 \frac{1}{v^s - 1}. \tag{22}$$

# B  Emergence of input-output correlations ($P_c$)

As discussed in the main text, the Random Hierarchy Model presents a characteristic sample size $P_c$ corresponding to the emergence of the input-output correlations. This sample size predicts the sample complexity of deep CNNs, as we also discuss in the main text. In this appendix, we prove that

$$P_c \xrightarrow{n_c, m \to \infty} n_c m^L. \tag{23}$$

## B.1  Estimating the Signal

The correlations between input features and the class label can be quantified via the conditional probability (over realizations of the RHM) of a data point belonging to class $\alpha$ conditioned on displaying the $s$-tuple $\boldsymbol{\mu}$ in the $j$-th input patch,

$$f_j(\alpha | \boldsymbol{\mu}) := \Pr\{\boldsymbol{x} \in \alpha | \boldsymbol{x}_j = \boldsymbol{\mu}\}, \tag{24}$$

where the notation $\boldsymbol{x}_j = \boldsymbol{\mu}$ means that the elements of the patch $\boldsymbol{x}_j$ encode the tuple of features $\boldsymbol{\mu}$. We say that the low-level features are correlated with the output if

$$f_j(\alpha | \boldsymbol{\mu}) \neq \frac{1}{n_c}, \tag{25}$$

and define a 'signal' as the difference $f_j(\alpha | \boldsymbol{\mu}) - n_c^{-1}$. In the following, we compute the statistics of the signal over realizations of the RHM.

**Occurrence of low-level features.** Let us begin by defining the joint occurrences of a class label $\alpha$ and a low-level feature $\mu_1$ in a given position of the input. Using the tree representation of the model, we will identify an input position with a set of $L$ indices $i_\ell = 1, \dots, s$, each indicating which branch to follow when descending from the root (class label) to a given leaf (low-level feature). These joint occurrences can be computed by combining the occurrences of the single rules introduced in [Appendix A](#) of this Appendix. With $L = 2$, for instance,

$$N_{i_1 i_2}^{(1 \to 2)}(\mu_1; \alpha) = \sum_{\mu_2 = 1}^{v} \left(m^{s-1} N_{i_1}^{(1)}(\mu_1; \mu_2)\right) \times N_{i_2}^{(2)}(\mu_2; \alpha), \tag{26}$$

where:

i) $N_{i_2}^{(2)}(\mu_2; \alpha)$[5] counts the occurrences of $\mu_2$ in position $i_2$ of the level-2 representations of $\alpha$, i.e. the $s$-tuples generated from $\alpha$ according to the second-layer composition rule;

---

[5]We are using the superscript $(\ell)$ to differentiate the occurrences of the different composition rules.

*ii)* $N_{i_1}^{(1)}(\mu_1; \mu_2)$ counts the occurrences of $\mu_1$ in position $i_1$ of the level-1 representations of $\mu_2$, i.e. $s$-tuples generated by $\mu_2$ according to the composition rule of the first layer;

*iii)* the factor $m^{s-1}$ counts the descendants of the remaining $s-1$ elements of the level-2 representation ($m$ descendants per element);

*iv)* the sum over $\mu_2$ counts all the possible paths of features that lead to $\mu_1$ from $\alpha$ across 2 generations.

The generalization of Eq. 26 is immediate once one takes into account that the multiplicity factor accounting for the descendants of the remaining positions at the $\ell$-th generation is equal to $m^{s^\ell}/m$ ($s^{\ell-1}$ is the size of the representation at the previous level). Hence, the overall multiplicity factor after $L$ generations is

$$1 \times \frac{m^s}{m} \times \frac{m^{s^2}}{m} \times \cdots \times \frac{m^{s^{L-1}}}{m} = m^{\frac{s^L-1}{s-1}-L}, \tag{27}$$

so that the number of occurrences of feature $\mu_1$ in position $i_1 \dots i_L$ of the inputs belonging to class $\alpha$ is

$$N_{i_{1\to L}}^{(1 \to L)}(\mu_1; \alpha) = m^{\frac{s^L-1}{s-1}-L} \sum_{\mu_2, \dots, \mu_L=1}^{v} N_{i_1}^{(1)}(\mu_1; \mu_2) \times \cdots \times N_{i_L}^{(L)}(\mu_L; \alpha), \tag{28}$$

where we used $i_{1\to L}$ as a shorthand notation for the tuple of indices $i_1, i_2, \dots, i_L$.

The same construction allows us to compute the number of occurrences of up to $s-1$ features within the $s$-dimensional patch of the input corresponding to the path $i_{2\to L}$. The number of occurrences of a whole $s$-tuple, instead, follows a slightly different rule, since there is only one level-2 feature $\mu_2$ which generates the whole $s$-tuple of level-1 features $\boldsymbol{\mu}_1 = (\mu_{1,1}, \dots, \mu_{1,s})$—we call this feature $g_1(\boldsymbol{\mu}_1)$, with $g_1$ denoting the first-layer composition rule. As a result, the sum over $\mu_2$ in the right-hand side of Eq. 28 disappears and we are left with

$$N_{i_{2\to L}}^{(1 \to L)}(\boldsymbol{\mu}_1; \alpha) = m^{\frac{s^L-1}{s-1}-L} \sum_{\mu_3, \dots, \mu_L=1}^{v} N_{i_2}^{(2)}(g_1(\boldsymbol{\mu}_1); \mu_3) \times \cdots \times N_{i_L}^{(L)}(\mu_L; \alpha). \tag{29}$$

Coincidentally, Eq. 29 shows that the joint occurrences of a $s$-tuple of low-level features $\boldsymbol{\mu}_1$ depend on the level-2 feature corresponding to $\boldsymbol{\mu}_1$. Hence, $N_{i_{2\to L}}^{(1 \to L)}(\boldsymbol{\mu}_1; \alpha)$ is invariant for the exchange of $\boldsymbol{\mu}_1$ with one of its synonyms, i.e. level-1 tuples $\boldsymbol{\nu}_1$ corresponding to the same level-2 feature.

**Class probability conditioned on low-level observations.** We can turn these numbers into probabilities by normalizing them appropriately. Upon dividing by the total occurrences of a low-level feature $\mu_1$ independently of the class, for instance, we obtain the conditional probability of the class of a given input, conditioned on the feature in position $i_1 \dots i_L$ being $\mu_1$.

$$f_{i_{1\to L}}^{(1 \to L)}(\alpha|\mu_1) := \frac{N_{i_{1\to L}}^{(1 \to L)}(\mu_1; \alpha)}{\sum_{\alpha'=1}^{n_c} N_{i_{1\to L}}^{(1 \to L)}(\mu_1; \alpha')} = \frac{\sum_{\mu_2, \dots, \mu_L=1}^{v} N_{i_1}^{(1)}(\mu_1; \mu_2) \times \cdots \times N_{i_L}^{(L)}(\mu_L; \alpha)}{\sum_{\mu_2, \dots, \mu_L=1}^{v} \sum_{\mu_{L+1}=1}^{n_c} N_{i_1}^{(1)}(\mu_1; \mu_2) \times \cdots \times N_{i_L}^{(L)}(\mu_L; \mu_{L+1})}. \tag{30}$$

Let us also introduce, for convenience, the numerator and denominator of the right-hand side of Eq. 30.

$$U_{i_{1\to L}}^{(1 \to L)}(\mu_1 \alpha) = \sum_{\mu_2, \dots, \mu_L=1}^{v} N_{i_1}^{(1)}(\mu_1; \mu_2) \times \cdots \times N_{i_L}^{(L)}(\mu_L; \alpha); \quad D_{i_{1\to L}}^{(1 \to L)}(\mu_1) = \sum_{\alpha=1}^{n_c} U_{i_{1\to L}}^{(1 \to L)}(\mu_1; \alpha). \tag{31}$$

### B.1.1 Statistics of the numerator $U$

We now determine the first and second moments of the numerator of $f_{i_{1\to L}}^{(1 \to L)}(\mu_1; \alpha)$. Let us first recall the definition for clarity,

$$U_{i_{1\to L}}^{(1 \to L)}(\mu_1; \alpha) = \sum_{\mu_2, \dots, \mu_L=1}^{v} N_{i_1}^{(1)}(\mu_1; \mu_2) \times \cdots \times N_{i_L}^{(L)}(\mu_L; \alpha) \tag{32}$$

**Level 1** $L = 1$. For $L = 1$, $U$ is simply the occurrence of a single production rule $N_i(\mu_1; \alpha)$,

$$\left\langle U^{(1)} \right\rangle = \frac{m}{v}; \tag{33}$$

$$\sigma^2_{U^{(1)}} := \mathrm{Var}\left[U^{(1)}\right] = \frac{m}{v}\frac{v-1}{v}\frac{v^s - m}{v^s - 1} \xrightarrow{v \gg 1} \frac{m}{v}; \tag{34}$$

$$c_{U^{(1)}} := \mathrm{Cov}\left[U^{(1)}(\mu_1; \alpha), U^{(1)}(\nu_1; \alpha)\right] = -\frac{\mathrm{Var}\left[U^{(1)}\right]}{(v-1)} = -\left(\frac{m}{v}\right)^2\frac{v^s - m}{v^s - 1}\frac{1}{m} \xrightarrow{v \gg 1} \left(\frac{m}{v}\right)^2\frac{1}{m}; \tag{35}$$

where the relationship between variance and covariance is due to the constraint on the sum of $U^{(1)}$ over $\mu_1$, see Eq. 17.

**Level 2** $L = 2$. For $L = 2$,

$$U^{(1 \to 2)}_{i_{1 \to 2}}(\mu_1; \alpha) = \sum_{\mu_2 = 1}^{v} N^{(1)}_{i_1}(\mu_1; \mu_2) \times N^{(2)}_{i_2}(\mu_3; \alpha) = \sum_{\mu_2 = 1}^{v} N^{(1)}_{i_1}(\mu_1; \mu_2) U^{(2)}_{i_2}(\mu_2; \alpha). \tag{36}$$

Therefore,

$$\left\langle U^{(1 \to 2)} \right\rangle = v\left(\frac{m}{v}\right) \times \left\langle U^{(1)} \right\rangle = v\left(\frac{m}{v}\right)^2; \tag{37}$$

$$\begin{aligned}
\sigma^2_{U^{(2)}} := \mathrm{Var}\left[U^{(1 \to 2)}\right] &= \sum_{\mu_2, \nu_1 = 1}^{v} \left(\left\langle N^{(1)}(\mu_1; \mu_2)N^{(1)}(\mu_1; \nu_2) \right\rangle \left\langle U^{(2)}(\mu_2; \alpha)U^{(2)}(\nu_2; \alpha) \right\rangle - \left\langle N \right\rangle^2 \left\langle U^{(1)} \right\rangle^2\right) \\
&= \sum_{\mu_2, \nu_2 = \mu_2} \cdots + \sum_{\mu_2} \sum_{\nu_2 \neq \mu_2} \cdots \\
&= v\left(\sigma^2_N \sigma^2_{U^{(1)}} + \sigma^2_N \left\langle U^{(1)} \right\rangle^2 + \sigma^2_{U^{(1)}} \left\langle N \right\rangle^2\right) + v(v-1)\left(c_{if} c_{U^{(1)}} + c_{if} \left\langle U^{(1)} \right\rangle^2 + c_{U^{(1)}} \left\langle N \right\rangle^2\right) \\
&= v\left(\sigma^2_N \sigma^2_{U^{(1)}} + (v-1)c_{if} c_{U^{(1)}}\right) + v\left\langle U^{(1)} \right\rangle^2 \left(\sigma^2_N + (v-1)c_{if}\right) + v\left\langle N \right\rangle^2 \left(\sigma^2_{U^{(1)}} + (v-1)c_{U^{(1)}}\right) \\
&= v\sigma^2_{U^{(1)}}\left(\sigma^2_N - c_{if}\right) + v\left\langle U^{(1)} \right\rangle^2 \left(\sigma^2_N + (v-1)c_{if}\right), \tag{38}
\end{aligned}$$

$$c_{U^{(2)}} = -\frac{\sigma^2_{U^{(2)}}}{(v-1)} \tag{39}$$

**Level L.** In general,

$$U^{(1 \to L)}_{i_{1 \to L}}(\mu_1; \alpha) = \sum_{\mu_2 = 1}^{v} N^{(1)}_{i_1}(\mu_1; \mu_2) U^{(2 \to L)}_{i_{2 \to L}}(\mu_2; \alpha). \tag{40}$$

Therefore,

$$\left\langle U^{(L)} \right\rangle = v\left(\frac{m}{v}\right) \times \left\langle U^{(L-1)} \right\rangle = v^{L-1}\left(\frac{m}{v}\right)^L; \tag{41}$$

$$\begin{aligned}
\sigma^2_{U^{(L)}} &= \sum_{\mu_2, \nu_1 = 1}^{v} \left(\left\langle N^{(1)}(\mu_2; \mu_2)N^{(1)}(\mu_1; \nu_2) \right\rangle \left\langle U^{(2 \to L)}(\mu_2; \alpha)U^{(2 \to L)}(\nu_1; \alpha) \right\rangle - \left\langle N \right\rangle^2 \left\langle U^{(1 \to (L-1))} \right\rangle^2\right) \\
&= \sum_{\mu_2, \nu_2 = \mu_2} \cdots + \sum_{\mu_2} \sum_{\nu_2 \neq \mu_2} \cdots \\
&= v\left(\sigma^2_N \sigma^2_{U^{(L-1)}} + \sigma^2_N \left\langle U^{(L-1)} \right\rangle^2 + \sigma^2_{U^{(L-1)}} \left\langle N \right\rangle^2\right) + v(v-1)\left(\sigma^2_{if} c_{U^{(L-1)}} + c_{if} \left\langle U^{(L-1)} \right\rangle^2 + c_{U^{(L-1)}} \left\langle N \right\rangle^2\right) \\
&= v\sigma^2_{U^{(L-1)}}\left(\sigma^2_N - c_{if}\right) + v\left\langle U^{(L-1)} \right\rangle^2 \left(\sigma^2_N + (v-1)c_{if}\right), \tag{42}
\end{aligned}$$

$$c_{U^{(L)}} = -\frac{\sigma^2_{U^{(L)}}}{(v-1)} \tag{43}$$

**Concentration for large $m$.** In the large multiplicity limit $m \gg 1$, the $U$'s concentrate around their mean value. Due to $m \leq v^{s-1}$, large $m$ implies large $v$, thus we can proceed by setting $m = qv^{s-1}$, with $q \in (0,1]$ [6] and studying the $v \gg 1$ limit. From Eq. 41,

$$\left\langle U^{(L)} \right\rangle = q^L v^{L(s-1)-1}. \tag{44}$$

In addition,

$$\sigma_N^2 \xrightarrow{v \gg 1} \frac{m}{v} = qv^{(s-1)-1}, \quad c_{if} \xrightarrow{v \gg 1} -\left(\frac{m}{v}\right)^2 \frac{1}{v^{s-1}} = -q^2 v^{(s-1)-2}, \tag{45}$$

so that

$$\sigma_{U^{(L)}}^2 = v\sigma_{U^{(L-1)}}^2 \left(\sigma_N^2 - \sigma_{if}^2\right) + v\left\langle U^{(L-1)} \right\rangle^2 \left(\sigma_N^2 + (v-1)\sigma_{if}^2\right)$$

$$\xrightarrow{v \gg 1} \sigma_{U^{(L-1)}}^2 qv^{(s-1)} + \sigma_{U^{(L-1)}}^2 q^2 v^{(s-1)-1} + q^{2L-1}(1-q)v^{(2L-1)(s-1)-2} \tag{46}$$

The second of the three terms is always subleading with respect to the first, so we can discard it for now. It remains to compare the first and the third terms. For $L=2$, since $\sigma_{U^{(1)}}^2 = \sigma_N^2$, the first term depends on $v$ as $v^{2(s-1)-1}$, whereas the third is proportional to $v^{3(s-1)-2}$. For $L \geq 3$ the dominant scaling is that of the third term only: for $L=3$ it can be shown by simply plugging the $L=2$ result into the recursion, and for larger $L$ it follows from the fact that replacing $\sigma_{U^{(L-1)}}^2$ in the first term with the third term of the precious step always yields a subdominant contribution. Therefore,

$$\sigma_{U^{(L)}}^2 \xrightarrow{v \gg 1} \begin{cases} q^2 v^{2(s-1)-1} + q^3(1-q)v^{3(s-1)-2}, & \text{for } L=2, \\ q^{2L-1}(1-q)v^{(2L-1)(s-1)-2}, & \text{for } L \geq 3. \end{cases} \tag{47}$$

Upon dividing the variance by the squared mean we get

$$\frac{\sigma_{U^{(L)}}^2}{\left\langle U^{(L)} \right\rangle^2} \xrightarrow{v \gg 1} \begin{cases} \dfrac{1}{q^2} \dfrac{1}{v^{2(s-1)-1}} + \dfrac{1-q}{q} \dfrac{1}{v^{(s-1)}}, & \text{for } L=2, \\ \dfrac{1-q}{q} \dfrac{1}{v^{(s-1)}}, & \text{for } L \geq 3, \end{cases} \tag{48}$$

whose convergence to 0 guarantees the concentration of the $U$'s around the average over all instances of the RHM.

### B.1.2 Statistics of the denominator $D$

Here we compute the first and second moments of the denominator of $f_{i_1 \to L}^{(1 \to L)}(\mu_1; \alpha)$,

$$D_{i_1 \to L}^{(1 \to L)}(\mu_1) = \sum_{\mu_2, \dots, \mu_L}^{v} \sum_{\mu_{L+1}=1}^{n_c} N_{i_1}^{(1)}(\mu_1; \mu_2) \times \cdots \times N_{i_L}^{(L)}(\mu_L; \mu_{L+1}) \tag{49}$$

**Level 1 $L=1$.** For $L=1$, $D$ is simply the sum over classes of the occurrences of a single production rule, $D^{(1)} = \sum_\alpha N_i(\mu_1; \alpha)$,

$$\left\langle D^{(1)} \right\rangle = n_c \frac{m}{v}; \tag{50}$$

$$\sigma_{D^{(1)}}^2 := \text{Var}\left[D^{(1)}\right] = n_c \sigma_N^2 + n_c(n_c-1)c_{if} = n_c\left(\frac{m}{v}\right)^2 \frac{v-1}{v^s-1}\left(\frac{v^s}{m} - n_c\right)$$

$$\xrightarrow{v \gg 1} n_c\left(\frac{m}{v}\right)^2 \left(\frac{v}{m} - \frac{n_c}{v^{s-1}}\right); \tag{51}$$

$$c_{D^{(1)}} := \text{Cov}\left[D^{(1)}(\mu_1), D^{(1)}(\nu_0)\right] = -\frac{\text{Var}\left[D^{(1)}\right]}{(v-1)} = n_c c_N + n_c(n_c-1)c_g, \tag{52}$$

where, in the last line, we used the identities $\sigma_N^2 + (v-1)c_N = 0$ from Eq. 16 and $c_{if} + (v-1)c_g = 0$ from Eq. 22.

---

[6]The minimum $m$ is 1, which corresponds to $q = v^{1-s}$, but actually there is no stochasticity in the $U$'s and $D$'s in that case. Thus the minimal $q$ is actually $2v^{1-s}$.

**Level 2** $L = 2$. For $L = 2$,

$$D_{i_1 \to 2}^{(1\to2)}(\mu_1) = \sum_{\mu_2}^{v} \sum_{\mu_3=1}^{n_c} N_{i_1}^{(1)}(\mu_1;\mu_2) \times N_{i_2}^{(2)}(\mu_2;\mu_3) = \sum_{\mu_2=1}^{v} N_{i_1}^{(1)}(\mu_1;\mu_2) D_{i_2}^{(2)}(\mu_2). \tag{53}$$

Therefore,

$$\left\langle D^{(1\to2)} \right\rangle = v \left( \frac{m}{v} \right) \times \left\langle D^{(1)} \right\rangle = \frac{n_c}{v} m^2; \tag{54}$$

$$\sigma_{D^{(2)}}^2 := \mathrm{Var}\left[D^{(1\to2)}\right] = \sum_{\mu_2,\nu_1=1}^{v} \left( \left\langle N^{(1)}(\mu_1;\mu_2) N^{(1)}(\mu_1;\nu_1) \right\rangle \left\langle D^{(2)}(\mu_2) D^{(2)}(\nu_1) \right\rangle - \langle N \rangle^2 \left\langle D^{(1)} \right\rangle^2 \right)$$

$$= \sum_{\mu_2,\nu_1=\mu_2} \cdots + \sum_{\mu_2} \sum_{\nu_1 \neq \mu_2} \cdots$$

$$= v \left( \sigma_N^2 \sigma_{D^{(1)}}^2 + \sigma_N^2 \left\langle D^{(1)} \right\rangle^2 + \sigma_{D^{(1)}}^2 \langle N \rangle^2 \right) + v(v-1) \left( c_{if} c_{D^{(1)}} + c_{if} \left\langle D^{(1)} \right\rangle^2 + c_{D^{(1)}} \langle N \rangle^2 \right)$$

$$= v \left( \sigma_N^2 \sigma_{D^{(1)}}^2 + (v-1) c_{if} c_{D^{(1)}} \right) + v \left\langle D^{(1)} \right\rangle^2 \left( \sigma_N^2 + (v-1) c_{if} \right) + v \langle N \rangle^2 \left( \sigma_{D^{(1)}}^2 + (v-1) c_{D^{(1)}} \right)$$

$$= v \sigma_{D^{(1)}}^2 \left( \sigma_N^2 - c_{if} \right) + v \left\langle D^{(1)} \right\rangle^2 \left( \sigma_N^2 + (v-1) c_{if} \right), \tag{55}$$

$$c_{D^{(2)}} = -\frac{\sigma_{D^{(2)}}^2}{(v-1)}. \tag{56}$$

**Level L.** In general,

$$D_{i_1 \to L}^{(1\to L)}(\mu_1) = \sum_{\mu_2=1}^{v} N_{i_1}^{(1)}(\mu_1;\mu_2) D_{i_2 \to L}^{(2\to L)}(\mu_2). \tag{57}$$

Therefore,

$$\left\langle D^{(L)} \right\rangle = v \left( \frac{m}{v} \right) \times \left\langle D^{(L-1)} \right\rangle = \frac{n_c}{v} m^L; \tag{58}$$

$$\sigma_{D^{(L)}}^2 = \sum_{\mu_2,\nu_1=1}^{v} \left( \left\langle N^{(1)}(\mu_1;\mu_2) N^{(1)}(\mu_1;\nu_1) \right\rangle \left\langle D^{(2\to L)}(\mu_2) D^{(2\to L)}(\nu_1;\alpha) \right\rangle - \langle N \rangle^2 \left\langle D^{(1\to(L-1))} \right\rangle^2 \right)$$

$$= \sum_{\mu_2,\nu_1=\mu_2} \cdots + \sum_{\mu_2} \sum_{\nu_1 \neq \mu_2} \cdots$$

$$= v \left( \sigma_N^2 \sigma_{D^{(L-1)}}^2 + \sigma_N^2 \left\langle D^{(L-1)} \right\rangle^2 + \sigma_{D^{(L-1)}}^2 \langle N \rangle^2 \right) + v(v-1) \left( c_{if} c_{D^{(L-1)}} + c_{if} \left\langle D^{(L-1)} \right\rangle^2 + c_{D^{(L-1)}} \langle N \rangle^2 \right)$$

$$= v \sigma_{D^{(L-1)}}^2 \left( \sigma_N^2 - c_{if} \right) + v \left\langle D^{(L-1)} \right\rangle^2 \left( \sigma_N^2 + (v-1) c_{if} \right), \tag{59}$$

$$c_{D^{(L)}} = -\frac{\sigma_{D^{(L)}}^2}{(v-1)}. \tag{60}$$

**Concentration for large $m$.** Since the $D$'s can be expressed as a sum of different $U$'s, their concentration for $m \gg 1$ follows directly from that of the $U$'s.

### B.1.3 Estimate of the conditional class probability

We can now turn back to the original problem of estimating

$$f_{i_1 \to L}^{(1\to L)}(\alpha|\mu_1) = \frac{\displaystyle\sum_{\mu_2,\ldots,\mu_L=1}^{v} N_{i_1}^{(1)}(\mu_1;\mu_2) \times \cdots \times N_{i_L}^{(L)}(\mu_L;\alpha)}{\displaystyle\sum_{\mu_2,\ldots,\mu_L=1}^{v} \sum_{\mu_{L+1}=1}^{n_c} N_{i_1}^{(1)}(\mu_1;\mu_2) \times \cdots \times N_{i_L}^{(L)}(\mu_L;\mu_{L+1})} = \frac{U_{i_1 \to L}^{(1\to L)}(\mu_1;\alpha)}{D_{i_1 \to L}^{(1\to L)}(\mu_1)}. \tag{61}$$

Having shown that both numerator and denominator converge to their average for large $m$, we can expand for small fluctuations around these averages and write

$$f_{i_{1\to L}}^{(1\to L)}(\alpha|\mu_1) = \frac{v^{-1}m^L\left(1 + \frac{U_{i_{1\to L}}^{(1\to L)}(\mu_1;\alpha) - m^L/v}{m^L/v}\right)}{n_c v^{-1} m^L\left(1 + \frac{D_{i_{1\to L}}^{(1\to L)}(\mu_1) - n_c m^L/v}{m^L}\right)}$$

(62)

$$= \frac{1}{n_c} + \frac{1}{n_c}\frac{U_{i_{1\to L}}^{(1\to L)}(\mu_1;\alpha) - m^L/v}{m^L/v} - \frac{1}{n_c}\frac{D_{i_{1\to L}}^{(1\to L)}(\mu_1) - n_c m^L/v}{m^L/v}$$

$$= \frac{1}{n_c} + \frac{v}{n_c m^L}\left(U_{i_{1\to L}}^{(1\to L)}(\mu_1;\alpha) - \frac{1}{n_c}D_{i_{1\to L}}^{(1\to L)}(\mu_1)\right).$$

(63)

Since the conditional frequencies average to $n_c^{-1}$, the term in brackets averages to zero. We can then estimate the size of the fluctuations of the conditional frequencies (i.e. the 'signal') with the standard deviation of the term in brackets.

It is important to notice that, for each $L$ and position $i_{1\to L}$, $D$ is the sum over $\alpha$ of $U$, and the $U$ with different $\alpha$ at fixed low-level feature $\mu_1$ are identically distributed. In general, for a sequence of identically distributed variables $(X_\alpha)_{\alpha=1,\dots,n_c}$,

$$\left\langle\left(\frac{1}{n_c}\sum_{\beta=1}^{v}X_\beta\right)^2\right\rangle = \frac{1}{n_c^2}\sum_{\beta=1}^{n_c}\left(\langle X_\beta\rangle^2 + \sum_{\beta'\neq\beta}\langle X_\beta X_{\beta'}\rangle\right) = \frac{1}{n_c}\left(\langle X_\beta\rangle^2 + \sum_{\beta'\neq\beta}\langle X_\beta X_{\beta'}\rangle\right).$$

(64)

Hence,

$$\left\langle\left(X_\alpha - \frac{1}{n_c}\sum_{\beta=1}^{n_c}X_\beta\right)^2\right\rangle = \langle X_\alpha^2\rangle + n_c^{-2}\sum_{\beta,\gamma=1}^{n_c}\langle X_\beta X_\gamma\rangle - 2n_c^{-1}\sum_{\beta=1}^{n_c}\langle X_\alpha X_\beta\rangle$$

$$= \langle X_\alpha^2\rangle - n_c^{-1}\left(\langle X_\alpha\rangle^2 + \sum_{\beta\neq\alpha}\langle X_\alpha X_\beta\rangle\right)$$

$$= \langle X_\alpha^2\rangle - n_c^{-2}\left\langle\left(\sum_{\beta=1}^{n_c}X_\beta\right)^2\right\rangle.$$

(65)

In our case

$$\left\langle\left(U_{i_{1\to L}}^{(1\to L)}(\mu_1;\alpha) - \frac{1}{n_c}D_{i_{1\to L}}^{(1\to L)}(\mu_1)\right)^2\right\rangle = \left\langle\left(U_{i_{1\to L}}^{(1\to L)}(\mu_1;\alpha)\right)^2\right\rangle - n_c^{-2}\left\langle\left(D_{i_{1\to L}}^{(1\to L)}(\mu_1)\right)^2\right\rangle$$

$$= \sigma_{U^{(L)}}^2 - n_c^{-2}\sigma_{D^{(L)}}^2,$$

(66)

where, in the second line, we have used that $\langle U^{(L)}\rangle = \langle D^{(L)}\rangle/n_c$ to convert the difference of second moments into a difference of variances. By Eq. 41 and Eq. 58,

$$\sigma_{U^{(L)}}^2 - n_c^{-2}\sigma_{D^{(L)}}^2 = v\sigma_{U^{(L-1)}}^2\left(\sigma_N^2 - \sigma_{if}^2\right) + v\left\langle U^{(L-1)}\right\rangle^2\left(\sigma_N^2 + (v-1)\sigma_{if}^2\right)$$

$$- \frac{v}{n_c^2}\sigma_{D^{(L-1)}}^2\left(\sigma_N^2 - \sigma_{if}^2\right) - \frac{v}{n_c^2}\left\langle D^{(L-1)}\right\rangle^2\left(\sigma_N^2 + (v-1)\sigma_{if}^2\right)$$

$$= v\left(\sigma_N^2 - \sigma_{if}^2\right)\left(\sigma_{U^{(L-1)}}^2 - n_c^{-2}\sigma_{D^{(L-1)}}^2\right),$$

(67)

having used again that $\langle U^{(L)}\rangle = \langle D^{(L)}\rangle/n_c$. Iterating,

$$\sigma_{U^{(L)}}^2 - n_c^{-2}\sigma_{D^{(L)}}^2 = \left[v\left(\sigma_N^2 - \sigma_{if}^2\right)\right]^{L-1}\left(\left(\sigma_{U^{(1)}}^2 - n_c^{-2}\sigma_{D^{(1)}}^2\right)\right).$$

(68)

Since

$$\sigma^2_{U^{(1)}} = \frac{m}{v}\frac{v-1}{v}\frac{v^s-m}{v^s-1} \xrightarrow{v\gg 1} \frac{m}{v},$$

$$n_c^{-2}\sigma^2_{D^{(1)}} = n_c^{-1}\sigma_N^2 + n_c^{-1}(n_c-1)\sigma^2_{if} \xrightarrow{v\gg 1} n_c^{-1}\left(\frac{m}{v}\right)^2\left(\frac{v}{m}-\frac{n_c}{v^{s-1}}\right) = \frac{1}{n_c}\frac{m}{v}\left(1-\frac{mn_c}{v^s}\right), \tag{69}$$

One has

$$\sigma^2_{U^{(L)}} - n_c^{-2}\sigma^2_{D^{(L)}} \xrightarrow{v\gg 1} \frac{m^L}{v}\left(1-\frac{1-n_c v/v^s}{n_c}\right), \tag{70}$$

so that

$$\mathrm{Var}\left[f^{(1\to L)}_{i_{1\to L}}(\alpha|\mu_1)\right] = v^2\frac{\left\langle\left(U^{(1\to L)}_{i_{1\to L}}(\mu_1;\alpha)-\frac{1}{n_c}D^{(1\to L)}_{i_{1\to L}}(\mu_1)\right)^2\right\rangle}{n_c^2 m^{2L}} \xrightarrow{v, n_c \gg 1} \frac{v}{n_c}\frac{1}{n_c m^L}. \tag{71}$$

## B.2 Introducing sampling noise due to the finite training set

In a supervised learning setting where only $P$ of the total data are available, the occurrences $N$ are replaced with their empirical counterparts $\hat{N}$. In particular, the empirical joint occurrence $\hat{N}(\mu; \alpha)$[7] coincides with the number of successes when sampling $P$ points without replacement from a population of $P_{\max}$ where only $N(\mu; \alpha)$ belong to class $\alpha$ and display feature $\mu$ in position $j$. Thus, $\hat{N}(\mu; \alpha)$ obeys a hypergeometric distribution where $P$ plays the role of the number of trials, $P_{\max}$ the population size, and the true occurrence $N(\mu; \alpha)$ the number of favorable cases. If $P$ is large and $P_{\max}$, $N(\mu; \alpha)$ are both larger than $P$, then

$$\hat{N}(\mu; \alpha) \to \mathcal{N}\left(P\frac{N(\mu;\alpha)}{P_{\max}}, P\frac{N(\mu;\alpha)}{P_{\max}}\left(1-\frac{N(\mu;\alpha)}{P_{\max}}\right)\right), \tag{72}$$

where the convergence is meant as a convergence in probability and $\mathcal{N}(a,b)$ denotes a Gaussian distribution with mean $a$ and variance $b$. The statement above holds when the ratio $N(\mu; \alpha)/P_{\max}$ is away from 0 and 1, which is true with probability 1 for large $v$ due to the concentration of $f(\alpha|\mu)$. In complete analogy, the empirical occurrence $\hat{N}(\mu)$ obeys

$$\hat{N}(\mu) \to \mathcal{N}\left(P\frac{N(\mu)}{P_{\max}}, P\frac{N(\mu)}{P_{\max}}\left(1-\frac{N(\mu)}{P_{\max}}\right)\right). \tag{73}$$

We obtain the empirical conditional frequency by the ratio of Eq. 72 and Eq. 73. Since $N(\mu) = P_{\max}/v$ and $f(\alpha|\mu) = N(\mu; \alpha)/N(\mu)$, we have

$$\hat{f}(\alpha|\mu) = \frac{\frac{f(\alpha|\mu)}{v} + \xi_P\sqrt{\frac{1}{P}\frac{f(\alpha|\mu)}{v}\left(1-\frac{f(\alpha|\mu)}{v}\right)}}{\frac{1}{v} + \zeta_P\sqrt{\frac{1}{P}\frac{1}{v}\left(1-\frac{1}{v}\right)}}, \tag{74}$$

where $\xi_P$ and $\zeta_P$ are correlated zero-mean and unit-variance Gaussian random variables over independent drawings of the $P$ training points. By expanding the denominator of the right-hand side for large $P$ we get, after some algebra,

$$\hat{f}(\alpha|\mu) \simeq f(\alpha|\mu) + \xi_P\sqrt{\frac{vf(\alpha|\mu)}{P}\left(1-\frac{f(\alpha|\mu)}{v}\right)} - \zeta_P f(\alpha|\mu)\sqrt{\frac{v}{P}\left(1-\frac{1}{v}\right)}. \tag{75}$$

Recall that, in the limit of large $n_c$ and $m$, $f(\alpha|\mu) = n_c^{-1}(1+\sigma_f\xi_{\mathrm{RHM}})$ where $\xi_{\mathrm{RHM}}$ is a zero-mean and unit-variance Gaussian variable over the realizations of the RHM, while $\sigma_f$ is the 'signal', $\sigma_f^2 = v/m^L$ by Eq. 71. As a result,

$$\hat{f}(\alpha|\mu) \xrightarrow{n_c, m, P \gg 1} \frac{1}{n_c}\left(1+\sqrt{\frac{v}{m^L}}\xi_{\mathrm{RHM}} + \sqrt{\frac{vn_c}{P}}\xi_P\right). \tag{76}$$

---

[7]For ease of notation, we drop level and positional indices in this subsection.

### B.3 Sample complexity

From Eq. 76 it is clear that for the 'signal' $\hat{f}$, the fluctuations due to noise must be smaller than those due to the random choice of the composition rules. Therefore, the crossover takes place when the two nose terms have the same size, occurring at $P = P_c$ such that

$$\sqrt{\frac{v}{m^L}} = \sqrt{\frac{v n_c}{P_c}} \Rightarrow P_c = n_c m^L. \tag{77}$$

## C One-Step Gradient Descent (GD)

We will consider a simplified but tractable setting, where we generate an instance of the RHM and then train a one-hidden-layer fully-connected network only on the first $s$-dimensional patch of the input. Since there are many data having the same first $s$-dimensional patch but a different label, this network does not have the capacity to fit the data. Nevertheless, in the case where the $s$-dimensional patches are orthogonalized, neural networks can learn the synonymic invariance of the RHM if trained on at least $P_c$ data.

**GD on Cross-Entropy Loss.** More specifically, let us first sample an instance of the RHM, then $P$ input-label pairs $(\boldsymbol{x}_k, \alpha_k)$ with $\alpha_k := \alpha(\boldsymbol{x}_k)$ for all $k = 1, \ldots, P$. For any datum $\boldsymbol{x}$, we denote with $\boldsymbol{\mu}_1(\boldsymbol{x})$ the $s$-tuple of features in the first patch and with $\boldsymbol{\delta_\mu}$ the one-hot encoding of the $s$-tuple $\boldsymbol{\mu}$ (with dimension $v^s$). The fully-connected network acts on the one-hot encoding of the $s$-tuples with ReLU activations $\sigma(x) = \max(0, x)$,

$$\mathcal{F}_{\text{NN}}(\boldsymbol{\delta_\mu}) = \frac{1}{H} \sum_{h=1}^{H} a_h \sigma(\boldsymbol{w}_h \cdot \boldsymbol{\delta_\mu}), \tag{78}$$

where the inner-layer weights $\boldsymbol{w}_h$'s have the same dimension as $\boldsymbol{\delta_\mu}$ and the top-layer weights $a_h$'s are $n_c$-dimensional. The top-layer weights are initialized as i.i.d. Gaussian with zero mean and unit variance and fixed. The $\boldsymbol{w}_h$'s are trained by Gradient Descent (GD) on the cross-entropy loss,

$$\mathcal{L} = \hat{\mathbb{E}}_{\boldsymbol{\delta}} \left[ -\sum_{\beta=1}^{n_c} \delta_{\beta, \alpha(\boldsymbol{x})} \log \left( \frac{e^{(\mathcal{F}_{\text{NN}}(\boldsymbol{\delta}))_\beta}}{\sum_{\beta'=1}^{n_c} e^{(\mathcal{F}_{\text{NN}}(\boldsymbol{\delta}))_{\beta'}}} \right) \right], \tag{79}$$

where $\delta_{\beta, \alpha(\boldsymbol{x})}$ stems from the one-hot encoding of the class label $\alpha(\boldsymbol{x})$ and $\hat{\mathbb{E}}$ denotes expectation over the training set. For simplicity, we consider the mean-field limit $H \to \infty$, so that $\mathcal{F}_{\text{NN}}^{(0)} = 0$ identically, and initialize all the inner-layer weights to $\mathbf{1}$ (the vector with all elements set to 1)[8].

**Update of the Hidden Representation.** In this setting, with enough training points, one step of gradient descent is sufficient to build a representation invariant to the exchange of synonyms. Due to the one-hot encoding, $(\boldsymbol{w}_h \cdot \boldsymbol{\delta_\mu})$, namely the $h$-th component of the hidden representation of the $s$-tuple $\boldsymbol{\mu}$, coincides with the $\boldsymbol{\mu}$-th component of the weight $\boldsymbol{w}_h$. This component, which is set to 1 at initialization, is updated by (minus) the corresponding component of the gradient of the loss in Eq. 79. Recalling that at initialization the predictor is 0 and all the components of the inner-layer weights are 1, we get

$$-\nabla_{(\boldsymbol{w}_h)_{\boldsymbol{\mu}}} \mathcal{L} = \frac{1}{P} \sum_{\alpha=1}^{n_c} a_{h,\alpha} \left( \hat{N}_1(\boldsymbol{\mu}; \alpha) - \frac{1}{n_c} \hat{N}_1(\boldsymbol{\mu}) \right), \tag{80}$$

where $\hat{N}_1(\boldsymbol{\mu})$ is the empirical occurrence of the $s$-tuple $\boldsymbol{\mu}$ in the first patch of the $P$ training points and $\hat{N}_1(\boldsymbol{\mu}; \alpha)$ is the (empirical) joint occurrence of the $s$-tuple $\boldsymbol{\mu}$ and the class label $\alpha$. As $P$ increases, the empirical occurrences $\hat{N}$ converge to the true occurrences $N$, which are invariant for the exchange of synonym $s$-tuples $\boldsymbol{\mu}$. Hence, the hidden representation is also invariant for the exchange of synonym $s$-tuples in this limit.

---

[8]These two assumptions can be relaxed by extending the tools developed in [22].

## C.1 Extension to a one-hidden-layer CNN

The same argument can be carried out by considering a one-hidden-layer CNN with weight sharing and global average pooling:

$$\mathcal{F}_{\text{CNN}}(\boldsymbol{x}) = \frac{1}{Hs^{L-1}} \sum_{h=1}^{H} \sum_{j=1}^{s^{L-1}} \boldsymbol{a}_h \sigma(\boldsymbol{w}_h \cdot \boldsymbol{x}_j), \tag{81}$$

where we added an average over all input patches $j$. The gradient updates now read,

$$-\nabla_{(\boldsymbol{w}_h)_\mu} \mathcal{L} = \frac{1}{P} \sum_{\alpha=1}^{n_c} a_{h,\alpha} \frac{1}{s^{L-1}} \sum_{j=1}^{s^{L-1}} \left( \hat{N}_j(\boldsymbol{\mu}; \alpha) - \frac{1}{n_c} \hat{N}_j(\boldsymbol{\mu}) \right), \tag{82}$$

hence, synonymic invariance can now be inferred from the average occurrences over patches. This average results in a reduction of both the signal and noise term by the same factor $\sqrt{s^{L-1}}$. Consequently, analogously to the case without weight sharing, the hidden representation becomes insensitive to the exchange of synonymic features for $P \gg P_c = n_c m^L$.

# D Improved Sample Complexity via Clustering

In [Section 4.C](#) of the main text and [Appendix 4](#), we showed that the hidden representation of a one-hidden layer fully-connected network trained on the first patch of the RHM inputs becomes insensitive to exchanges of synonyms at $P = P^* = P_c = n_c m^L$. Here we consider the maximal dataset case $n_c = v$ and $m = v^{s-1}$, and show that a distance-based clustering method acting on these hidden representations would identify synonyms at $P \simeq \sqrt{n_c} m^L$, much smaller than $P_c$ in the large-$n_c$ limit.

Let us then imagine feeding the representations updates $\Delta f_h(\boldsymbol{\mu})$ of [Eq. 82](#) to a clustering algorithm aimed at identifying synonyms. This algorithm is based on the distance between the representations of different tuples of input features $\boldsymbol{\mu}$ and $\boldsymbol{\nu}$,

$$\|\Delta f(\boldsymbol{\mu}) - \Delta f(\boldsymbol{\nu})\|^2 := \frac{1}{H} \sum_{h=1}^{H} \left( \Delta f_h(\boldsymbol{\mu}) - \Delta f_h(\boldsymbol{\nu}) \right)^2, \tag{83}$$

where $H$ is the number of hidden neurons. By defining

$$\hat{g}_\alpha(\boldsymbol{\mu}) := \frac{\hat{N}_1(\boldsymbol{\mu}; \alpha)}{P} - \frac{1}{n_c} \frac{\hat{N}_1(\boldsymbol{\mu})}{P}, \tag{84}$$

and denoting with $\hat{\boldsymbol{g}}(\boldsymbol{\mu})$ the $n_c$-dimensional sequence having the $\hat{g}_\alpha$'s as components, we have

$$\|\Delta f(\boldsymbol{\mu}) - \Delta f(\boldsymbol{\nu})\|^2 = \sum_{\alpha,\beta=1}^{n_c} \left( \frac{1}{H} \sum_{h}^{H} a_{h,\alpha} a_{h,\beta} \right) (\hat{g}_\alpha(\boldsymbol{\mu}) - \hat{g}_\alpha(\boldsymbol{\nu})) (\hat{g}_\beta(\boldsymbol{\mu}) - \hat{g}_\beta(\boldsymbol{\nu}))$$

$$\xrightarrow{H \to \infty} \sum_{\alpha=1}^{n_c} (\hat{g}_\alpha(\boldsymbol{\mu}) - \hat{g}_\alpha(\boldsymbol{\nu}))^2 = \|\hat{\boldsymbol{g}}(\boldsymbol{\mu}) - \hat{\boldsymbol{g}}(\boldsymbol{\nu})\|^2, \tag{85}$$

where we used the i.i.d. Gaussian initialization of the readout weights to replace the sum over neurons with $\delta_{\alpha,\beta}$.

Due to the sampling noise, from [Eq. 72](#) and [Eq. 73](#), when $1 \ll P \ll P_{\max}$,

$$\hat{g}_\alpha(\boldsymbol{\mu}) = g_\alpha(\boldsymbol{\mu}) + \sqrt{\frac{1}{n_c m v P}} \eta_\alpha(\boldsymbol{\mu}), \tag{86}$$

where $\eta_\alpha(\boldsymbol{\mu})$ is a zero-mean and unit-variance Gaussian noise and $g$ without hat denotes the $P \to P_{\max}$ limit of $\hat{g}$. In the limit $1 \ll P \ll P_{\max}$, the noises with different $\alpha$ and $\boldsymbol{\mu}$ are independent of each other. Thus,

$$\|\hat{\boldsymbol{g}}(\boldsymbol{\mu}) - \hat{\boldsymbol{g}}(\boldsymbol{\nu})\|^2 = \|\boldsymbol{g}(\boldsymbol{\mu}) - \boldsymbol{g}(\boldsymbol{\nu})\|^2 + \frac{1}{n_c m v P} \|\boldsymbol{\eta}(\boldsymbol{\mu}) - \boldsymbol{\eta}(\boldsymbol{\nu})\|^2 + \frac{2}{\sqrt{n_c m v P}} \left( \boldsymbol{g}(\boldsymbol{\mu}) - \boldsymbol{g}(\boldsymbol{\nu}) \right) \cdot \left( \boldsymbol{\eta}(\boldsymbol{\mu}) - \boldsymbol{\eta}(\boldsymbol{\nu}) \right). \tag{87}$$

If $\boldsymbol{\mu}$ and $\boldsymbol{\nu}$ are synonyms, then $\boldsymbol{g}(\boldsymbol{\mu}) = \boldsymbol{g}(\boldsymbol{\nu})$ and only the noise term contributes to the right-hand side of Eq. 87. If this noise is sufficiently small, then the distance above can be used to cluster tuples into synonymic groups.

By the independence of the noises and the Central Limit Theorem, for $n_c \gg 1$,

$$\|\boldsymbol{\eta}(\boldsymbol{\mu}) - \boldsymbol{\eta}(\boldsymbol{\nu})\|^2 \sim \mathcal{N}(2n_c, \mathcal{O}(\sqrt{n_c})), \tag{88}$$

over independent samplings of the $P$ training points. The $g$'s are also random variables over independent realizations of the RHM with zero mean and variance proportional to the variance of the conditional probabilities $f(\alpha|\boldsymbol{\mu})$ (see Eq. 62 and Eq. 71),

$$\text{Var}\left[g_\alpha(\boldsymbol{\mu})\right] = \frac{1}{n_c m v n_c m^L} = \frac{1}{n_c m v P_c}. \tag{89}$$

To estimate the size of $\|\boldsymbol{g}(\boldsymbol{\mu}) - \boldsymbol{g}(\boldsymbol{\nu})\|^2$ we must take into account the correlations (over RHM realizations) between $g$'s with different class label and tuples. However, in the maximal dataset case $n_c = v$ and $m = v^{s-1}$, both the sum over classes and the sum over tuples of input features of the joint occurrences $N(\boldsymbol{\mu}; \alpha)$ are fixed deterministically. The constraints on the sums allow us to control the correlations between occurrences of the same tuple within different classes and of different tuples within the same class, so that the size of the term $\|\boldsymbol{g}(\boldsymbol{\mu}) - \boldsymbol{g}(\boldsymbol{\nu})\|^2$ for $n_c = v \gg 1$ can be estimated via the Central Limit Theorem:

$$\|\boldsymbol{g}(\boldsymbol{\mu}) - \boldsymbol{g}(\boldsymbol{\nu})\|^2 \sim \mathcal{N}\left(\frac{2n_c}{n_c m v P_c}, \frac{\mathcal{O}(\sqrt{n_c})}{n_c m v P_c}\right). \tag{90}$$

The mixed term $(\boldsymbol{g}(\boldsymbol{\mu}) - \boldsymbol{g}(\boldsymbol{\nu})) \cdot (\boldsymbol{\eta}(\boldsymbol{\mu}) - \boldsymbol{\eta}(\boldsymbol{\nu}))$ has zero average (both with respect to training set sampling and RHM realizations) and can also be shown to lead to relative fluctuations of order $\mathcal{O}(\sqrt{n_c})$ in the maximal dataset case.

Tu sum up, we have that, for synonyms,

$$\|\hat{\boldsymbol{g}}(\boldsymbol{\mu}) - \hat{\boldsymbol{g}}(\boldsymbol{\nu})\|^2 = \|\boldsymbol{\eta}(\boldsymbol{\mu}) - \boldsymbol{\eta}(\boldsymbol{\nu})\|^2 \sim \frac{1}{m v P}\left(1 + \frac{1}{\sqrt{n_c}}\xi_P\right), \tag{91}$$

where $\xi_P$ is some $\mathcal{O}(1)$ noise dependent on the training set sampling. If $\boldsymbol{\mu}$ and $\boldsymbol{\nu}$ are not synonyms, instead,

$$\|\hat{\boldsymbol{g}}(\boldsymbol{\mu}) - \hat{\boldsymbol{g}}(\boldsymbol{\nu})\|^2 \sim \frac{1}{m v P}\left(1 + \frac{1}{\sqrt{n_c}}\xi_P\right) + \frac{1}{m v P_c}\left(1 + \frac{1}{\sqrt{n_c}}\xi_{\text{RHM}}\right), \tag{92}$$

where $\xi_{\text{RHM}}$ is some $\mathcal{O}(1)$ noise dependent on the RHM realization. In this setting, the signal is the deterministic part of the difference between representations of non-synonymic tuples. Due to the sum over class labels, the signal is scaled up by a factor $n_c$, whereas the fluctuations (stemming from both sampling and model) are only increased by $\mathcal{O}\left(\sqrt{n_c}\right)$. Therefore, the signal required for clustering emerges from the sampling noise at $P = P_c/\sqrt{n_c} = \sqrt{n_c}\,m^L$, equal to $v^{1/2 + L(s-1)}$ in the maximal dataset case. This prediction is tested for $s = 2$ in Fig. 12, which shows the error achieved by a layerwise algorithm which alternates single GD steps to clustering of the resulting representations [22, 27]. More specifically, the weights of the first hidden layer are updated with a single GD step while keeping all the other weights frozen. The resulting representations are then clustered, so as to identify groups of synonymic level-1 tuples. The centroids of the ensuing clusters, which correspond to level-2 features, are orthogonalized and used as inputs of another one-step GD protocol, which aims at identifying synonymic tuples of level-2 features. The procedure is iterated $L$ times.

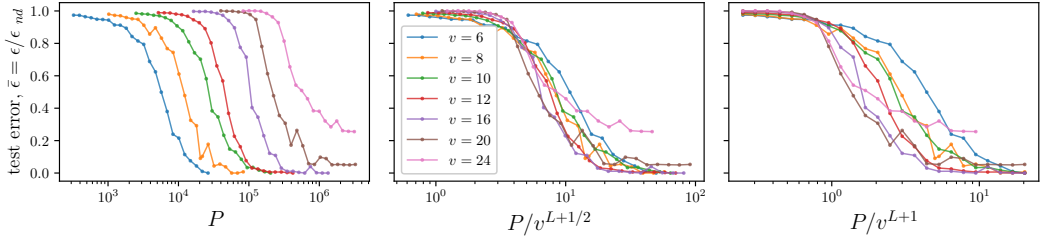

Figure 12: **Sample complexity for layerwise training,** $m = n_c = v$, $L = 3$, $s = 2$. Training of a $L$-layers network is performed layerwise by alternating one-step GD as described in Section 4.C and clustering of the hidden representations. Clustering of the $mv = v^2$ representations for the different one-hot-encoded input patches is performed with the $k$-means algorithms. Clustered representations are then orthogonalized and the result is given to the next one-step GD procedure. Left: Test error vs. number of training points. Different colors correspond to different values of $v$. Center: collapse of the test error curves when rescaling the $x$-axis by $v^{L+1/2}$. Right: analogous, when rescaling the $x$-axis by $v^{L+1}$. The curves show a better collapse when rescaling by $v^{L+1/2}$, suggesting that these layerwise algorithms as an advantage of a factor $\sqrt{v}$ over end-to-end training with deep CNNs, for which $P^* = v^{L+1}$.

# E  Instances of the Homogeneous Feature Model (HFM) in the Random Hierarchy Model (RHM)

Given that the rules in the RHM are chosen uniformly at random, a non-zero probability exists that an HFM, where no input-output correlations exist, is sampled as an instance of the RHM. In these instances, semantic invariance, and good generalization, would be impossible to learn from correlations, as illustrated in the main text. In this appendix we show that such specific instances of the RHM are sampled with vanishing probability, for increasing values of the RHM parameters.

## E.1  $m = v^{s-1}$ case

The number of rules for the RHM made by $L$ layers for generic values of $m$ and $v$ is given by

$$\left[ \frac{(v^s)!}{(m!)^v (v^s - vm)!} \right]^L \left( \frac{1}{v!} \right)^{L-1}. \tag{93}$$

For $m = v^{s-1}$ it becomes $\left( \frac{v^s!}{v^{s-1}!)^v (v^s - v \cdot v^{s-1})!} \right)^L \left( \frac{1}{v!} \right)^{L-1}$. The number of HFM rules, defined by having $N_i(\mu; \alpha) = v^{s-2}$ independently of $\mu$ in each of the single-layer rules, can be computed as follows. Let's look at a given feature of the previous layer, for example the symbol 1. We want to assign to it $m = v^{s-1}$ $s$−tuples

$$(1, \alpha_1^{1,1}, ..., \alpha_1^{1,s-1}), ..., (1, \alpha_{v^{s-2}}^{1,1}, ..., \alpha_{v^{s-2}}^{1,s-1}), ..., (v, \alpha_{v^{s-1}-(v^{s-2}-1)}^{1,1}, ..., \alpha_{v^{s-1}-(v^{s-2}-1)}^{1,s-1}), ..., (v, \alpha_{v^{s-1}}^{1,1}, ..., \alpha_{v^{s-1}}^{1,s-1}), \tag{94}$$

where the first $v^{s-2}$ tuples has as first symbol 1, the second $v^{s-2}$ the symbol 2 and we continue up to the last $v^{s-2}$ tuples with first symbol $v$. The numbers $\{\alpha_i^{1,t}\}_{i=1,...,v^{s-1}}$ are permutations of the set $\{1, ..., v\}^{v^{s-2}}$ for feature 1 and location $t$. In these rules, there is no symbol that occurs more than the others at a given location. Consequently, the network cannot exploit any correlation between the presence of a symbol at a given location and the label to solve the task. With regard to the other features $j$ of the previous layer, to any of these we will assign the $v^{s-1}$ tuples $(1, \alpha_1^{j,1}, ..., \alpha_1^{j,s-1}), ..., (v, \alpha_{v^{s-1}}^{j,1}, ..., \alpha_{v^{s-1}}^{j,s-1})$, with the numbers $\{\alpha_i^{j,t}\}$ being the same of $\alpha_i^{1,t}$ for any $t$ but shifted forward of $(j-1)v$ positions. For example, the numbers related to the "block" of tuples with first element 1 for feature 1 will be the same related to the "block" of tuples with first element 2 for feature. In formulae: $\alpha_i^{j,t} = \alpha_{i-(j-1)v}^{1,t}$, with $i - (j-1)v$ being equivalent to $(v^{s-1} - (v-i) + 1)$ (periodic boundary conditions). The number of such uncorrelated rules is just the number $(v^{s-1}!)$ of permutations of the numbers $\{\alpha_i^{1,t}\}_{i=1,...,v^{s-1}}$ for a fixed tuple location $t$, elevated to the number of positions $s-1$. Consequently, the fraction of uncorrelated rules for $L$ layers is:

$$f_{\text{HFM}} = \frac{1}{v!} \left( \frac{(v^{s-1}!)^{(s-1)}}{(v^{v^{s-1}\cdots v^{s-1}})^{\frac{1}{v!}}} \right)^L. \tag{95}$$

We now want to show that $f_{\text{HFM}}$ vanishes for large $v$. We implement the Stirling approximation in (95) getting

$$f_{\text{HFM}} \approx \frac{1}{v!} \left( \frac{v^{(s-1)(v^{s-1}+\frac{1}{2})} e^{-v^{s-1}(s-1)}}{v^{s(v^s+\frac{1}{2})} e^{-v^s}} v^{v(s-1)(v^{s-1}+\frac{1}{2})} e^{-v^s} \right)^L, \tag{96}$$

yielding the following limit behavior for large $v$:

$$f_{\text{HFM}} \approx \frac{1}{v!} \left( \frac{e^{-v^{s-1}(s-1)}}{v^{v^s}} \right)^L, \tag{97}$$

which is vanishing for large $v$ and large $L$.

## E.2 Generic $m$ case

Let's characterize the uncorrelated rules in the case of generic $m$. For each single-layer rule, we assign $m$ $s-$tuples to each symbol of the previous layer. Let's consider a given symbol $j$. To this symbol, we assign $m$ $s-$tuples of the type $(\alpha_1^{j,1},...,\alpha_1^{j,s}),(\alpha_2^{j,1},...,\alpha_2^{j,s}),...,(\alpha_m^{j,1},...,\alpha_m^{j,s})$, with the $m$ numbers $(\alpha_i^{j,t})_i$ at fixed location $t$ being a permutation of a subset of $\{1,...,v\}^{v^{s-2}}$ such that, if we call $m_q$ the number of items in the subset equal to $q \in \{1,...,v\}$, each $m_q$ is either 0 or we have that $m_{q_1} = m_{q_2}$ for $q_1$ and $q_1$ such that $m_{q_1} = m_{q_2} > 0$ and $\sum_{q=1}^{v} m_q = m$. Moreover, since each symbol $q$ can appear at most $v^{s-2}$ times, we have that $m_q \leq v^{s-2}$. We take the $m$ numbers $(\alpha_i^{j,1})_{i=1,...,m}$ at the first location $t=1$ ordered in increasing order. Note that these tuples can be picked just once across different symbols $j$, imposing then constraints on the numbers $\alpha_i^{j,t}$ for different features $j$.

As in the case of $m = v^{s-1}$ in Sec. E.1, we want to show that the probability of occurrence $f_{\text{HFM}}$ of such uncorrelated rules, given by the number of these rules divided by the number of total rules, is vanishing for large $v$ and/or large $L$. To count the number of uncorrelated rules, that we call $\#_{\text{HFM}}$, we first count the number $\#_{j,t}$ of possible series of numbers $\{\alpha_i^{j,t}\}_{i=1,...,m}$ for a fixed feature $j$ and position $t > 1$, and for a single-layer rule. In other words, we have to count the number of possible subsets made by $m$ elements of $\{1,...,v\}^{v^{s-2}}$ such that each symbol $q \in \{1,...,v\}$ appears $m_q$ times, with the $m_q$ satisfying the constraints above. We introduce the quantity $v_0$ which is the number of symbols $q$ which appears 0 times in a given subset. Once we fix $v_0$, from the constraint $\sum_{q=1}^{v} m_q = m$ we get that the features with $m_q > 0$ appear $\bar{m} = \frac{m}{v-v_0}$ if $\frac{m}{v-v_0}$ is a positive integer, otherwise, there are no subsets with that $v_0$. Consequently, the number $\#_{j,t}$ is given by:

$$\#_{j,t} = \sum_{v_0=0}^{v-1} \binom{v}{v_0} \mathbb{I}\left[ \frac{m}{v-v_0} \in \mathbb{N}_{>0} \right] \mathbb{I}\left[ \frac{m}{v-v_0} \leq v^{s-2} \right] m!, \tag{98}$$

where $(i)$ $\binom{v}{v_0}$ counts the number of choices of the features with 0 appearances and $(ii)$ $m!$ counts the number of permutations of the $m$ numbers $\{\alpha_i^{j,t}\}_i$. Since we are interested in proving that $f_{\text{HFM}}$ is vanishing for large $v$ and $L$, we upper bound it relaxing the constraint that $\frac{m}{v-v_0} \in \mathbb{N}_{>0}$ and using that $\binom{v}{v_0} \leq \binom{v}{\lceil v/2 \rceil}$:

$$\#_{j,t} \leq \left( v - \frac{m}{v^{s-2}} \right) \binom{v}{\lceil v/2 \rceil} m! \tag{99}$$

Considering all the $s$ locations, we get

$$\#_j \leq \left( v - \frac{m}{v^{s-2}} \right)^s \binom{v}{\lceil v/2 \rceil}^s (m!)^{s-1}, \tag{100}$$

where $\#_j$ is defined similarly as $\#_{f,t}$. Notice that for the first location $t = 1$ there is not a factor $m!$ since we are ordering the numbers $\alpha_i^{j,1}$ in ascending order there.

If we want to count $\#_{\mathrm{HFM}}$, we have to take into account that we are sampling without replacement from the space of $s-$tuples, and hence two different symbols $j$ cannot share the same $s-$tuples, hence increasing the number of possible rules. To upper bound $\#_{\mathrm{HFM}}$, we relax this constraint, hence sampling the tuples with replacement. Consequently, we have:

$$\#_{\mathrm{HFM}} \leq \left[ \left( v - \frac{m}{v^{s-2}} \right)^s \binom{v}{\lceil v/2 \rceil}^s (m!)^{s-1} \right]^v, \tag{101}$$

since the choice of the $\alpha_i^{j,t}$ is independent between different features $j$. Consequently for a $L-$layer rule:

$$f_{\mathrm{HFM}} \leq \left[ \left( v - \frac{m}{v^{s-2}} \right)^s \binom{v}{\lceil v/2 \rceil}^s (m!)^{s-1} \right]^{vL} \Bigg/ \left[ \binom{v^s}{m \ldots m}^L \left( \frac{1}{v!} \right)^{L-1} \right] \tag{102}$$

We now assume $m \sim v^\alpha$, with $0 < \alpha < (s-1)$ and for large $n$ we implement the Stirling approximation[9], getting:

$$f_{\mathrm{HFM}} \leq \frac{1}{v!} \left( v - v^{\alpha-(s-2)} \right)^{svL} 2^{(v+1)vL} v^{\alpha(v^\alpha+1/2)(s-1)vL+vL/2} e^{-v^\alpha(s-1)vL - vL}$$
$$\Bigg/ v^{L(v^s+v/2-sv/2+s/2)} e^{-\frac{L}{2v^{s-1}} \left( v(v^\alpha - v^{s-1}) + (v^s - v^{1+\alpha} - v^{s-1}) \right)} \tag{103}$$

where we approximated $\lceil v/2 \rceil$ with $v/2$. At the leading order for large $v$ we get, using the fact that $(\alpha+1) < s$

$$f_{\mathrm{HFM}} \leq \frac{1}{v!} \frac{e^{-(s-1)Lv^{\alpha+1}}}{v^{(v^s)L}} 2^{(v+1)vL}, \tag{104}$$

hence the probability of occurrence of a parity-like rule is vanishing for large $n$ and $L$.

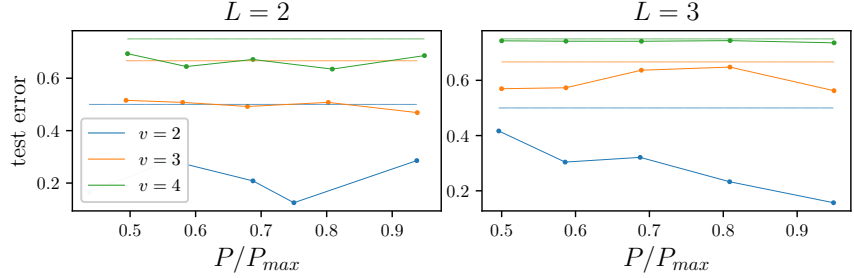

Figure 13: Test error of deep CNNs on the Homogeneous Feature Model for different $v$ and $L$. Horizontal dashed lines stand for the test error $(v-1)/v$ given by guessing labels uniformly at random. For $v = 2$ the networks can generalize to a number of training points $P$ which scales with the total size of the dataset $P_{max}$. Increasing $v$, performance is very close to chance already at $v = 4$.

## F  Intrinsic Dimensionality of Data Representations

In deep learning, the representation of data at each layer of a network can be thought of as lying on a manifold in the layer's activation space. Measures of the *intrinsic dimensionality* of these manifolds can provide insights into how the networks lower the dimensionality of the problem layer by layer [14, 15]. However, such measurements have challenges. One key challenge is that it assumes that real data exist on a smooth manifold, while in practice, the dimensionality is estimated based on a discrete set of points. This leads to counter-intuitive results such as an increase in the intrinsic dimensionality with depth, especially near the input. An effect that is impossible for continuous smooth manifolds. We resort to an example to illustrate how this increase with depth can result from spurious effects. Consider a manifold of a given intrinsic dimension that undergoes a transformation where one of the coordinates is multiplied by a large factor. This operation would result in an elongated manifold that appears one-dimensional. The measured intrinsic dimensionality would consequently be one, despite the higher dimensionality of the manifold. In the context of neural networks, a network that operates on such an elongated manifold could effectively 'reduce' this extra, spurious dimension. This could result in an increase in the observed intrinsic dimensionality as a function of network depth, even though the actual dimensionality of the manifold did not change.

---

[9]The Stirling approximation for the multinomial $\binom{n}{a_1 \ldots a_k}$ for $n \to \infty$ and integers $a_i$ such that $\sum_{i=1}^k = n$ is given by $\binom{n}{a_1 \ldots a_k} \sim (2\pi n)^{(1/2-k/2)} k^{n+k/2} \exp\{ -\frac{k}{2n} \sum_{i=1}^k (a_i - n/k)^2 \}$. In our case $n = v^s$, $k = (v+1)$ and $a_i = m$ for $i \in 1, \ldots, v$ and $a_{k+1} = (v^s - vm)$.

In the specific case of our data, the intrinsic dimensionality of the internal representations of deep CNNs monotonically decreases with depth, see Fig. 14, consistently with the idea proposed in the main text that the CNNs solve the problem by reducing the effective dimensionality of data layer by layer. We attribute this monotonicity to the absence of spurious or noisy directions that might lead to the counter-intuitive effect described above.

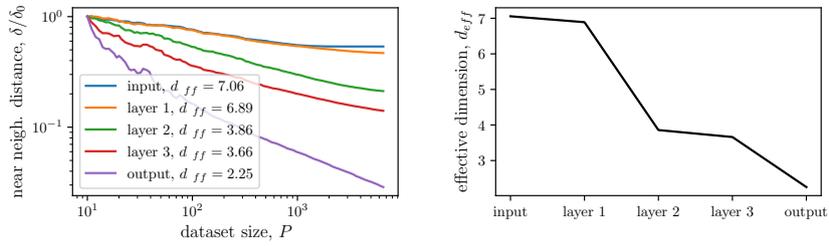

Figure 14: Effective dimension of the internal representation of a CNN trained on one instance of the RHM with $m = n_c = v$, $L = 3$ resulting in $P_{max} = 6'232$. Left: average nearest neighbor distance of input or network activations when probing them with a dataset of size $P$. The value reported on the $y$-axis is normalized by $\delta_0 = \delta(P = 10)$. The slope of $\delta(P)$ is used as an estimate of the effective dimension. Right: effective dimension as a function of depth. We observe a monotonic decrease, consistent with the idea that the dimensionality of the problem is reduced by DNNs with depth.

## G  Additional Figures

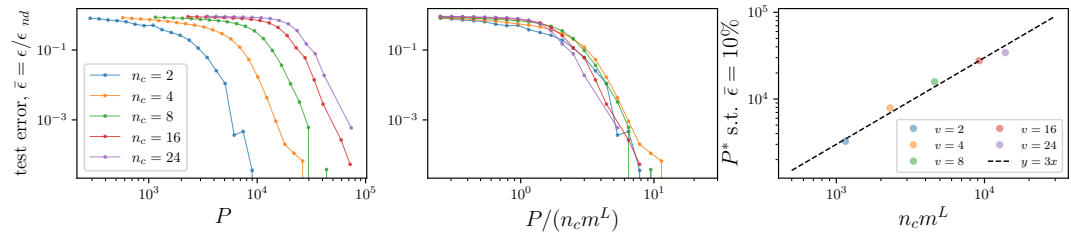

Figure 15: **Sample complexity for deep CNNs, $m = v = 24$, $L = s = 2$ and different values of $n_c$.** Left: Test error vs. number of training points. Different colors correspond to different numbers of classes $n_c$. Center: collapse of the test error curves when rescaling the $x$-axis by $n_c m^L$. Right: sample complexity $P^*$ corresponding to a test error $\epsilon^* = 0.1$.

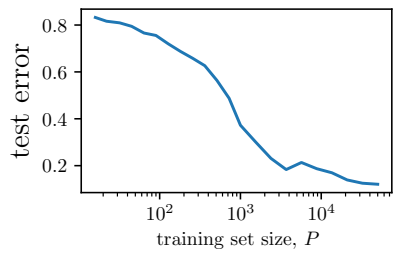

Figure 16: Test error vs. number of training points for a ResNet18 [44] trained on subsamples of the CIFAR10 dataset. Results are the average of 10 jointly different initializations of the networks and dataset sampling.

# Conclusions and Perspectives Part IV

# 7 Conclusion

The central takeaway from this thesis is that the power of deep learning indeed resides in its ability to adapt to the structure of the data. In particular, we put forward the hypothesis that the curse of dimensionality——a major challenge in processing high-dimensional data——can be substantially mitigated by learning representations that are *invariant* to aspects of the data that are irrelevant for the task, extending the ideas of Bruna and Mallat (2013); Mallat (2016) to other invariances beyond smooth deformations and image tasks. Our findings demonstrate that neural networks are able to learn such representations, provided they have the right architecture and are trained in the feature learning regime. Crucially, we have shown that this learning of invariances is fundamental for achieving good performance.

The first section (7.1) of this concluding chapter presents a summarizing table that encapsulates our primary results, systematically categorizing the invariances, neural network architectures, and training regimes that were examined. One of the central aspects of our research is the introduction of empirical tools to characterize invariance learning in neural networks in order to connect it to performance. The second section (7.2) of this conclusion elaborates on these tools, providing a unified formulation for all the invariances we studied, and placing these tools in context with existing literature. In section 7.3 we discuss more in detail the results encapsulated in the table and the key lessons drawn from them; section 7.4 provides the big picture that emerges from our work. We close with the limitations of our work and the ongoing and future investigations that our findings open up to (section 7.5).

## 7.1 Table of Results

We provide a compact summary of our results in Table 7.1. On the rows, one finds the invariances we considered:

1. **Linear Invariance.** This invariance could model the pixels at the corner of an image,

| Arch. + regime<br><br>Data Structure | shallow FCN<br>(lazy) | | shallow FCN<br>(feature) | | deep FCN<br>(feature) | | deep CNN<br>(feature) | |
|---|---|---|---|---|---|---|---|---|
| 1. Linear Inv. 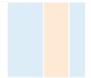 |  Algo: GF<br>$\beta = \dfrac{d}{3d-2} \xrightarrow{d \gg 1} \dfrac{1}{3}$ | |  Algo: GF<br>$\beta = \dfrac{3d-d'}{2(3d-2)} \xrightarrow{d \gg 1} \dfrac{1}{2}$ | | | | | |
| 2. Rotation Inv. 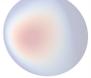 |  Algo: GD<br>$\beta = \dfrac{2d}{d-1} \xrightarrow{d \gg 1} 2$ | |  Algo: GD<br>$\beta = \dfrac{d+2}{d-1} \xrightarrow{d \gg 1} 1$ | | | | | |
| 3. Deformation Inv. <br>(image classification) 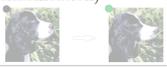 |  Algo: SVM<br>$R_f \approx 1$ | |  Algo: GF<br>$R_f > 1$ | |  Algo: GD<br>$R_f > 1$ | |  Algo: SGD<br>$R_f < 1$ | |
| 4. Deformation Inv. <br>(scale detection) 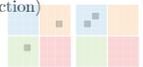 | | | | | | | 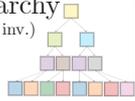 Algo: SGD<br>$R_f < 1$ | |
| 5. Hierarchy <br>(synonymic inv.)  |  Algo: SVM<br>$P^* \propto \exp(d)$ | |  Algo: SGD<br>$P^* \propto \exp(d)$ | | Algo: SGD<br>$\mathrm{poly}(d) \leq P^* \ll \exp(d)$<br>(preliminary results) | |  Algo: SGD<br>$P^* \propto \mathrm{poly}(d)$ | |

**Table 7.1: Summary of Results.** The rows represent the various data structures considered, categorized by data invariances, while the columns detail the types of neural networks and training regimes studied, including 2-layer neural networks in both the lazy and feature regimes, deep fully-connected networks, and deep convolutional networks. Gray cells indicate configurations that were not expressly addressed in our research. For each invariance, symbols provide a ranking of the performance of the architecture and training regime we investigated (green: best, orange: intermediate, red: worst).

The performance is more precisely reported by the predicted scaling exponent $\beta$, when the test error follows a power law, or by the sample complexity $P^*$. When predictions of test error are not available (i.e. for deformation invariance), we indicate the qualitative value of relative sensitivity $R_f$, which correlates closely with test error.

The algorithms used in each setting are also indicated, namely Support Vector Machine SVM, (Stochastic) Gradient Descent (S)GD, and Gradient Flow GF, that is GD with adaptive learning rate as defined in Geiger et al. (2020b).

likely uncorrelated to the task. It generically includes target functions of the form:

$$f^*(\boldsymbol{x}) = g(A\boldsymbol{x}) \quad \text{where} \quad A : \mathbb{R}^d \to \mathbb{R}^{d'} \quad \text{and} \quad d' \ll d. \tag{7.1}$$

2. **Rotation Invariance.** This invariance is introduced as a simple model of non-linear invariance. In this setting, data points $\boldsymbol{x}_i$ are sampled uniformly at random from the $d$-dimensional unit sphere $\mathbb{S}^{d-1}$. The target $f^*$ is a Gaussian random function of controlled smoothness:

$$\mathbb{E}\|f^*(\boldsymbol{x}) - f^*(\boldsymbol{z})\|^2 = O\left(\|\boldsymbol{x} - \boldsymbol{z}\|^{2\nu_t}\right), \quad \text{as} \quad \boldsymbol{x} \to \boldsymbol{z}, \tag{7.2}$$

where the exponent $\nu_t$ controls the smoothness of $f^*$ and hence its stability with respect to rotations of the input. For $\nu_t \to \infty$, $f^*$ is the constant function that is invariant to rotations. The results reported in the second row of Table 7.1 are valid for large $\nu_t$ (cf. chapter 4, Equation 3.4).

3. **Deformation Invariance in Images.** Given $\tau$ an operator that applies a small deformation to 2D images, we have that the target function defining the class of an image $\boldsymbol{x}$ satisfies

$$f^*(\boldsymbol{x}) \approx f^*(\tau\boldsymbol{x}), \tag{7.3}$$

in the sense that $\|f^*(\boldsymbol{x}) - f^*(\tau\boldsymbol{x})\|$ is small, if the norm of $\tau$ is small.

4. **Scale-detection tasks.** We've designed artificial tasks as a model for scenarios where deformation invariance plays a role to better characterize it. These include models of 2D images where only two pixels $i$ and $j$ are active, and the target is given by

$$f^*(\boldsymbol{x}) = \text{sign}(\|i - j\| - \xi). \tag{7.4}$$

The task hence consists of classifying whether the two pixels are closer or not than a given distance or *scale* $\xi$ and moving one active pixel to a neighboring position is unlikely to affect the class.

5. **Synonymic Invariance of Hierarchical Tasks.** This invariance is related to hierarchically compositional and local tasks. Consider, for instance:

$$f^*(\boldsymbol{x}) = g_3(g_2(g_1(x_1, x_2), g_1(x_3, x_4)), g_2(g_1(x_5, x_6), g_1(x_7, x_8))), \tag{7.5}$$

with $\boldsymbol{x} = (x_1, \ldots, x_8)$—see also graphical representation in Figure 7.1. Here, we have a function $f^*$ of a $d$-dimensional input ($d = 8$), written as a hierarchy of $L = 3$ levels of constituent functions. Each of the constituent functions is local in the sense that it only depends on a number of variables $s \ll d$, with $s = 2$ in this case. Synonymic invariance stems from the fact that, for each constituent function, different tuples e.g. $(x_1, x_2), (x_1', x_2')$ with identical meaning may exist, i.e. $g_{11}(x_1, x_2) = g_{11}(x_1', x_2')$. We call these tuples *synonyms*.

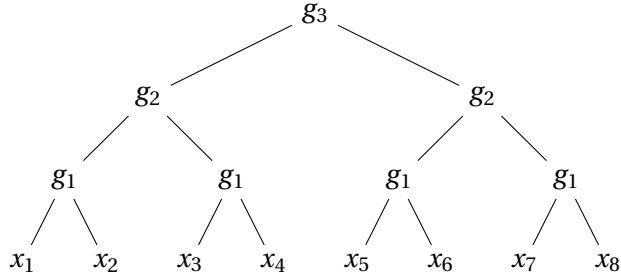

Figure 7.1: **Graphical Representation of the Hierarchically Compositional and Local Function of Equation 7.5.** The tree represents the structure of the hierarchical function with $L = 3$ layers, whose constituent functions ($g$'s) are local on patches of size $s = 2$. The leaves correspond to input variables.

In the model of the hierarchical data we introduced in chapter 6, each input takes values from a finite vocabulary $V$, with $v = |V|$, and the constituents $g_l : V^s \to V$ are surjective functions that map sets of $m$ inputs of the domain to each one of the $v$ outputs. The functions $g_l$ are chosen by drawing the sets of $m$ synonyms uniformly at random from all the possible $s$-tuples of inputs.

Notice that the results reported in the fifth row of Table 7.1, are valid when the input distribution has full support, in the sense that all possible $v^d$ input data are generated— this is the case when $m$ is large. If only a fraction of input data was generated ($m = O(1)$), then even kernel methods could beat the curse of dimensionality, as they are known to be adaptive to a low-dimensional input support Bach (2022).

## 7.2 Empirical Characterization of Invariants Learning

Observables to measure the invariance of neural networks to input transformations that leave the label unchanged are crucial for the findings of this thesis, as they are needed to quantitatively link invariance learning and performance.

Our main contribution in this context consists in proposing to characterize the learning of invariances through the relative measure of the sensitivity of the network activations with respect to input transformations that the target function is invariant to. Relative sensitivities generally take the form,

$$S_f = \frac{\mathbb{E}_{\boldsymbol{x},T}\|f(\boldsymbol{x}) - f(T\boldsymbol{x})\|^2}{\mathbb{E}_{\boldsymbol{x},G}\|f(\boldsymbol{x}) - f(G\boldsymbol{x})\|^2}, \tag{7.6}$$

where $f$ is the neural network output or internal activations, the $\boldsymbol{x}$'s are test samples, $T$ is an operator applying the invariant transformation to the inputs, and $G$ is an operator applying a *generic* transformation of the same magnitude, i.e. $\mathbb{E}_{\boldsymbol{x},T}\|Tx - x\| = \mathbb{E}_{\boldsymbol{x},G}\|Gx - x\|$.

This definition is such that $S_f \approx 1$ if the function $f$ does not distinguish between the invariant transformation $T$ and a generic one $G$, otherwise $S_f \ll 1$. Such measures of sensitivity have proven to be successful to characterize the learning of invariance in different contexts, and are shown to have a remarkable correlation with performance.

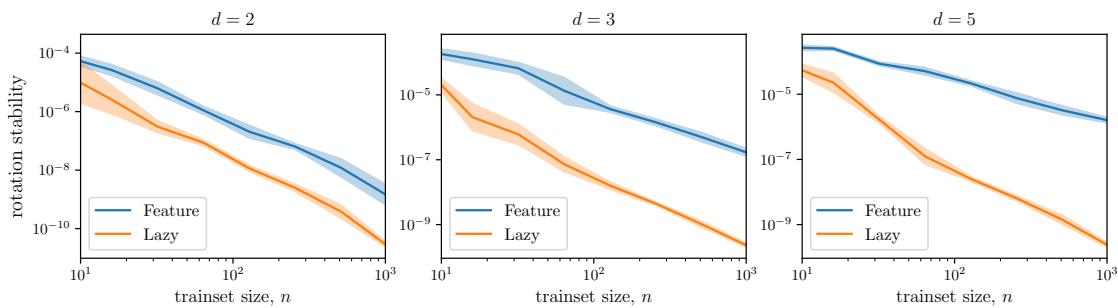

Figure 7.2: Rotation sensitivity of the predictor when learning the constant function on the sphere (cf. chapter 4), in the feature and lazy regime, for a varying number of training points. Results for different input-space dimensions are reported in each column. To see the correlation between sensitivity and performance, these curves are to be compared with Figure 4 in chapter 4.

Our sensitivity measurements have drawn inspiration from the definition of deformation stability Bruna and Mallat (2013); Mallat (2016). Yet, as found in chapter 3, measures of relative stability more effectively correlate with performance in practical settings. Furthermore, we adapt these observables to encompass other invariances beyond just diffeomorphisms. While Goodfellow et al. (2009) has also proposed methods to assess the invariance of deep networks to various input changes, their approach is based on the assumption that individual neurons are specialized for specific invariances. This assumption is challenged by the empirical evidence in Szegedy et al. (2014), which indicates that the entire activation space, rather than individual neurons/coordinates, carries semantic information about the task.

We summarize here the invariant and generic transformations specific to each invariance and task we considered in this thesis, and our main results on relative sensitivity measurements.

1. For **linear invariance**, $T$ corresponds to Gaussian noise added to the $d - d'$ irrelevant directions, and $G$ to Gaussian noise added to all directions. We have shown that this invariance can be learned by 2-layers neural networks in the feature regime both in anisotropic targets (chapter 3, Figure 7) and image classification tasks (chapter 4, Figure H.1), where linear invariance is associated with pixels at the boundary of images.

2. For **rotation invariance**, $T$ corresponds to a small rotation of $\boldsymbol{x}$ over the input coordinates and $G$ to Gaussian noise. We report results for the sensitivity of feature and lazy predictor when regressing the constant function on the $d$-sphere in Figure 7.2. The separation in sensitivity to rotations between the two regimes is consistent with the one of test error as reported in chapter 4.

4./5. For **deformation invariance** the transformation $T$ correspond to a small deformation. To generate deformations of controlled norm we introduced an ensemble of maximum-

original        deformed

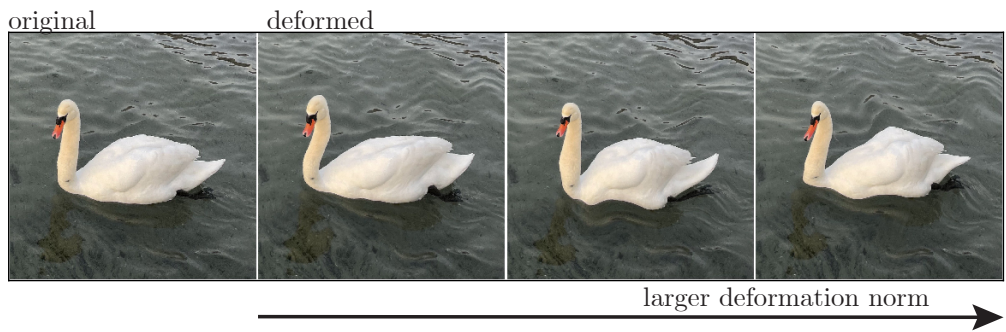

larger deformation norm

Figure 7.3: Image of a swan deformed with the ensemble of max-entropy diffeomorphisms we introduced in chapter 3. Starting from the original image on the left, deformations of increasingly larger magnitudes are applied while going to the right.

entropy diffeomorphisms that allows us to uniformly sample from the distribution of all diffeomorphisms having the same norm. Samples from this distribution as a function the deformation norm are reported in Figure 7.3. In this case, $G$ corresponds to isotropic additive noise. This invariance can be profitably learned by deep CNNs (chapter 3 and chapter 5), but even deteriorates when learning features in shallow architectures (chapter 4).

6. For **synonymic invariance**, $T$ corresponds to the exchange of synonymous features at a given level of the hierarchy, $G$ to the substitution of a feature at the same level with a different one taken at random. This invariance can be learned by deep but not shallow networks (chapter 6 and Figure 7.6).

Overall, our findings suggest that, if a certain invariance is relevant for the task, then $S_f$ of a given network predictor correlates with its test error. In deep networks, we observe that these invariances are gradually learned, layer by layer, an observation in line with previous findings within the information bottleneck framework reviewed in subsection 1.2.1. As for the measures of intrinsic dimension that we discussed in subsection 1.2.1, our results are consistent with the decrease in intrinsic dimension in later layers, but hint that the increase in early layers might actually be a side effect due to the inherent difficulties of intrinsic dimension estimation, as described in subsection 1.2.1. This is because such an increase is not observed for any of the invariances considered, and measuring the intrinsic dimension of representation in deep CNNs trained on the Random Hierarchy Model does not reveal this effect either (cf. chapter 6).

In summary, our sensitivity measures advance existing tools for assessing dimensionality reduction and allow establishing quantitative relationships to performance.

## 7.3 Lessons Learned

To understand which elements in deep neural networks are responsible for learning which kind of invariances, we broke down such architectures and studied their different aspects separately. Which benefits come from feature learning and/or architectural advancements like depth and convolutional filters?

**Shallow Neural Networks** In our study of shallow neural networks we established that *(i)* they are able to learn linear invariances in the feature regime, *(ii)* they are bad for certain non-linear invariances with a performance deterioration in the feature regime with respect to lazy and *(iii)* they are not able to learn hierarchical tasks in all regimes. In particular,

*(i)* In the case of 2-layer neural networks trained on classification tasks that exhibit linear invariance (as in Equation 7.1), we established that in the feature learning regime, fully-connected neural networks can adapt effectively to data structure by orienting their weights and develop an invariance to irrelevant input directions. However, we show this is not the case when the networks are trained in the lazy regime, which leads to a performance difference that we quantified via scaling exponents of generalization error versus the number of training points, and have shown these exponents to be practically tight——a finding scarce in existing literature. We highlight that here the target is still given by the sign of a smooth function, and for this reason kernels are not cursed by dimensionality. Indeed, bounds to the scaling exponent $\beta$ for a problem of this kind would predict $\beta \geq 1/4$ Bartlett and Mendelson (2001). In our setting, such bound is not tight as, for large $d$, we find and $\beta_{\text{lazy}} = 1/3$, as also highlighted in Paccolat et al. (2021b).

Finally, to draw qualitative parallels between the learning of linear invariances by neural networks and that of more complex invariances of deep CNNs on real-world tasks, we studied the neural tangent kernel and the one computed from the last layer activations, after training. We found these kernels to retain the properties of the neural network, displaying only a few non-negligible eigenvalues, whose corresponding eigenvectors have a large projection on the target function. This finding showed that dimensionality reduction takes place in the feature regime both in the presence of linear invariance for shallow FCNs, and more complex invariances for deep CNNs. Determining the exact nature of these complex invariances remained an open question at this stage. The low-dimensionality of the space of the final layer activations has been further characterized in the context of the *neural collapse* phenomenon Papyan et al. (2020). Moreover, more recent works Guth et al. (2023) report that low-dimensional structures can be found not only at the last, but at all layers of deep networks.

*(ii)* In which cases learning features in shallow networks is instead a disadvantage? Building on the results of Geiger et al. (2020b); Lee et al. (2020), we empirically characterized the performance vs. number of training points of shallow networks trained on image classification tasks and showed a systematic gap between feature and lazy regime, emphasizing the presence of drawbacks of learning features in shallow fully-connected networks.

We proposed to rationalize these drawbacks by arguing that shallow FCNs, when trained in the feature regime, tend to overfit irrelevant input directions in the absence of linear invariances. We quantitatively characterized this overfitting in the case of rotationally invariant tasks as defined in Equation 7.2. In this setting, we computed tight generalization error rates with the number of training points for 2-layer FCNs and show a performance gap in favor of the lazy regime. Notice that, consistently with our discussion on generalization in kernel methods of subsection 1.1.3, the curse of dimensionality can only be beaten if the target smoothness $\nu_t$ is proportional to the input dimension (cf. chapter 4, Equation 3.4). If that is the case, for $d \gg 1$ we find $\beta_{\text{feature}} = 1$ and $\beta_{\text{lazy}} = 2$, highlighting that the optimal generalization bound for kernel methods is not tight in this setting, as it would predict $\beta_{\text{lazy}} = 1/2$.

In practical applications, these results suggest that when the selected architecture is not ideally suited to the task, gradient descent could potentially lead to overfitting solutions. In such cases, resorting to a kernel method might be a more prudent choice.

*(iii)* In the case of hierarchically compositional tasks exhibiting synonymic invariance, we showed that 2-layers networks are not able to learn the invariance and hence are cursed by dimensionality, with a sample complexity scaling exponentially with the input dimension.

These results underscore a limitation of studying the benefits of feature learning by considering 2-layer neural networks. Indeed, in real-world settings, 2-layer fully-connected neural networks do not perform well, especially for more complex datasets (Figure 7.4). Surprisingly, the lazy regime often yields better performance in these settings. These observations suggest that linear invariance might not be the most relevant aspect of data structure that enables the learnability of real data. Thus, while building a robust understanding of 2-layer networks is a necessary first step to making further progress, the study of more complex architectures appears necessary to explain the success of deep learning. In particular, what is the role of depth? What is the role of local filters implemented through convolutional layers?

**The Role of Depth and Locality**   *Deep* learning has demonstrated an ability to capture a hierarchy of increasingly abstract representations with depth Zeiler and Fergus (2014); Yosinski et al. (2015); Olah et al. (2017). Our work demonstrated that depth is indispensable for learning hierarchical tasks in the feature learning regime. In particular, we have shown that *(i)* for a toy model of hierarchically compositional tasks, deep networks are able to perform well by learning the associated synonymic invariance and *(ii)* for real-world tasks, specifically image classification, deeper networks generally perform better as they are able to better learn deformation invariance. Alongside depth, the utilization of local filters is crucial for image classification tasks. However, the advantage of using local filters for tasks that are purely hierarchical, such as the ones we introduced, remains a subject for further exploration and is discussed in section 7.5 regarding open questions.

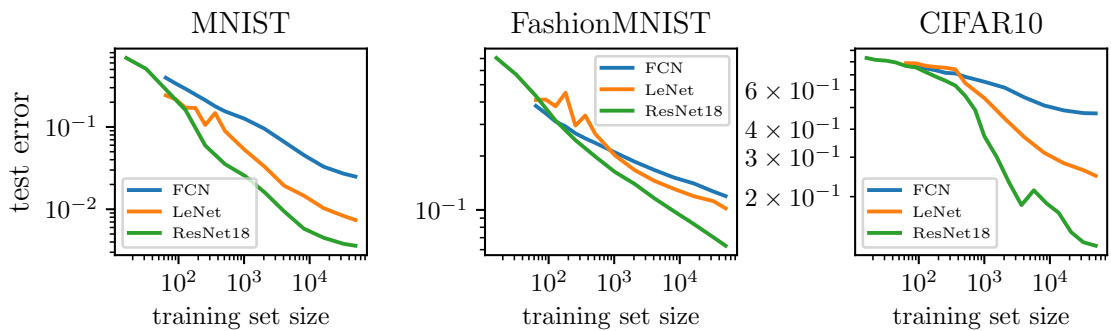

Figure 7.4: Test error rates of Fully-Connected Networks (FCN), LeNet LeCun et al. (1989), and ResNet18 He et al. (2016) with increasing number of training points, for different image datasets. As depicted, a widening performance gap emerges between FCN and the two convolutional neural networks (LeNet and ResNet18) with the increase in dataset complexity (left: black and white digits, center: black and white clothes, right: colored objects).

More specifically,

*(i)* We developed the *Random Hierarchy Model* to mimic the hierarchical structure of real data and study synonymic invariance. We find that deep, but not shallow, neural networks are able to profitably learn this invariance, and beat the curse of dimensionality. Interestingly, we have shown that the possibility to detect input-output correlations is crucial for deep neural networks trained by gradient descent to learn, as previously noted in related but different settings Arora et al. (2014); Malach and Shalev-Shwartz (2018, 2020). While our quantitative characterization of the sample complexity in this setting offers an estimate of the data required to learn a task based on its hierarchical structure, how to fit the parameters of the model ($m, v, s, L$) to real data remains an open question.

*(ii)* In real data, these features, other than being hierarchically composed, also need to be identified in space. In particular, their exact position may not matter, in the sense that they can be slightly displaced without changing the label. This property gives rise to deformation invariance. For image classification tasks we found that neural networks that are more invariant to input deformations are also the ones that perform best, as hypothesized in Bruna and Mallat (2013); Mallat (2016). Notably, neural networks employing local filters are better than fully connected ones in learning the invariance, and deeper CNNs are better than shallow ones. By examining the mechanisms responsible for granting deformation invariance, we found that deep CNNs often achieve it by utilizing low-frequency filters.

Complementary to our findings, other works have explored the benefits of depth and locality when features are not learned, i.e. in the lazy regime, and regression tasks Bietti and Bach (2021); Favero et al. (2021); Bietti (2021); Cagnetta et al. (2022); Xiao (2022); Misiakiewicz and Mei (2022). In particular, for fully connected networks, the NTK of a deep network has no advantages over its shallow counterpart Bietti and Bach (2021). Instead, when filters are local, as for a deep CNN, then the NTK can learn targets that are local on patches of different sizes, without suffering from the curse of dimensionality Cagnetta et al. (2022).

For classification tasks, preliminary results obtained from the Random Hierarchy Model (not shown) suggest that the NTK of a deep CNN cannot learn tasks that are hierarchical if the target function has full support, exhibiting a sample complexity exponential in $d$. These preliminary findings further emphasize the fundamental importance of learning features for the success of deep neural networks.

## 7.4 The Big Picture

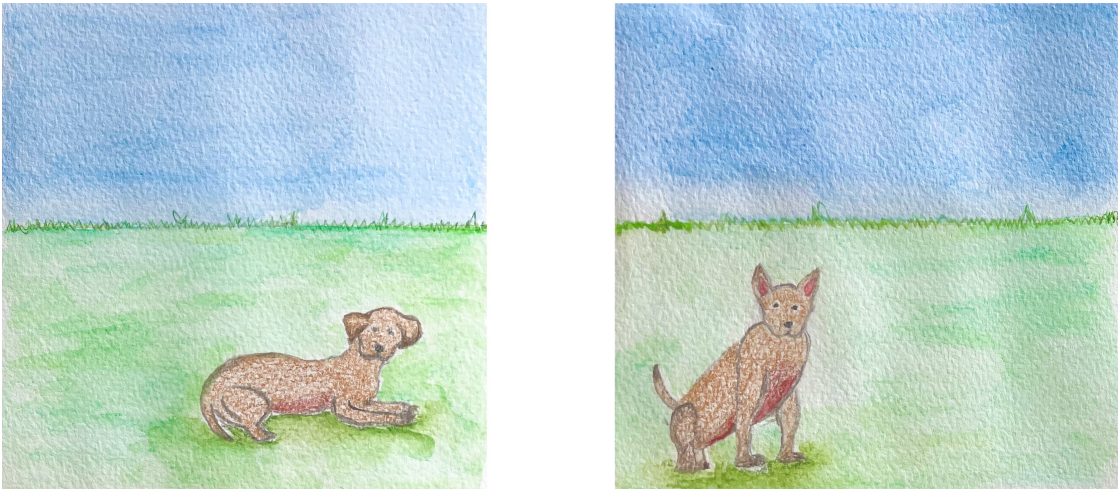

Figure 7.5: **The task of recognizing an object in a visual scene.** In both drawings we are able to recognize a dog even if some features of the dog come in different positions and shapes (e.g. the legs), or in different realizations (e.g. the ears). In this thesis, we argue that the invariance of the task to such changes is crucial for neural networks to learn it, as it allows for reducing the dimensionality of the problem and breaking the curse of dimensionality. Illustration courtesy of A.P.

The task of recognizing objects within a visual scene, such as identifying a dog as depicted in Figure 7.5, involves several steps.

Firstly, *(i)* we must discern which parts of the scene are irrelevant to our task, such as pixels in the upper corners. Secondly, *(ii)* it becomes apparent that solving our task does not require an exact match to a specific, idealized image of a dog. Instead, the features that constitute the dog can manifest in a multitude of positions, and the precise relative location of these features is not vital to our task. Lastly, *(iii)* upon recognizing that the exact placement of these features is unimportant, we begin to compose them to form a coherent picture of the object. We discern edges and textures, realizing that they form components of the dog such as legs, eyes, and mouth. Gradually, these elements combine, allowing us to recognize the dog as a whole. It's noteworthy that these elements can appear, not only in various positions or shapes, but also in various *synonymous* realizations, that are not just deformations or rearrangings of the constituent features in space; for example, the ears might be pictured from their back and appear droopy and brown when relaxed Figure 7.5(left), or show their pink interior when perked up if the dog is alert (right).

This thesis posits that these three aspects of real data structure are relevant in solving practical tasks like image classification, with particular emphasis on the significance of *(ii)* and *(iii)*. If this is the case, then unraveling how neural networks tackle these properties is fundamental to understanding their remarkable success. Specifically, we argue that these properties can be more precisely defined in terms of data invariances: *linear invariance* for *(i)*, *deformation invariance* for *(ii)*, and *synonymic invariance* for *(iii)*. We propose that in order to effectively leverage these invariances, neural networks must learn them from data since they are not typically built into the network's architecture from the start. As such, feature learning and adaptation are essential for the optimal performance of neural networks. Yet, we also demonstrate that the architectural selection of the network is equally important for proper feature learning. A suboptimal architectural choice could result in learning the wrong features, leading to poor performance. Thus, implementing proven architectural advancements, such as the use of local filters through convolutional layers and the incorporation of multiple layers, plays a critical role in successful feature learning.

## 7.5   Some Open Questions

This final section outlines the limitations of our study and suggests potential avenues for future exploration. We first discuss how to unify the two main lines of inquiry in this thesis: the roles of deformation and synonymic invariance both *(i)* in the Random Hierarchy Model and *(ii)* in real-world tasks. Then *(iii)* we show novel results aimed at disentangling the role of depth and local filters in learning hierarchical tasks. *(iv)* We propose extensions to the Random Hierarchy Model to get the power-law behavior of test error that is often observed in practice. We conclude *(v)* by discussing practical applications of our ensemble of diffeomorphisms, specifically for data augmentation and confidence estimation.

*(i)*  Our study on deformation invariance currently does not provide a theoretical framework for understanding the observed correlation with test error. Such an understanding would allow for making quantitative predictions on how many training points would be needed to learn the invariance, hence the task. While some progress in this direction has been made in the study of kernel methods Bietti et al. (2021), the sample complexity of algorithms that are allowed to learn features is still unknown.

One way to make progress consists in unifying the two main lines of inquiry of this thesis by creating a data model that can be invariant to both deformations and to the exchange of synonymous features. This can be done by adding a notion of *sparsity in space* to the Random Hierarchy Model (RHM) in such a way that, at every hierarchy level, the relevant features can equivalently appear at different locations of a given patch, with the rest of the patch being empty or noisy, irrelevant pixels. This construction is such that exchanging the position of a relevant feature with an irrelevant pixel within a patch does not change the task, hence modeling invariance to small deformations. Some preliminary results in this direction (to appear in *Tomasini et al. 2023*) show that synonymic and deformation invariance are learned together in deep CNNs. From these results, we can derive the following interpretation. The RHM can be efficiently learned only by understanding the structure of the problem at all levels, which involves achieving synonymic invariance: a model that did not learn synonymic invariance could only memorize the input samples, and would need to see all of them to solve the task. Hence a correlation between synonymic invariance and performance comes naturally in this setting. The preliminary findings we report here—specifically the correlation between deformation and synonymic invariance—suggest that deformation invariance is indicative of the goodness of a network representation on various aspects, hence rationalizing its strong correlation to performance. Estimating the sample complexity as a function of both invariances is still an ongoing work.

*(ii)* A related and complementary direction would include studying synonymic invariance in real-world tasks: when learning to predict e.g. age from face pictures, is the invariance to exchange of synonymous features (e.g. eyes with or without sunglasses) correlated to the performance of a given neural network, as predicted from our findings in chapter 6? Are synonymic and deformation or 2D/3D rotation Goodfellow et al. (2009) invariance learned together in real-world tasks as well? Moreover, the transformer architecture Vaswani et al. (2017) has been the most recent technological breakthrough in the context of natural language processing. The study of synonymic invariance in real tasks may include understanding how transformers learn that the exchanges of two synonyms do not change e.g. the sentiment of a sentence, and whether measures of the invariance of a transformer to such exchanges are also correlated to its performance.

*(iii)* In our study of hierarchical tasks we mainly focus on shallow FCNs and deep CNNs. One important question though is how locality is learned, when not imposed from the start, or "What is the performance of deep FCNs?". In the case of a target function just depending on one of the input patches, locality corresponds to the linear invariance studied in chapter 2. However, when the function depends on different patches independently, as in

$$f^*(\boldsymbol{x}) = g_3(g_1(\boldsymbol{x}_1), g_2(\boldsymbol{x}_2)), \tag{7.7}$$

where $\boldsymbol{x}_1$ and $\boldsymbol{x}_2$ are two input patches, the picture is different. In this case, the constituent functions $g_1$ and $g_2$ are invariant to all the input except $\boldsymbol{x}_1$ and $\boldsymbol{x}_2$, respectively, but the target function is not. As a consequence, to solve the task, different neurons need to specialize to different parts of the input. Understanding when and how non-local neurons can achieve such specialization in practice to learn compositionally local functions is still an open question Neyshabur (2020); Pellegrini and Biroli (2021); Ingrosso and Goldt (2022).

We believe that the Random Hierarchy Model (RHM) provides a practical setting to start addressing this question. To this end, we conducted some preliminary experiments on deep fully-connected networks trained on the RHM, see results in Figure 7.6. Neurons are indeed able to achieve specialization: they clearly separate into distinct groups, each responding to one of the different input patches. Interestingly, (not shown) we observe that this specialization occurs at all hierarchy levels, with neurons in successive layers specializing to *patches of patches*, and so on. These experiments led to the following insight: Deep FCNs can learn the Random Hierarchy Model with a sample complexity that is orders of magnitude smaller than the one of shallow FCNs, corresponding to the total number of data points. These preliminary results suggest that depth is crucial for learning tasks that are solely hierarchical, but imposing locality from the start is not. However, additional experiments with varying RHM parameters ($L$ in particular) are needed to validate this finding and characterize the scaling of the sample complexity of deep FCNs. Our current hypothesis is that $P^*_{\mathrm{dFCN}} = C(d)P^*_{\mathrm{dCNN}}$, with $C(d)$ a prefactor polynomial in $d$, that could be large for large $d$. Hence, even if both deep FCNs and CNNs beat the curse of dimensionality, deep CNNs may still have a relevant edge on performance. Ultimately, while we observe neuron specialization in practice, a comprehensive theoretical understanding of this phenomenon remains an open challenge.

*(iv)* The Random Hierarchy Model serves as a toy model, and while it may provide useful insights, it clearly does not perfectly reflect real-world data. Future investigations could profitably extend to more sophisticated and realistic models. We already discussed how to include the sparsity of features in space in the model. One other possible extension drawn from the context of natural languages is the incorporation of *Zipf's law* Zipf (1935), an empirical principle in linguistics, stating that the frequency of a word in a natural language is a power law of its rank in the frequency table.

Currently, in the Random Hierarchy Model, the features are uniformly distributed and hence all appear with the same probability. As a consequence, there exists a well-defined scale $P^* = n_c m^L$ at which all correlations between input features and labels emerge from sampling noise. If, instead of being uniform, the features were distributed as a power law, then correlations of rarer features would need an increasing number of training points to emerge from sampling noise. As a consequence, the sharp transitions we observe in our setting around $P^*$ for the test error versus number of training points could manifest as power-law behaviors when Zipf's law is incorporated into the model. Indeed, power-law behaviors have been observed in large language models Kaplan et al. (2020).

*(v)* We conclude by discussing the practical application of the ensemble of diffeomorphisms we introduced to data augmentation and confidence estimation.

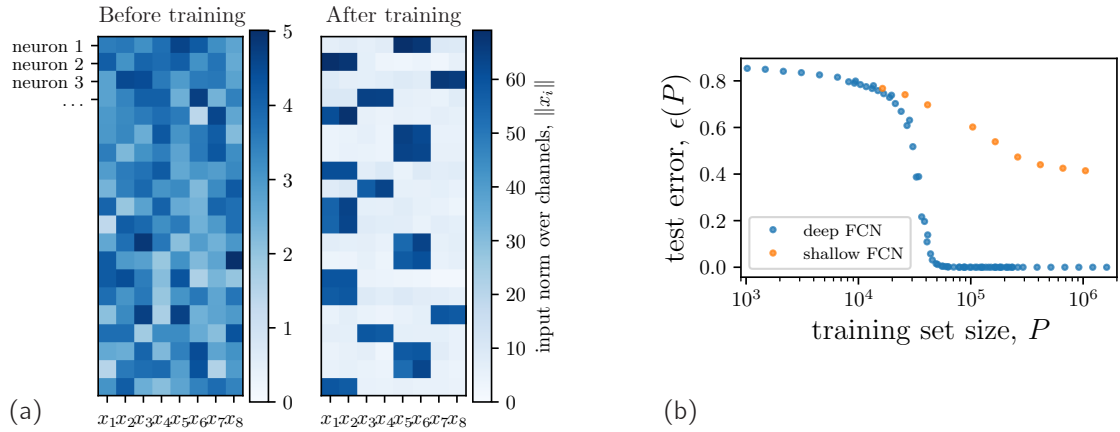

(a)                                                                                    (b)

Figure 7.6: **Deep FCNs Learn Hierarchical Tasks.** 3-hidden layers fully-connected network trained on the Random Hierarchy Model with three hierarchy levels ($L = 3$) and patches of size $s = 2$, resulting in input dimension $d = 8$. The number of classes equals number of features and multiplicity $n_c = v = m = 8$, resulting in a predicted sample complexity for deep CNNs of $P^*_{\text{dCNN}} = n_c m^L = 4'096$ and a total data set size $P_{\max} = n_c m^{d-1} \approx 10^8$. (a) On the rows, there are different neurons, on the columns different input locations $x_1, \ldots, x_d$. The color indicates the weights norm across the $v$ input channels. Left and right correspond to before and after training with $P^*_{\text{dCNN}} \ll P \ll P_{\max}$. FCN learns locality in the sense that different neurons specialize to different input patches $(x_1, x_2)$, $(x_3, x_4)$, etc. with training. A similar behavior can be observed for the following layers focalizing on the next level of the hierarchy, i.e. *patches of patches*. (b) Test error vs. number of training points for Deep (blue) vs Shallow (orange) FCNs. Deep networks are able to reach zero test error with a number of training points $P$ that satisfies $P^*_{\text{dCNN}} < P \ll P_{\max}$, while shallow do not.

Data augmentation consists in creating additional training samples by applying transformations to existing data that do not change the label, thereby enhancing the networks' ability to generalize or robustness to adversarial attacks Ortiz-Jiménez et al. (2021). Although we found that diffeomorphisms didn't outperform standard data augmentation methods when trying to improve generalization, successful applications of our maximum-entropy transformations in this domain of adversarial robustness have appeared in the literature Modas et al. (2022), suggesting this may be a promising direction for further future work.

Confidence estimation is the process of quantifying the certainty or confidence level of a model's predictions Guo et al. (2017). The goal is to assess how likely the model's prediction is to be correct, providing a measure of trustworthiness for the output that is crucial in high-stakes applications like healthcare or finance. Our ensemble of diffeomorphisms can be employed to design better confidence estimators: by exploring the value of the predictor in the neighborhood of data points, other than *on* the data points alone, this could provide more accurate confidence estimations. Current work, that I'm co-supervising, involves this approach. Preliminary results indicate that averaging standard confidence estimators over max-entropy diffeomorphisms of a specific input yields improved confidence estimators, that are more informative than standard ones, and better calibrated. These findings will appear in Hasler et al. (2023).

# Leonardo**Petrini**
PhD Student, Physics and Machine Learning @ EPFL

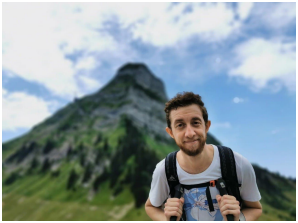

## about
**Currently** in Lausanne, CH

website: leopetrini.me
leo.petrini@outlook.com
github/leonardopetrini
slides.com/leopetrini
twitter.com/leopetrini_

## languages
**italian** native
**english** full proficiency
**french** fluent

## programming
**Python** advanced (6yrs)
(Numpy, sklearn, Pandas)
**PyTorch** advanced (4yrs)
**Julia** beginner (1yr)

## soft skills
curiosity
team work
communication

## interests
food
deep learning
(personal) finance
mountains
photography

## higher education

**2019-present** **PhD Student @ Physics of Complex Systems Lab**    EPFL, Lausanne CH
Deep Learning Theory

**summer '18** **CERN Summer Student Program - ATLAS Experiment**    CERN, Meyrin CH
Project: Classification and Regression Studies for Flavour Tagging

**2017 - 2019** **Master in Physics @ EPFL (GPA: 5.7/6)**    EPFL, Lausanne CH
*Minor in Computational Science and Engineering*
Master Thesis: *Replicated Affinity Propagation Algorithm.*
Supervisor: Prof. Riccardo Zecchina, Artificial Intelligence Lab–Bocconi University

**2016 - 2017** **ETH Exchange program**    ETHZ, Zurich CH
Visiting student

**2014 - 2017** **Bachelor cum laude (110L/110)**    Politecnico di Torino, Turin IT
Physical Engineering and Young Talents Program (Progetto Giovani Talenti)

## publications

**2023** **How Deep Neural Networks Learn Compositional Data:
The Random Hierarchy Model**
Paper under review
LP, F. Cagnetta, U.M. Tomasini, A. Favero, M. Wyart (arXiv link)

**2022** **How deep convolutional neural networks lose spatial information with training**
Paper @ *ICLR 2023 Workshop, Physics for ML*
U.M. Tomasini, **LP**, F. Cagnetta, M. Wyart (arXiv link)

**2022** **Learning sparse features can lead to overfitting in neural networks**
Paper @ *NeurIPS 2022*
LP, F. Cagnetta, E. Vanden-Eijnden, M. Wyart (OpenReview link)

**2021** **Relative stability toward diffeomorphisms indicates performance in deep nets**
Paper @ *NeurIPS 2021*
LP, A. Favero, M. Geiger, M. Wyart (OpenReview link)

**2020** **Landscape and training regimes in deep learning**
Paper @ *Physics Reports*
M. Geiger, LP, M. Wyart

**2020** **Geometric compression of invariant manifolds in neural networks**
Paper @ *Journal of Statistical Mechanics: Theory and Experiment*
J. Paccolat, LP, M. Geiger, K. Tyloo, M. Wyart

## teaching and reviewing

- Teaching assistant for Statistical Physics II and III, 2019 to 2022.
- Teacher and supervisor of semester and master projects, 2019 to 2023.
- Reviewer for the Journal of Machine Learning Research (JMLR), 2022.
- Workshop on the Theory of Overparameterized Machine Learning (TOPML), 2022
- Reviewer for the Conference on Neural Information Processing Systems (NeurIPS), 2023.

## conferences and schools

February '23   **NeuroStatPhys Workshop [poster]**
Ecole de Physique des Houches, FR

August '22   **IAIFI PhD Summer School and Workshop [poster]**
Institute for Artificial Intelligence and Fundamental Interactions, Boston, US

Apr. '22   **Workshop on the Theory of Overparameterized Machine Learning [talk]**
https://topml.rice.edu/

Sept. '21   **On Future Synergies for Stochastic and Learning Algorithms [poster]**
CIRM Marseille, FR

June '21   **Statistical Mechanics and Emergent Phenomena in Biology [poster]**
The Beg Rohu Summer School, FR

June '21   **Youth in High Dimensions Conference [poster]**
ICTP, Trieste, IT

March '21   **How neural nets compress invariant manifolds [talk]**
Americal Physical Society, March Meeting, US

August '20   **Statistical Physics and Machine Learning Workshop [talk]**
Ecole de Physique des Houches, FR

## secondary education

2018   **National High School Model United Nations NHSMUN**   New York, NY
Faculty Advisor for Liceo Scientifico G. Galilei, Macerata

2014   **National Physics Olympiad, Italy**   Senigallia, IT
Winner of 1st and 2nd level competitions, admitted to the national stage

2014   **National Mathematical Olimpiad, Italy**   Cesenatico, IT
Team competition

2013   **National Physics Olympiad, Italy**   Senigallia, IT
Winner of 1st and 2nd level competitions, admitted to the national stage

2013   **National High School Model United Nations NHSMUN**   New York, NY
Nigeria delegation, Legal Committee

2013-2014   **Scientific degrees program, University of Camerino**   Camerino, IT
Progetto lauree scientifiche, physics division

2012   **National Mathematical Olimpiad, Italy**   Cesenatico, IT
Team competition

2009-2014   **Liceo Scientifico G.Galilei Macerata**   Macerata, IT
PNI, Piano Nazionale Informatica *(National program for informatics)*